\RequirePackage{fix-cm}
\documentclass{svjour3}                     
\smartqed  
\usepackage{graphicx}
\journalname{MTAP}

\usepackage{cite}

\usepackage{color}
\usepackage[table,xcdraw]{xcolor}

\usepackage{float}
\usepackage{subfloat}
\graphicspath{{figs/}}
\graphicspath{{charts/}}
\usepackage{subfig}
\usepackage{amsmath}
\usepackage{amssymb}
\usepackage{amstext}

\usepackage[numbers]{natbib}

\usepackage{algorithm}
\usepackage{listings}

\usepackage{grffile}

\usepackage{amsmath}

\usepackage{placeins}

\usepackage{url}
\usepackage{hyperref}
\usepackage{gensymb}

\floatname{algorithm}{Algorithm}
\lstdefinelanguage{myPascal}[]{Pascal}{
	morekeywords={end,while,for,if,else}
}

\newcommand{\rtorres}[1]{#1}

\newcommand{\edit}[1]{#1}
\newcommand{\editt}[1]{#1}

\begin{document}

\title{A Genetic Algorithm Approach for Image Representation Learning through Color Quantization}

\author{Erico M. Pereira$^{1}$ \and Ricardo da S. Torres$^{2,3}$ \and Jefersson A. dos Santos$^{1*}$}

\institute{
	$^1$Department of Computer Science, Universidade Federal de Minas Gerais -- Av. Ant\^{o}nio Carlos, 6627, Belo Horizonte, MG - Brazil -- 31270-010\\ \email{\{emarco.pereira, jefersson\}@dcc.ufmg.br}\\
	$^2$Institute of Computing, University of Campinas -- Av. Albert Einstein, 1251, Campinas, SP - Brazil\\
	$^3$Department of ICT and Natural Sciences, NTNU -- Norwegian University of Science and Technology -- {\AA}lesund -- Norway\\
	\email{ricardo.torres@ntnu.no}\\
	$^*$Corresponding Author: jefersson@dcc.ufmg.br
}

\date{Received: date / Accepted: date}

\maketitle


\begin{abstract}
Over the last decades, hand-crafted feature extractors have been used to encode image visual properties into feature vectors. 
Recently, data-driven feature learning approaches have been successfully explored as alternatives for producing more representative visual features. In this work, we combine both research venues, focusing on the color quantization problem. We propose two data-driven approaches to learn image representations through the search for optimized quantization schemes, which lead to more effective feature extraction algorithms and compact representations. Our strategy employs Genetic Algorithm, a soft-computing apparatus successfully utilized in Information-retrieval-related optimization problems. We hypothesize that changing the quantization affects the quality of image description approaches, leading to effective and efficient representations. We evaluate our approaches in content-based image retrieval tasks, considering eight well-known datasets with different visual properties. Results indicate that the approach focused on representation effectiveness outperformed baselines in all tested scenarios. The other approach, which also considers the size of created representations, produced competitive results keeping or even reducing the dimensionality of feature vectors up to 25\%.


\end{abstract}

\vfill

\section{Introduction}
\label{sec:introduction}

It is known that the form in which multimedia data, especially images, are represented can highly impact the performance of machine learning methods typically used in visual pattern recognition tasks, such as content-based image retrieval (CBIR)~\cite{Smeulders:2000}, object detection~\cite{Yilmaz:2006}, remote sensing image analysis~\cite{Davis:1978}, and image classification~\cite{Lu:2007}.
In the last years, representation learning~\cite{Bengio:2013}, which consists in the process of using pattern recognition algorithms to find representations optimized for a given data domain and/or task at focus, has become a tendency. In fact, the current state-of-the-art methods for representation learning, which are based on deep learning~\cite{Lecun:2015} techniques, in many cases present considerable gains in terms of the image content description quality. 

However, the use of these methods presents serious drawbacks, such as
the broad range of hyper-parameters and possible architectures,
the huge computational workload spent to train existing models,
the big amount of labeled data required to produce effective models,
and the need of specific expertise or training for properly designing, optimizing, and evaluating promising solutions.

Representation learning methods usually employ one of two main approaches: those that learn representations from a feature set provided by a hand-crafted extractor and those that completely compose new ones without any prior feature extraction (from scratch). 
The latter approach often leads to the usage of more complex and consequently costly methodologies, such as deep learning.
Such complexity, however, used to be avoided in the generation of representative features.
A few years ago, before the arising of deep neural networks, hand-crafted feature extractors were used to encode image visual properties (e.g., color, texture, or shape) into effective representations~\cite{Penatti:2012,DosSantos:2010, Penatti:2008}. In general, those solutions rely on less costly algorithms and do not depend on previously annotated datasets or time-consuming learning steps. On the other hand, these feature extractors are application-dependent, being less generalizable.

In this paper, we propose a hybrid scheme, focused on color quantization, which aims to take advantage of both research venues. We propose data-driven color quantization schemes, which improve the effectiveness of hand-crafted feature extractors\edit{, as it allows for the identification of discriminative visual features.}
Our representation learning scheme exploits a particular characteristic of the current image context, its color distribution\edit{, a simple but yet suitable visual cue in several applications~\cite{DBLP:journals/mta/LiLPZCZX19,9004613,8768511}}. \edit{Our hypothesis is that data-driven quantization optimizations are able to positively impact the quality of image content description approaches, leading to effective and efficient representations. In this paper, we investigate how these optimizations can be performed effectively and efficiently and to what extent.}

Our color quantization optimization relies on a soft computing framework, implemented using genetic algorithms (GA). GA is an evolutionary algorithm widely used to solve optimization problems. According to its formulation, a population of individuals, representing possible solutions of a problem, evolves over generations, subjected to genetic operations. The goal is \edit{to} find the best individuals, i.e., the best solutions for the problem. In our color quantization problem, a GA individual encodes how color channels should be divided in order to improve the effectiveness of feature extractors.
To the best of our knowledge, this is the first work to use GA to model the representation learning problem.

In summary, the main contributions of this work are:

\begin{enumerate}
	\item We show that different color quantizations impact the effectiveness performance of feature extractors;
    \item We model the search of suitable color quantization using a soft computing apparatus based on the genetic algorithm;
	\item We introduce two approaches for supervised representation learning capable of providing compact and more effective representations through color quantization optimization.
\end{enumerate}

\edit{In summary, the main novelty of our work relies on the presentation of an integrative framework for the implementation of effective image search systems that combines several concepts, approaches and techniques, such as, Genetic Algorithm optimization, Color Quantization, Representation Learning, Feature Extraction, and Content-Based Image Retrieval.} 

We conducted a series of experiments in order to evaluate the robustness of the proposed approaches in content-based image retrieval tasks, considering eight well-known datasets containing images with different visual properties.
Experimental results indicate that the approach focused on the representation effectiveness outperformed the baselines in all tested scenarios. The other approach, which focuses not only on the effectiveness, but also on the size of the generated feature vectors, was able to produce competitive results by keeping or even reducing the final feature vector dimensionality up to 25\%.

The remainder of this paper is organized as follows. 
Section~{\ref{sec:related_work}} presents related work.
Section~{\ref{sec:backgound}} provides a background upon base methods used in our work.
Section~{\ref{sec:methodology}} describes the proposed color quantization schemes and its use in CBIR tasks. 
Section~{\ref{sec:experimental_setup}} details the experiments performed to assess the effectiveness and efficiency of the proposed methods.
Section~{\ref{sec:results}}, in turn, presents and discusses achieved results.
Finally, Section~{\ref{sec:conclusions}} presents our main conclusions and outlines possible future research directions.

\section{Related Work}
\label{sec:related_work} 


Image representation learning (a.k.a. feature learning) consists in automatically discovering the representations needed for object detection or classification from raw images. It is a set of approaches that aim at making it easier to extract useful information when building classifiers or other predictors~\cite{Bengio2013TPAMI}. 
In other words, feature learning allows to find the most suitable or discriminative representation from the raw data according to some constraint imposed by the target application. Thus, it is also commonly known as data-driven features because of its contraposition to engineered or hand-crafted features.

Although feature learning has been an active research area for a long time, the development of effective techniques (mainly based on deep learning) has been boosted in the last decade mainly due to the spread use of powerful computational resources, which were motivated by the development of graphical processing units (GPUs). Many successful recent feature representation approaches are based on deep belief nets~\cite{Hinton:2006}, denoising auto-encoders~\cite{Vincent:2008}, deep Boltzmann machines~\cite{Salakhutdinov:2009}, K-Means-based feature learning~\cite{Coates:2011}, hierarchical matching pursuit~\cite{Bo:2011}, and sparse coding~\cite{Yu:2011}.
Regarding image representation learning, the most successful approaches are based on the Convolutional Neural Networks (CNNs)~\cite{Krizhevsky2012nips}.

Although, by definition, a large number of techniques perform feature learning, the term is most commonly employed by the community that develops methods based on deep learning or probabilistic graphical models.  These methods are the basis for most of the state-of-the-art approaches for pattern recognition and computer vision. 
Despite the recent great success of these approaches, they still have several limitations, such as a large number of parameters for optimization and the difficulty in designing network architectures.

Evolutionary algorithms are meta-heuristic optimization techniques that use mechanisms inspired by biological evolution (e.g., reproduction, mutation, recombination, and selection). They have been widely employed in a myriad of frameworks developed for image analysis and retrieval usually for feature fusion~\cite{TORRES2009PR} or selection~\cite{Oh2004TPAMI,Nakamura2014TGRS}.
In the last few years, evolutionary algorithms have also been successfully employed for neural networks architecture search~\cite{Suganuma2017GECCO,Xie2017ICCV}. 
Nonetheless, we did not find other works that directly model feature learning as an evolutionary algorithm-based problem from the raw data.

In this work, we propose to learn image features from images via genetic algorithms by color quantization optimization.
Some works developed quantization learning using evolutive heuristics for image segmentation~\cite{Luccheseyz:2001}. Scheunders~\cite{Scheunders:1996} \rtorres{handles} the quantization problem as global image segmentation and proposes an optimal mean squared quantizer and a hybrid technique combining optimal quantization with a Genetic Algorithm modelling~\cite{Goldberg:1989}. Further, the same author~\cite{Scheunders:1996} presents a genetic c-means clustering algorithm (GCMA), which is a hybrid technique combining the c-means clustering algorithm (CMA) with Genetic Algorithm. Lastly, Omran et al.~\cite{Omran:2005} developed colour image quantization algorithm based on a combination of Particle Swarm Optimization (PSO) and K-means clustering.

Regarding the effects of colour quantization on image representations, Ponti et al.~\cite{Ponti:2016} approached the colour quantization procedure as a pre-processing step of feature extraction. 
They applied four fixed quantization methods -- Gleam, Intensity, Luminance, and a concatenation of the Most Significant Bits (MSB) -- over the images of three datasets and then used four feature extractors -- ACC, BIC, CCV, and Haralick-6 -- to compute representations intended to solve the tasks of Image Classification and Image Retrieval.
Their conclusions show that it is possible to obtain compact and effective feature vectors by extracting features from images with a reduced pixel depth
and how the feature extraction and dimensionality reduction are affected by different quantization methods.

\edit{New approaches based on deep learning developed in the last ten years have revolutionized the learning of representations from data.
Regarding the learning of representations for images, convolutional networks have established themselves as the most effective solution. However, its use still has some limitations, such as: (1) they require a large amount of data for training from scratch; (2) traditional networks have a large number of parameters.
Therefore, some works have been proposed in order to mitigate these limitations and produce more compact networks~\cite{Makhzani:2015,Zhang2018:CVPR}.
In this context, approaches based on nature-inspired/evolutionary algorithms have emerged as an alternative to optimize network architectures in various ways~\cite{Rodriguez2019:GPEM,Bharti2020:Conf}. 
Although color quantization approaches are less used nowadays than in the past for image representation, they are still an alternative to obtain compact and effective representation for some applications, such as color-based image retrieval~\cite{Zeng2016:Neurocomputing,Perez2019:AppInt,Sheng2018:EMC2,Bhunia2019:PAA}.}

\edit{To the best of our knowledge, our work is the unique that provides an application-driven way to learn compact representation from color quantization. 
Note that it is not comparable to modern neural network-based compact approaches because it does not take advantage of transfer learning strategies.}

\section{Background}
\label{sec:backgound} 

This section presents background concepts on feature extraction algorithms (Section~\ref{sec:color-quantization}) and genetic algorithms (Section~\ref{sec:ga}). The feature extraction algorithms described here refer to methods that are combined with the quantization scheme defined by GA in the performed experiments.

\subsection{Color Quantization-based Feature Extraction Algorithms}
\label{sec:color-quantization}

	\noindent\textbf{Border/Interior Classification (BIC).}
	Stehling et al. \cite{Stehling:2002} proposed BIC, a simple and fast approach for feature extraction which presented prominent results in web image retrieval \cite{Penatti:2012} and remote sensing image classification \cite{DosSantos:2010,Nogueira:2017}.
	This approach relies on \edit{an RGB} color-space uniformly quantized in $4\times4\times4=64$ \text{colors}. 
	After the quantization, the authors propose to apply a segmentation procedure, which classifies the image pixels according to a neighborhood criterion: a pixel is classified as interior if its 4-neighbours (right, left, top, and bottom) have the same quantized color; otherwise, it is classified as border.
	Then, two color histograms, one for border pixels and other for interior pixels, are computed and concatenated composing a 128-bin representation.
	In the end, the histograms undergo two normalizations: division by the maximum value, for image dimension invariance, and a transformation according to a discrete logarithmic function ($dLog$), aiming to smooth major discrepancies.
	\edit{When comparing BIC via L1 distance, it was observed that the dLog function is able to increase substantially the effectiveness of histogram-based CBIR approaches and also reduces by 50\% the space required to represent a histogram.}
	
	\noindent\textbf{Global Color Histogram (GCH).}	
	GCH~\cite{Swain:1991} is a widely used feature extractor that presents one of the simplest forms of encoding image information in a representation, a color histogram, which is basically the computation of the pixel frequencies of each color. It relies on the same uniformly quantized RGB color-space such as BIC and, consequently, produces a feature vector of 64 bins. After the histogram computation, it undergoes a normalization by the max value in order to avoid scaling bias. \edit{Additionally, for the same reasons as for BIC, dLog normalization is also applied to the final histogram.}

\subsection{Genetic Algorithm}
\label{sec:ga}

GA is a bio-inspired optimization heuristic that mimics natural genetic evolution to search the optimal in a solution space~\cite{Goldberg:1989}.
It models potential solutions for the problem as individuals of a population and subjects them to an iterative process of combinations and transformations towards an improved population, i.e., a population with better solutions for the target problem. 

\edit{At each step, GA randomly selects individuals from the current population, called parents, in an operation called tournament, in which individuals are grouped and only the best ones are selected. 
From this selection, GA exchanges genetic material of the individuals in order to produce new individuals of the next generation. This operation is known as cross-over.
Some individuals are also selected to undergo a mutation operation, which consists in randomly changing small pieces of the individual representation. 
This new individual is also integrated into the new generation~\cite{Davis1991}.
Typically a few of the best individuals of the population also compose the new one, a practice known as elitism.}
When a new generation is formed, its individuals are evaluated by means of a fitness function, which assesses the individual (solution) performance on the target problem. According to this function score, the algorithm selects the parent individuals that will generate the next population, simulating a natural selection process. At the end of the process, when the stopping condition is satisfied, the expected result is the best-performing individual, i.e., the one that best solves the target problem.

\edit{
\subsection{Winner-Take-All (WTA) Autoencoders}
\label{subsec:aes}
An autoencoder~{\cite{Hinton:1994}} is a framework 
that employs representation learning by optimizing an encoding that reconstructs as well as possible the entry data. 
It is specified by a explicitly defined feature-extraction function $f_{\theta}$, called encoder, which allows the computation of a representation $z = f_{\theta}(x)$ from a given input $x$, and a parametrised function  $g_{\theta}$, that maps the representation from feature space back to input space producing a reconstruction $r = g_{\theta}(x)$.
The set of parameters $\theta$ of the encoder and decoder are learned simultaneously by reconstructing the original input $x$
with the lowest possible discrepancy $L(x,r)$ between $x$ and $r$, employing a optimization process that minimizes:
\begin{equation}
	\Gamma_{AE}(\theta) = \sum_t L(x^{(t)}, g_{\theta}(f_{\theta}(x^{(t)})))
\end{equation}
where $x^{(t)}$ is a training sample.}

\edit{
It is crucial that an autoencoder presents good generalization, i.e., that the produced representations yield low reconstruction error for both train and test samples.
For this purpose, it is important that the training criterion or the  parametrisation prevents the auto-encoder from learning the identity function to the training samples, which presents zero reconstruction error. 
This is achieved by imposing different forms of regularisation in different versions of autoencoders. Regularized Autoencoders limit the representational capacity of $z$ provoking a bottleneck effect that does not allow the autoencoder to reconstruct the whole input and forces it to learn more meaningful features.
As a consequence, it is trained to reconstruct well the training samples and also present small reconstruction error on test samples, implying generalization. }

\edit{
The most common types of regularised autoencoders include:
Sparse autoencoders~{\cite{Ranzato:2007,Ng:2011,Makhzani:2013}}, which limit capacity by imposing a sparsity constraint on the learnt representation of the data; 
Denoising autoencoders~{\cite{Vincent:2008,Vincent:2010}}, which has the objective of removing noise of an artificially corrupted input, i.e. learning to reconstruct the clean version from a corrupted data;
Contractive Autoencoders~{\cite{Rifai:2011}}, which penalize the sensitivity of learned features to input variations producing more robust features;
and Variational Autoencoders~{\cite{Kingma:2013}}, which learn probabilistic latent spaces in order to generate artificial samples.}

\edit{
Winner-takes-all Autoencoders (WTA-AE)~{\cite{Makhzani:2015}} are sparse autoencoders that employ two types of sparsity constraints: 
\begin{itemize}
    \item A spatial sparsity constraint, 
which, rather than reconstructing the input from all of the representational hidden units, selects the single largest value within each feature map, and set the rest to zero. This results in a sparse representation whose sparsity level is the number of feature maps and in a reconstruction which uses only the active hidden units in the feature maps; 
    \item A \textit{winner-take-all} lifetime sparsity constraint, 
which maintains only the $k\%$ largest values of each feature map, and set others to zero,
considering the values selected spatial sparsity within an entire mini-batch.
\end{itemize}}

\edit{
We choose WTA as baseline because it is one of the most robust and efficient Sparse Enconders -- the most effective class of (non-generative) methods based on deep learning that are dedicated for feature extraction/representation learning.
WTA autoencoders were capable of aiming at any target sparsity rate, training very fast compared to other sparse autoencoders, and efficiently training all hidden units even under very aggressive sparsity rates (e.g., 1\%).
Furthermore, the usage of its sparsity properties allows the train of non-symmetrical architectures (different sizes for encoder and decoder) reducing computation and data resource consumption. }

\section{\rtorres{GA-based Color Quantization}}
\label{sec:methodology} 

\begin{figure}[!t]
	\centering
	\includegraphics[width=\textwidth]{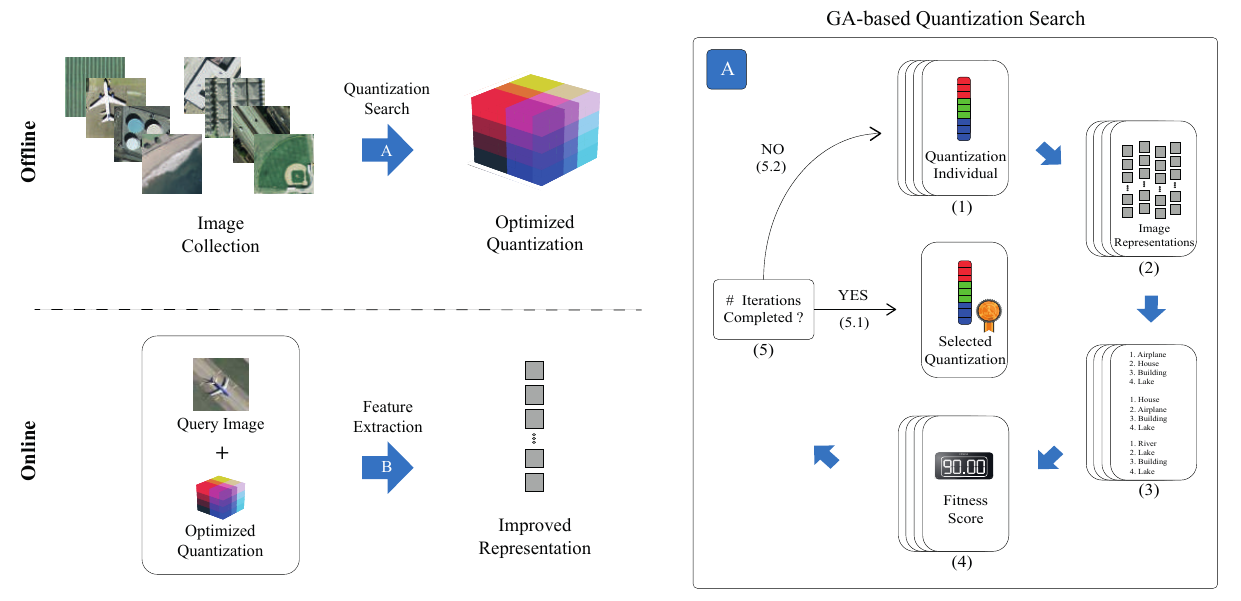}
	\caption{
Overview of the proposed approach. First, (A) we use Genetic Algorithm to search for an optimized color quantization. 
Later, (B) the resulting quantization is incorporated into the feature extractor to generate improved image representations.
The GA-based quantization search proceeds as follows: first, (1) a population of encoded color quantizations is randomly produced; second, (2) sets of image representations of the whole collection are produced being each one according to one quantization color space; third, (3) similarity rankings for all to all images are computed within each representation set; and \edit{fourth}, (4) a fitness score is computed to measure each retrieval effectiveness. Finally, \edit{if the stopping condition is met or} the total number of iterations is achieved, (5.1) the quantization of the highest fitness of the last population is selected as the optimized colour space, otherwise, (5.2) a new population is created, via crossover and mutation operations over the current population, initiating the next iteration.}	
	\label{fig:method}
\end{figure}

In this paper, we introduce the use of Genetic Algorithm to learn an optimized color quantization for a given image domain. Figure~\ref{fig:method} provides an overview of the entire process, which is composed of two main steps: (A) quantization search, and (B) feature extraction. These steps are described next.

\subsection{Quantization Search}

We propose the use of
Genetic Algorithm~\cite{Goldberg:1989} to learn the best color quantization for a given collection.
GA has been a widely used approach for finding near-optimal solutions for optimization problems.
One remarkable property of this optimization apparatus relies on its ability in performing parallel searches starting from multiple random initial search points and considering several candidate solutions simultaneously.
Consequently, it represents a fair alternative to an exhaustive search strategy, which would be unfeasible given the number of possible solutions.

According to this optimization algorithm, an individual corresponds to a representation of a potential solution to the problem that is being analyzed.
In our modeling, each individual represents a possible color quantization, as detailed in Section~\ref{subsec:encoding}.
During the evolution process, described in Section~\ref{subsec:search}, these individuals are gradually evolved. At the end of the evolutionary process, the best-performing individual, which encodes a quantization that leads to an improved representation, is selected.

\subsubsection{Quantization Encoding}
\label{subsec:encoding}

In our modeling, a quantization is represented in a GA individual as follows: 
Let $\mathbb{M}$ be a color model composed of three channels. Without loss of generality, we will assume the RGB color model from now on. Assume that each channel is divided into 256 discrete levels, i.e., eight bits can be used to define the number of colors in each channel. In the case of the traditional 24-bit RGB model, there are almost 17 million ($256 \times 256 \times 256$) different colors.

\begin{figure}[!t]
	\centering
	\includegraphics[width=\textwidth]{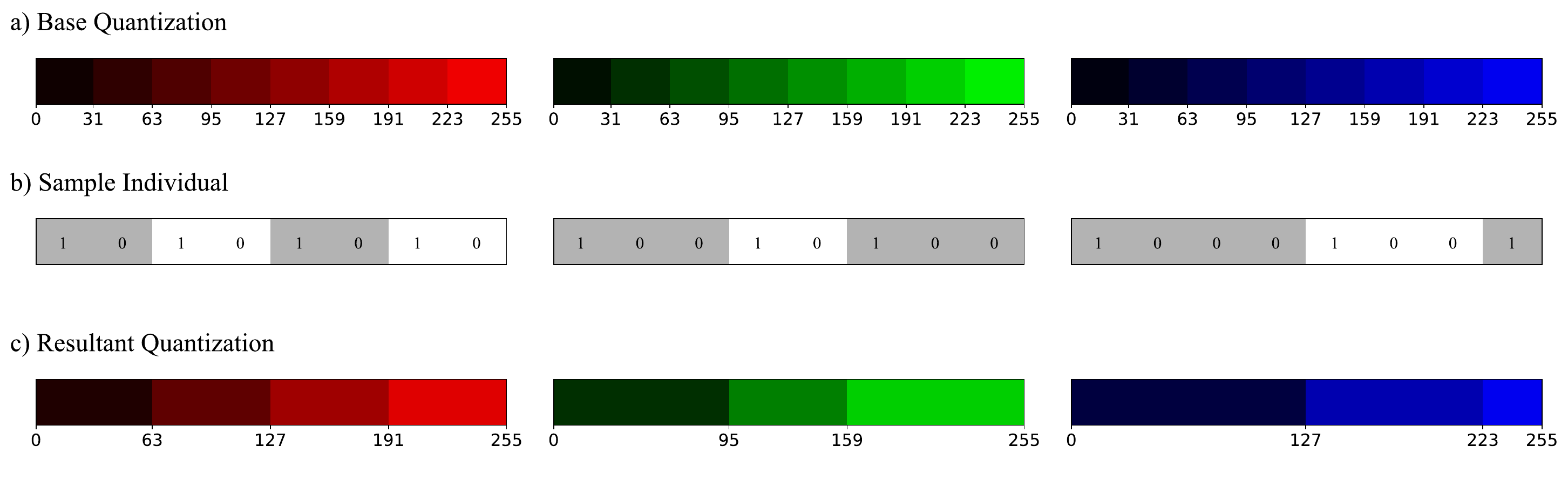}
	\caption{\edit{
	Our modeling takes reference from a base quantization (a) representing each interval of color tonalities as a bit in individuals implemented as binary arrays (b). 
	These bits dictate the union of intervals producing a new quantization (c): 
	if a bit is set, its respective interval has its own position, otherwise, it is aggregated to the immediate previous interval. The first bit of each color axis is forced to always be set.}}
	\label{fig:individual}
\end{figure}

In our formulation, a 24-bit long GA individual encodes the number of partitions of the different channels. Figure~\ref{fig:individual} (top) presents the typical 24-bit RGB channel partitioning. Figure~\ref{fig:individual} (middle) illustrates a possible  GA individual encoding how each channel should be divided. Figure~\ref{fig:individual} (bottom), in turn, illustrates the resulting color quantization after using the GA individual encoding.

\begin{center}
	\begin{figure}[h!]
		\subfloat[]{\includegraphics[width=0.5\textwidth]{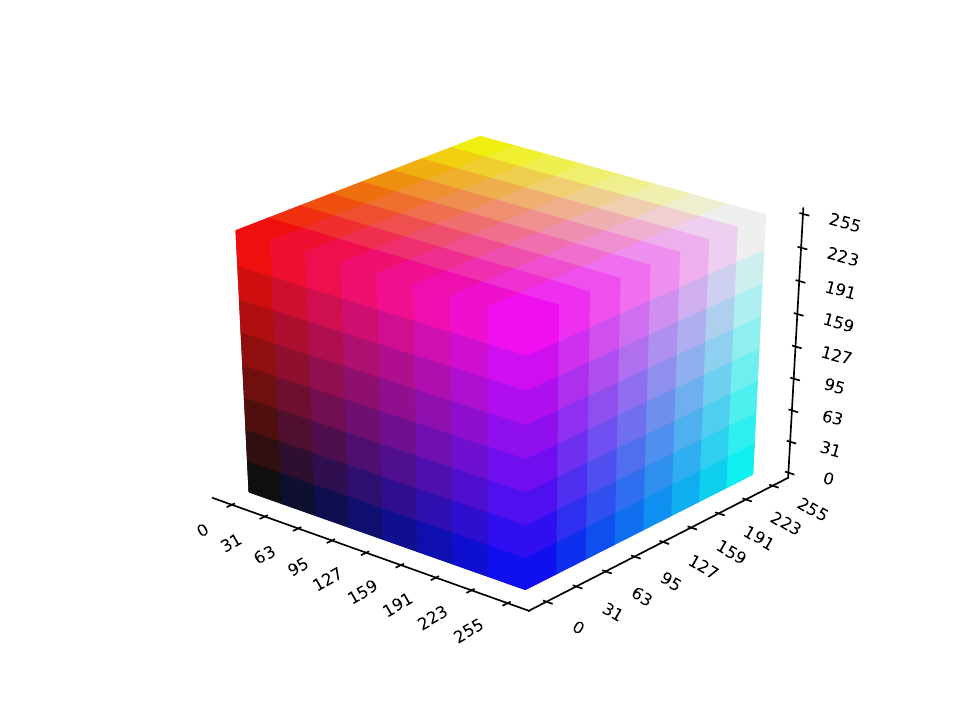}}
		\subfloat[]{\includegraphics[width=0.5\textwidth]{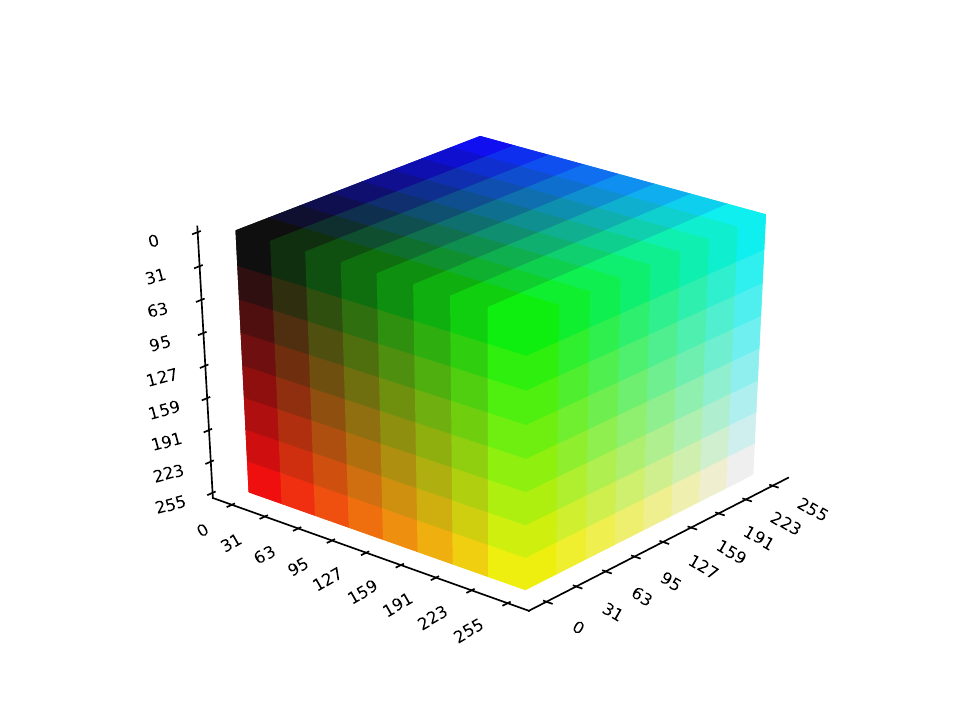}}\\
        \subfloat[]{\includegraphics[width=0.5\textwidth]{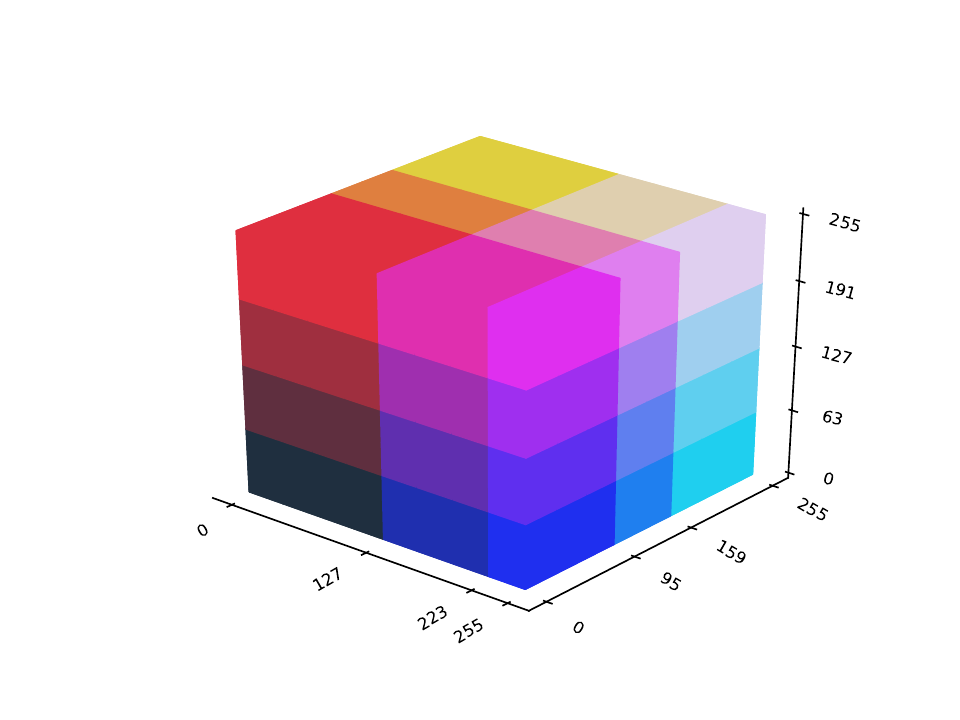}}
		\subfloat[]{\includegraphics[width=0.5\textwidth]{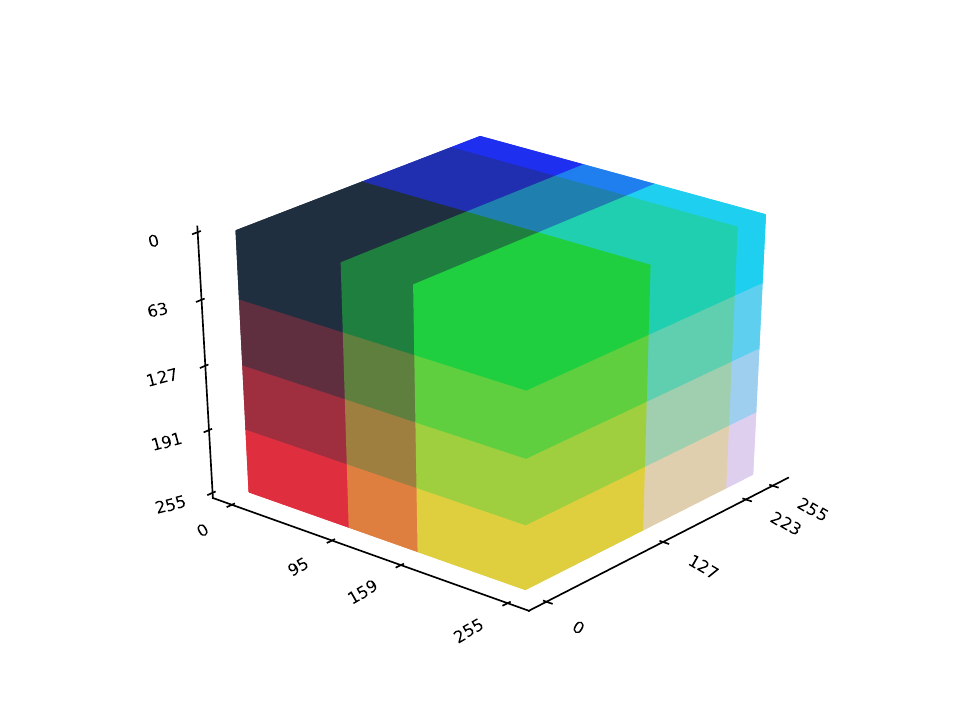}}
		\caption{\rtorres{ (a) RGB color space using the traditional 8-bit quantization per channel. (b) The same color space presented in (a), but rotated in 180º over the $Z$ axis. (c) Color space after applying the GA individual illustrated in Figure~\ref{fig:individual}  (middle). (d) The same color space presented in (c), but rotated in 180\degree over the $Z$ axis.}}
		\label{fig:widest_cube}
	\end{figure}
\end{center}

Figure~\ref{fig:widest_cube} presents the RGB color space before (a-b) and after (c-d) using the GA-based encoding defined in Figure~\ref{fig:individual}(middle). Figures~\ref{fig:widest_cube}(b) and~\ref{fig:widest_cube}(d) present different views of the same color space presented in Figures~\ref{fig:widest_cube}(a) and~\ref{fig:widest_cube}(c), respectively.

\FloatBarrier

\subsubsection{GA-based Quantization Search}
\label{subsec:search}

Algorithm~\ref{alg:quantization} illustrates the proposed GA-based quantization. 
The population starts with individuals created randomly (line 3).
The population evolves generation by generation through genetic operations (line 4). A function (described in Section~{\ref{subsec:fitness}}) is used to assign the fitness value for each individual (lines 5-7), i.e., to assess how well an individual solves the target problem.
\edit{According to the elitism operation, the top $k$ best individuals of the current generation are recorded (line 8).
Then, individuals from $P$ are selected according to a tournament operation of $n_t$-sized groupings (line 9).   
After that, the next generation is formed from the union of the resulting individuals from the operations of mutation and cross-over over the tournament selection and those selected in elitism (line 10).
If the stopping condition (discussed on Section~{\ref{subsec:parameters}}) 
were met, the iterations stop (lines 11-13).}
The last step is concerned with the selection of the best individual $q^\ast$ of all generations (line 15). 
The individual $q^\ast$ is used later to define the quantization used in the feature representation process.
\edit{For details regarding genetic operators (cross-over, mutation, tournament and elitism), refer to Section~{\ref{subsec:parameters}}.}

\begin{algorithm}[h!]
	\caption{GA-based quantization}
	\begin{lstlisting}
	'Let $T$ be a training set'
	'Let $P$, $S_e$ e $S_t$ be sets of pairs $(q,fitness_q)$, where $q$ and $fitness_q$ are an individual and its fitness, respectively'
	'$P \leftarrow$ Initial random population of individuals'
	'\textbf{For each} generation $g$ of $N_g$ generations \textbf{do}'
		'\textbf{For each} individual $q \in P$ \textbf{do}'
			'$fitness_q \leftarrow fitness(q,T)$'
		'\textbf{End For}'
		'$S_e \leftarrow elitism(k,P)$'
		'$S_t \leftarrow tournament(n_t, P)$'
		'$P \leftarrow S_e \cup mutation(S_t) \cup cross\-over(S_t)$'
		'\textbf{If} stopping condition is met'
			'Break outer loop'
		'\textbf{End If}'
	'\textbf{End For}'
	'Select the best individual $q^\ast = \arg\max\limits_{q \in P}(fitness_q)$'
	\end{lstlisting}
	\label{alg:quantization}
\end{algorithm}

\subsection{Feature Extraction}

\rtorres{In the second phase, the best individual, i.e., the one which leads to the best quantization $q^\ast$} is used with the feature extractor algorithm to produce a color image representation. 
In order to do that, it was necessary to implement a slightly modified version of the feature extractor, that incorporates the capacity of generating representations according to a specified color quantization. Equations 1, 2, and 3, where $M_c$ is the maximum color axis size and $q^\ast$ is the quantization individual, define how to calculate the new $R$, $G$, and $B$ (referred to as $R_{new}$, $G_{new}$, and $B_{new}$, respectively) values for each pixel.
\edit{In this work, according to empirical observations, $M_c$ was chosen as 8.}

\begin{eqnarray}
R_{new} = \left ( \sum_{i=0}^{r}  q^\ast[i] \right )\times\frac{|R_{axis}|}{256}, & & \\
\nonumber
& & where \quad r = R\times\frac{M_c}{256} ; \quad |R_{axis}| = \sum_{l=0}^{M_c} q^\ast[l] 
\end{eqnarray}

\begin{eqnarray}
G_{new} = \left ( \sum_{j=N}^{g+M_c}  q^\ast[i] \right )\times\frac{|G_{axis}|}{256}, & & \\
\nonumber
& & where \quad g = G\times\frac{M_c}{256} ; \quad |G_{axis}| = \sum_{m=M_c}^{2M_c} q^\ast[m] 
\end{eqnarray}

\begin{eqnarray}
B_{new} = \left ( \sum_{k=2M_c}^{b+2M_c}  q^\ast[i] \right )\times\frac{|B_{axis}|}{256}, & & \\
\nonumber
& & where \quad b = B\times\frac{M_c}{256} ; \quad |B_{axis}| = \sum_{n=2M_c}^{3M_c} q^\ast[n] 
\end{eqnarray}

%
%

\begin{figure}[!t]
	\centering
	\includegraphics[width=\textwidth]{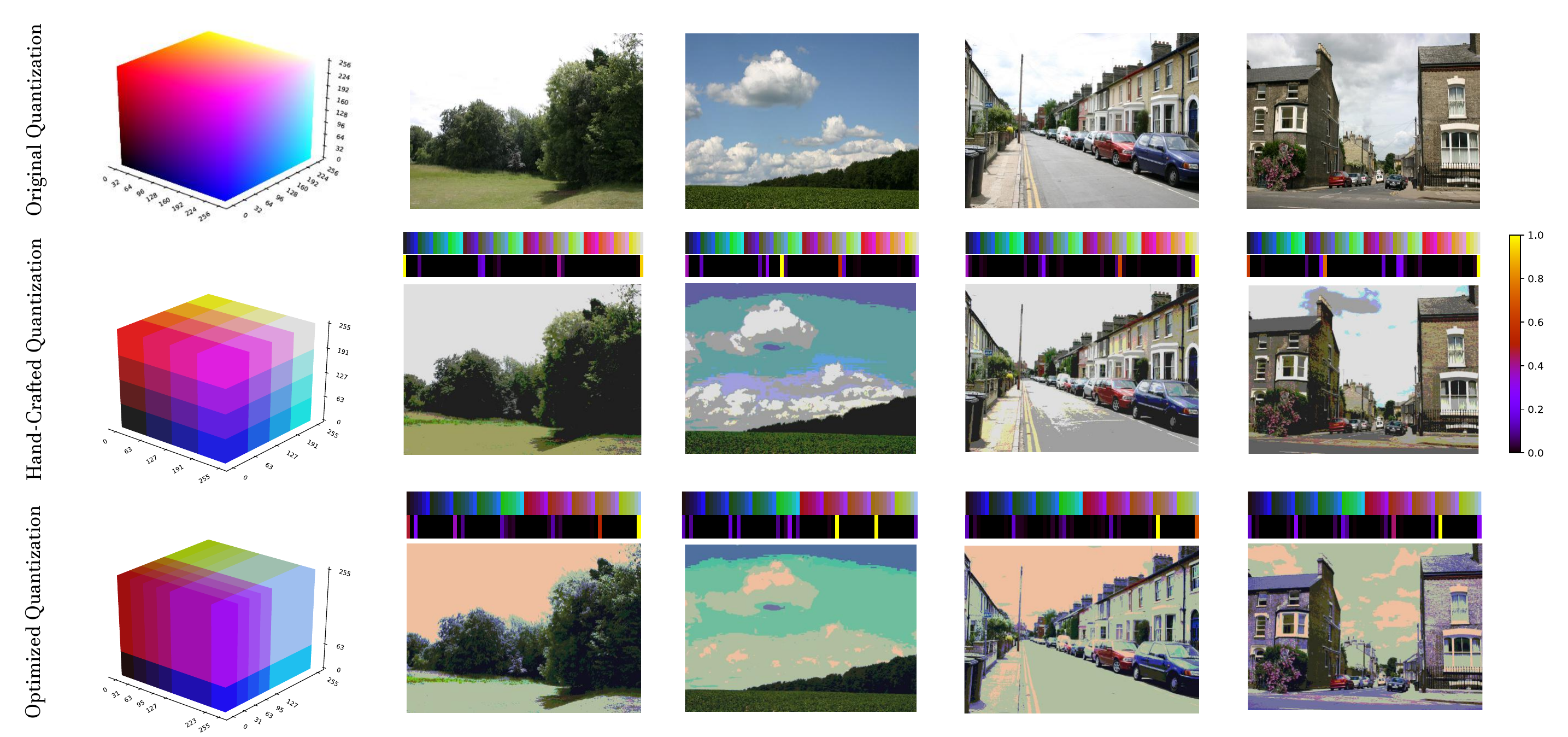}
	\caption{This figure$^1$ shows the visual effect of different color quantizations on different sample images. The \edit{first} column shows the RGB color spaces defined according to the specified quantizations: the original space in which the image is captured, a widely-used hand-crafted quantization scheme using 64 colors, and an example of optimized quantization defined by our method. The remaining columns show sample images after using each quantization scheme.}
	\label{fig:quant_effect}
\end{figure}

Figure~\ref{fig:quant_effect} shows the visual effect of different color quantizations on different sample images. The \edit{first} column shows the RGB color spaces defined according to the specified quantizations: the original space in which the image is captured, a widely-used hand-crafted quantization scheme using 64 colors, and an example of optimized quantization defined by our method. The remaining columns show sample images after using each quantization scheme. Original images are shown in the top line. Above each quantized image, we present the color spectrum and its respective histogram.

\footnotetext[1]{We recommend colourful printing for adequate visualization.}

\subsection{\rtorres{Individual Fitness Computation}}
\label{subsec:fitness}

The use of the proposed GA-based quantization leads to discriminative features, which may be useful in different applications, such as Image Classification~\cite{DosSantos:2010},  Image Retrieval~\cite{Penatti:2012}, and Object Recognition. In this paper, we opted for evaluating the method in the context of Content-Based Image Retrieval (CBIR)~\cite{Smeulders:2000} tasks. The goal of this task is to retrieve the most relevant images from a collection, given their similarity to a given query image. The similarity computation relies on the use of a distance (or similarity) function applied to feature vectors, which encode their content (in our case, their color properties).

We first extract feature vectors from all images within a collection, by taking into account feature extractors that benefit from the learned color quantization. Collection images are later ranked according to the distance of their feature vectors to the feature vector of a query using the Manhattan Distance (L1).
Two images belonging to the same class are assumed to be relevant to each other. Given a query image, our goal is to produce a ranked list with collection images of the same class of the query on top positions. The more relevant images on top positions, the more effective is the ranked list, i.e., the more effective is the description approach.

More formally, an image $img$ is firstly encoded through a feature extraction procedure, which allows quantifying the similarity  between images.
Let $C=\{img_1, img_2, \dots, img_n\}$ be a collection with $n$ images. Let $\mathcal{D}$ be a descriptor, which can be defined as a tuple $(\epsilon,\delta)$~\cite{Torres2006RITA}, where:

\begin{itemize}
\item $\epsilon$: $img_i$ $\rightarrow$ $\mathbb{R}^{d}$ is a function, which extracts a feature vector $v_{\hat{i}}$ from an image $img_i$;
\item $\delta$: $\mathbb{R}^{d} \times \mathbb{R}^{d} \rightarrow \mathbb{R}^+$ is a function that computes the distance between two images according to the distance between their corresponding feature vectors.
\end{itemize}
 
The distance between two images $img_{i}$, $img_{j}$ is computed as $\delta$($\epsilon(o_{i})$, $\epsilon(o_{j})$). 
 The Euclidean distance is commonly used to compute $\delta$, although the proposed ranking method is independent of distance measures.
 A similarity measure $\rho(img_i,img_j)$ can be computed based on distance function $\delta$ and used for ranking tasks.
We will use $\rho(i,j)$ from now on to simplify the notation.
 
The target task refers to retrieving multimedia objects (e.g., images, videos) from $C$ based on their content. Let $img_{q}$ be a query image. A ranked list $\tau_{q}$ can be computed in response to $img_{q}$ based on the similarity function $\rho$.
The ranked list $\tau_{q}$=$(img_{1}$, $img_{2}$, $\dots$, $img_{n})$ can be defined as a permutation of the collection $C$.
A permutation $\tau_q$ is a bijection from the set $C$ onto the set 
$[N]=\{1,2,\dots,n\}$. For a permutation $\tau_q$, we interpret $\tau_q (i)$ as the position (or rank) of the image
$img_i$ in the ranked list $\tau_q$. 
If $img_{i}$ is ranked before $img_{j}$ in the ranked list of $img_q$, i.e., $\tau_q(i) 
< \tau_q(j)$, then $\rho(q, i)$ $\geq$  $\rho(q, j)$.

Given a training set composed of a set of queries and their respective list of relevant objects, the fitness of an individual is measured as a function of the quality (effectiveness) of ranked lists produced for each query, considering the use of a feature extractor implemented using the GA-based quantization. The more relevant images found at top positions, the better the GA individual is.

\subsection{Computational Complexity of GA-based Quantization Search}
\editt{The GA training procedure takes $\mathcal{O}(N_g \times N_i \times F)$, where $N_g$ is the number of generations considered in the
evolution process, $N_i$ is the number of individuals in the population, and $F$ is the cost for evaluating the fitness function.}

\editt{The costs for computing $F$ depends on the number of training samples $N_s$ and the size of pre-computed histograms $S_h$. The later, in the worst case, is $k^3$, where $k$ is the number of bins in a color axis. As overlying detailed base color spaces does not improve the results, $k$ is typically small, making $S_h$ also small ($k=8$ and $S_h=k^3=512$ in our experiments).
As a consequence, $F$ takes $\mathcal{O}(N_s \times S_h)$ for feature extraction and $\mathcal{O}({N_s}^2 \times log N_s)$ for computing rankings, then $\mathcal{O}(F) = \mathcal{O}({N_s}^2 \times log N_s)$.}

\editt{Finally, the whole procedure takes $\mathcal{O}(N_g \times N_i \times {N_s}^2 \times log N_s)$ to find the final quantization.
Recall that the training process is performed offline.}

\subsection{Quantization Approaches}

In this paper, we propose two formulations of the GA-based quantization method. The first, named Unconstrained Approach (UA), is intended to provide a quantization focused on generating representations that have the best possible effectiveness performance. The second, named Size-Constrained Approach (SCA), focuses not only on effectiveness aspects, but also on the size of the representation. The goal is to find the best-performing individual, which leads to feature vectors with a pre-defined size, i.e., the target feature vector size is defined a priori. From the implementation point of view, the GA-based quantization approach assigns a negative fitness score for the individuals that present dimensions higher than the pre-defined feature vector size. As a consequence, this latter formulation tends to produce more compact representations.

\section{Experimental Setup}
\label{sec:experimental_setup}

In this section, we present the adopted experimental setup, which concerns the image datasets considered (Section~\ref{subsec:datasets}), the configuration of parameters of the method (Section~\ref{subsec:parameters}), the baselines used for comparative analysis (Section~\ref{subsec:baselines}), the metrics used to evaluate the effectiveness and compactness of the produced feature vectors (Section~\ref{subsec:metrics}), and the employed experimental protocol (Section~\ref{subsec:protocol}).

\subsection{Datasets}

\label{subsec:datasets}
\begin{table}[!t]
	\centering
	\caption{Image datasets and statistics}
	\label{tab:datasets}
	\begin{tabular}{lccc}
		\hline\noalign{\smallskip}
		Dataset & \# of samples & \# of classes & Images content\\
		\noalign{\smallskip}\hline\noalign{\smallskip}
		\textit{Coil-100}~{\cite{Nayar:1996}} & 7,200 & 100 & objects \\
		\textit{Corel-1566}~{\cite{Wang:2001}} & 1,566 & 43 & mixed (objects, landscapes etc)\\
		\textit{Corel-3906}~{\cite{Wang:2001}} & 3,906 & 85 & mixed (objects, landscapes etc)\\
		\textit{ETH-80}~{\cite{Leibe:2003}} & 3,280 & 80 & objects\\
		\textit{MSRCORID}~{\cite{Criminisi:2004}} & 4,320 & 20 & mixed (scenes and objects)\\
		\textit{Groundtruth}~{\cite{Li:2002, Li:2005}} & 1,285 & 21 & landscapes\\
		\textit{Supermarket Produce}~{\cite{Rocha:2010}} & 2,633 & 15 & fruits\\
		\textit{UC Merced Land-use}~{\cite{Yang:2010}} & 2,100 & 21 & aerial scenes\\
		\noalign{\smallskip}\hline
	\end{tabular}
\end{table}

In order to assess the effectiveness of the employed quantization approach, we conducted experiments using eight different image datasets, which are described next. \edit{For convenience, Table~{\ref{tab:datasets}} summarizes some important information about them.}

\begin{itemize} 
	\item \textbf{\textit{Coil-100}}:
	This dataset~{\cite{Nayar:1996}} comprises images of 100 everyday objects, being each one used to define  a different class.
	Pictures of each object were taken in 72 different poses composing a total set of 7,200 images.
	Some samples of this dataset are shown in Figure \ref{fig:coil_100}.
	
	\item \textbf{\textit{Corel-1566 and Corel-3906}}:
	These datasets~{\cite{Wang:2001}} correspond to two sets from a collection with 200,000 images from the Corel Gallery Magic–Stock Photo Library 2.
	The first (Fig.~\ref{fig:corel_1566}) contains 1,566 samples distributed among 43 classes, while the second (Fig.~\ref{fig:corel_3906}) contains 3,906 samples among 85 classes.
	Besides the images quantity, the main difference between them is that the latter presents more intra-class variability.
	
	\item \textbf{\textit{ETH-80}}:
	This dataset~{\cite{Leibe:2003}}  was originally tailored to the task of object categorization.
	It includes images of 80 objects from 8 basic-level categories.
	Each object is represented by 41 views over the upper viewing hemisphere, performing a total of 2,384 images.
	Some samples of this dataset are shown in Figure~\ref{fig:eth80}.
	
	\item \textbf{\textit{Groundtruth}}:
	This dataset~{\cite{Li:2002, Li:2005}} contains a variety of 1,285 scenes and objects grouped among 21 high-level concepts, such as: 
	Arbor Greens, Australia, Barcelona, Cambridge, Campus In Fall, Cannon Beach, Cherries, Columbia George, Football, Geneva, Green Lake, Greenland, Indonesia, Iran, Italy, Japan, Leafless Trees, San Juans, Spring Flowers, Swiss Mountains, Yellow Stone. 
	Figure \ref{fig:groundtruth} depicts some of its classes.
	
	\item \textbf{\textit{Microsoft Research Cambridge Object Recognition Image Database (MSRCORID)}}:
	This collection~\cite{Criminisi:2004} contains a set of 4,320 images of scenes, objects and landscapes. Its images are grouped into 20 categories: Aeroplanes, Cows, Sheep, Benches and Chairs, Bicycles, Birds, Buildings, Cars, Chimneys, Clouds, Doors, Flowers, Kitchen Utensils, Leaves, Scenes Countryside, Scenes Office, Scenes Urban, Signs, Trees, Windows. Some samples of this dataset are shown in Figure~\ref{fig:msrcorid}.
	
	\item \textbf{\textit{Supermarket Produce}}:
	This dataset~{\cite{Rocha:2010}} contains images of fruits and vegetables collected from a local distribution center. 
	It comprises 2,633 images distributed into 15 different categories: Plum, Agata Potato, Asterix Potato, Cashew, Onion, Orange, Tahiti Lime, Kiwi, Fuji Apple, Granny-Smith Apple, Watermelon, Honeydew Melon, Nectarine, Williams Pear, and Diamond Peach.
	Figure \ref{fig:fruits} depicts some samples of its categories.
	
	\item \textbf{\textit{UC Merced Land-use}}:
	This dataset~{\cite{Yang:2010}} is composed of 2,100 aerial scene images divided into 21 classes selected from the United States Geological Survey (USGS) National Map. Its 21 categories are Agricultural, Airplane, Baseball Diamond, Beach, Buildings, Chaparral, Dense Residential, Forest, Freeway, Golf Course, Harbor, Intersection, Medium Density Residential, Mobile Home Park, Overpass, Parking Lot, River, Runway, Sparse Residential, Storage Tanks, and Tennis Courts. 
	Some samples of this dataset are shown in Figure~\ref{fig:ucmerced}.
\end{itemize}

{
\begin{figure}[!h]
	\subfloat[Forest]{
		\includegraphics[width=0.16\textwidth]{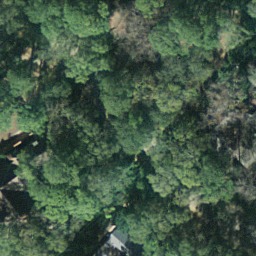}
		\hspace{-0.4em}
		\includegraphics[width=0.16\textwidth]{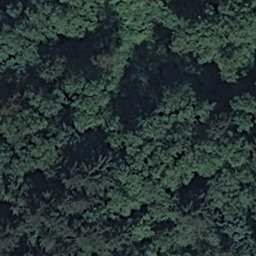}
	}
	\subfloat[Beach]{
		\includegraphics[width=0.16\textwidth]{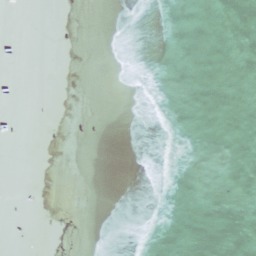}
		\hspace{-0.4em}
		\includegraphics[width=0.16\textwidth]{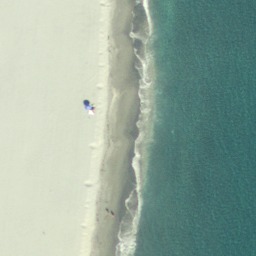}
	}
	\subfloat[Tennis Court]{
		\includegraphics[width=0.16\textwidth]{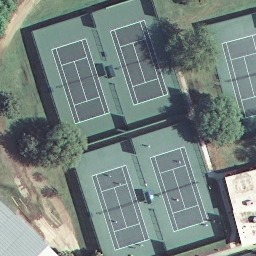}
		\hspace{-0.4em}
		\includegraphics[width=0.16\textwidth]{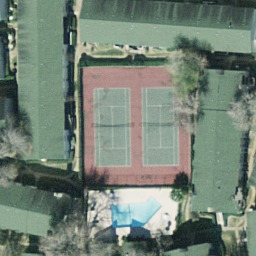}
	}
	\vspace{-1em}\\
	\subfloat[Dense Residential]{
		\includegraphics[width=0.16\textwidth]{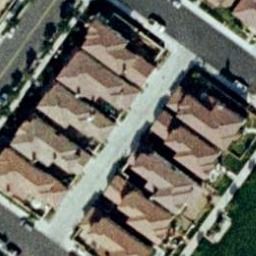}
		\hspace{-0.4em}
		\includegraphics[width=0.16\textwidth]{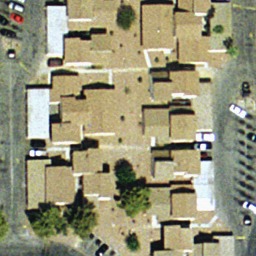}
	}
	\subfloat[Medium Residential]{
		\includegraphics[width=0.16\textwidth]{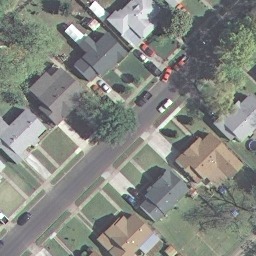}
		\hspace{-0.4em}
		\includegraphics[width=0.16\textwidth]{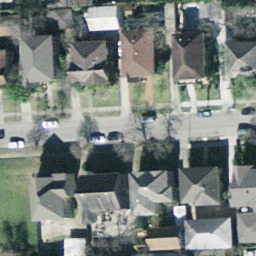}
	}
	\subfloat[Sparse Residential]{
		\includegraphics[width=0.16\textwidth]{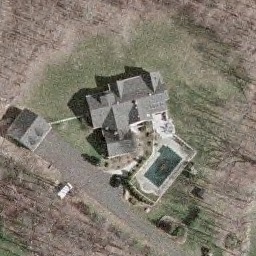}
		\hspace{-0.4em}
		\includegraphics[width=0.16\textwidth]{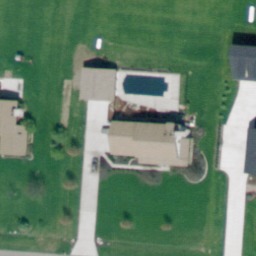}
	}\\
	\vspace{-2em}\\
	\caption{Examples of the \textit{UC Merced Land-use} dataset.}
	\label{fig:ucmerced}
\end{figure}

\begin{figure}[!h]
	\subfloat[Object 13]{
		\includegraphics[width=0.16\textwidth]{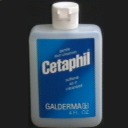}
		\hspace{-0.4em}
		\includegraphics[width=0.16\textwidth]{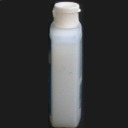}
		\hspace{-0.4em}
		\includegraphics[width=0.16\textwidth]{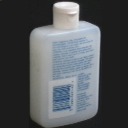}
	}
	\subfloat[Object 14]{
		\includegraphics[width=0.16\textwidth]{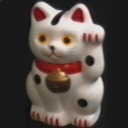}
		\hspace{-0.4em}
		\includegraphics[width=0.16\textwidth]{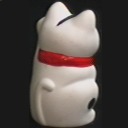}
		\hspace{-0.4em}
		\includegraphics[width=0.16\textwidth]{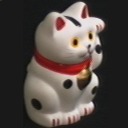}
	}\\
	\vspace{-1em}\\
	\subfloat[Object 87]{
		\includegraphics[width=0.16\textwidth]{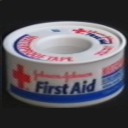}
		\hspace{-0.4em}
		\includegraphics[width=0.16\textwidth]{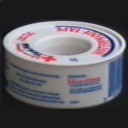}
		\hspace{-0.4em}
		\includegraphics[width=0.16\textwidth]{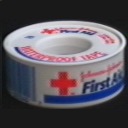}
	}
	\subfloat[Object 81]{
		\includegraphics[width=0.16\textwidth]{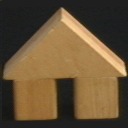}
		\hspace{-0.4em}
		\includegraphics[width=0.16\textwidth]{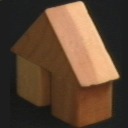}
		\hspace{-0.4em}
		\includegraphics[width=0.16\textwidth]{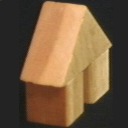}
	}\\
	\vspace{-2em}\\
	\caption{Examples of the \textit{COIL-100} dataset.}
	\label{fig:coil_100}
\end{figure}

\begin{figure}[!t]
	\subfloat[A6140]{
		\includegraphics[width=0.16\textwidth]{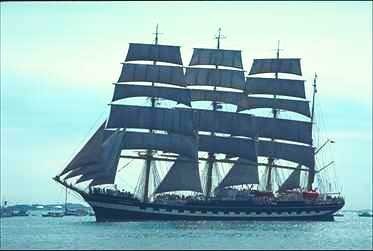}
		\hspace{-0.4em}
		\includegraphics[width=0.16\textwidth]{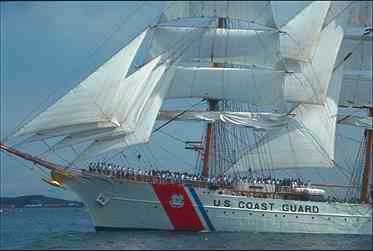}
	}
	\subfloat[A14935]{
		\includegraphics[width=0.16\textwidth]{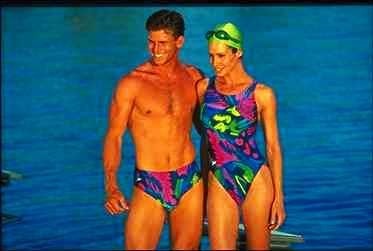}
		\hspace{-0.4em}
		\includegraphics[width=0.16\textwidth]{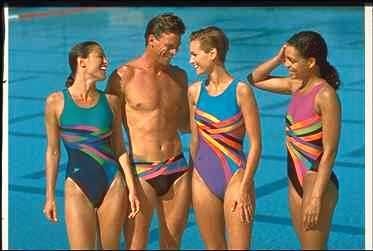}
	}
	\subfloat[A2231]{
		\includegraphics[width=0.16\textwidth]{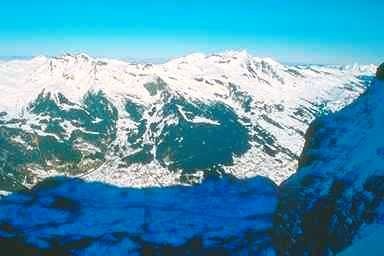}
		\hspace{-0.4em}
		\includegraphics[width=0.16\textwidth]{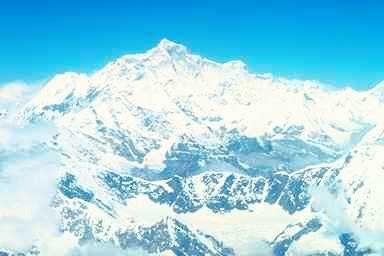}
	}\\
	\vspace{-1em}\\
	\subfloat[A0908]{
		\includegraphics[width=0.115\textwidth]{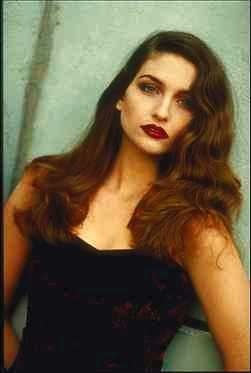}
		\hspace{-0.4em}
		\includegraphics[width=0.115\textwidth]{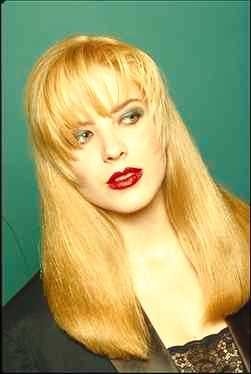}
	}
	\subfloat[A7840]{
		\includegraphics[width=0.115\textwidth]{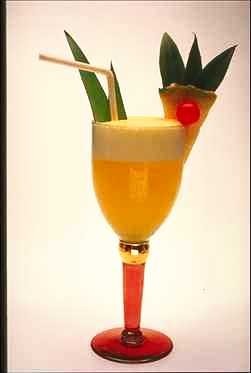}
		\hspace{-0.4em}
		\includegraphics[width=0.115\textwidth]{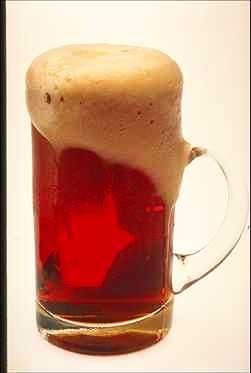}
	}
	\subfloat[A0004]{
		\includegraphics[width=0.115\textwidth]{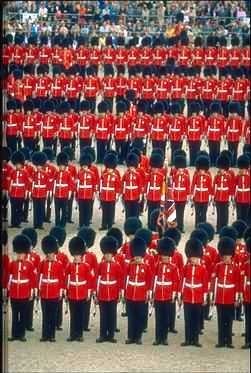}
		\hspace{-0.4em}
		\includegraphics[width=0.115\textwidth]{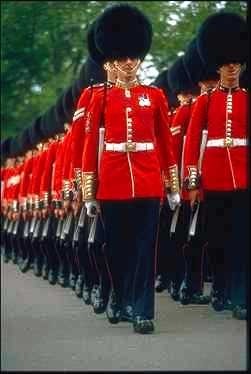}
	}
	\subfloat[A12147]{
		\includegraphics[width=0.115\textwidth]{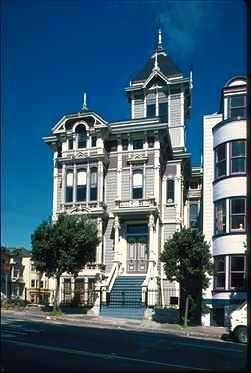}
		\hspace{-0.4em}
		\includegraphics[width=0.115\textwidth]{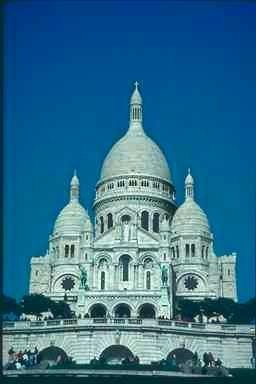}
	}\\
	\vspace{-2em}\\
	\caption{Examples of the \textit{COREL-1566} dataset.}
	\label{fig:corel_1566}
\end{figure}

\begin{figure}[!t]
	\subfloat[A0628]{
		\includegraphics[width=0.16\textwidth]{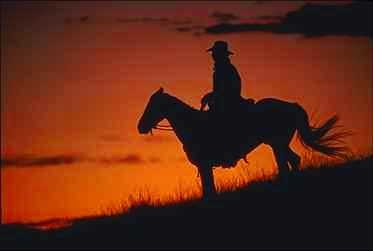}
		\hspace{-0.4em}
		\includegraphics[width=0.16\textwidth]{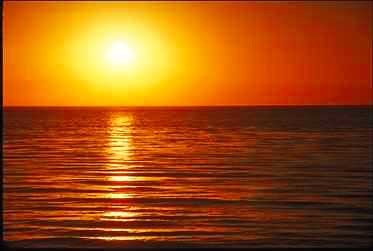}
	}
	\subfloat[A1401]{
		\includegraphics[width=0.16\textwidth]{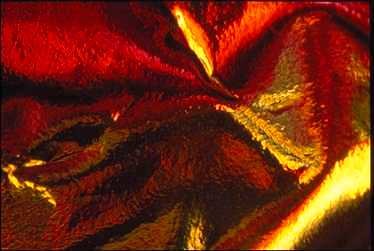}
		\hspace{-0.4em}
		\includegraphics[width=0.16\textwidth]{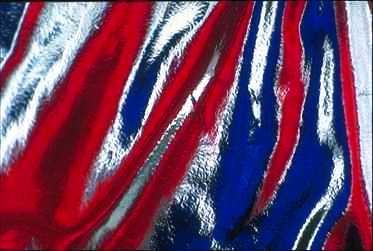}
	}
	\subfloat[A4604]{
		\includegraphics[width=0.16\textwidth]{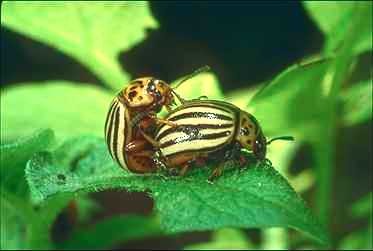}
		\hspace{-0.4em}
		\includegraphics[width=0.16\textwidth]{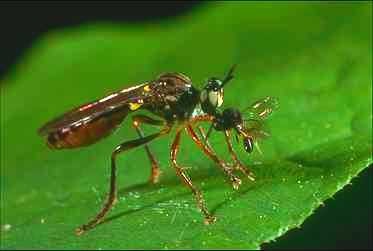}
	}
	\vspace{-1em}\\
	\subfloat[A4932]{
		\includegraphics[width=0.16\textwidth]{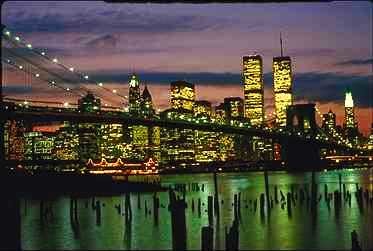}
		\hspace{-0.4em}
		\includegraphics[width=0.16\textwidth]{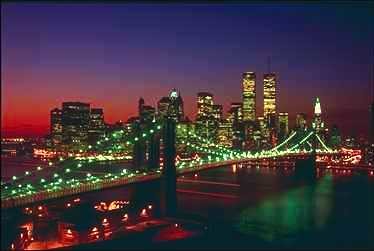}
	}
	\subfloat[A7601]{
		\includegraphics[width=0.16\textwidth]{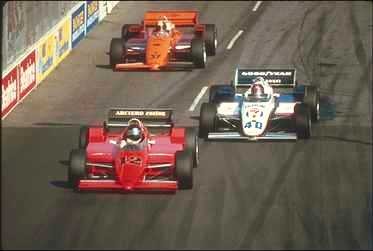}
		\hspace{-0.4em}
		\includegraphics[width=0.16\textwidth]{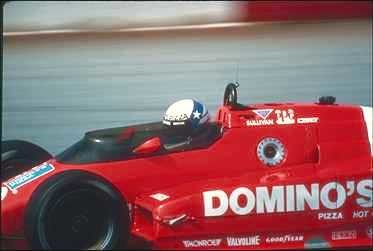}
	}
	\subfloat[A14208]{
		\includegraphics[width=0.16\textwidth]{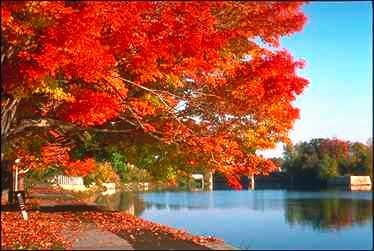}
		\hspace{-0.4em}
		\includegraphics[width=0.16\textwidth]{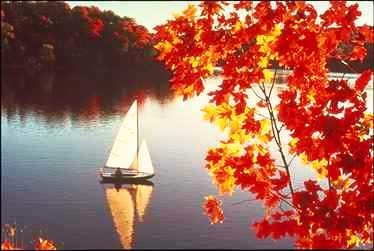}
	}\\
	\vspace{-2em}\\
	\caption{Examples of the \textit{COREL-3906} dataset.}
	\label{fig:corel_3906}
\end{figure}

\begin{figure}[!t]
	\includegraphics[width=0.16\textwidth]{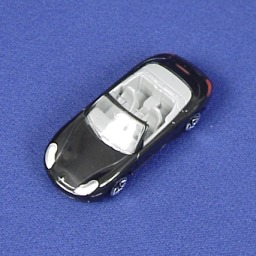}
	\hspace{-0.4em}
	\includegraphics[width=0.16\textwidth]{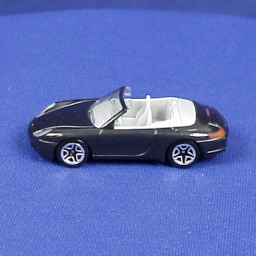}
	\hspace{0.1em}
	\includegraphics[width=0.16\textwidth]{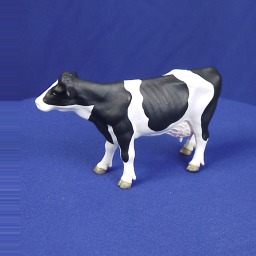}
	\hspace{-0.4em}
	\includegraphics[width=0.16\textwidth]{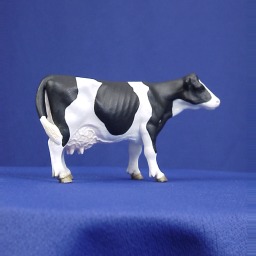}
	\hspace{0.1em}
	\includegraphics[width=0.16\textwidth]{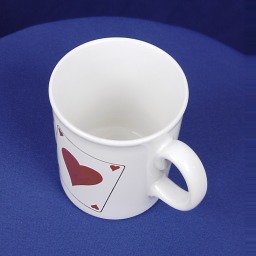}
	\hspace{-0.4em}
	\includegraphics[width=0.16\textwidth]{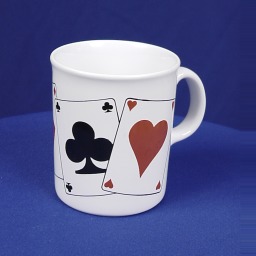}
	\vspace{-1em}\\
	\includegraphics[width=0.16\textwidth]{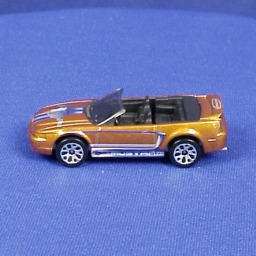}
	\hspace{-0.4em}
	\includegraphics[width=0.16\textwidth]{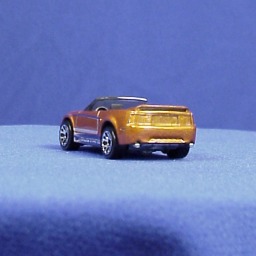}
	\hspace{0.1em}
	\includegraphics[width=0.16\textwidth]{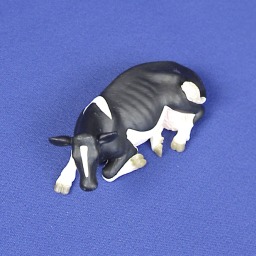}
	\hspace{-0.4em}
	\includegraphics[width=0.16\textwidth]{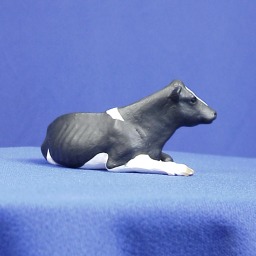}
	\hspace{0.1em}
	\includegraphics[width=0.16\textwidth]{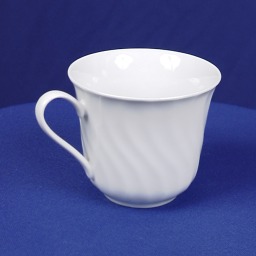}
	\hspace{-0.4em}
	\includegraphics[width=0.16\textwidth]{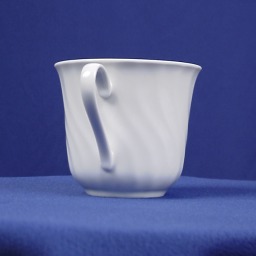}
	\vspace{-1em}\\
	\hspace{0.1em}
	\includegraphics[width=0.16\textwidth]{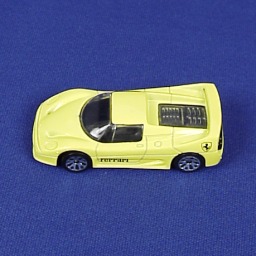}
	\hspace{-0.4em}
	\includegraphics[width=0.16\textwidth]{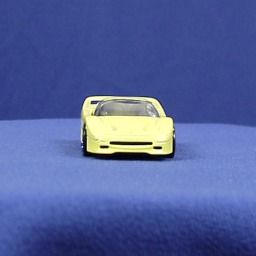}
	\hspace{0.1em}
	\includegraphics[width=0.16\textwidth]{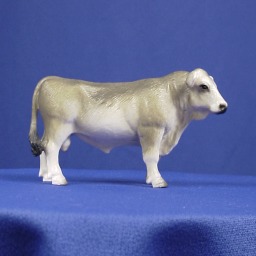}
	\hspace{-0.4em}
	\includegraphics[width=0.16\textwidth]{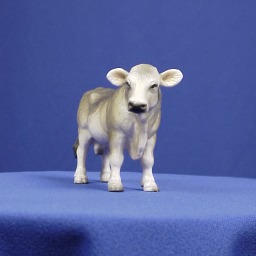}
	\hspace{0.1em}
	\includegraphics[width=0.16\textwidth]{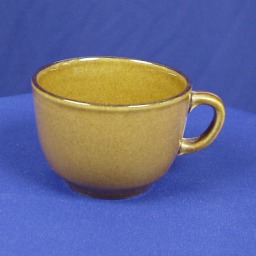}
	\hspace{-0.35em}
	\includegraphics[width=0.16\textwidth]{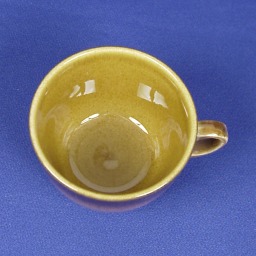}\\
	.\hspace{1.1cm}(a) Car \hspace{2.9cm}(b) Cow \hspace{2.9cm}(c) Cup\\
	\vspace{-2.5em}\\
	\caption{Examples of the \textit{ETH-80} dataset.}
	\label{fig:eth80}
\end{figure}

\begin{figure}[!t]
	\subfloat[Windowns]{
		\includegraphics[width=0.16\textwidth]{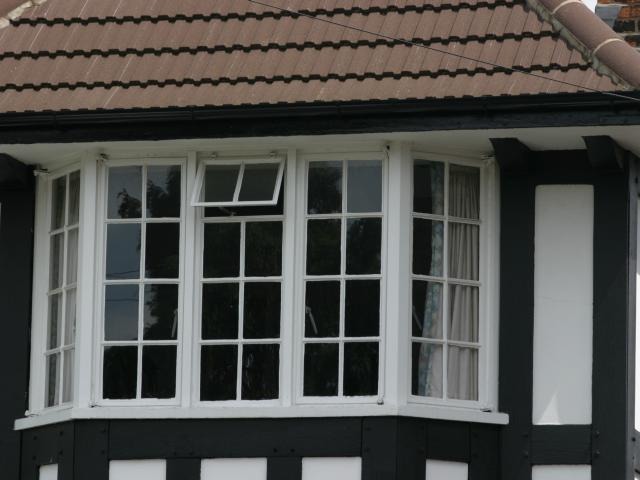}
		\hspace{-0.4em}
		\includegraphics[width=0.16\textwidth]{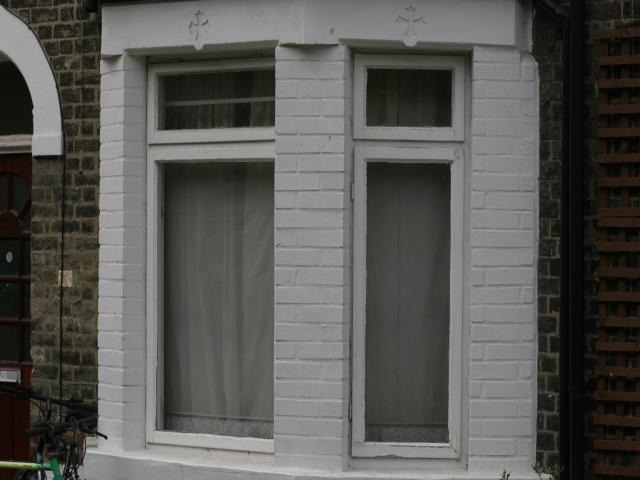}
	}
	\subfloat[Trees]{
		\includegraphics[width=0.16\textwidth]{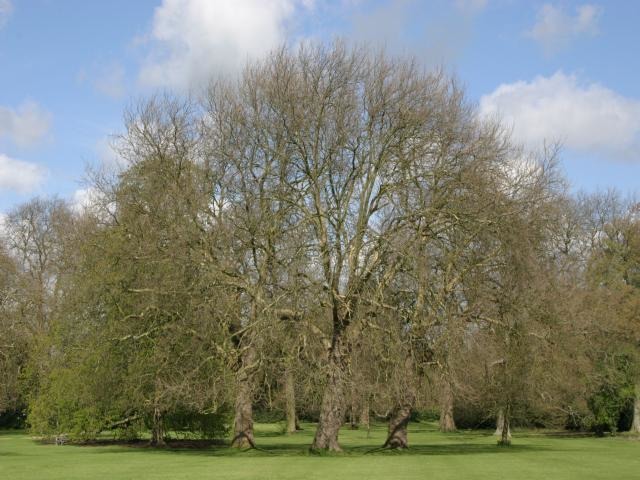}
		\hspace{-0.4em}
		\includegraphics[width=0.16\textwidth]{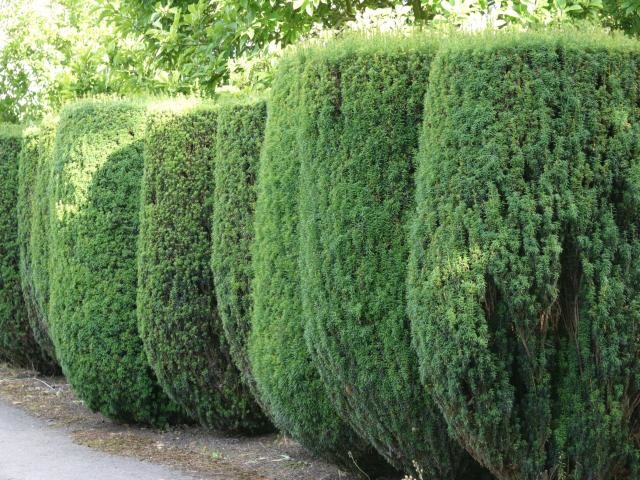}
	}
	\subfloat[Kitchen Utensils]{
		\includegraphics[width=0.16\textwidth]{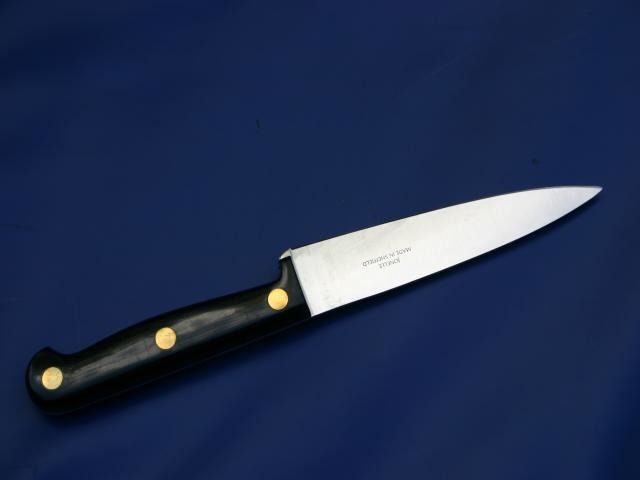}
		\hspace{-0.4em}
		\includegraphics[width=0.16\textwidth]{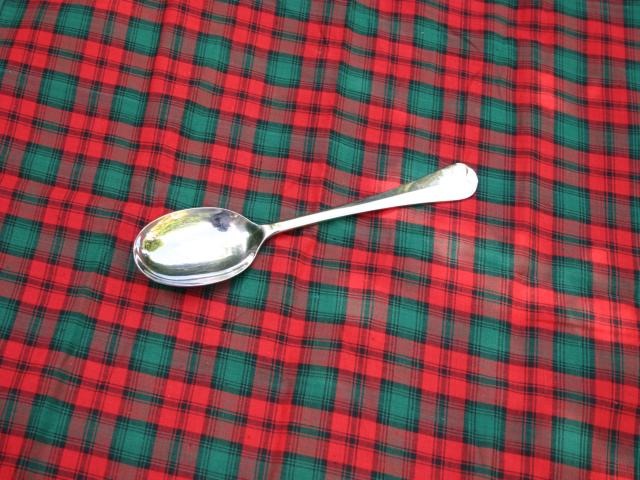}
	}
	\vspace{-1em}\\
	\subfloat[Scenes Office]{
		\includegraphics[width=0.16\textwidth]{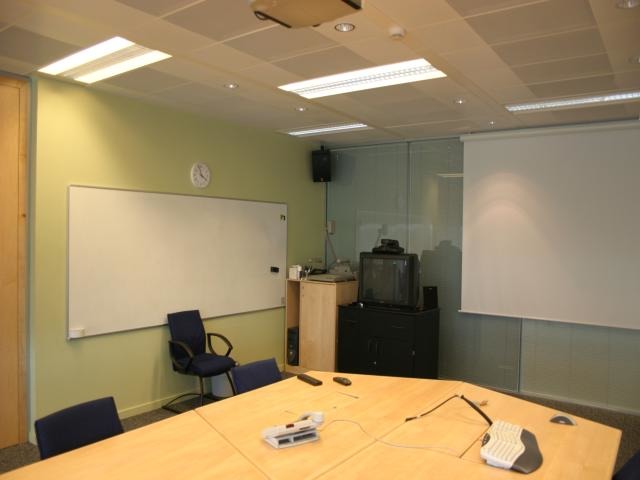}
		\hspace{-0.4em}
		\includegraphics[width=0.16\textwidth]{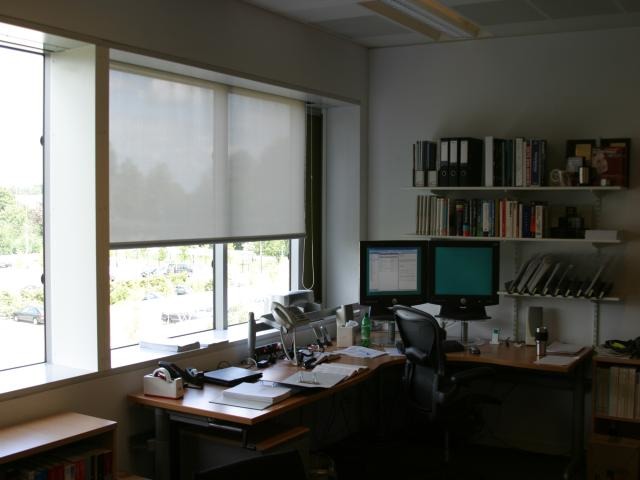}
	}
	\subfloat[Scenes Countryside]{
		\includegraphics[width=0.16\textwidth]{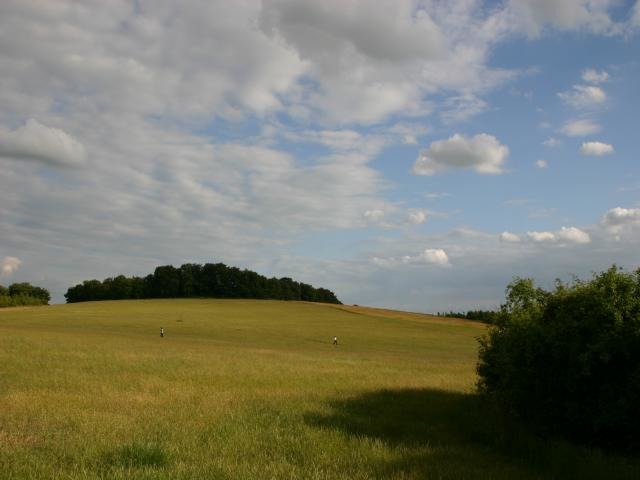}
		\hspace{-0.4em}
		\includegraphics[width=0.16\textwidth]{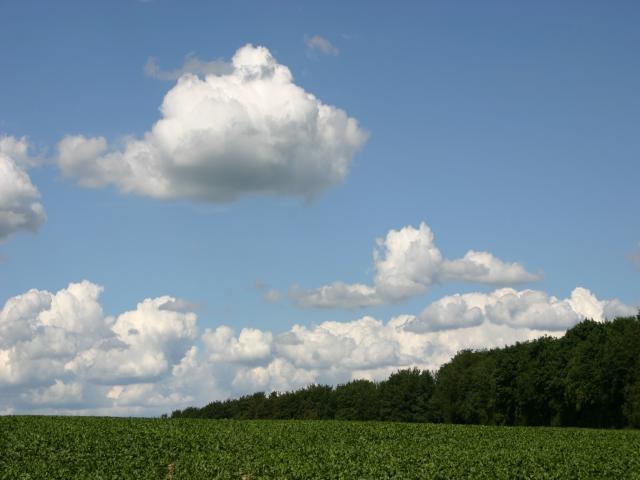}
	}
	\subfloat[Buildings]{
		\includegraphics[width=0.16\textwidth]{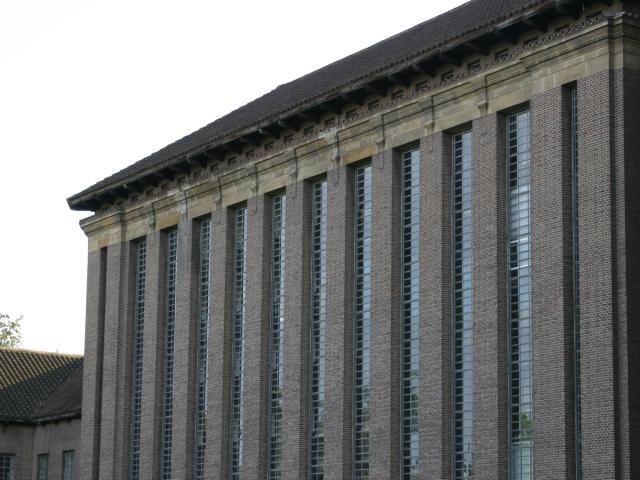}
		\hspace{-0.4em}
		\includegraphics[width=0.16\textwidth]{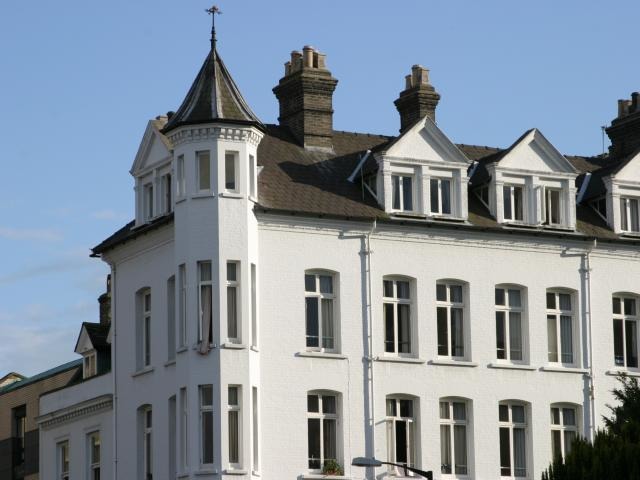}
	}\\
	\vspace{-2em}\\
	\caption{Examples of the \textit{MSRCORID} dataset.}
	\label{fig:msrcorid}
\end{figure}

\begin{figure}[!t]
	\subfloat[Agata Potato]{
		\includegraphics[width=0.16\textwidth]{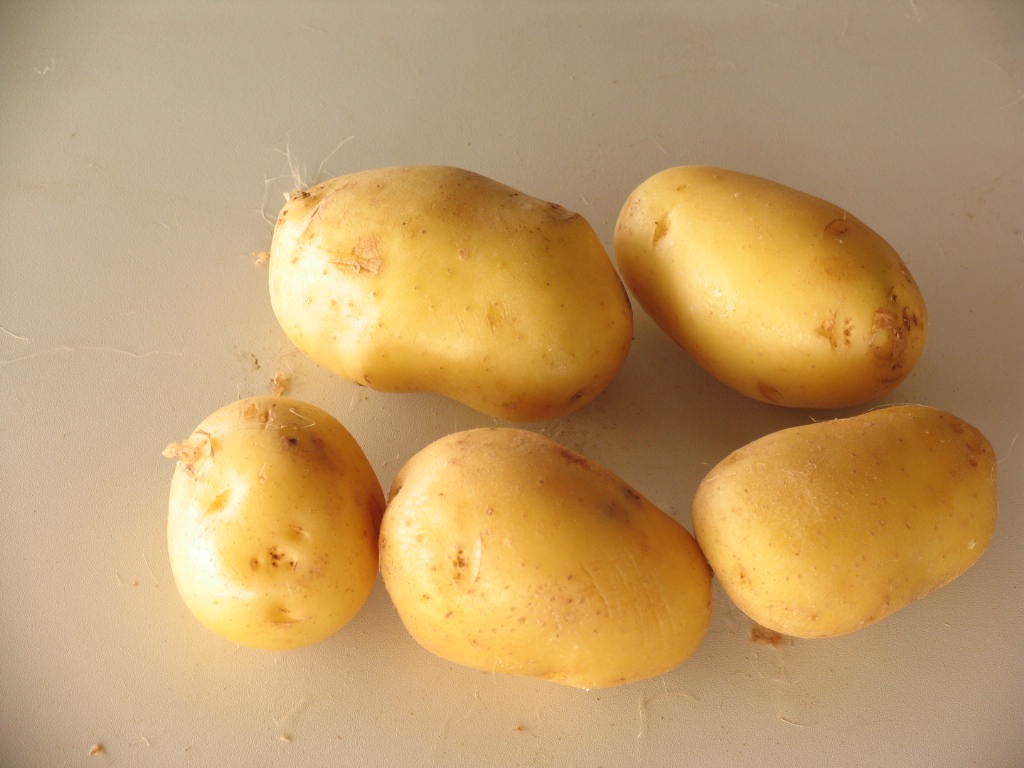}
		\hspace{-0.4em}
		\includegraphics[width=0.16\textwidth]{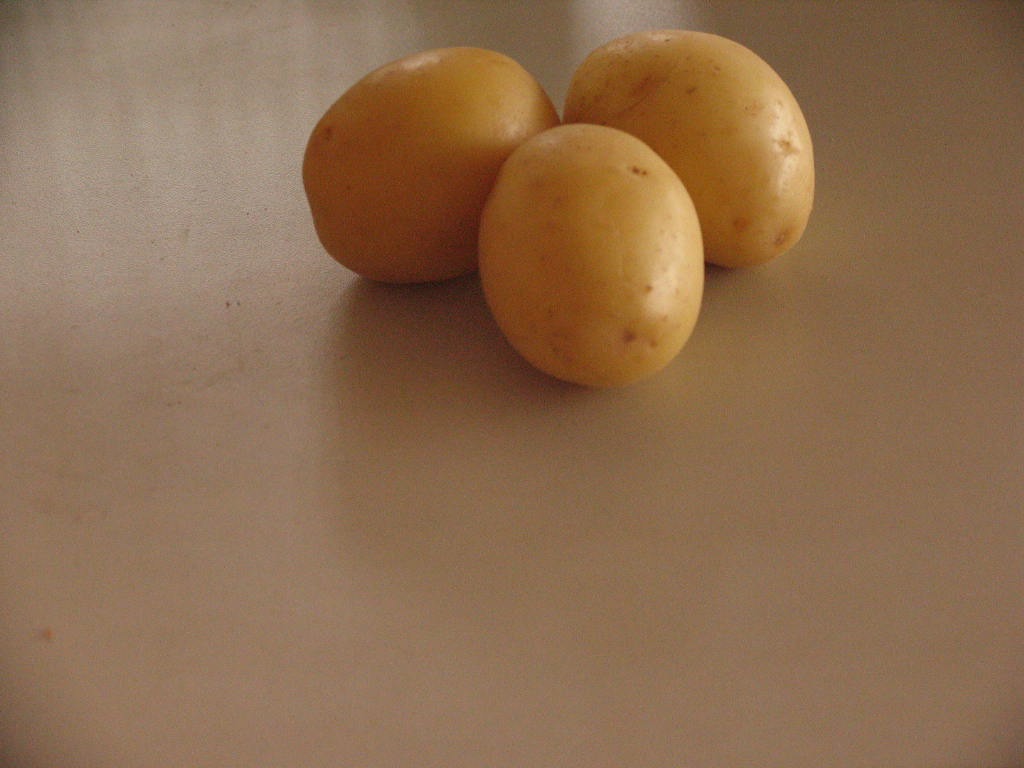}
	}
	\subfloat[Honneydew Melon]{
		\includegraphics[width=0.16\textwidth]{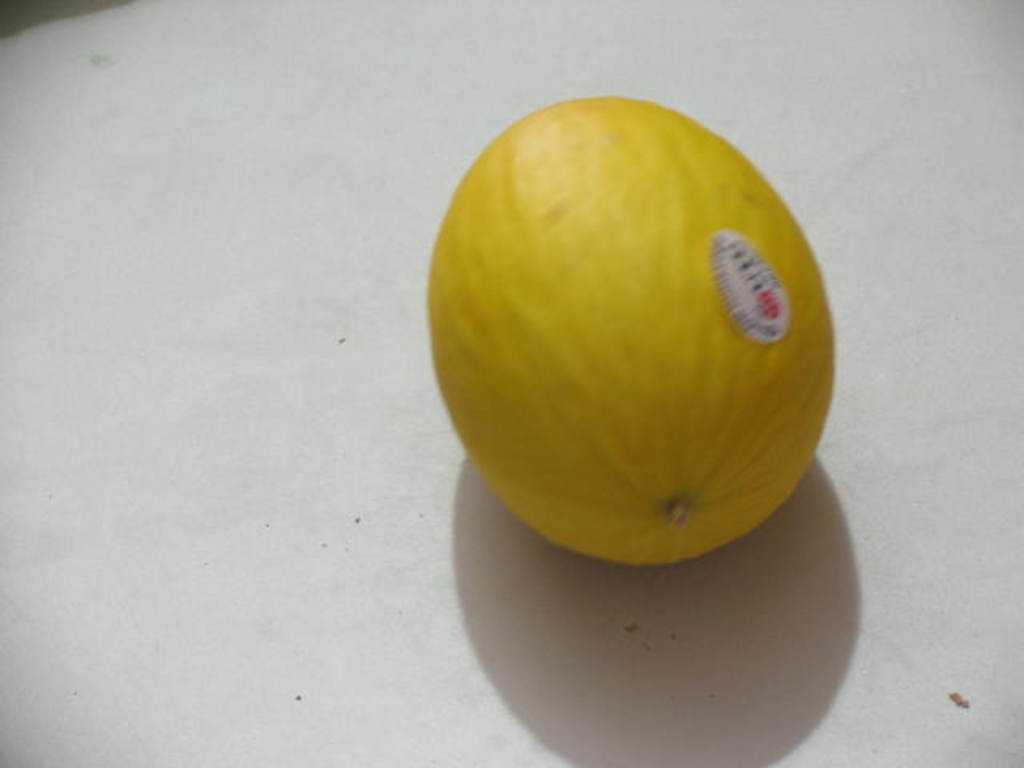}
		\hspace{-0.4em}
		\includegraphics[width=0.16\textwidth]{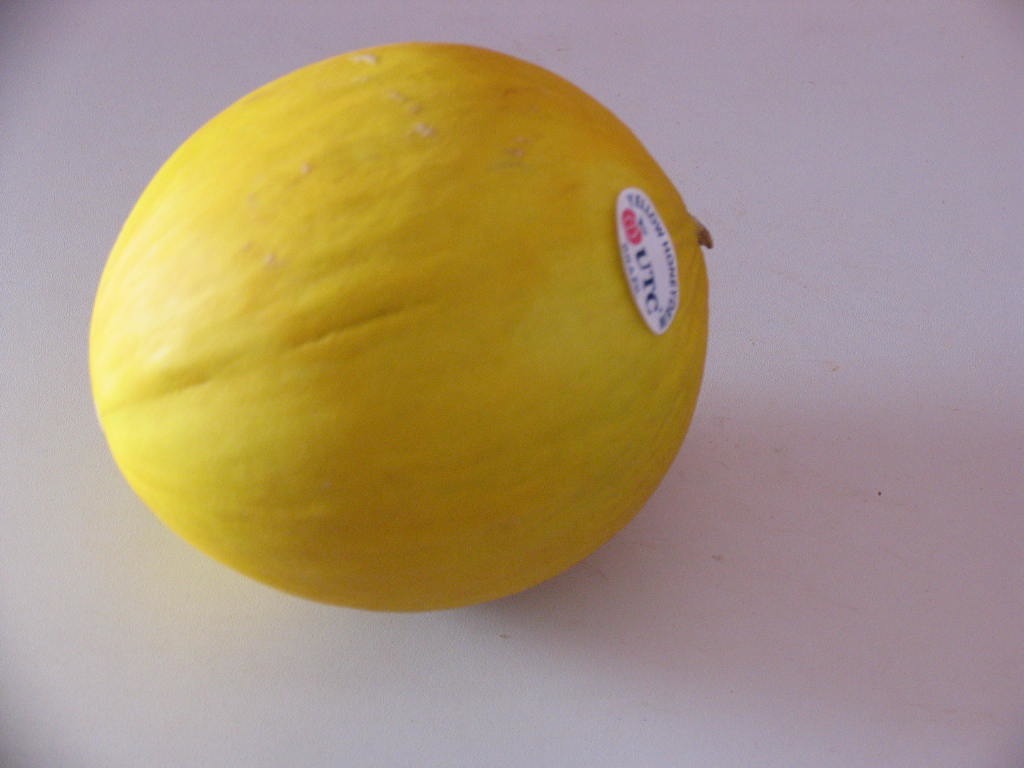}
	}
	\subfloat[Granny Smith Apple]{
		\includegraphics[width=0.16\textwidth]{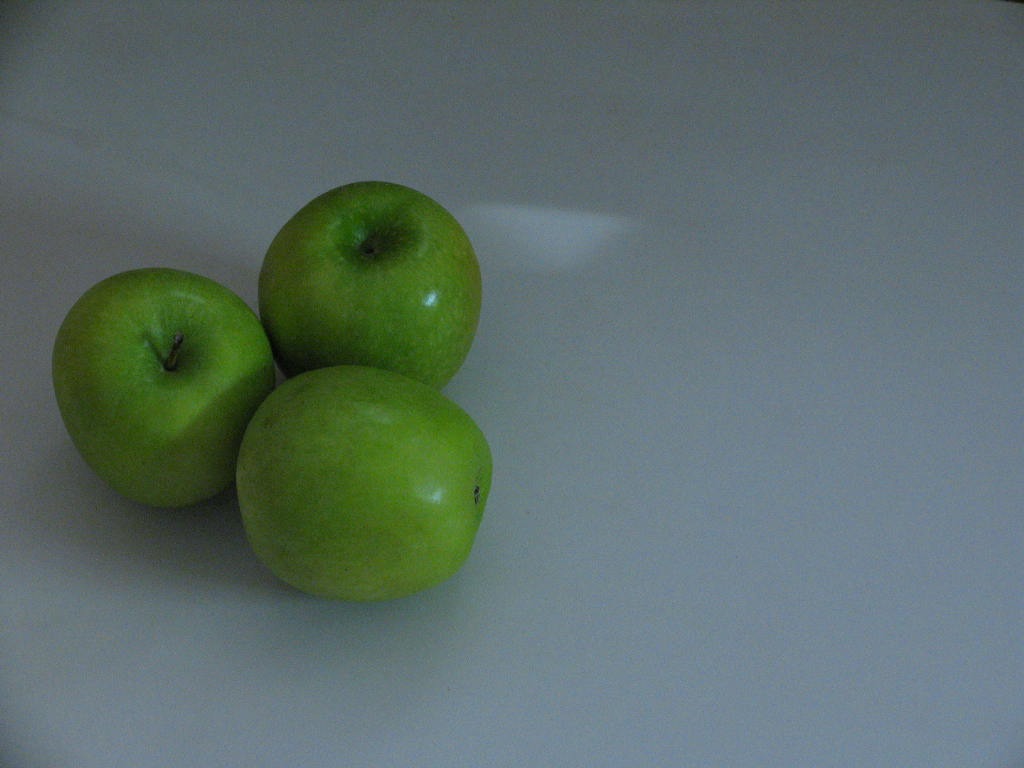}
		\hspace{-0.4em}
		\includegraphics[width=0.16\textwidth]{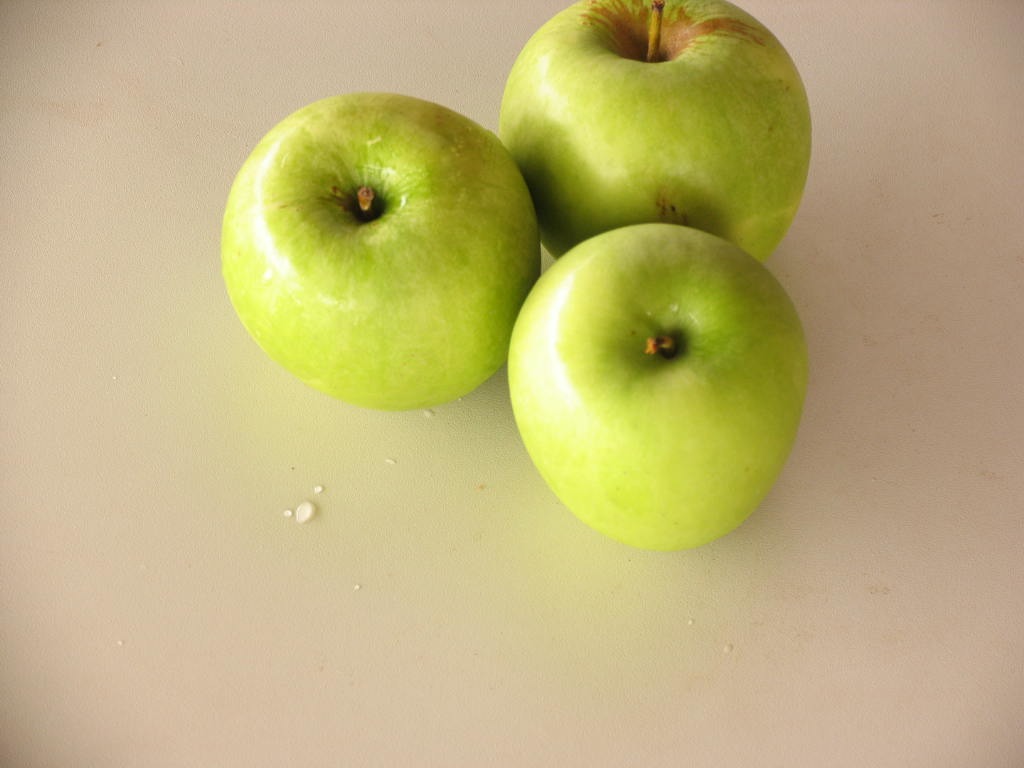}
	}
	\vspace{-1em}\\
	\subfloat[Fuji Apple]{
		\includegraphics[width=0.16\textwidth]{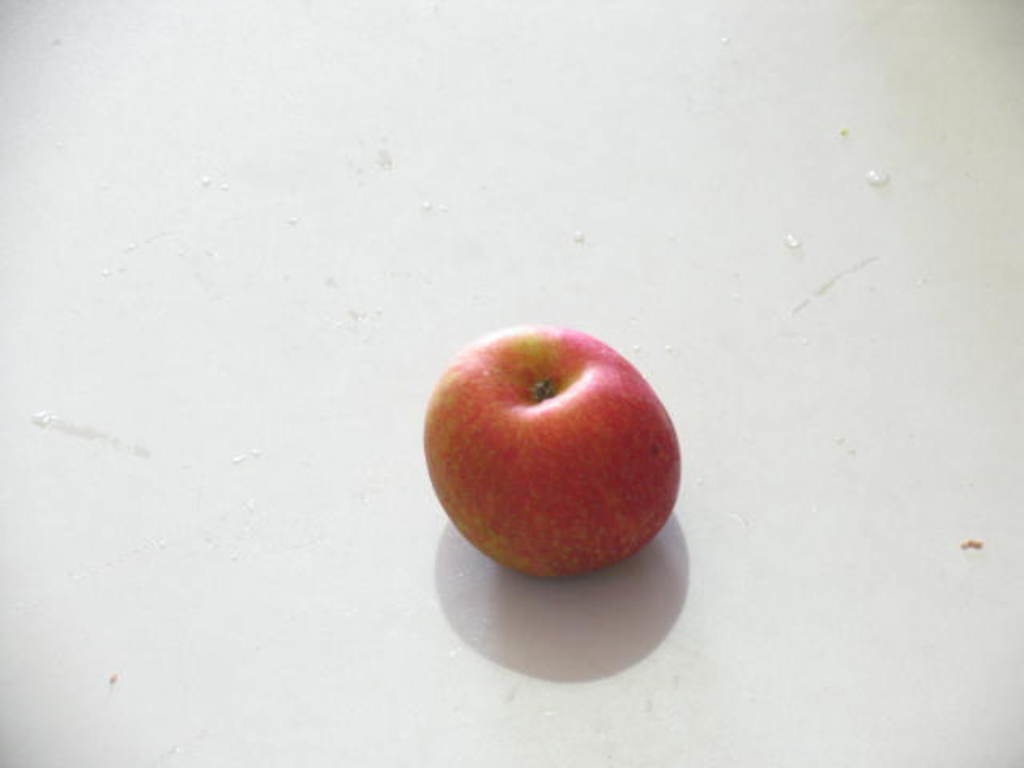}
		\hspace{-0.4em}
		\includegraphics[width=0.16\textwidth]{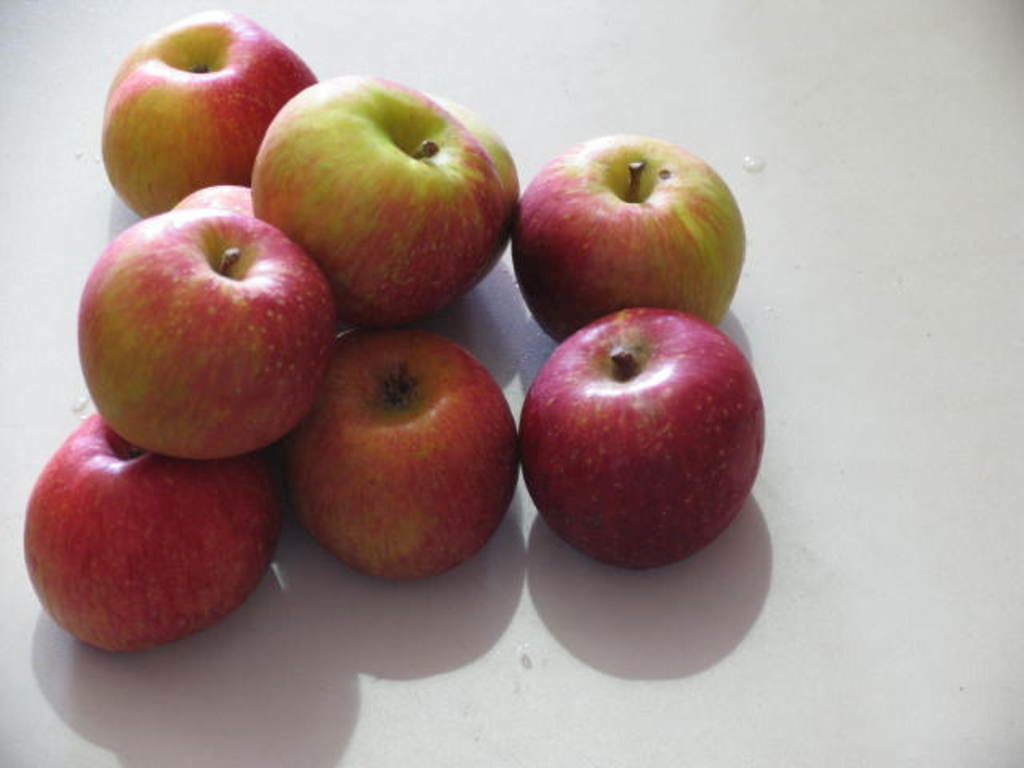}
	}
	\subfloat[Plum]{
		\includegraphics[width=0.16\textwidth]{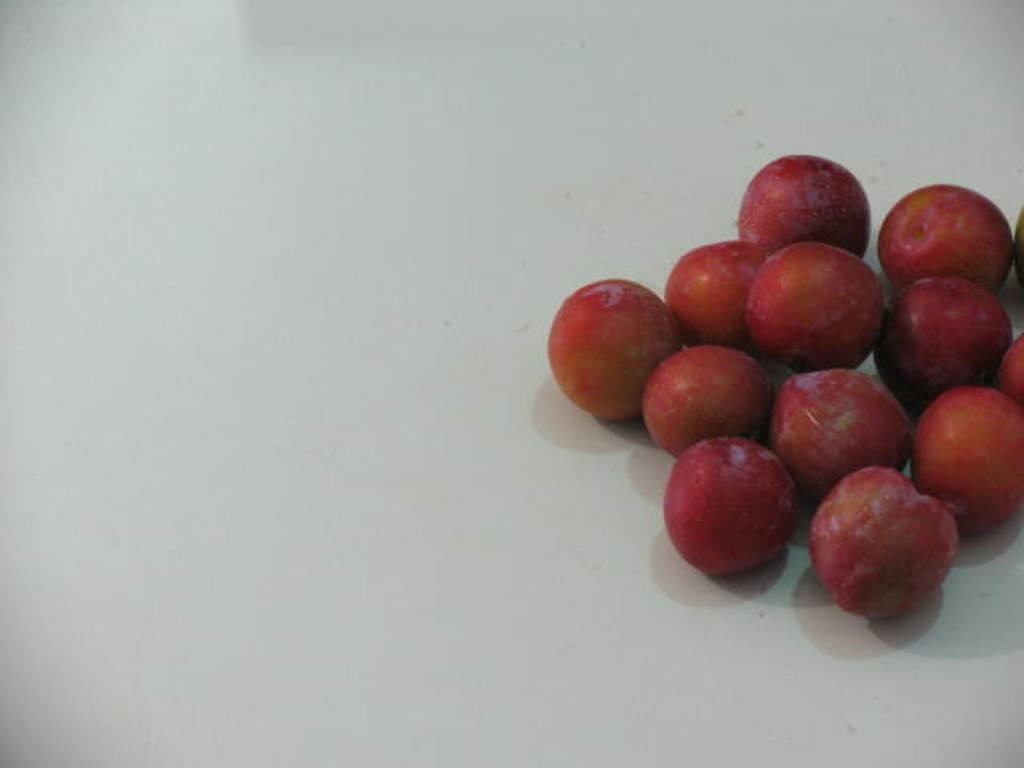}
		\hspace{-0.4em}
		\includegraphics[width=0.16\textwidth]{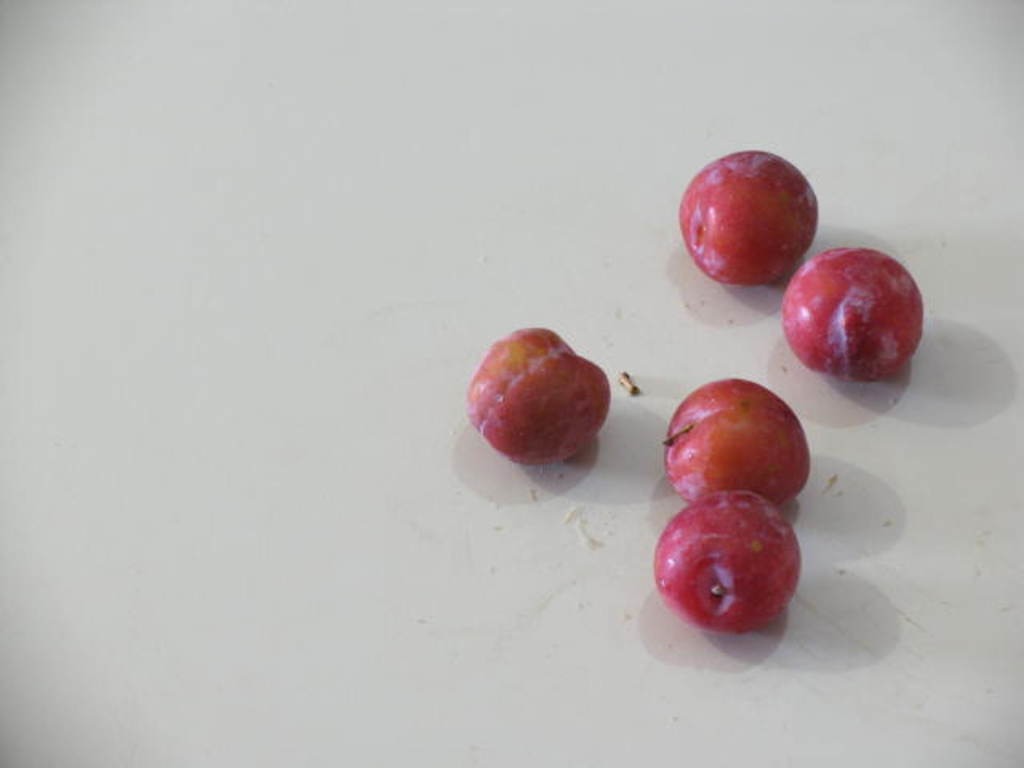}
	}
	\subfloat[Nectarine]{
		\includegraphics[width=0.16\textwidth]{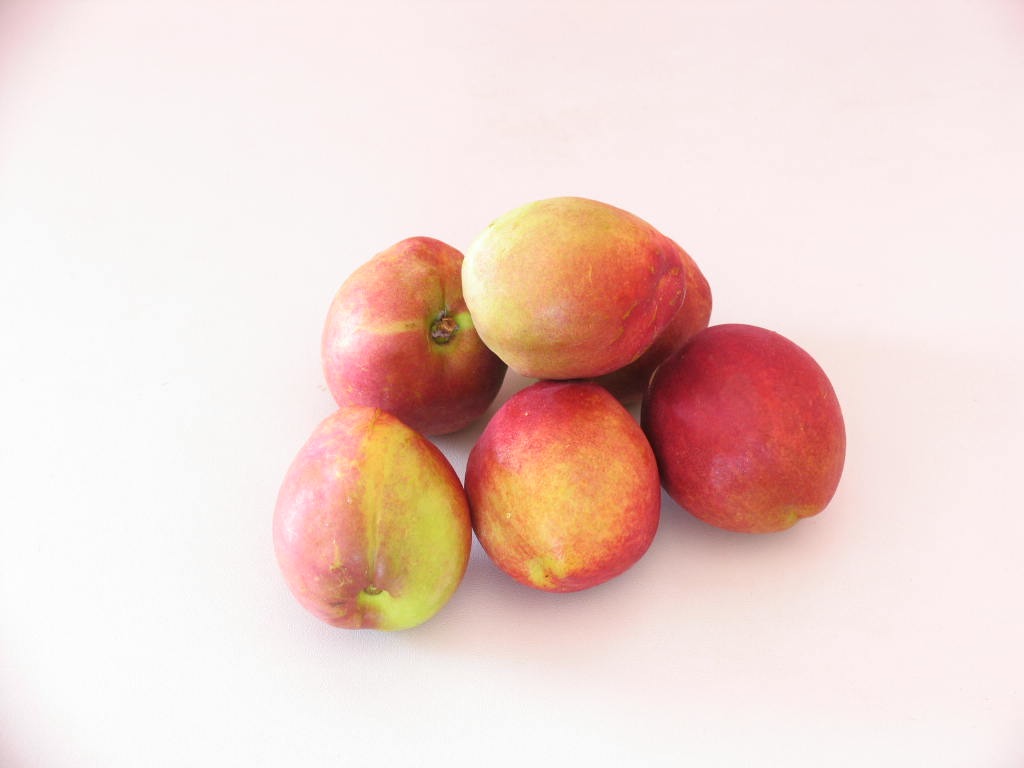}
		\hspace{-0.4em}
		\includegraphics[width=0.16\textwidth]{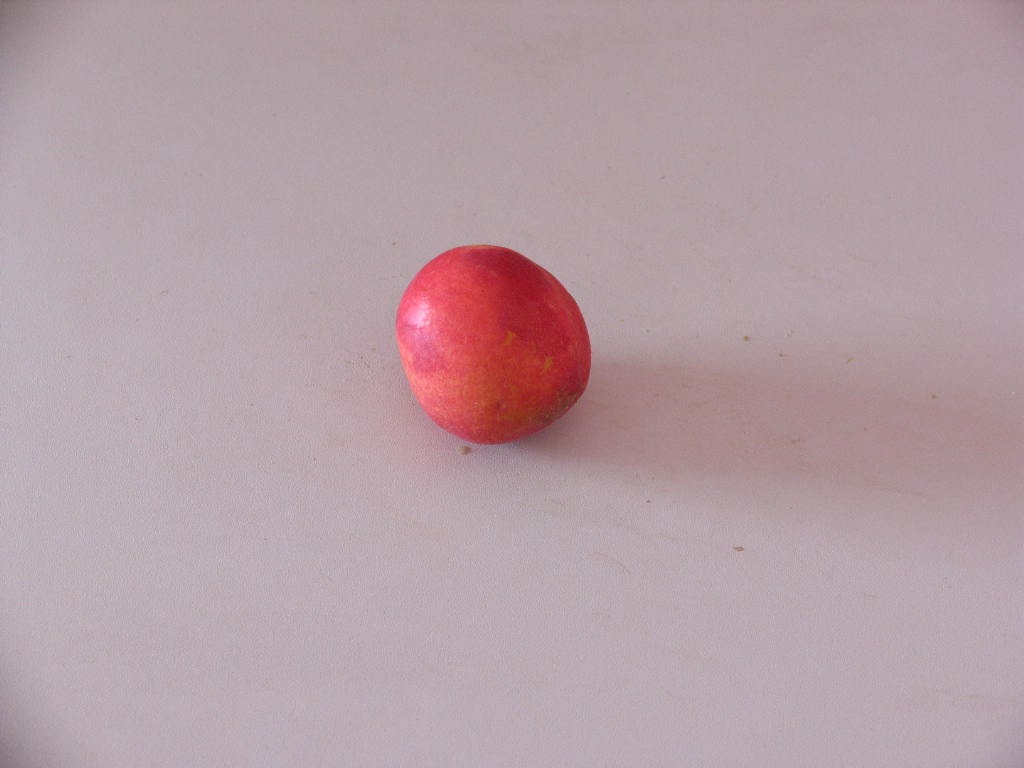}
	}\\
	\vspace{-2em}\\
	\caption{Examples of the \textit{Supermarket Produces} dataset.}
	\label{fig:fruits}
\end{figure}

\begin{figure}[!t]
	\subfloat[Cambridge]{
		\includegraphics[width=0.15\textwidth]{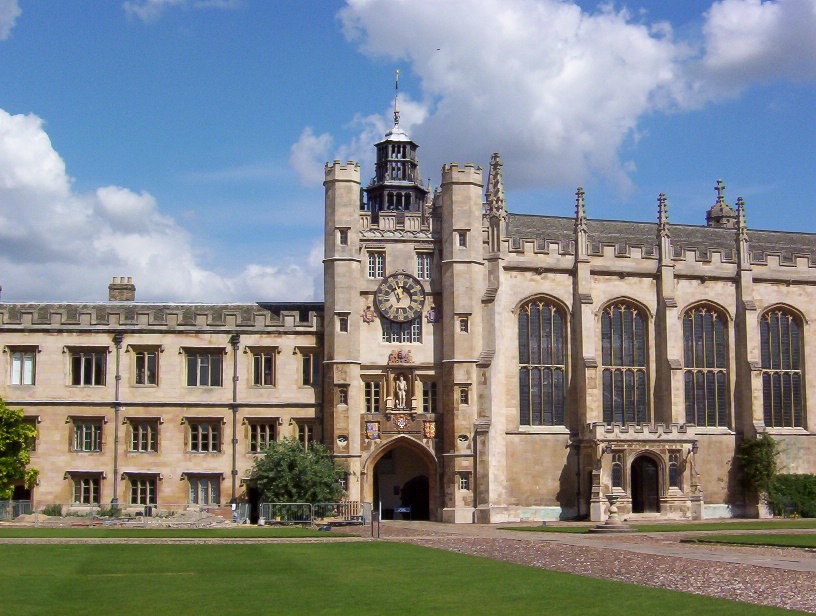}
		\hspace{-0.4em}
		\includegraphics[width=0.15\textwidth]{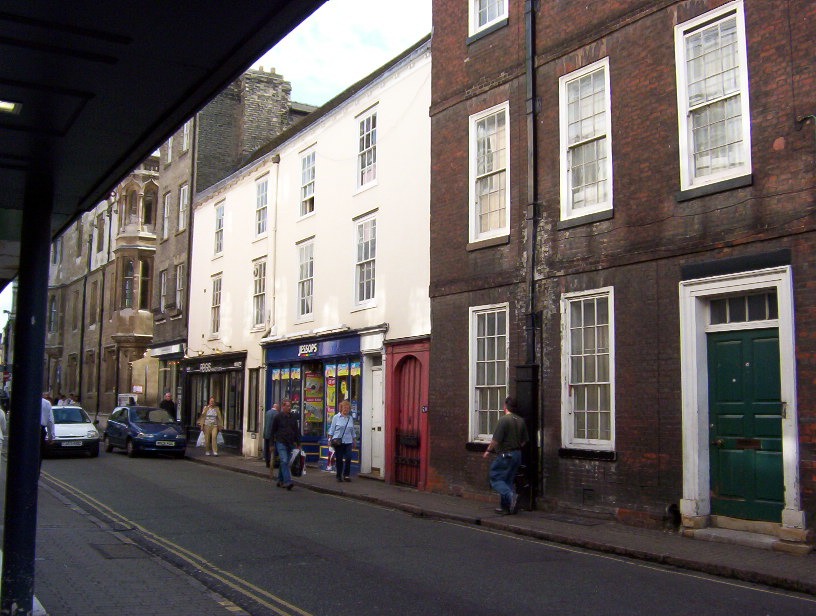}
	}
	\subfloat[Australia]{
		\includegraphics[width=0.17\textwidth]{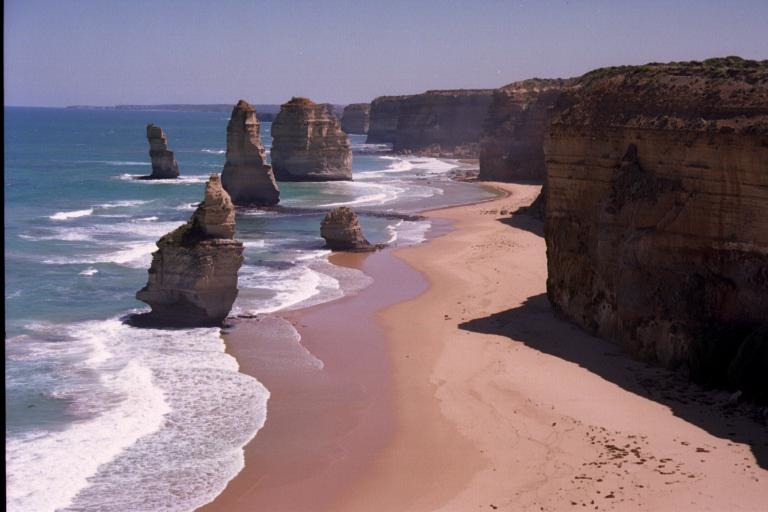}
		\hspace{-0.4em}
		\includegraphics[width=0.17\textwidth]{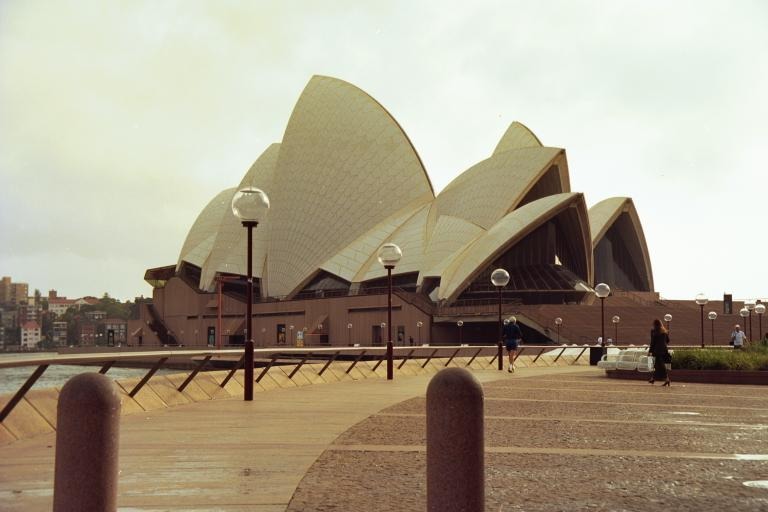}
	}
	\subfloat[Columbia Gorge]{
		\includegraphics[width=0.15\textwidth]{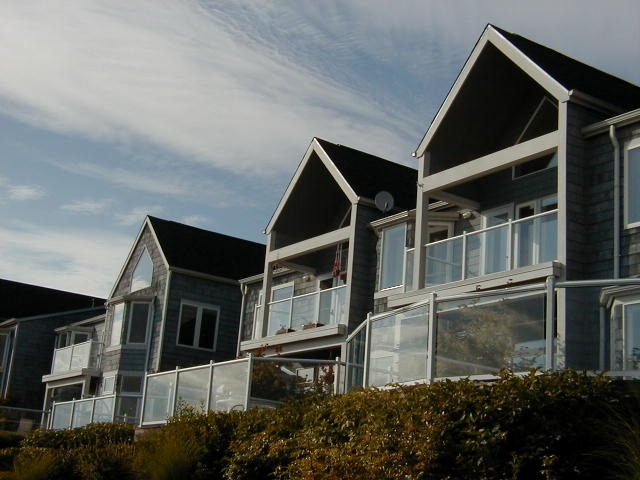}
		\hspace{-0.4em}
		\includegraphics[width=0.15\textwidth]{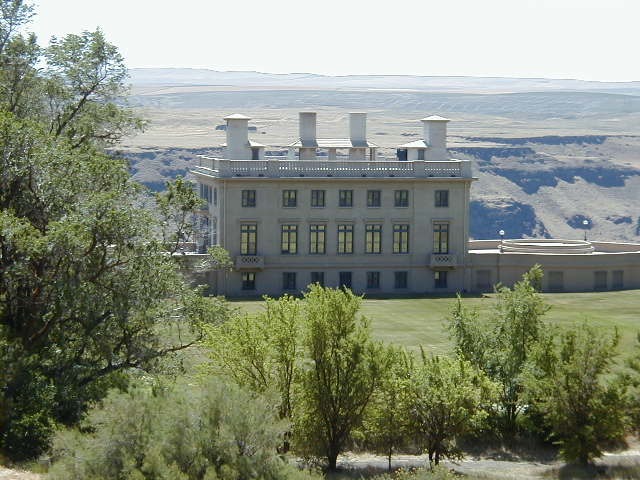}
	}\\
	\vspace{-1em}\\
	\subfloat[Barcelona]{
		\includegraphics[width=0.115\textwidth]{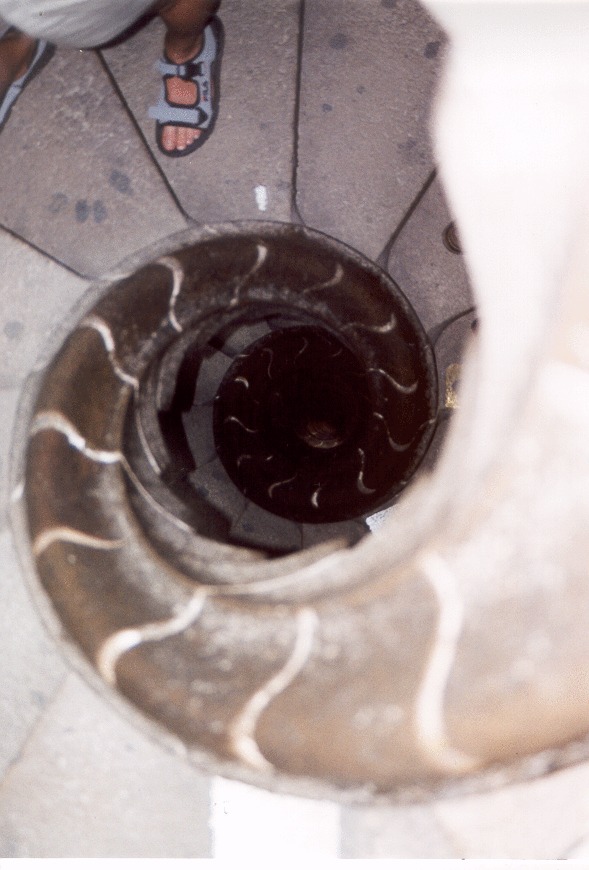}
		\hspace{-0.4em}
		\includegraphics[width=0.115\textwidth]{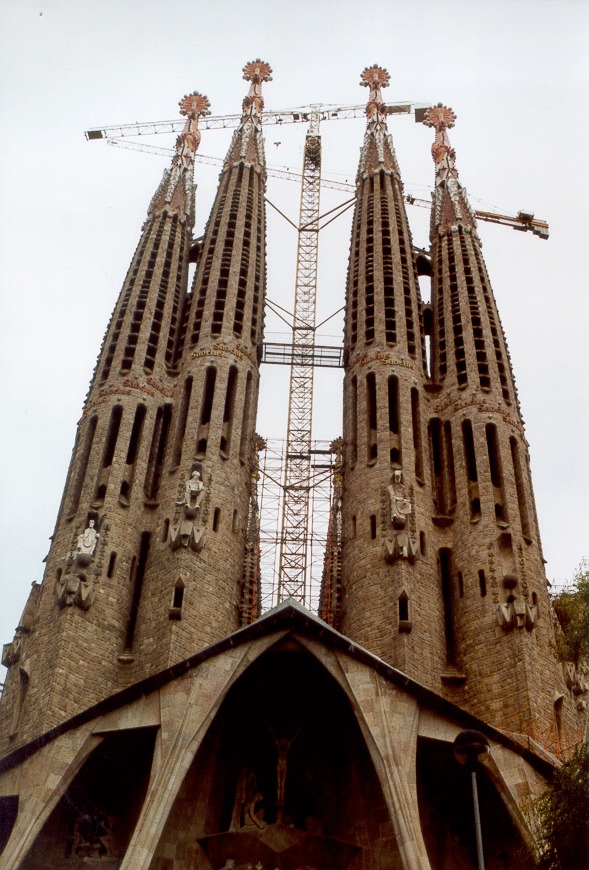}
	}
	\subfloat[Indonesia]{
		\includegraphics[width=0.115\textwidth]{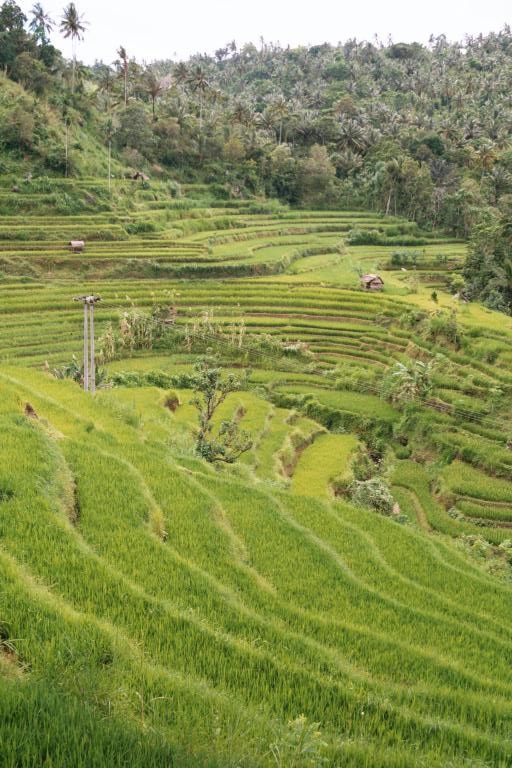}
		\hspace{-0.4em}
		\includegraphics[width=0.115\textwidth]{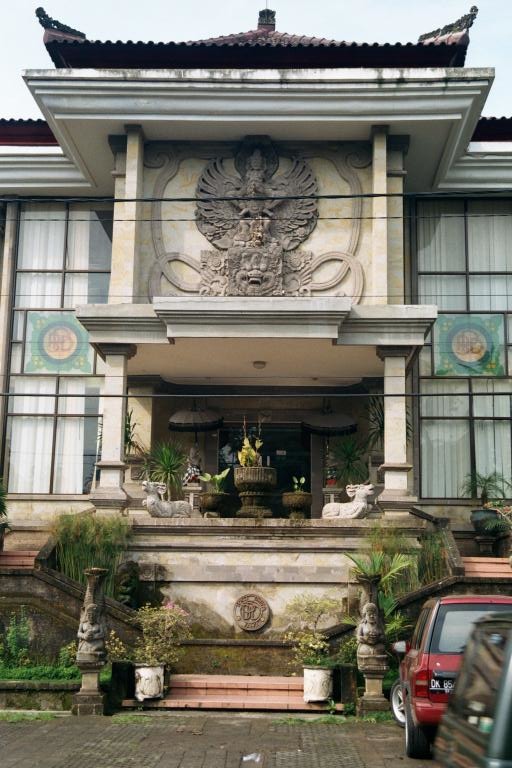}
	}
	\subfloat[Geneva]{
		\includegraphics[width=0.115\textwidth]{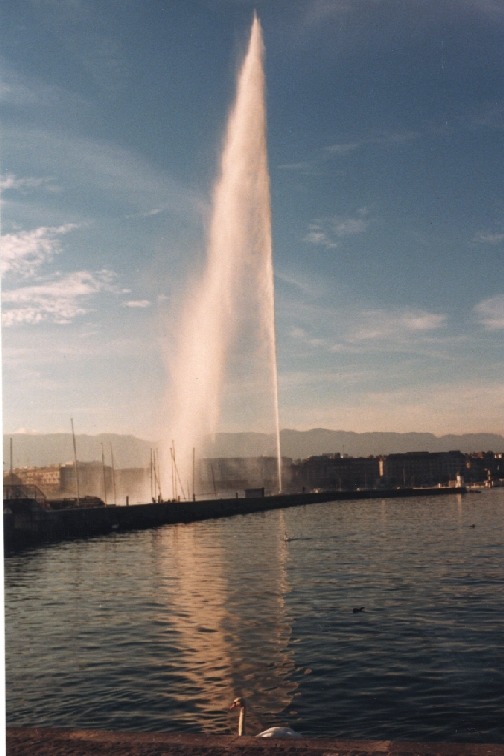}
		\hspace{-0.4em}
		\includegraphics[width=0.115\textwidth]{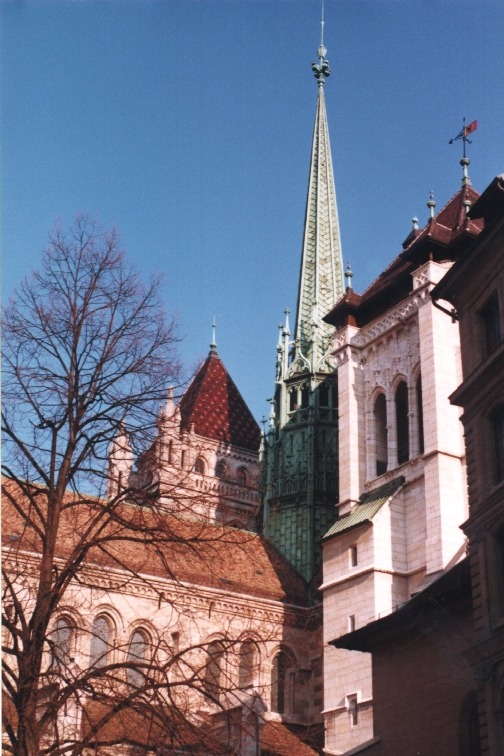}
	}
	\subfloat[Japan]{
		\includegraphics[width=0.115\textwidth]{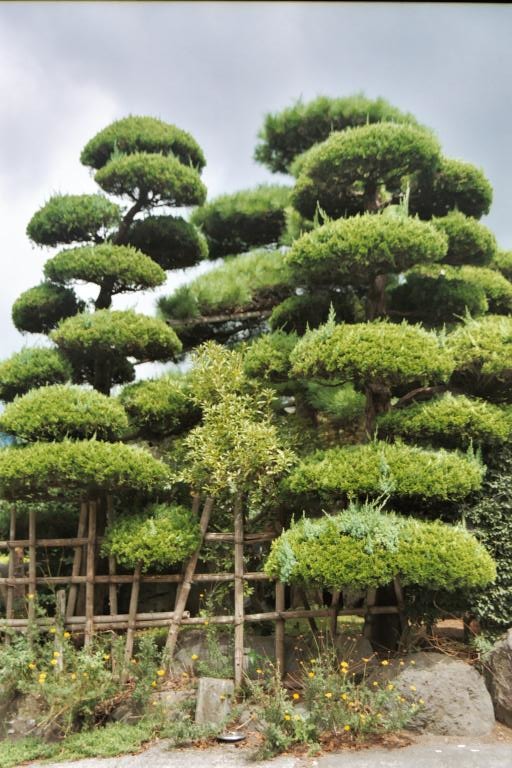}
		\hspace{-0.4em}
		\includegraphics[width=0.115\textwidth]{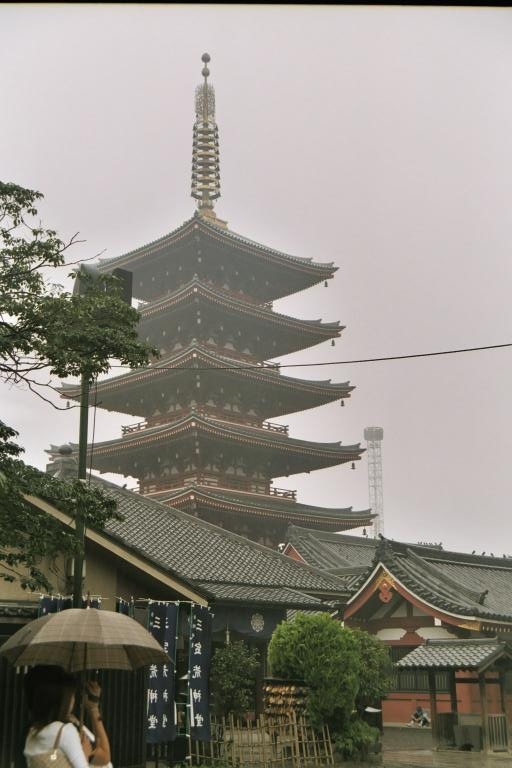}
	}\\
	\vspace{-2em}\\
	\caption{Examples of the \textit{Groundtruth} dataset.}
	\label{fig:groundtruth}
\end{figure}

\FloatBarrier	
}

\subsection{Baselines}
\label{subsec:baselines}

\subsubsection{Feature Extraction Algorithms}
In order to demonstrate the impact of using the learned quantizations in the generation of more effective image representations, we compare GA-based feature extractors with similar formulations {\em without} any quantization procedure. We use the BIC and the GCH original formulations (see Section~\ref{sec:color-quantization}) as baselines.

\edit{
\subsubsection{\textit{Winner-Take-All} Autoencoder}
We also perform comparisons with autoencoders (see Section~{\ref{subsec:aes}}), 
a class of methods based on Deep Learning -- state-of-the-Art framework for computer vision -- dedicated to perform representation learning and, consequently. They are, therefore, suitable recent approaches for comparison purposes.}

\edit{
According to Makhzani et al.~{\cite{Makhzani:2013}}, Sparse Autoencoders (SAE) yield the best performance than other types, such as Denoising Autoencoders, for feature extraction in tasks such as image classification.
Among SAEs, we selected Winner-Take-All Autoencoders WTA-AE (see Section~{\ref{subsec:aes}}) 
which hold some advantages in comparison with other SAEs including the capability of aiming any sparsity rate, efficient training and resource consumption besides allowing the use of reduced architectures.}

\edit{
Among the WTA-AE proposed configurations, we selected the CONV-WTA autoencoder, which is a non-symmetric architecture where the encoder consists of a stack of three 256-units ReLU convolutional layers ($5 \times 5$ filters) and the decoder is a 256-units linear deconvolutional layer of larger size ($11 \times 11$ filters): 256conv3-256conv3-256conv3-256deconv7. It also maintains $N_u$ hidden representation units between the encoder e decoder.
This is the same architecture used by Makhzani et al.~{\cite{Makhzani:2015}} in experiments for the CIFAR-10 dataset~{\cite{Krizhevsky:2009}}, an image dataset of domain similar to the ones used in our experiments.}

\edit{
Following the instructions of Makhzani et al., with the purpose of composing representations adequate to being used on an image classification/retrieval setting, we employed, after training, max-pooling on the last $N_u$ feature maps of the encoder, over $6 \times 6$ regions at strides of 4 pixels to obtain the final representation of $N_u \times 8 \times 8 = N_u \times 64$ total size.
In order to allow a fair performance comparison between the different-sized representations of WTA-AE and SCA, we employed Principal Components Analysis~{\cite{Wold:1987}} -- a well-known data projection algorithm -- on the WTA-AE representation as dimensionality reduction procedure where the number of dimensions corresponded to the imposed representation size limits.}

\subsection{Parameters}
\label{subsec:parameters}

Table~\ref{tab:parameters} presents the values adopted for the GA-based quantization learning process. The values chosen for population size, cross-over, mutation, elitism, and tournament parameters were defined empirically, but all of them represent typical values employed in GA-based optimization solutions.
\editt{Initially, it was applied a parameter search according to a $2^k$ Fractional Factorial Design (please refer to item 16.3.3 of \cite{Bukh:1992}) over a portion of the dataset. For the parameters which presented major sensitivities, a binary search was employed for exploration of different values.}

\edit{The total number of generations was defined aiming to ensure convergence of the evolutionary algorithm. However, we empirically observed that typically the best fitness value is not significantly improved after remaining unchanged for more than 50 iterations. Thus, one might impose a stopping condition regarding the fitness value as an option to avoid unnecessary iterations.}

\begin{table}[!t]
	\centering
	\caption{Genetic algorithm parameters. The indicated variables refer to Algorithm~{\ref{alg:quantization}}.}
	\label{tab:parameters} 
	\begin{tabular}{lc}
		\noalign{\smallskip}\hline\noalign{\smallskip}
		Two-point Cross-over Probability & 60\% \\
		One-point Mutation Probability & 40\% \\
		Number of Generations ($N_g$) & 200\\
		Population Size & 200 \\
		Tournament ($n_t$) & 5 \\
		Elitism ($k$) & 1\% \\
		\noalign{\smallskip}\hline
	\end{tabular}
\end{table}

We assess the quality of ranked lists defined by image representations obtained by means of a GA individual \edit{through} the FFP4 function~\cite{Fan:2004}. This score is defined for a given query image $q$ as:
\begin{eqnarray}
FFP4_{q} = \sum_{i=1}^{|D|} r_q(d_i) \times k_8 \times k_9^i
\end{eqnarray}
where $D$ is the image dataset; $r_q(d) \in [0, 1]$ is the relevance score for the image $d_i$ associated to the query, it being 1 if relevant and 0 otherwise; and $k_8$ and $k_9$ are two scaling factors adjusted to 7 e 0.982 respectively. The final fitness score is computed as the mean $FFP4$ for all images $q \in D$.

\editt{As Fan et al.~\cite{Fan:2004} explain, FFP4 is a utility function based on the idea that the utility of a relevant document decreases with its ranking order.
More formally, we need a utility function $U(x)$ which satisfies the condition $U(x_1) > U(x_2)$ for two ranks $x_1$ and $x_2$ which $x_1 < x_2$. Although there are many possible functions $U(x)$, we decided to use FFP4 as it presents good results in previous works~\cite{Torres:2009} applying this measure on similar evolutionary approaches that address rank-based tasks. Fan et al. report that this function and its associated parameters were chosen after exploratory data analysis.
}

\edit{
For the baseline WTA-AE, 
we set the parameters as: number of hidden representation units $N_u = 1024$ and \textit{winner-take-all} lifetime sparsity $k = 40\%$
empirically selecting them within the ranges \{64, 128, 256, 512, 1024\} and \{5\%, 10\%, 20\%, 30\%, 40\%, 50\%, 80\%\}, respectively.
}

\subsection{Evaluation Metrics}
\label{subsec:metrics}
	
	\subsubsection{Precision-Recall Curves}
	The most traditional measures to evaluate retrieval effectiveness over a set of queries are Precision and Recall~{\cite{baeza:1999}}.
	Precision measures the proportion of relevant images regarding the answer set, while Recall measures the proportion of relevant images retrieved in the answer set regarding all relevant images existing in the database. 
	
	A perfect system would provide a Precision equal to 1 (all the retrieved images are relevant) and a Recall also equal to 1 (all the relevant images were retrieved).
	In practice, there is an inverse relationship between them: the more items the system returns, the higher the likelihood that relevant documents will be retrieved (increasing recall). However, this comes at the cost of also retrieving many irrelevant documents (decreasing precision). Therefore, in general, it is necessary to define a compromise between them.
	
	In our case, we chose a measurement that considers Precision and Recall as functions of each other,
	generating interpolated Precision-Recall curves (11 points) whose the precision points $P$ given by
	\begin{eqnarray}
	    P(r_i) = \max_{\forall j | r_i \leq r_j} P(r_j)
	\end{eqnarray}
	where $i,j \in {0, 1, ..., 10}$ represent recall levels.
	
	In order to evaluate the retrieval effectiveness over a set of query images Q, an averaged Precision-Recall curve is computed according to
	\begin{eqnarray}
	    \overline{P}(r_i) = \sum_{q=1}^{|Q|} \frac{1}{|Q|} P_q(r_i)
	\end{eqnarray}
	\edit{where $P_q$ corresponds to the precision of the q-th query image.}
	
	\subsubsection{MAP: Mean Average Precision}
	In some cases, the Precision-Recall curves appear occluded or inter-crossed, restraining a proper visual comparison.
	Because of the compromise between Precision and Recall, it is possible to employ a combination of the two measures as a single metric.  
	This is the case of Mean Average Precision~{\cite{baeza:1999}} which provides a convenient measure to quantitatively compare Precision-Recall curves and is defined as
	\begin{eqnarray}
	MAP = \frac{1}{|Q|} \sum_{q=1}^{|Q|} AP_q
	\end{eqnarray}
	\begin{eqnarray}
	AP_q = \frac{1}{|R_q|} \sum_{k=1}^{|R_q|} P(R_q[k])
	\end{eqnarray}
	where $R_q$ is the set of relevant images in the dataset $Q$ for each image $q$.
	
	\subsubsection{P@10}
	As observed on real-world applications of CBIR, the user gives prior attention for a small group of the top answers, corresponding to the first page of results, usually preferring to reformulate the query instead of checking the next pages.
	The \textit{Precision-Recall} curves and \textit{MAP} do not provide an adequate measurement for the effectiveness of these top results as they generally consider longer portions of the ranking. 
	In order to address this issue, we also measured the precision at the top-10 results (P@10)~\cite{baeza:1999}.

    Due to the proximity of some measures and aiming to provide accurate comparisons between the methods and its baselines, we used the Student's Paired t-Test~{\cite{Kim:2015}} (p-value $<$ 0.05) to statistically verify the results of Precision-Recall, MAP, and P@10.
	
	\subsubsection{Representation Size}
	In order to evaluate the descriptions dimensionality and possibly detect occurrence compactness regarding the previous methods, we measured the representation size, defined as the total number of bins that compose the histogram representations.

\subsection{Experimental Protocol}
\label{subsec:protocol}

In order to evaluate the proposed method, we conducted a $k$-fold cross-validation. According to this protocol, the dataset is randomly split into k mutually exclusive samples subset (folds) of approximated size. Then, the $k - 1$ subsets are chosen as training set, and the remaining one as test set. The execution is repeated $k$ times, and for each time, a different subset (without replacement) is chosen as the current test set and the remaining compose the training set.

We carried out all experiments considering $k = 5$ folds. As a consequence, for each experiment, the method was executed 5 times using 80\% of the dataset as training set and 20\% as test set.

\section{Results and Discussion}
\label{sec:results} 

This section compares the results of the proposed methods and baselines according to the evaluation measures.

First, we present the results of the UA methods with regard to Precision-Recall (Figs. \ref{chart:pr_bic_nla} and \ref{chart:pr_gch_nla}), P@10 (Figs. \ref{chart:precision_bic_nla} and \ref{chart:precision_gch_nla}), MAP (Figs. \ref{chart:map_bic_nla} and \ref{chart:map_gch_nla}), and representation size (Fig. \ref{chart:size_nla}) for all datasets and feature extractors.
Next, we present charts comparing the SCA results for Precision-Recall (Figs. \ref{chart:pr_bic_la1}-\ref{chart:pr_gch_la2}), P@10 (Figs. \ref{chart:precision_bic_la} and \ref{chart:precision_gch_la}), MAP (Figs. \ref{chart:map_bic_la} and \ref{chart:map_gch_la}), and representation size (Fig.~\ref{chart:size_la}). In the figures, the symbols above each pair of measures indicate whether the proposed method yields statistically 
	better \includegraphics[height=0.9em]{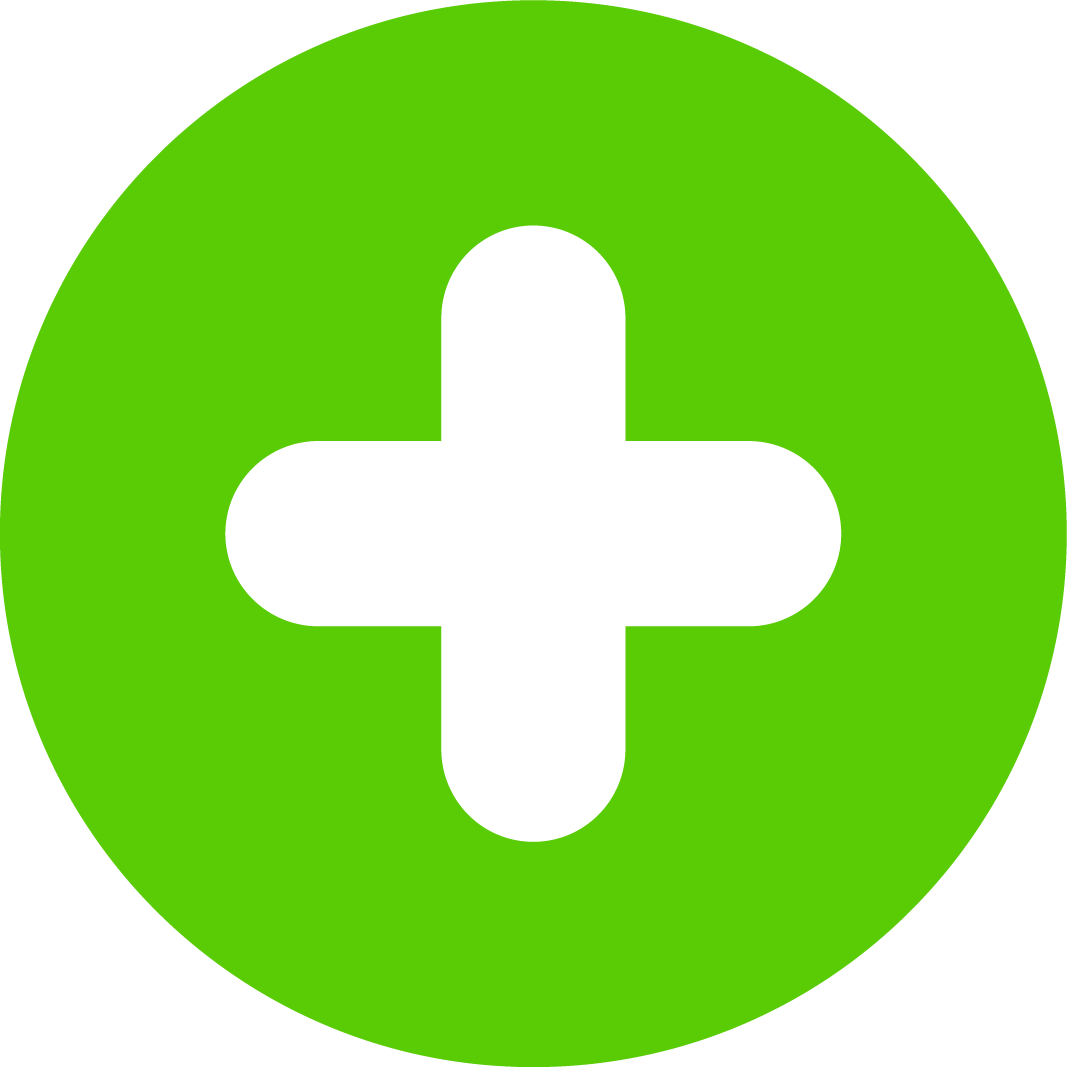}, 
	worse \includegraphics[height=0.9em]{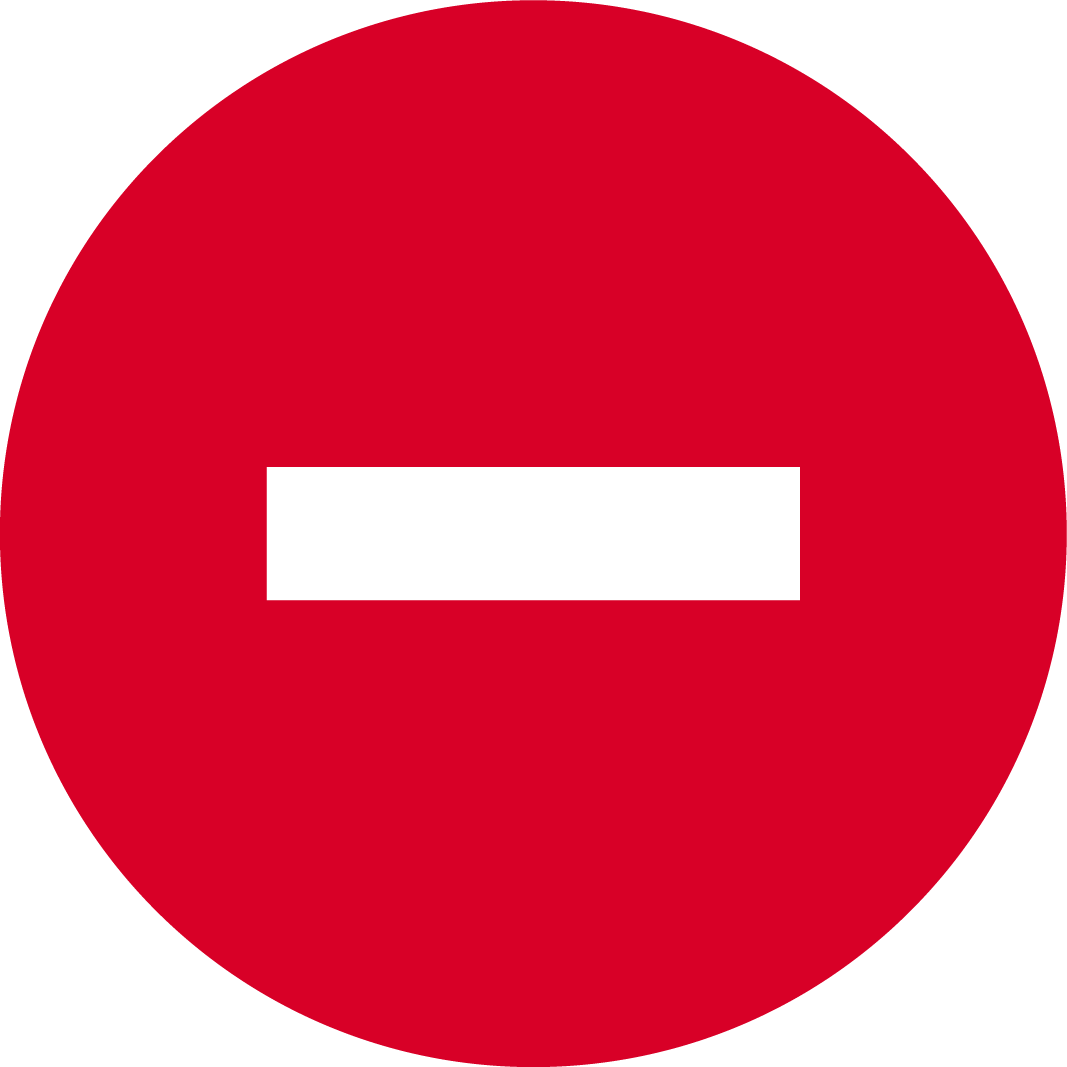}, 
	or similar \includegraphics[height=0.9em]{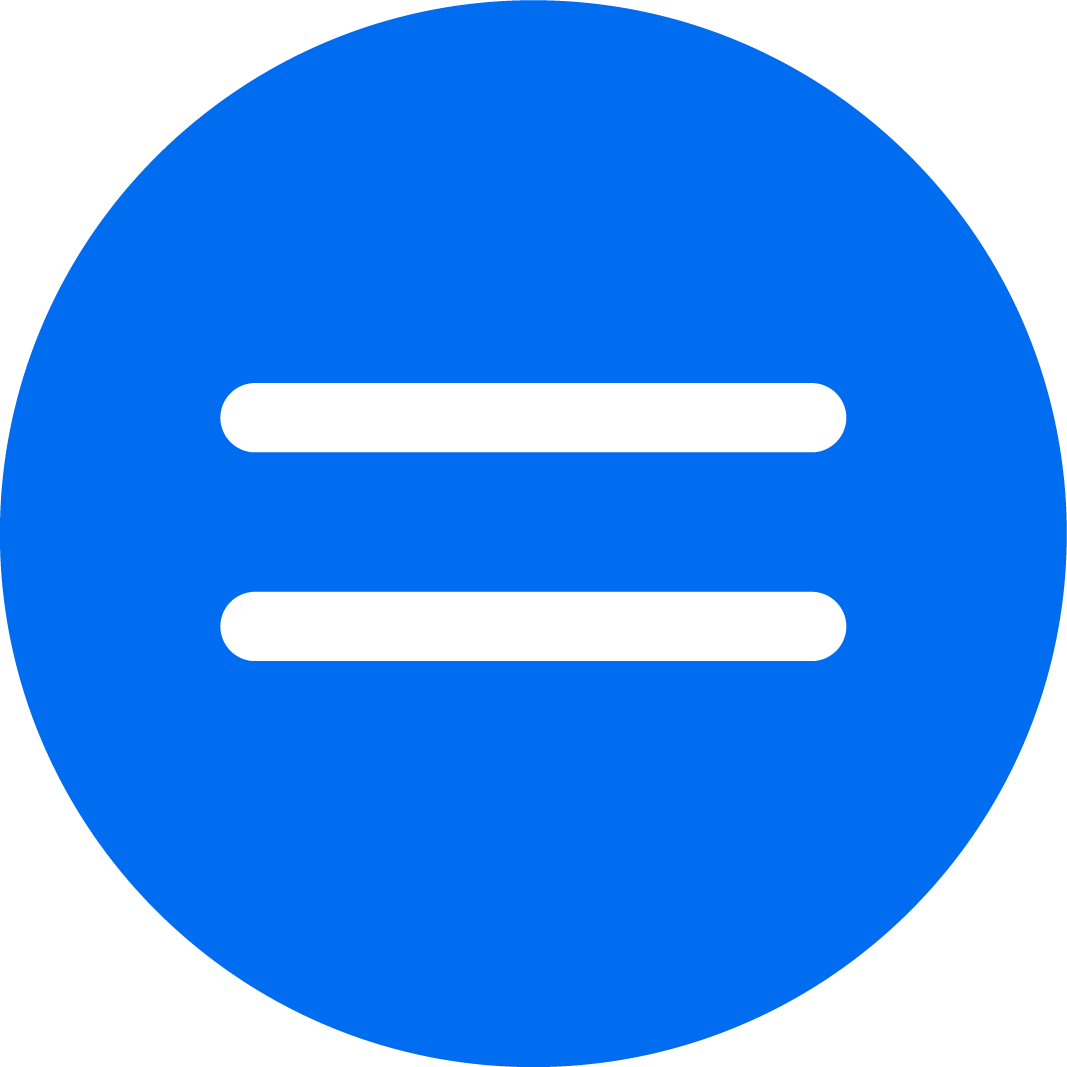} results to those observed for the baselines 
	\edit{(the minimum between BIC/GCH and WTA-AE)}, 
	considering rejection of the null hypothesis when p-value $<$ 0.05.

The following sections present and discuss the experimental results and provides comparisons between these two proposed approaches and baselines.

\subsection{Unconstrained Approach}
\label{ssec:nonlimited}

\begin{figure}[!t]
	\centering
	\subfloat[Groundtruth]{\includegraphics[width=0.5\textwidth]{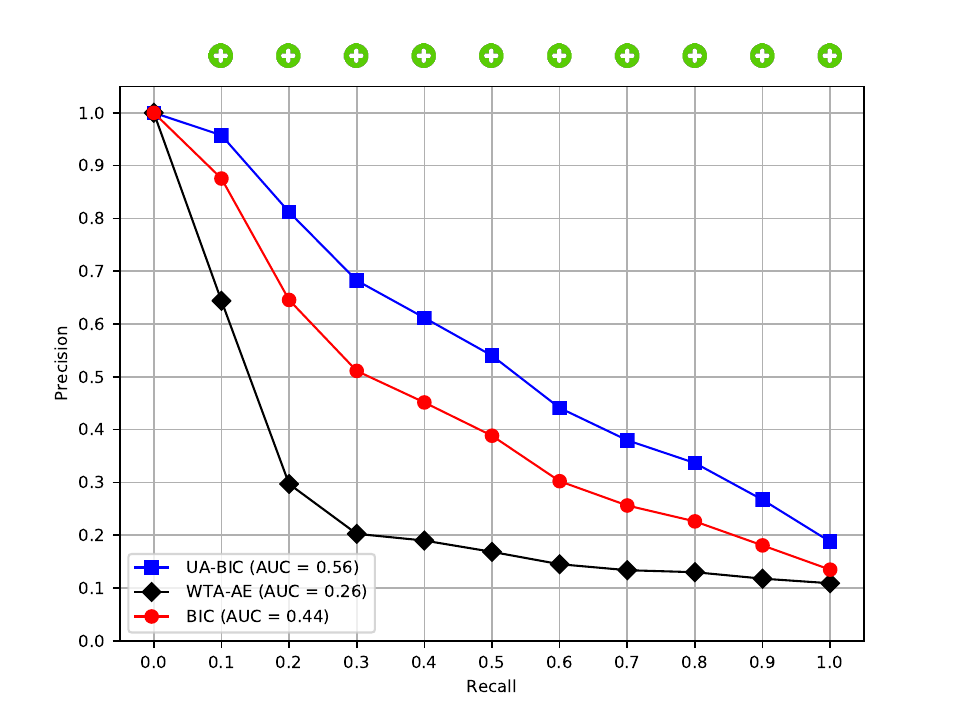}\label{chart:pr_bic_nla_coffe}}
	\subfloat[Coil-100]{\includegraphics[width=0.5\textwidth]{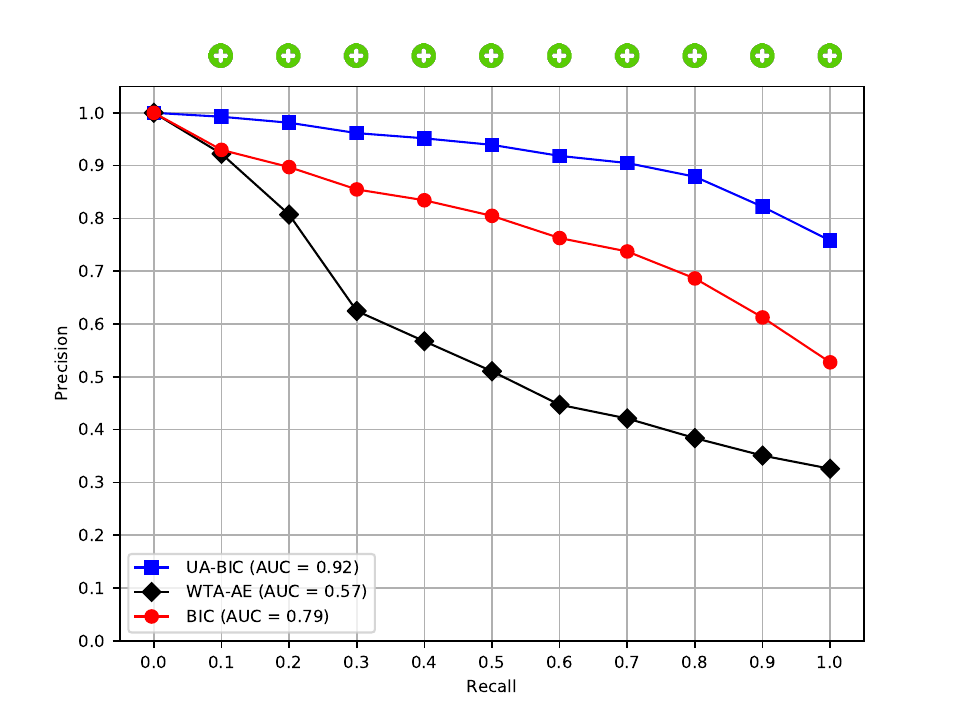}\label{chart:pr_bic_nla_coil100}}\\
	\subfloat[Corel-1566]{\includegraphics[width=0.5\textwidth]{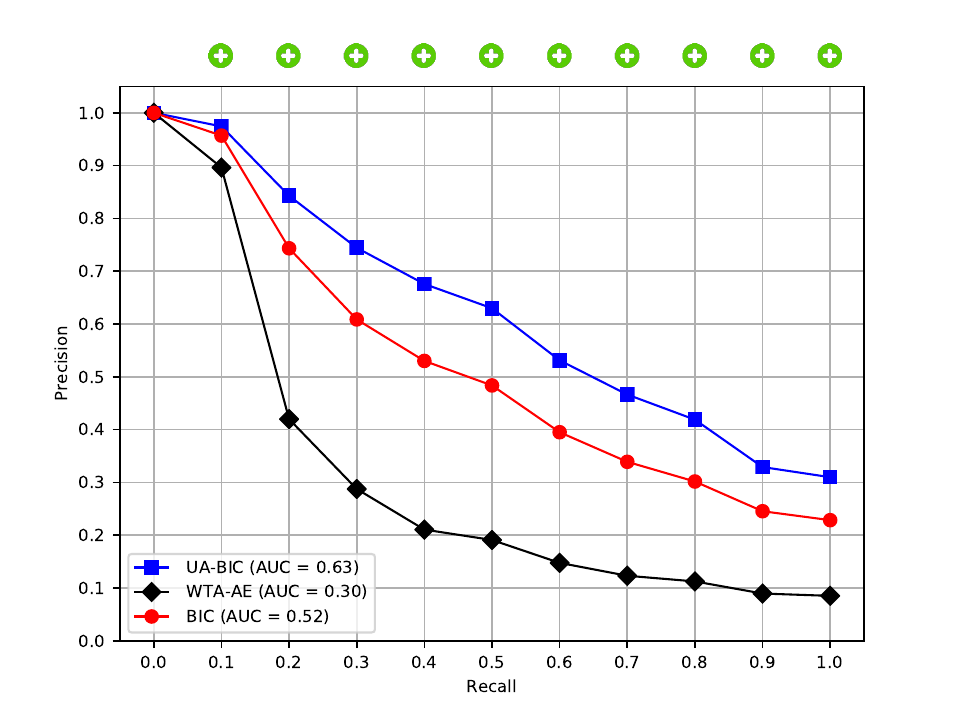}\label{chart:pr_bic_nla_corel1566}}
	\subfloat[Corel-3906]{\includegraphics[width=0.5\textwidth]{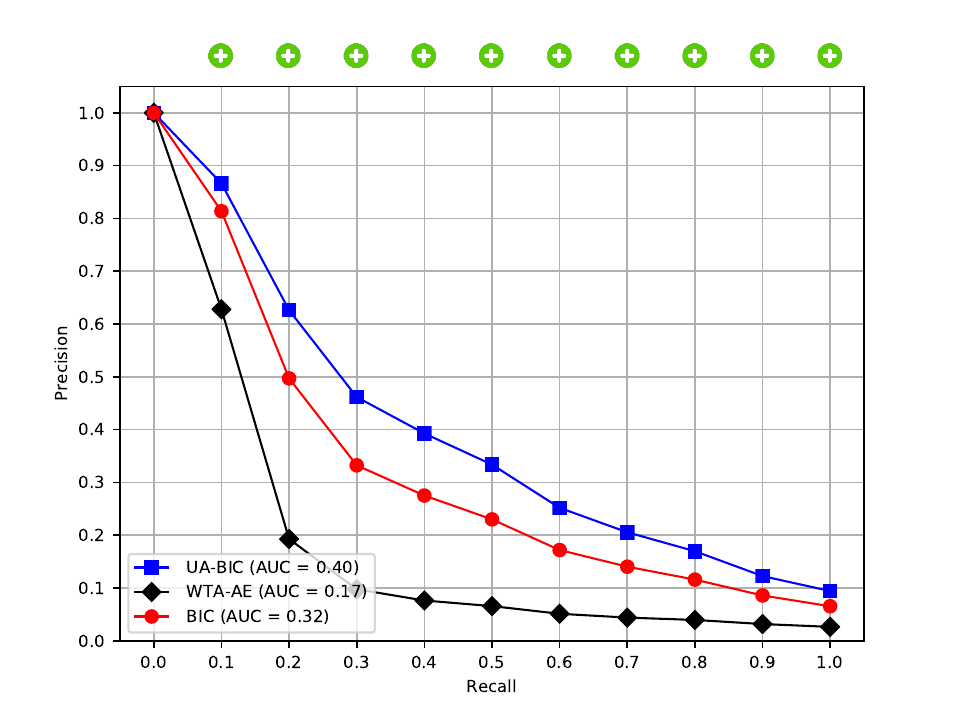}\label{chart:pr_bic_nla_corel3909}}\\
	\subfloat[ETH-80]{\includegraphics[width=0.5\textwidth]{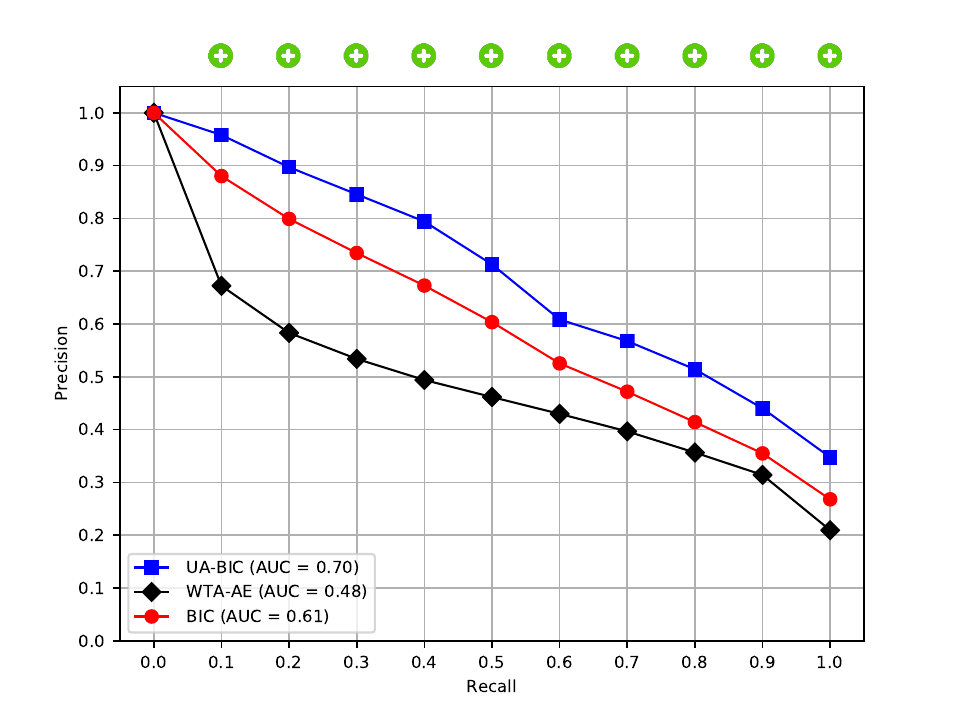}\label{chart:pr_bic_nla_eth80}}
	\subfloat[Supermarket P.]{\includegraphics[width=0.5\textwidth]{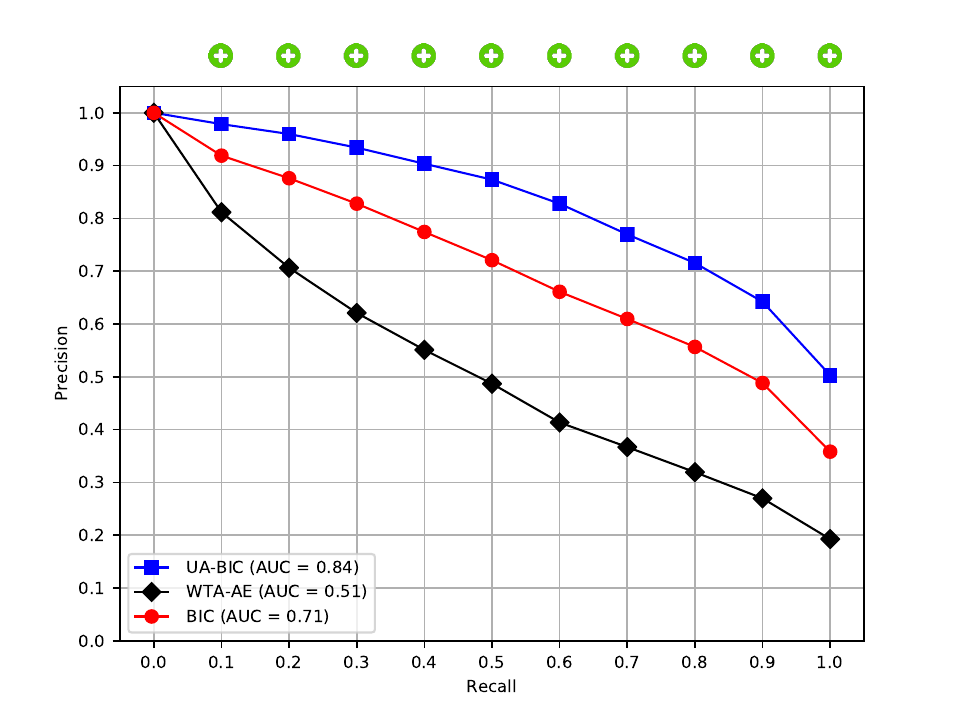}\label{chart:pr_bic_nla_fruits}}\\
	\subfloat[MSRCORID]{\includegraphics[width=0.5\textwidth]{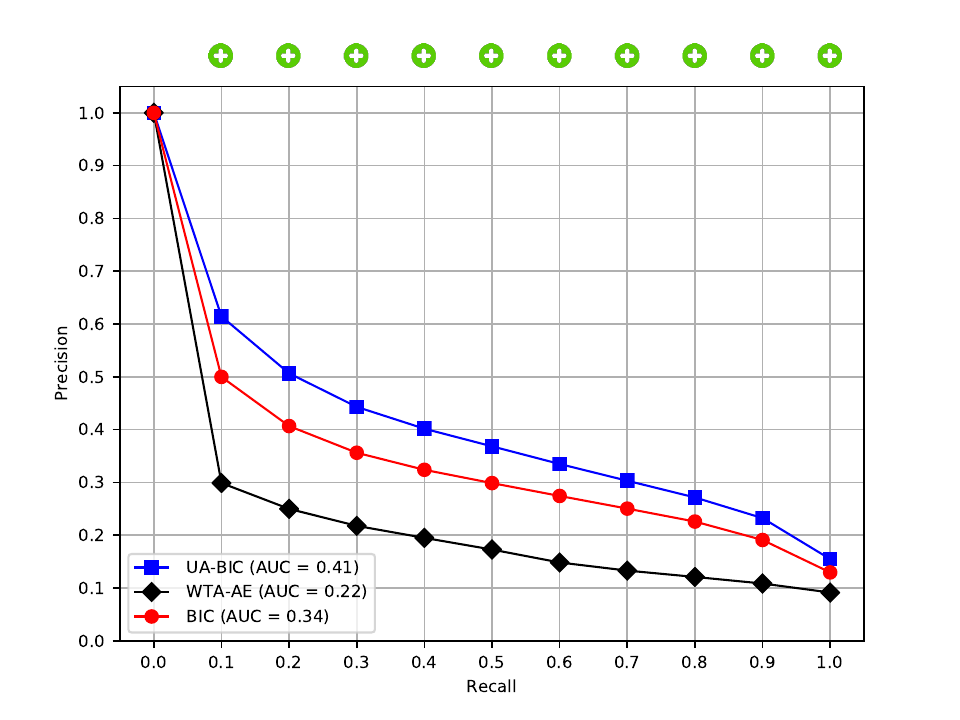}\label{chart:pr_bic_nla_msrcorid}}
	\subfloat[UCMerced Land-use]{\includegraphics[width=0.5\textwidth]{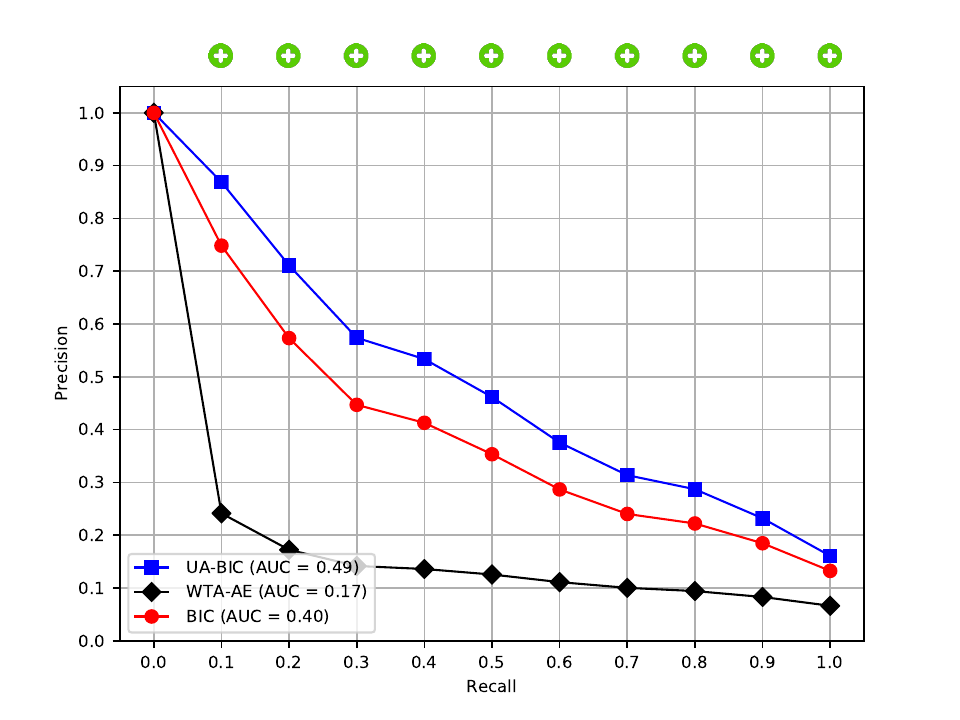}\label{chart:pr_bic_nla_ucmerced}}
	\caption{Comparison between the Precision-Recall Curves of the UA method, WTA Autoencoder and the BIC feature extractor.}
	\label{chart:pr_bic_nla}
\end{figure}

\begin{center}
	\begin{figure}[!t]
		\subfloat[P@10]{\includegraphics[width=0.5\textwidth]{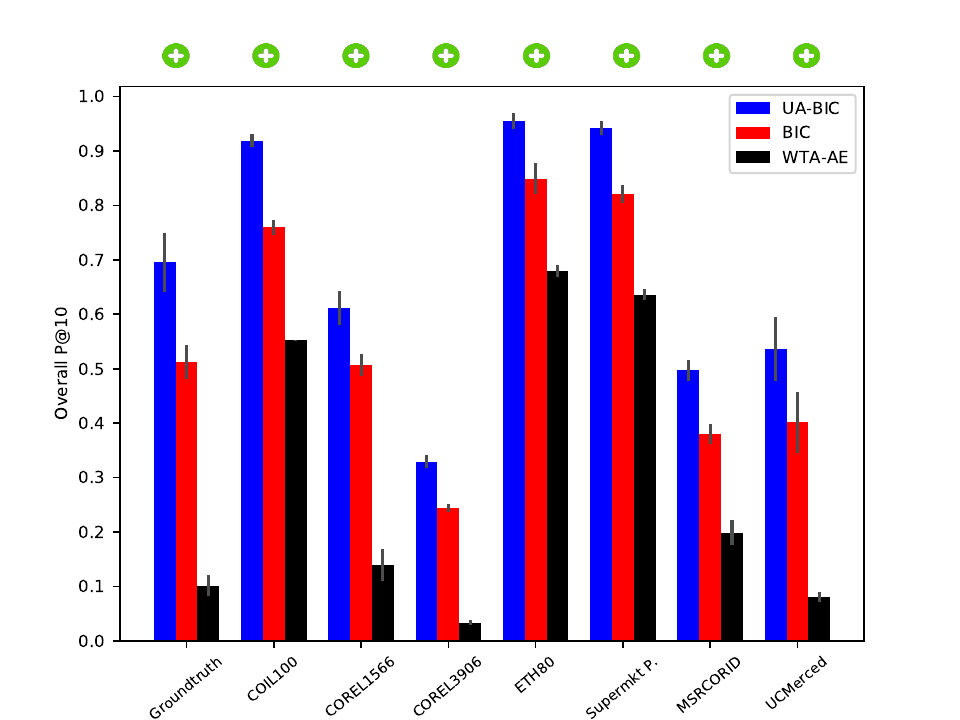}\label{chart:precision_bic_nla}}
		\subfloat[MAP]{\includegraphics[width=0.5\textwidth]{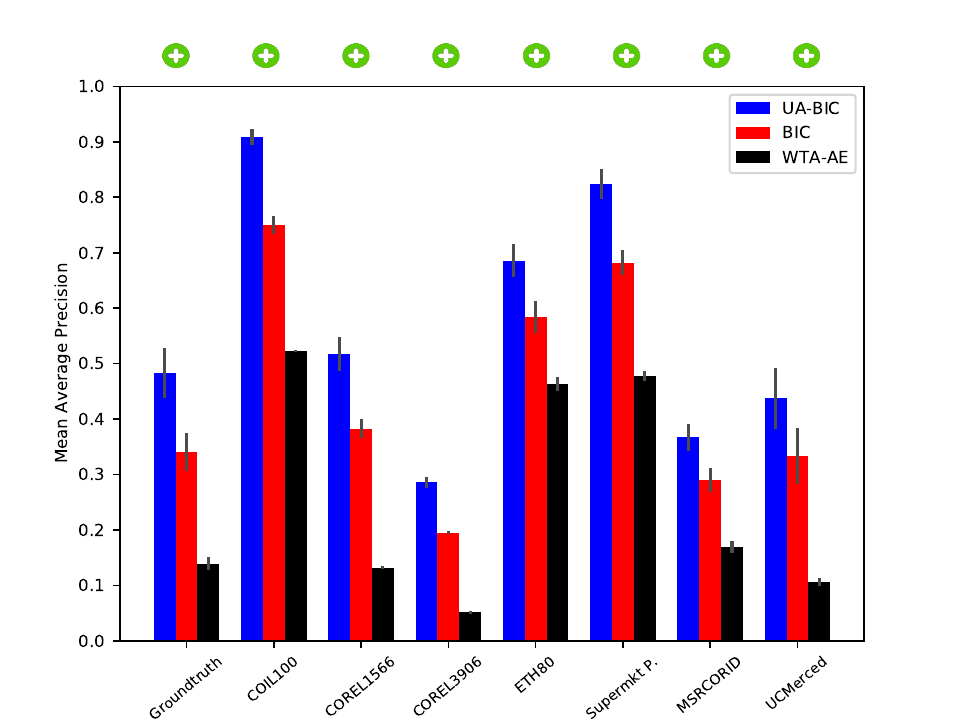}\label{chart:map_bic_nla}}
        
  \caption{Comparison between the (a) P@10 and (b) MAP results of UA, WTA Autoencoder and the {\bf BIC} feature extractor.}
  	\label{chart:p10_map_bic_nla}      
	\end{figure}
\end{center}

Observing the Precision-Recall curves for the BIC feature extractor (Fig. \ref{chart:pr_bic_nla}), 
the UA outperforms its baselines for all datasets. According to the P@10 measurements (Fig.~\ref{chart:precision_bic_nla}), the method also presents, on average, more relevant results in the first positions of the ranking for all datasets.
The superior MAP results (Fig.~\ref{chart:map_bic_nla}) confirm the superiority of UA, as this measure takes into account the performance of the evaluated methods for the whole Precision-Recall curve.

\begin{figure}[!t]
	\centering
	\subfloat[Groundtruth]{\includegraphics[width=0.5\textwidth]{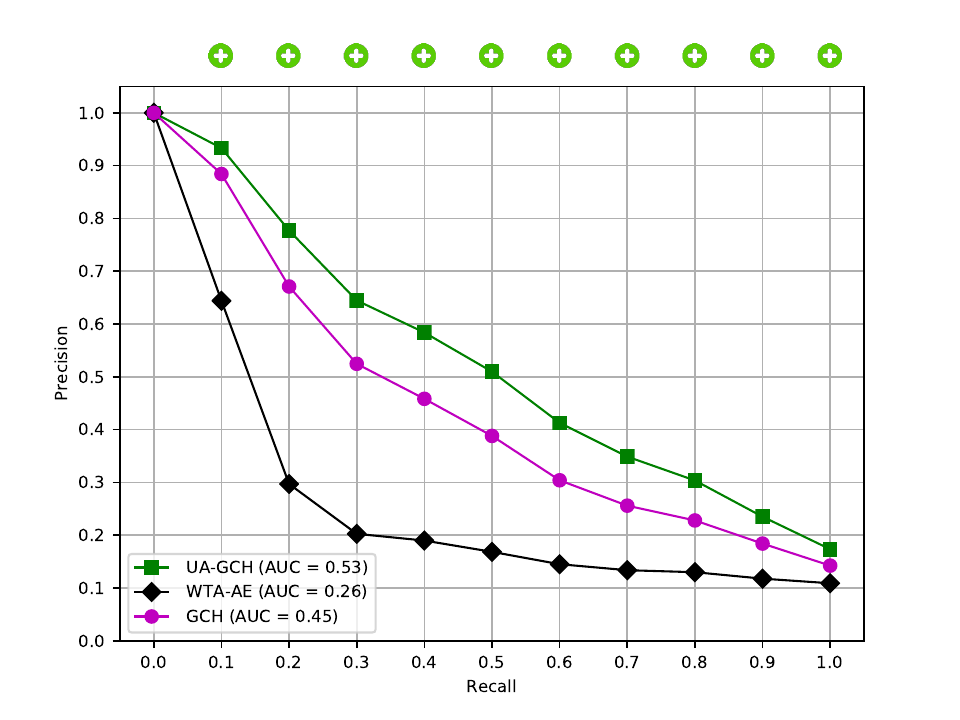}\label{chart:pr_gch_nla_coffe}}
	\subfloat[Coil-100]{\includegraphics[width=0.5\textwidth]{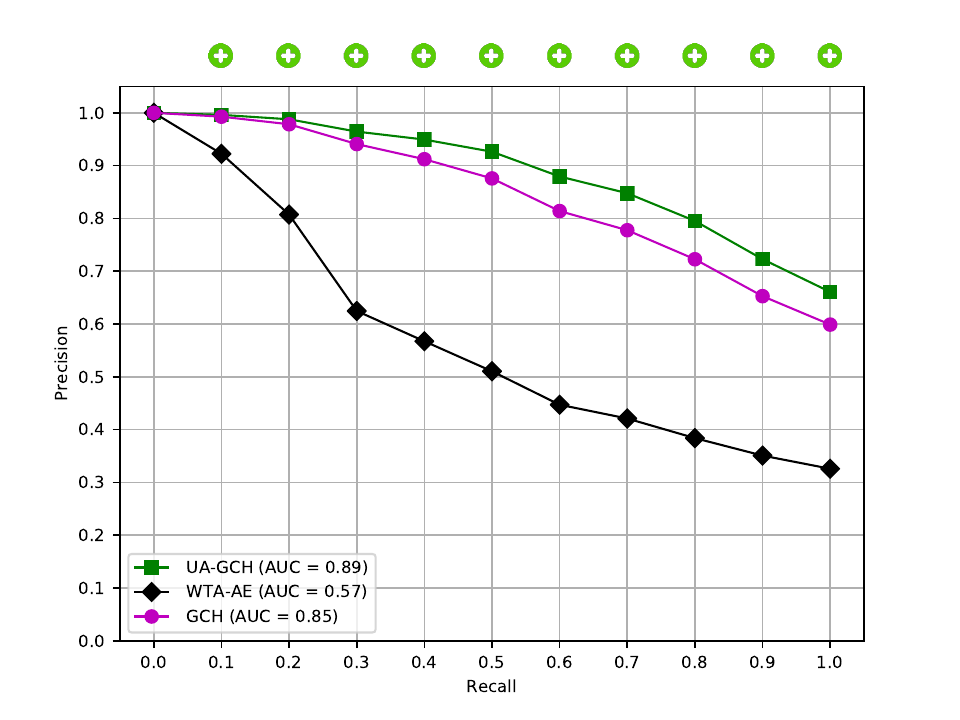}\label{chart:pr_gch_nla_coil100}}\\
	\subfloat[Corel-1566]{\includegraphics[width=0.5\textwidth]{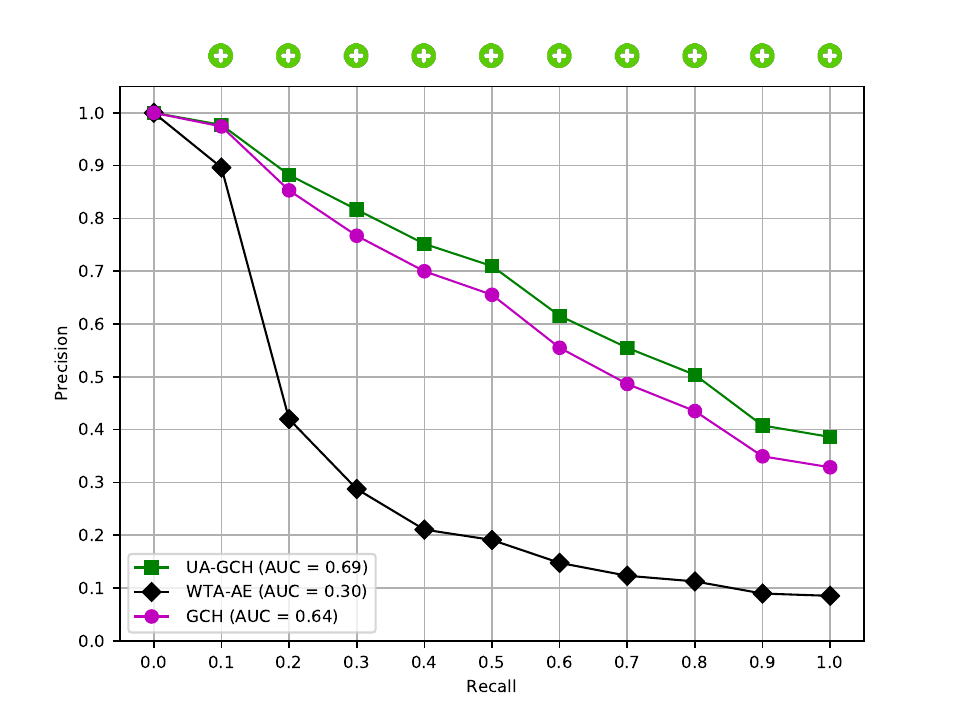}\label{chart:pr_gch_nla_corel1566}}
	\subfloat[Corel-3906]{\includegraphics[width=0.5\textwidth]{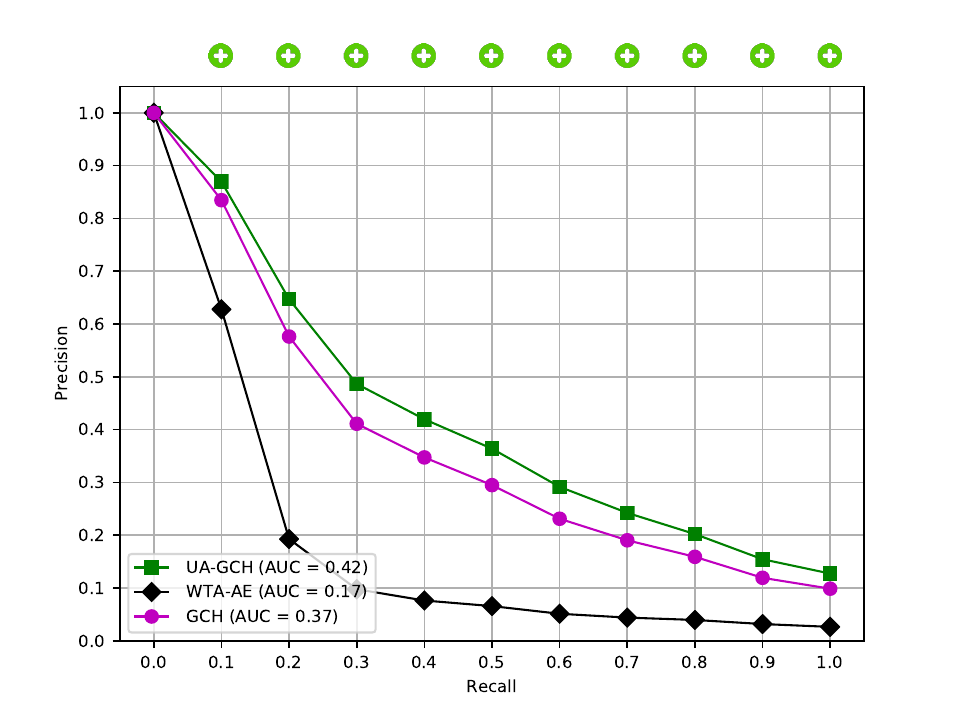}\label{chart:pr_gch_nla_corel3909}}\\
	\subfloat[ETH-80]{\includegraphics[width=0.5\textwidth]{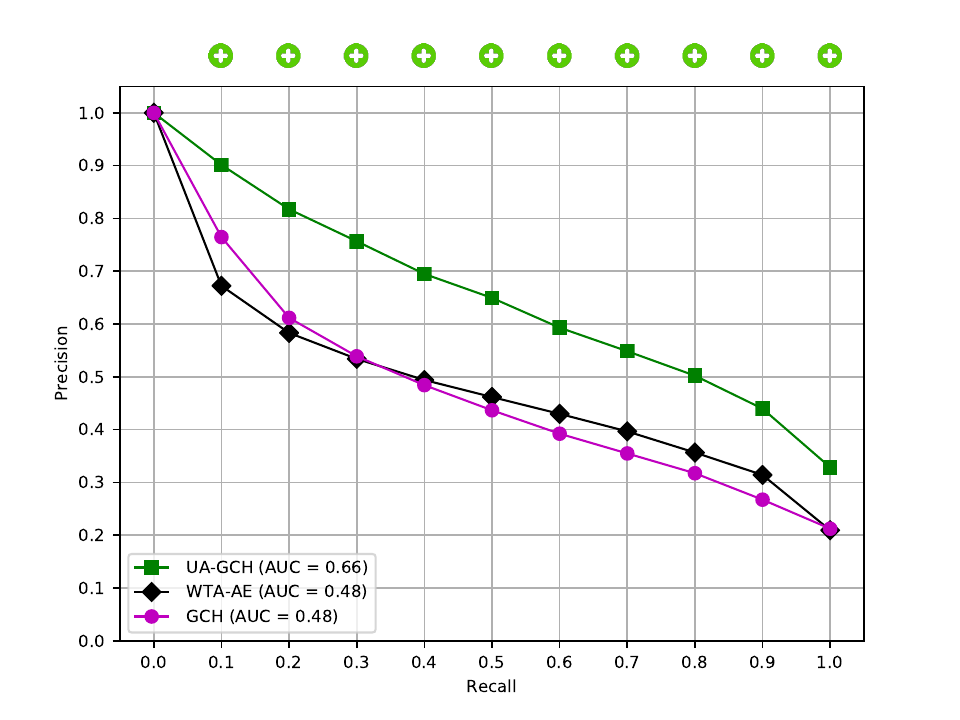}\label{chart:pr_gch_nla_eth80}}
	\subfloat[Supermarket P.]{\includegraphics[width=0.5\textwidth]{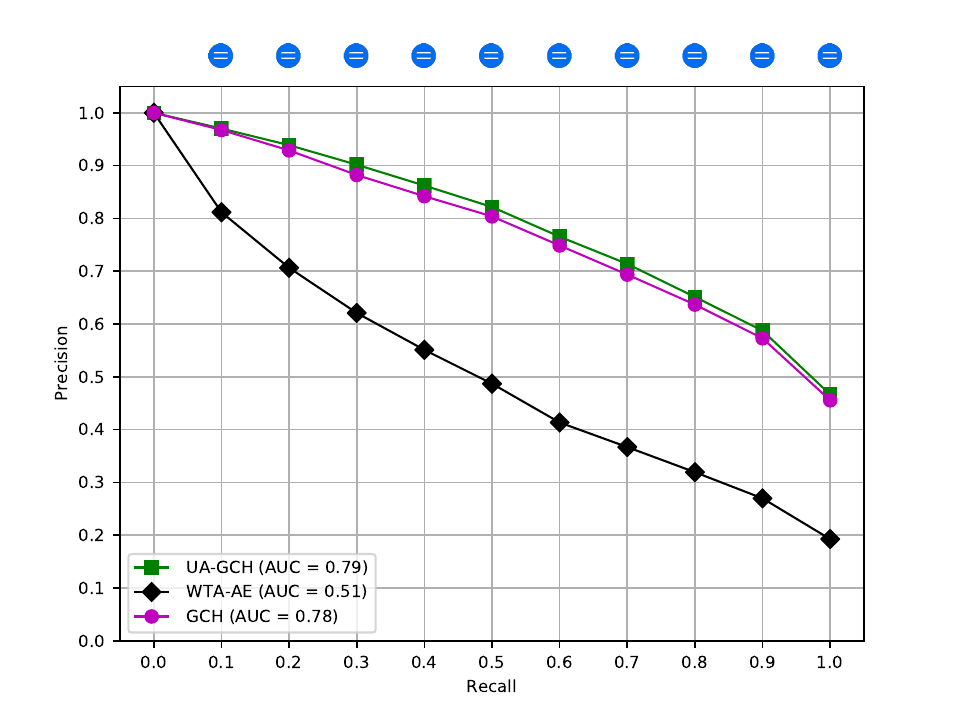}\label{chart:pr_gch_nla_fruits}}\\
	\subfloat[MSRCORID]{\includegraphics[width=0.5\textwidth]{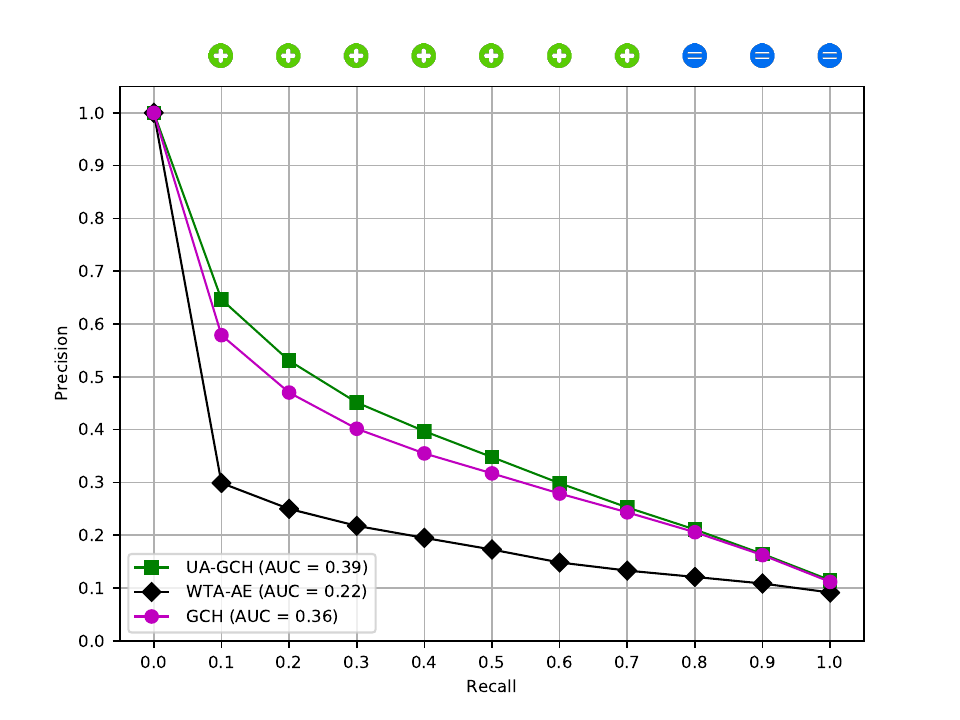}\label{chart:pr_gch_nla_msrcorid}}
	\subfloat[UCMerced Land-use]{\includegraphics[width=0.5\textwidth]{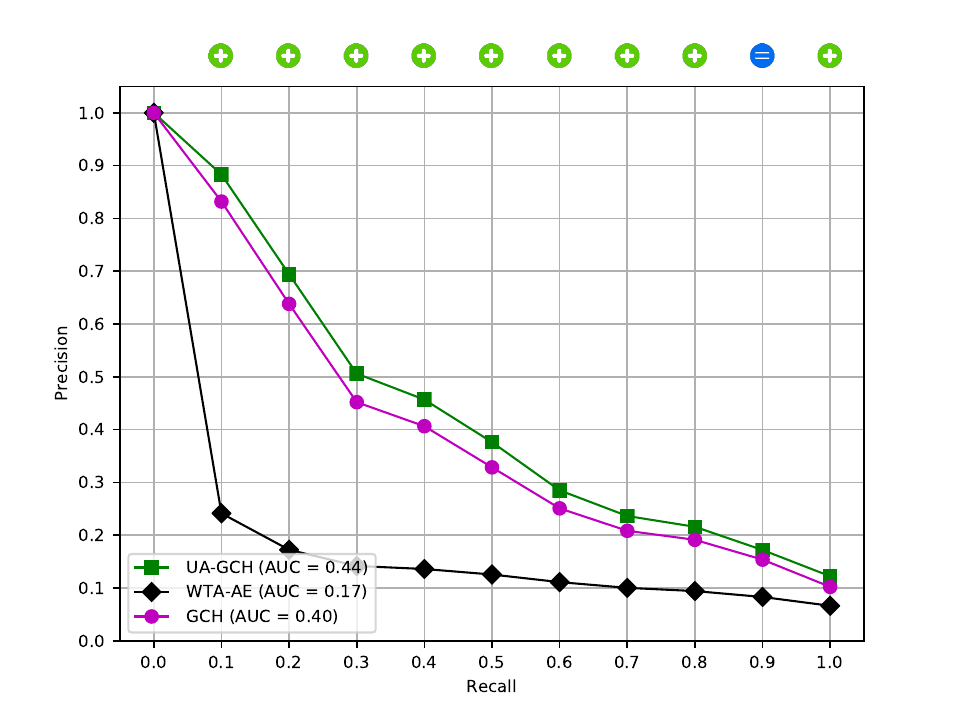}\label{chart:pr_gch_nla_ucmerced}}
	\caption{Comparison between the Precision-Recall curves of UA, WTA Autoencoder and GCH feature extractor.}
	\label{chart:pr_gch_nla}
\end{figure}

\begin{figure}[!t]
        \centering
        \subfloat[P@10]{\includegraphics[width=0.5\textwidth]{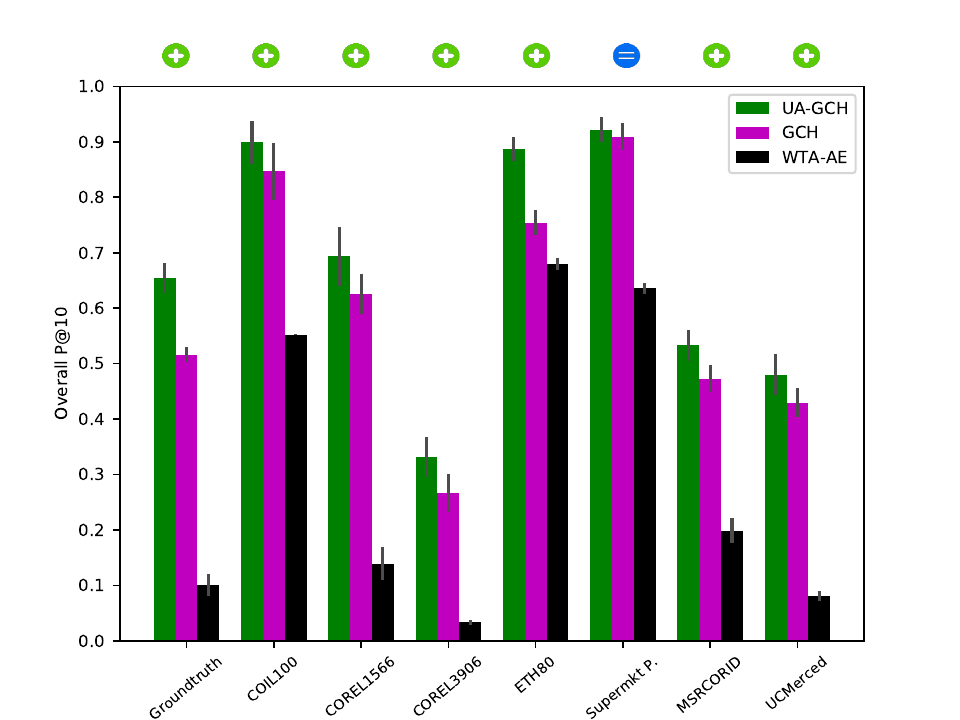}\label{chart:precision_gch_nla}}
		\subfloat[MAP]{\includegraphics[width=0.5\textwidth]{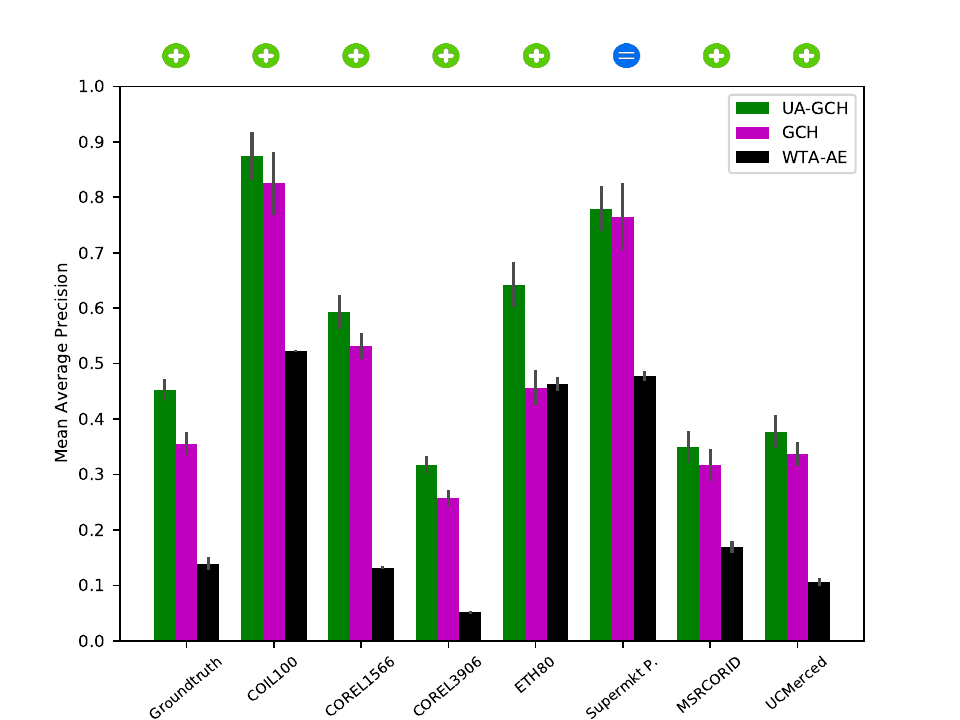}\label{chart:map_gch_nla}}
		\caption{Comparison between the (a) P@10 and (b) MAP results of UA, WTA Autoencoder and the {\bf GCH} feature extractor.}
	\end{figure}

Similar results were observed when the GCH feature extractor is considered. Figs.~\ref{chart:pr_gch_nla},~\ref{chart:precision_gch_nla}, and~\ref{chart:map_gch_nla} provide the effectiveness results in terms of Precision-Recall, MAP, and P@10, respectively. For all datasets, but for the \textit{Supermarket Produce}, the proposed UA approach yielded better results \edit{than} those of the baseline. For the \textit{Supermarket Produce} dataset, no statistical difference was observed.

\begin{figure}[!t]
	\centering
		\subfloat[BIC]{\includegraphics[width=0.5\textwidth]{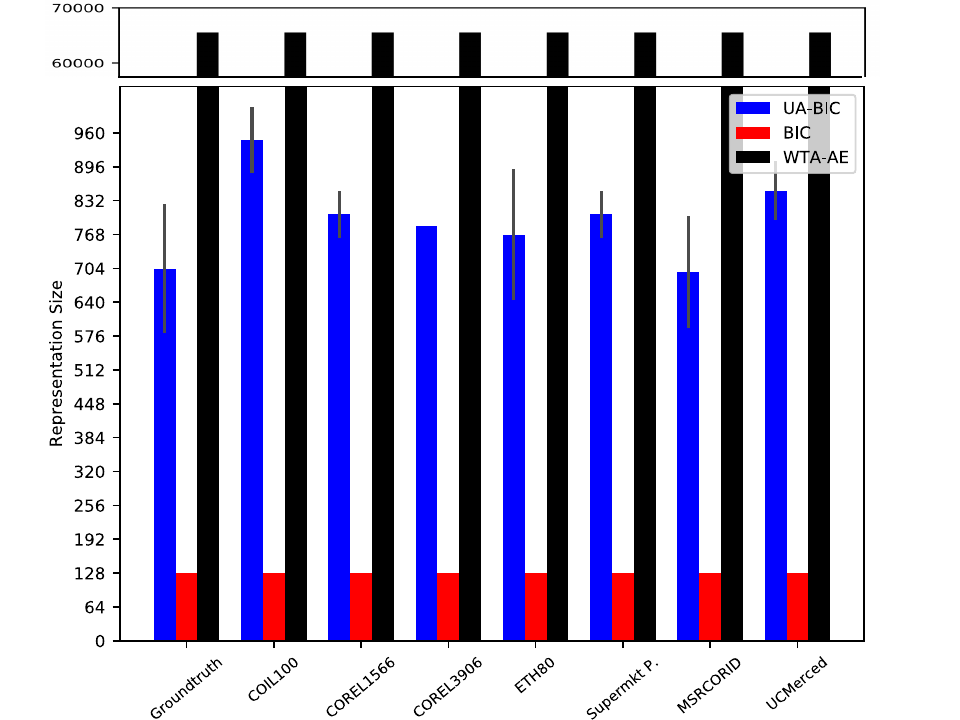}\label{chart:size_bic_nla}}
		\subfloat[GCH]{\includegraphics[width=0.5\textwidth]{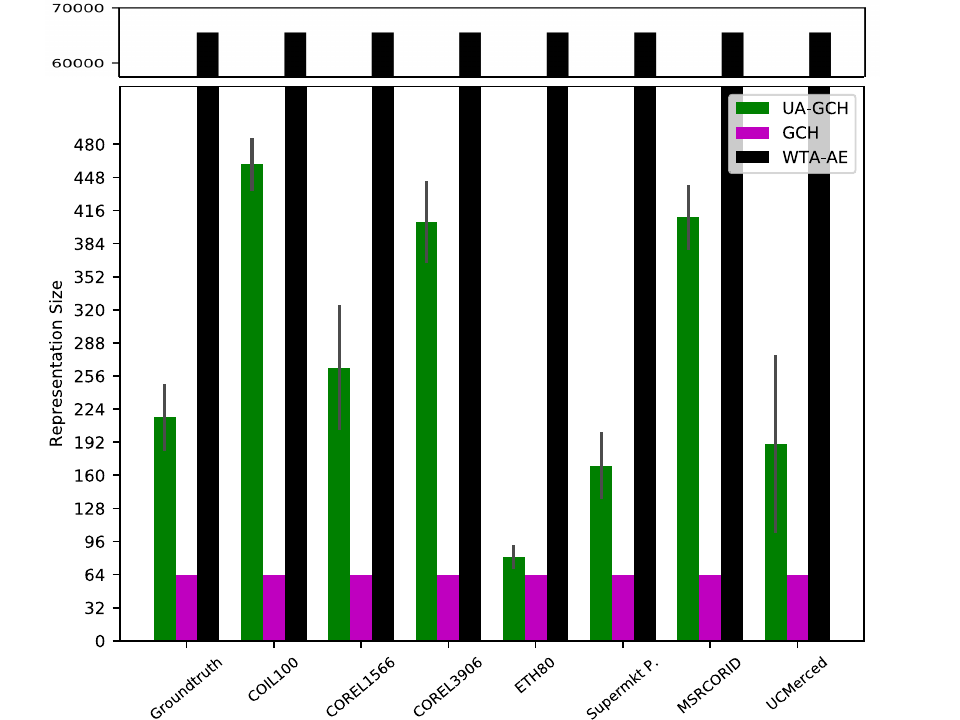}\label{chart:size_gch_nla}}
		\caption{Comparison between the representation size results of UA, WTA Autoencoder and the feature extractors: (a) BIC and (b) GCH. The upper windows show a cut of the highest columns while the lower windows show a view of the bottom.}
		\label{chart:size_nla}
	\end{figure}

With regard to the representation sizes (Fig.~\ref{chart:size_nla}), the differences between the proposed methods and the baselines are very high. Comparing to the feature extractors, the representations produced by our method approach were, on average, \edit{around 521\% larger for BIC (Fig.~\ref{chart:size_bic_nla}) and around 328\% larger for GCH (Fig.~\ref{chart:size_gch_nla})}. A possible reason relies on the fact that the fitness function used for evaluating the genetic algorithm individuals prioritizes the representation effectiveness performance on the retrieval task, i.e., the optimization process is not guided to guarantee compact representations.  In this scenario, the proposed method quantized more regions in the color space, leading to representation with higher dimensions. The Size-Constrained Approach (SCA), whose results are discussed next, addresses this issue.

\subsection{Size-Constrained Approach}
\label{ssec:limited}

In the evaluation of the SCA approach, we varied the number of bins in the ranges $\{16, 32, 64, 96, 128, 256, 384\}$ and $\{8, 16, 32, 48, 64, 128, 192\}$ for BIC and GCH approaches, respectively.
\edit{These ranges were defined based on a logarithmic sequence of proportions 
(12,5\%, 25\%, 50\%, 100\%, 200\%) of the baselines vector sizes and some additional points among them (75\% and 300\%) to provide a clearer view of the performances behaviour.}
Figs.~\ref{chart:pr_bic_la1} and \ref{chart:pr_bic_la2} present the Precision-Recall curves for the BIC-based approaches and WTA-AE for all datasets, considering these different feature vector sizes.
We can observe that the proposed method yielded comparable or better results than those observed for the baselines for feature vectors whose size is higher than 96 for the majority of the datasets. In fact, the smaller the feature vector size, the worse the results of SCA when compared to the baselines. Similar results were observed for the GCH-based approaches at Figs.~\ref{chart:pr_gch_la1} and \ref{chart:pr_gch_la2}.

\begin{figure}[h]
	\centering
	\subfloat[Groundtruth]{\includegraphics[width=0.5\textwidth]{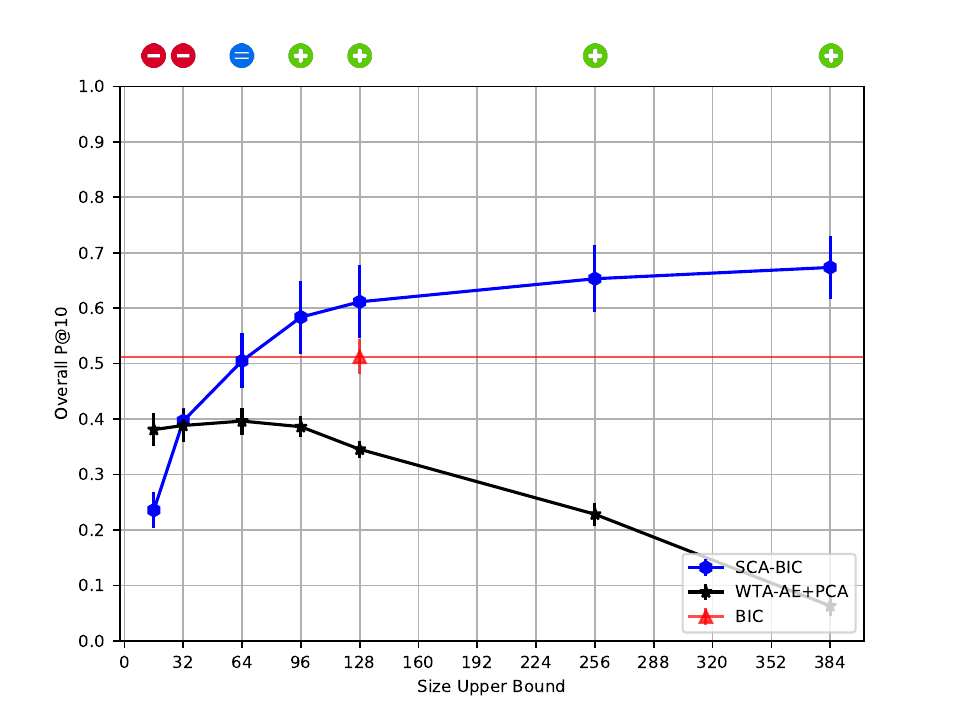}\label{chart:precision_bic_la_coffe}}
	\subfloat[Coil-100]{\includegraphics[width=0.5\textwidth]{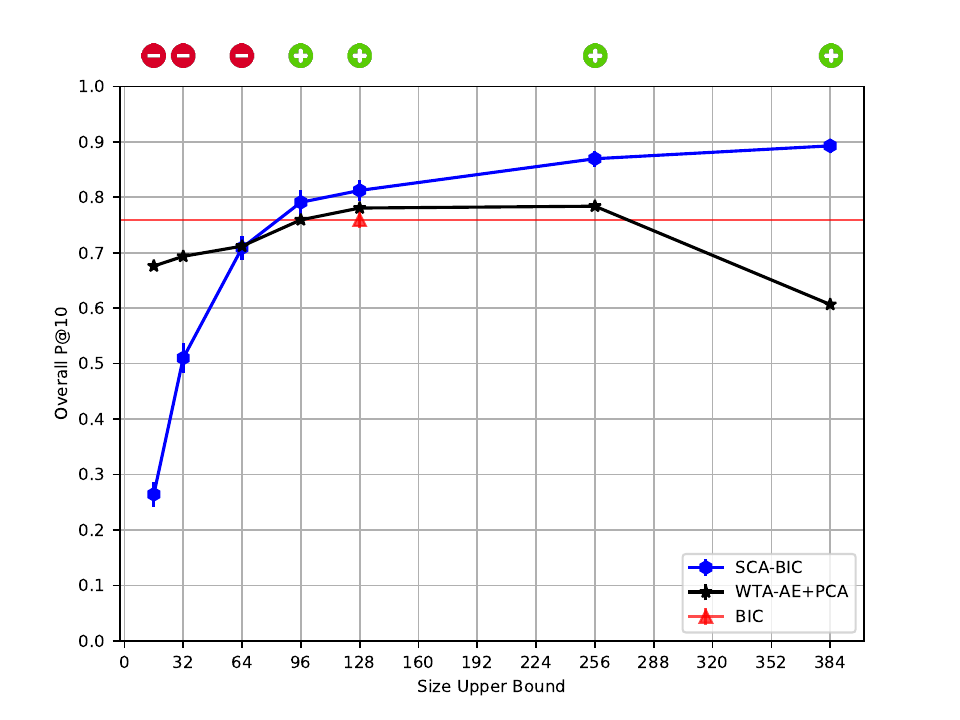}\label{chart:precision_bic_la_coil100}}\\
	\subfloat[Corel-1566]{\includegraphics[width=0.5\textwidth]{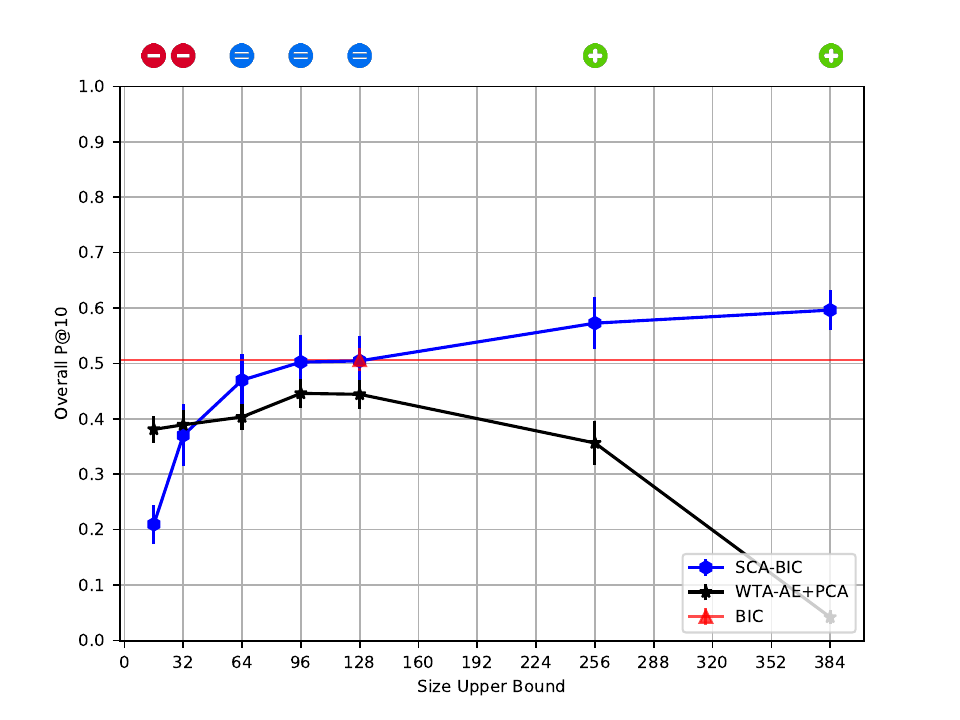}\label{chart:precision_bic_la_corel1566}}
	\subfloat[Corel-3906]{\includegraphics[width=0.5\textwidth]{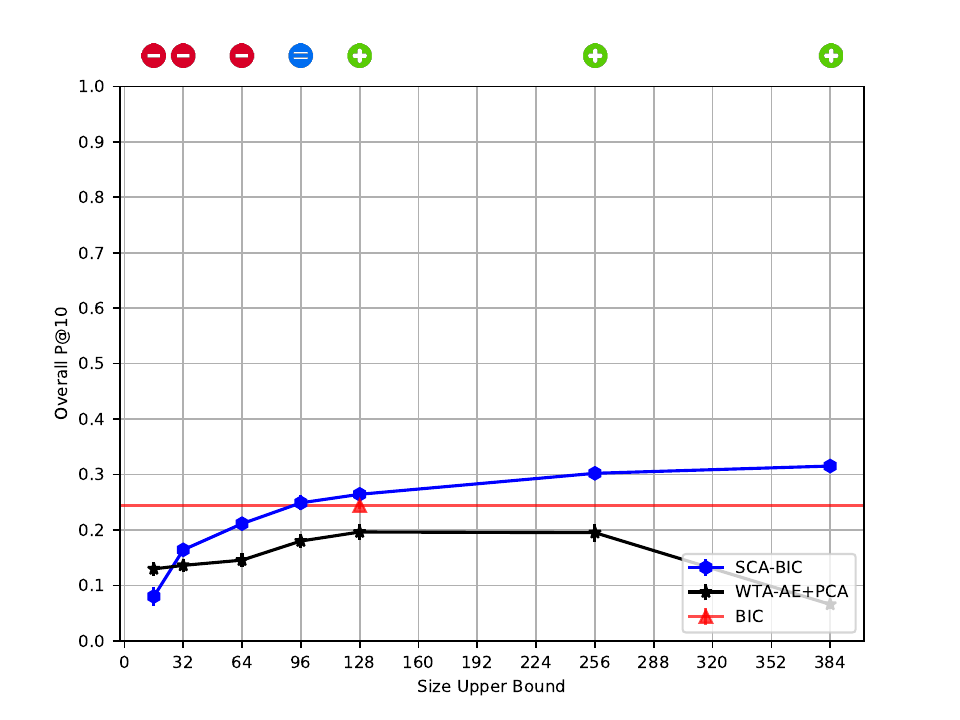}\label{chart:precision_bic_la_corel3909}}\\
	\subfloat[ETH-80]{\includegraphics[width=0.5\textwidth]{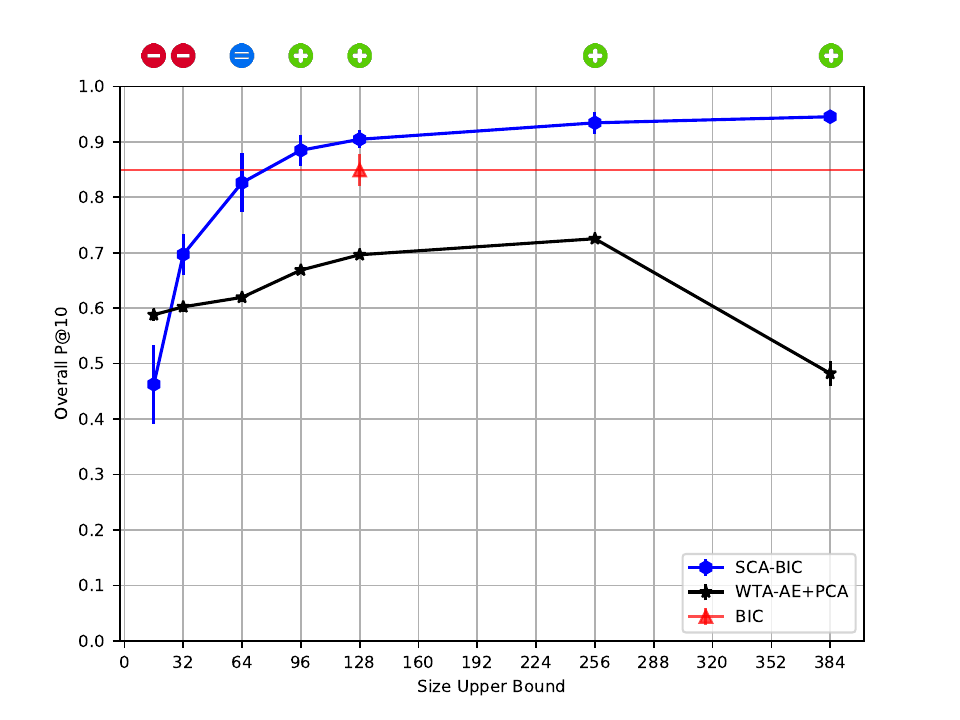}\label{chart:precision_bic_la_eth80}}
	\subfloat[Supermarket Produce]{\includegraphics[width=0.5\textwidth]{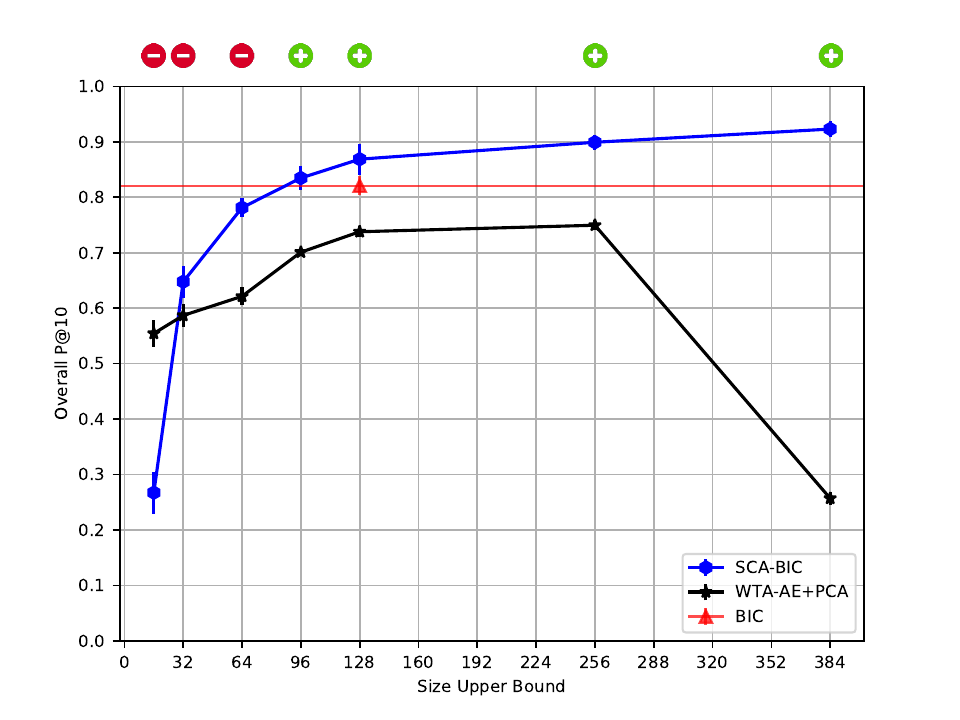}\label{chart:precision_bic_la_fruits}}\\
	\subfloat[MSRCORID]{\includegraphics[width=0.5\textwidth]{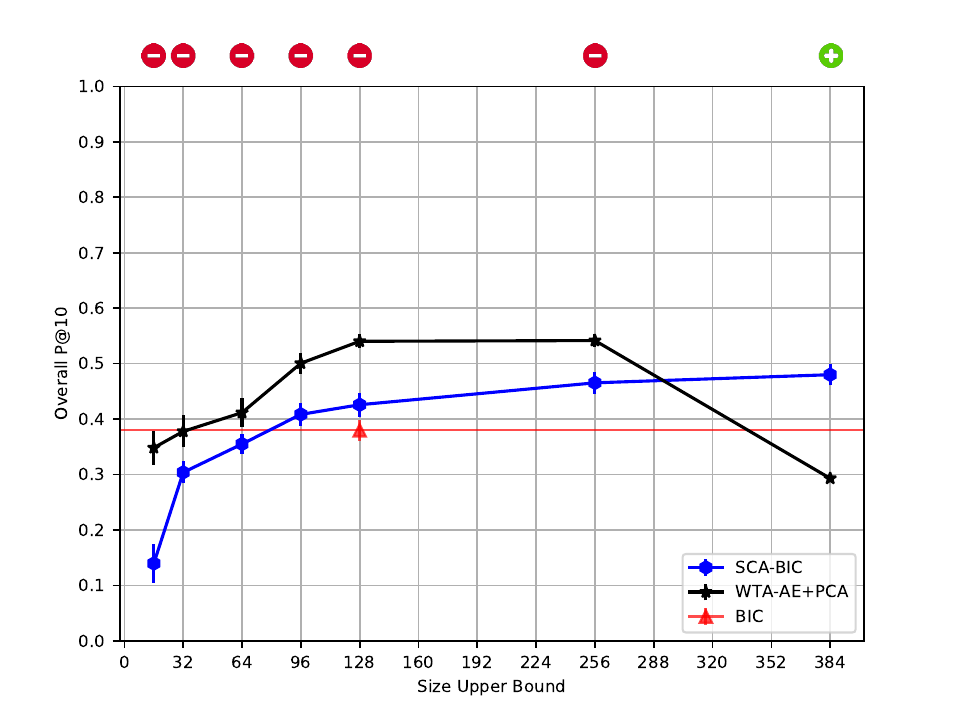}\label{chart:precision_bic_la_msrcorid}}
	\subfloat[UCMerced Land-use]{\includegraphics[width=0.5\textwidth]{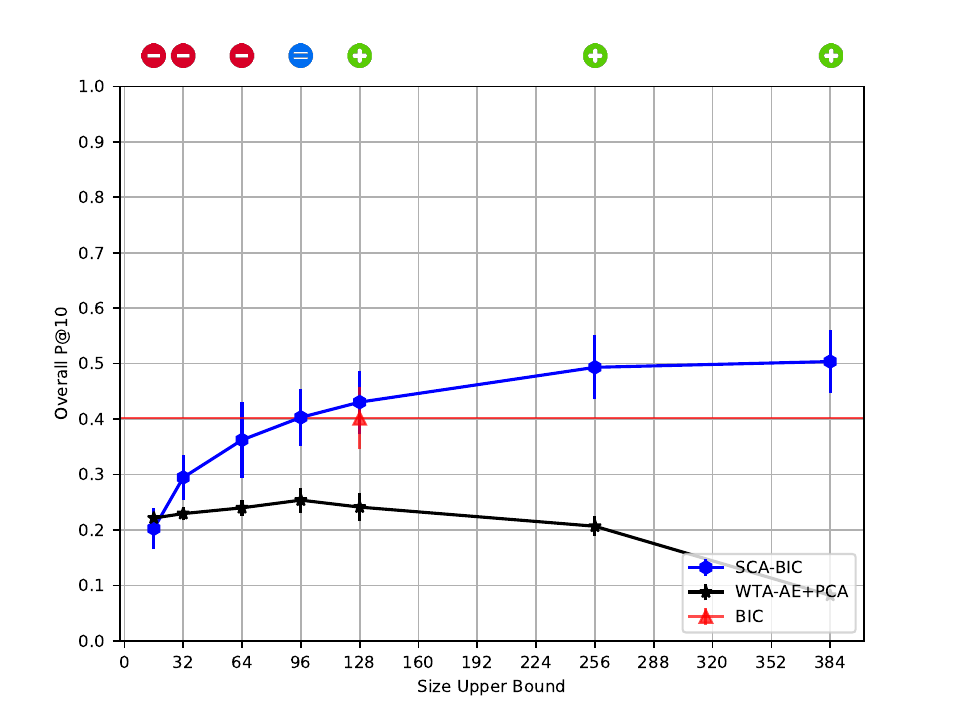}\label{chart:precision_bic_la_ucmerced}}
	\caption{Comparison between the P@10 results of SCA, WTA Autoencoder and BIC feature extractor}
	\label{chart:precision_bic_la}
\end{figure}

\begin{figure}[h]
	\centering
	\subfloat[Groundtruth]{\includegraphics[width=0.5\textwidth]{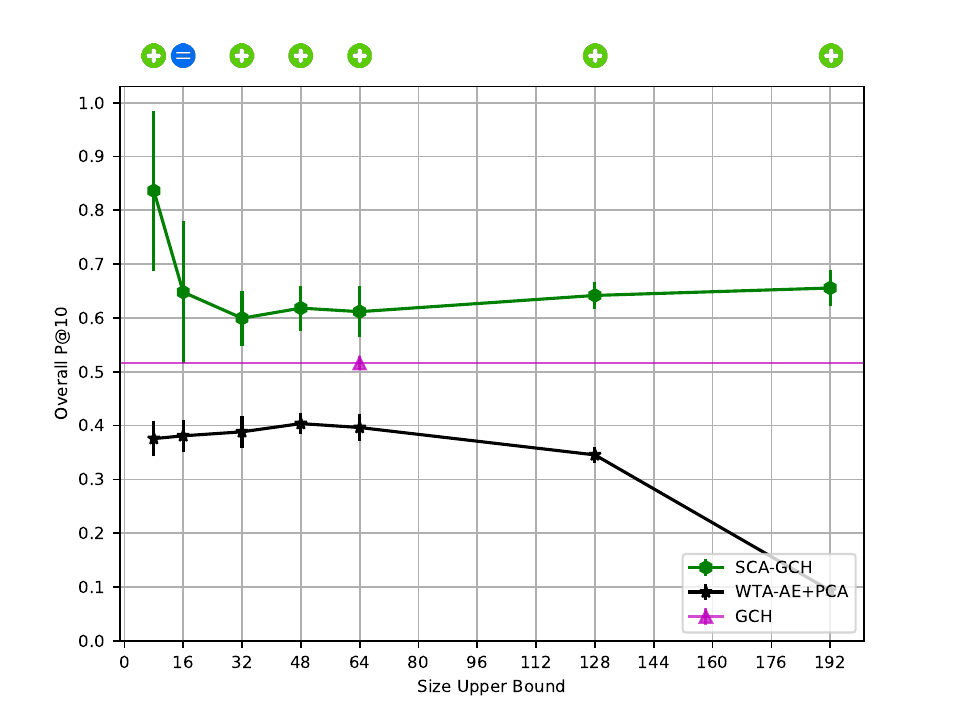}\label{chart:precision_gch_la_coffe}}
	\subfloat[Coil-100]{\includegraphics[width=0.5\textwidth]{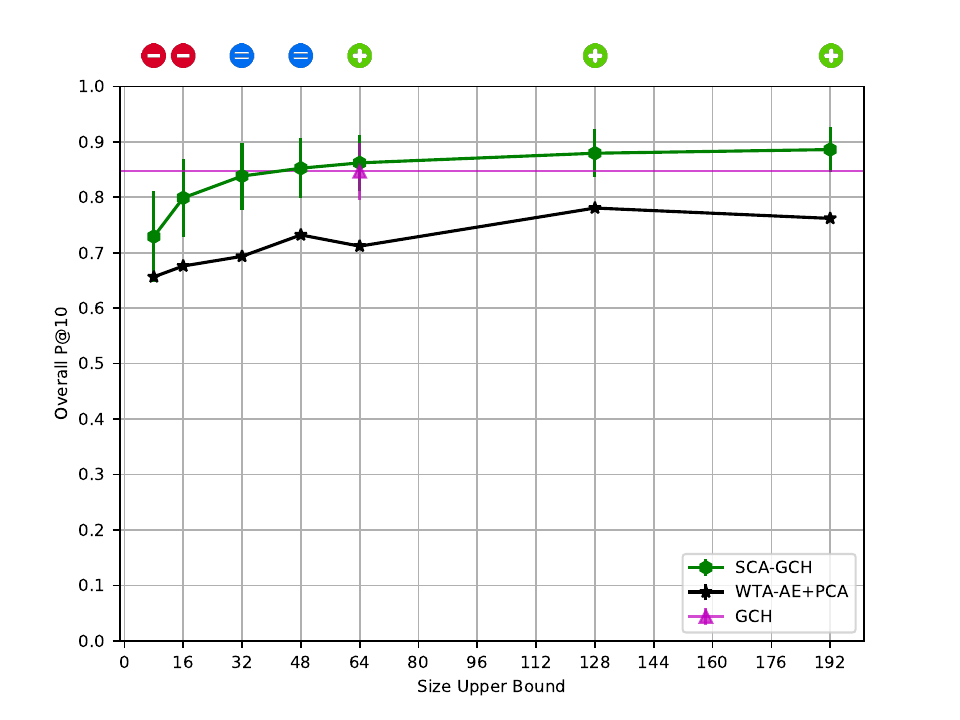}\label{chart:precision_gch_la_coil100}}\\
	\subfloat[Corel-1566]{\includegraphics[width=0.5\textwidth]{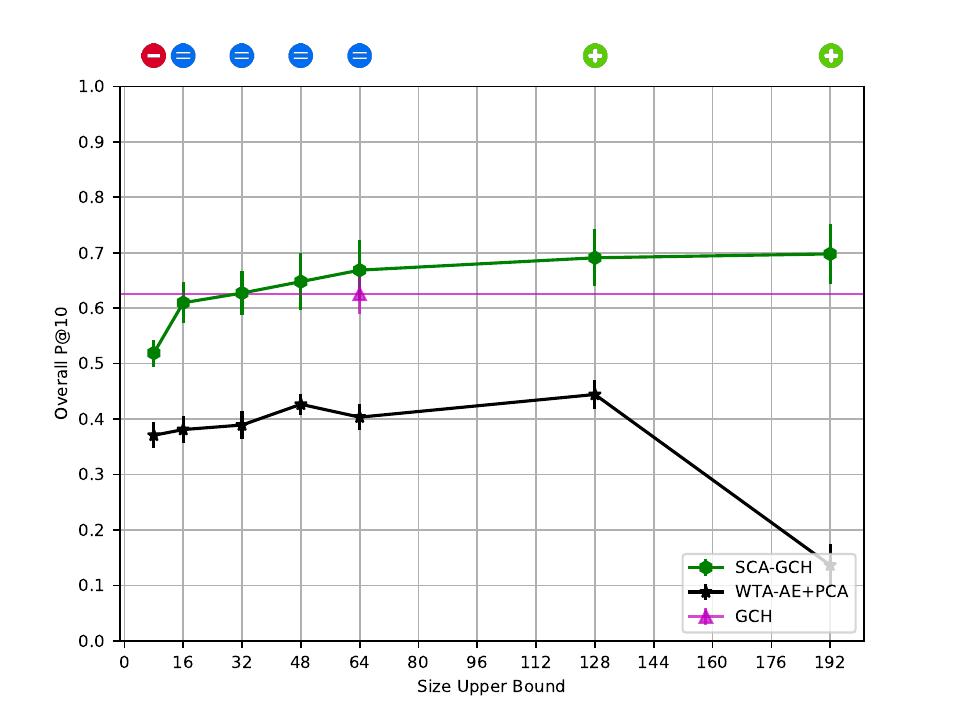}\label{chart:precision_gch_la_corel1566}}
	\subfloat[Corel-3906]{\includegraphics[width=0.5\textwidth]{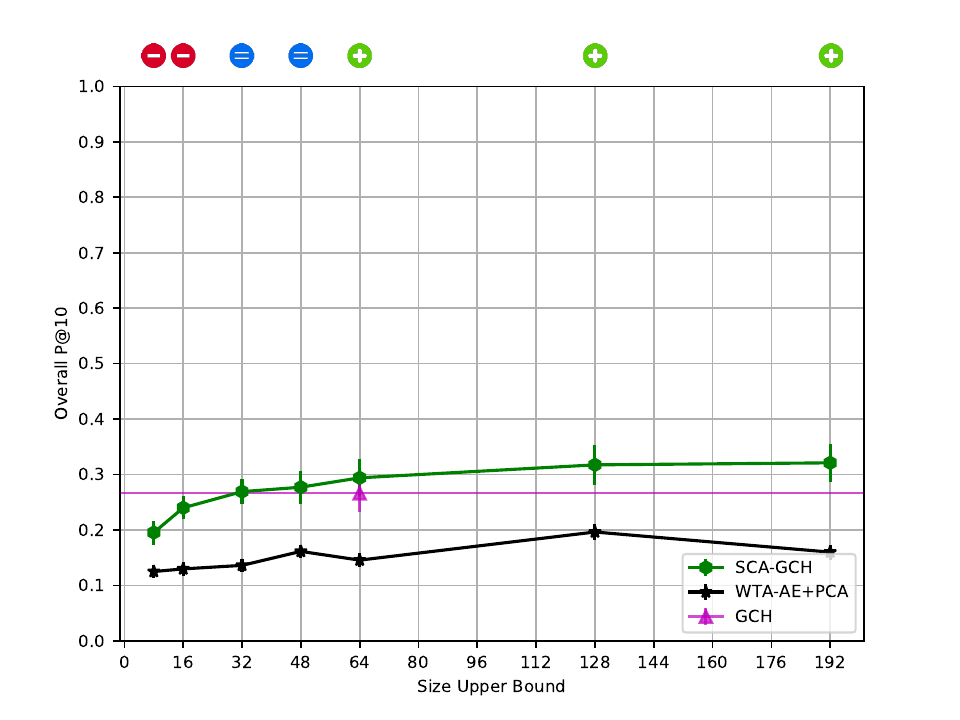}\label{chart:precision_gch_la_corel3909}}\\
	\subfloat[ETH-80]{\includegraphics[width=0.5\textwidth]{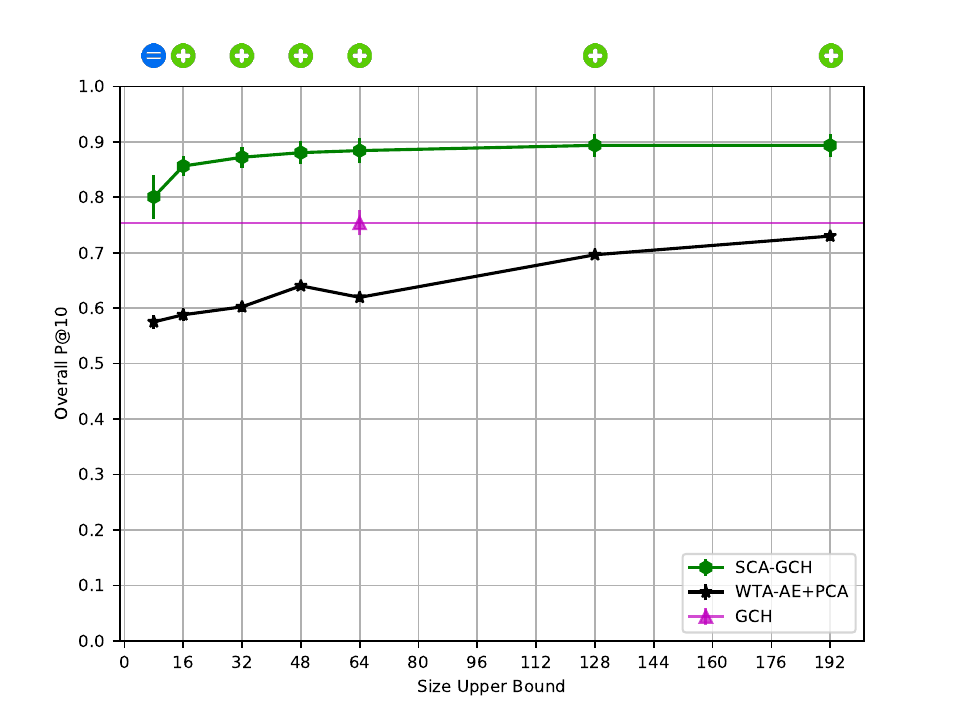}\label{chart:precision_gch_la_eth80}}
	\subfloat[Supermarket Produce]{\includegraphics[width=0.5\textwidth]{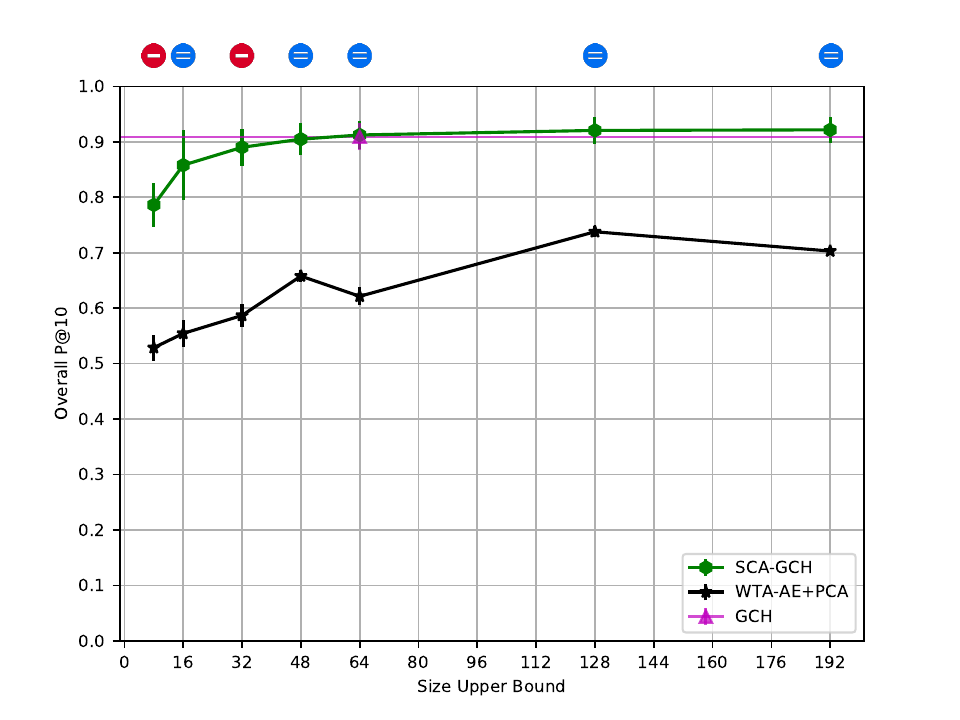}\label{chart:precision_gch_la_fruits}}\\
	\subfloat[MSRCORID]{\includegraphics[width=0.5\textwidth]{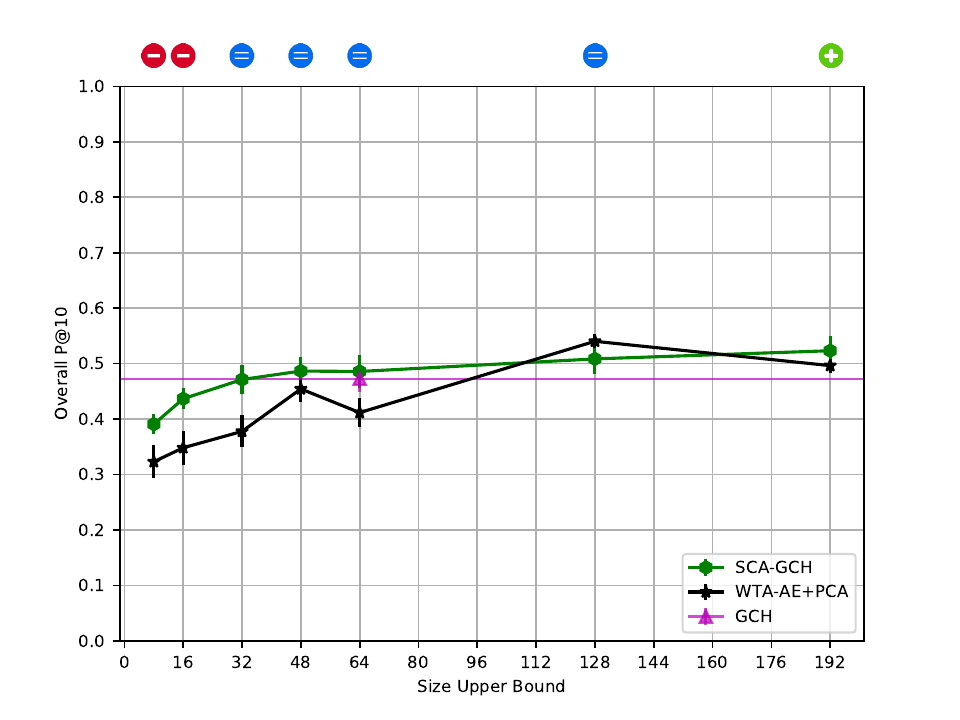}\label{chart:precision_gch_la_msrcorid}}
	\subfloat[UCMerced Land-use]{\includegraphics[width=0.5\textwidth]{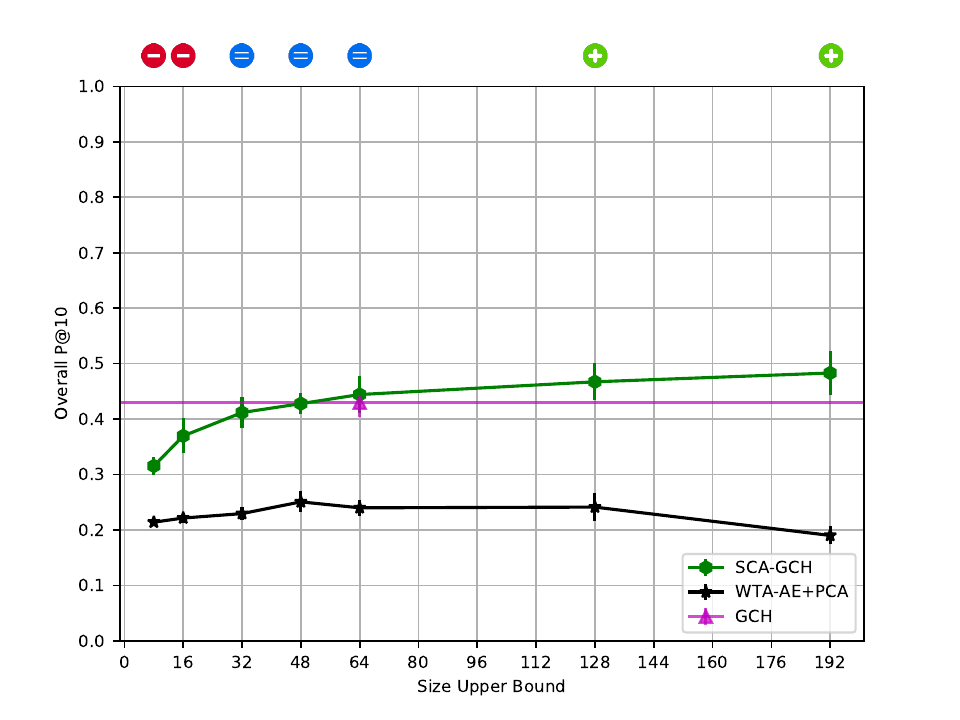}\label{chart:precision_gch_la_ucmerced}}
	\caption{Comparison between the P@10 results of SCA, WTA Autoencoder and GCH feature extractor}
	\label{chart:precision_gch_la}
\end{figure}

\begin{figure}[h]
	\centering
	\subfloat[Groundtruth]{\includegraphics[width=0.5\textwidth]{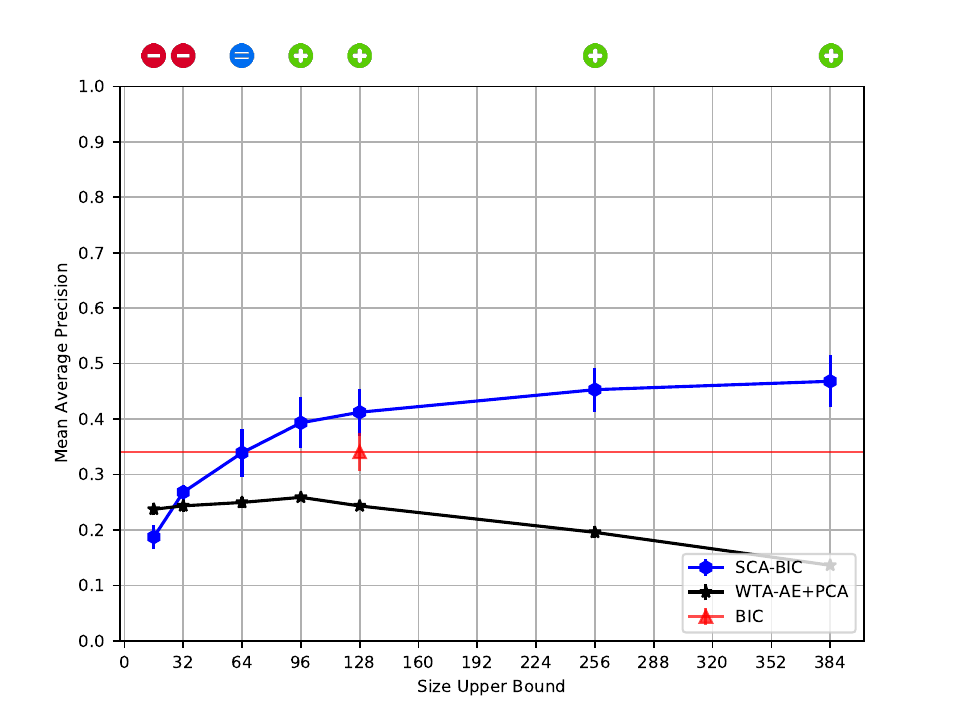}\label{chart:map_bic_la_coffe}}
	\subfloat[Coil-100]{\includegraphics[width=0.5\textwidth]{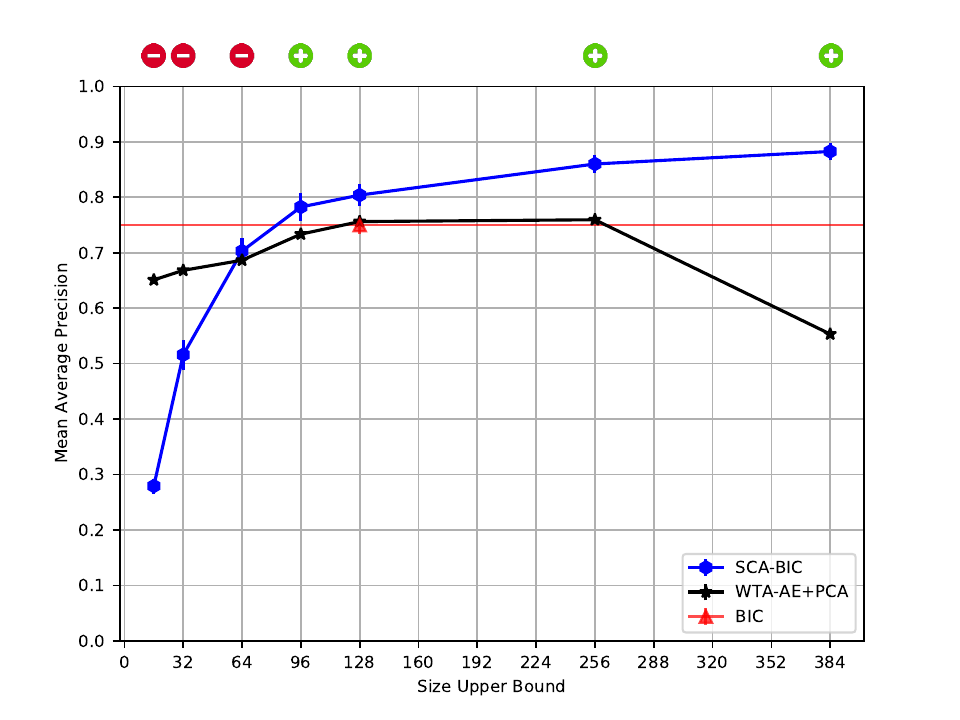}\label{chart:map_bic_la_coil100}}\\
	\subfloat[Corel-1566]{\includegraphics[width=0.5\textwidth]{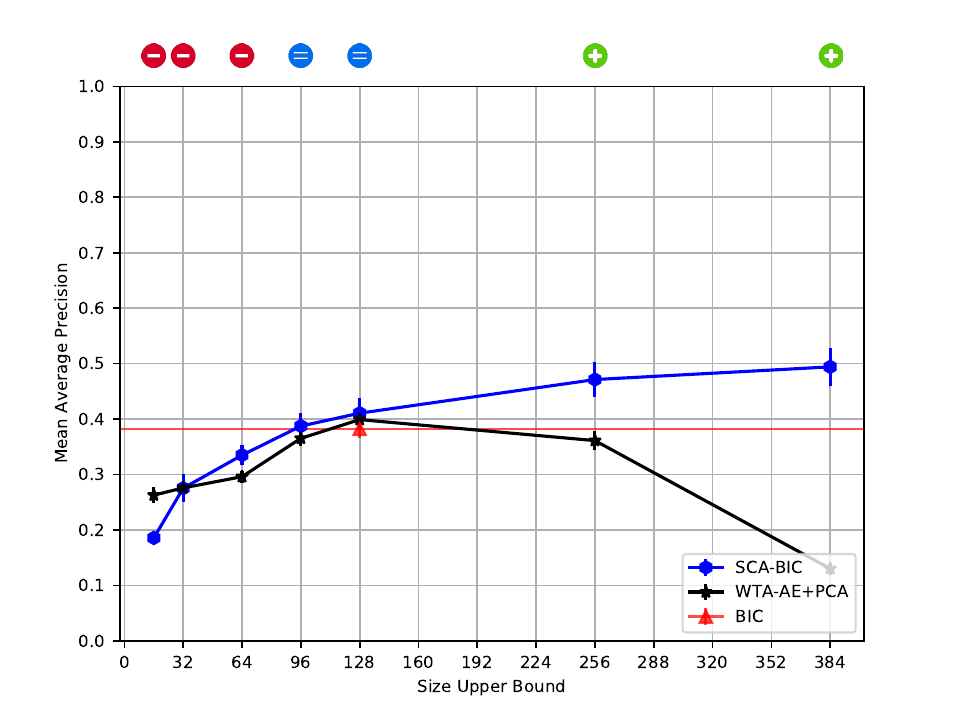}\label{chart:map_bic_la_corel1566}}
	\subfloat[Corel-3906]{\includegraphics[width=0.5\textwidth]{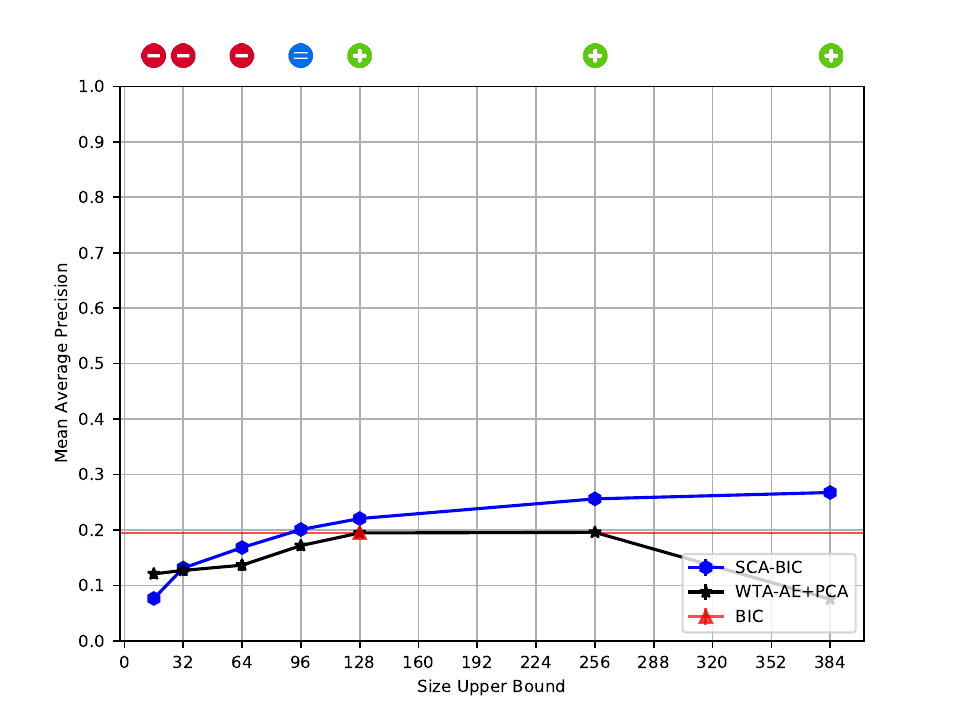}\label{chart:map_bic_la_corel3909}}\\
	\subfloat[ETH-80]{\includegraphics[width=0.5\textwidth]{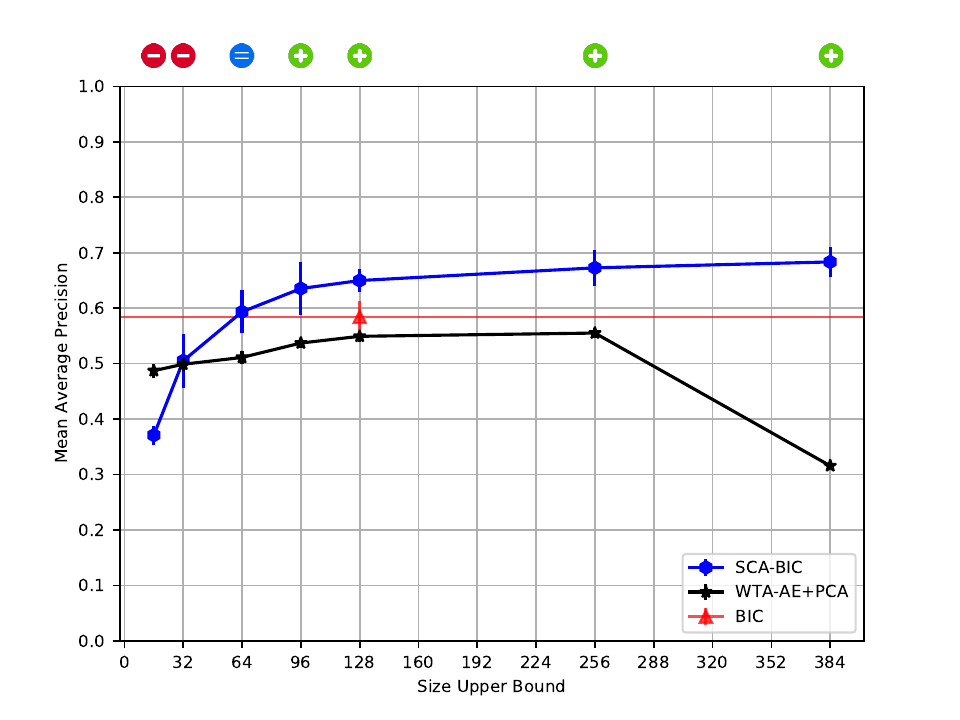}\label{chart:map_bic_la_eth80}}
	\subfloat[Supermarket Produce]{\includegraphics[width=0.5\textwidth]{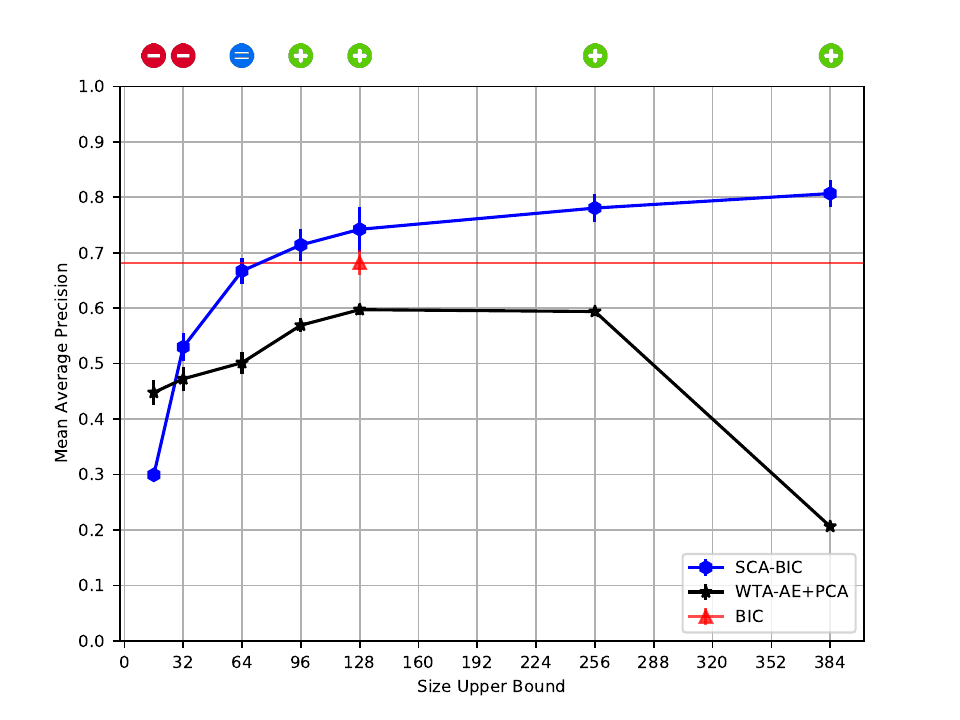}\label{chart:map_bic_la_fruits}}\\
	\subfloat[MSRCORID]{\includegraphics[width=0.5\textwidth]{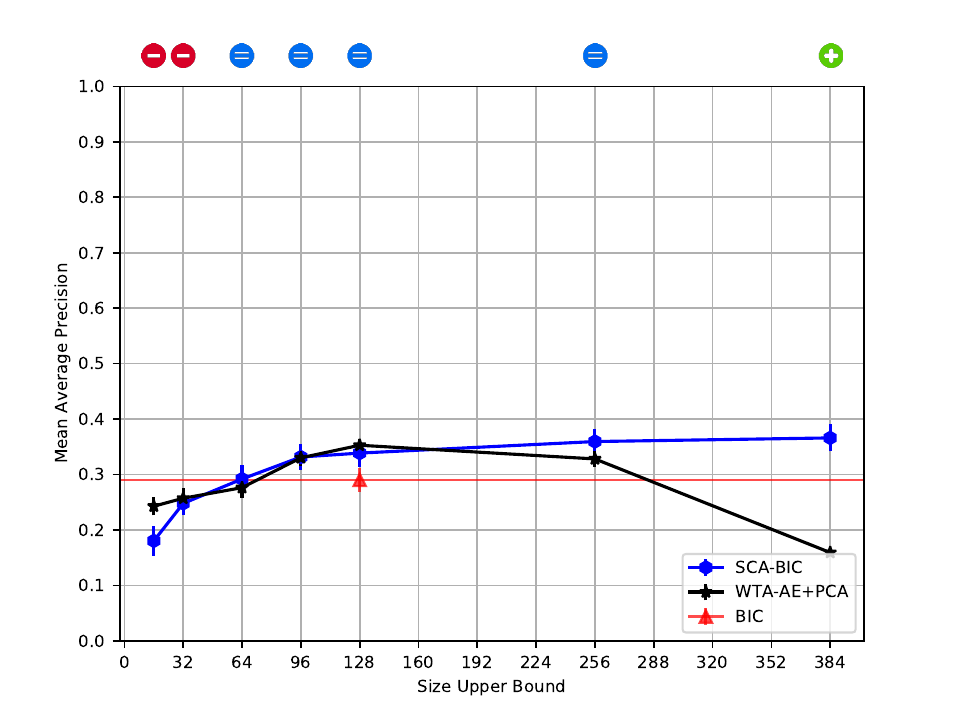}\label{chart:map_bic_la_msrcorid}}
	\subfloat[UCMerced Land-use]{\includegraphics[width=0.5\textwidth]{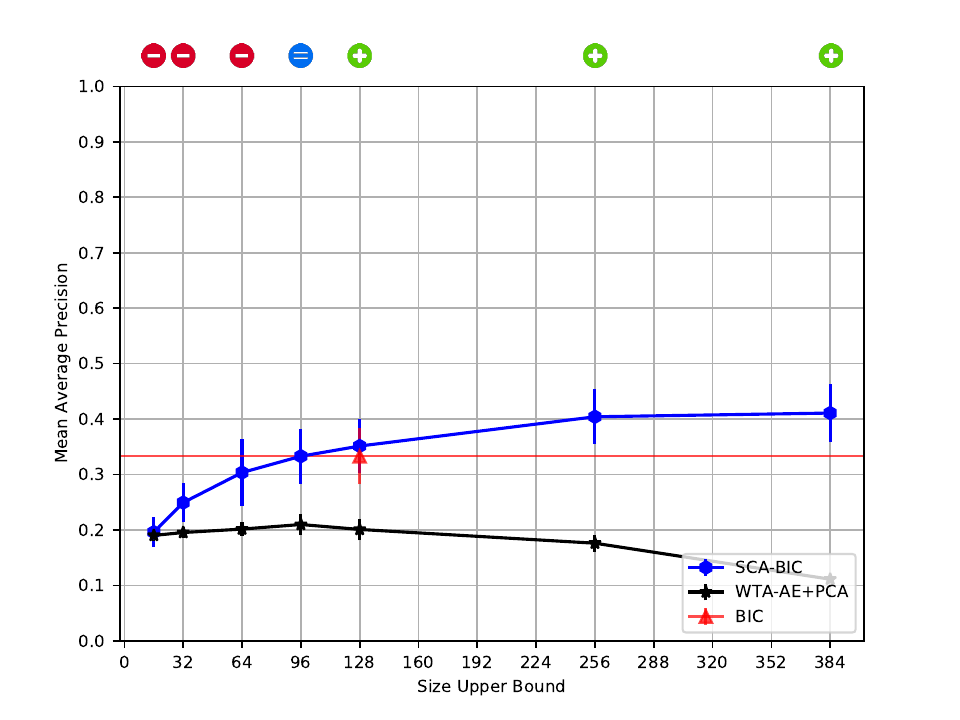}\label{chart:map_bic_la_ucmerced}}
	\caption{Comparison between the MAP results of SCA, WTA Autoencoder and BIC feature extractor}
	\label{chart:map_bic_la}
\end{figure}

\begin{figure}[h]
	\centering
	\subfloat[Groundtruth]{\includegraphics[width=0.5\textwidth]{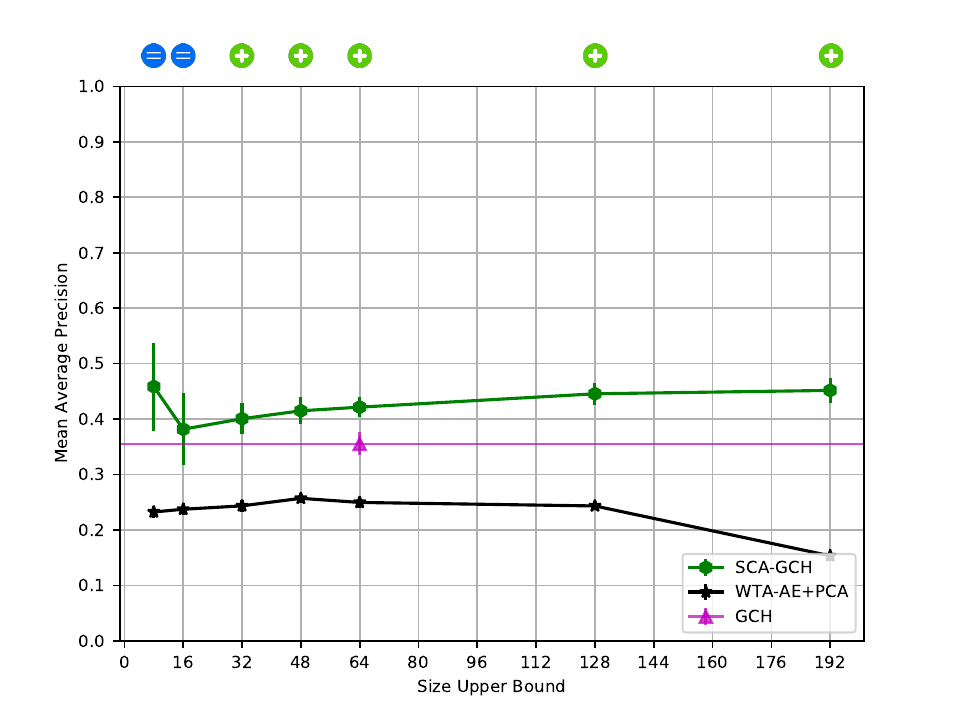}\label{chart:map_gch_la_coffe}}
	\subfloat[Coil-100]{\includegraphics[width=0.5\textwidth]{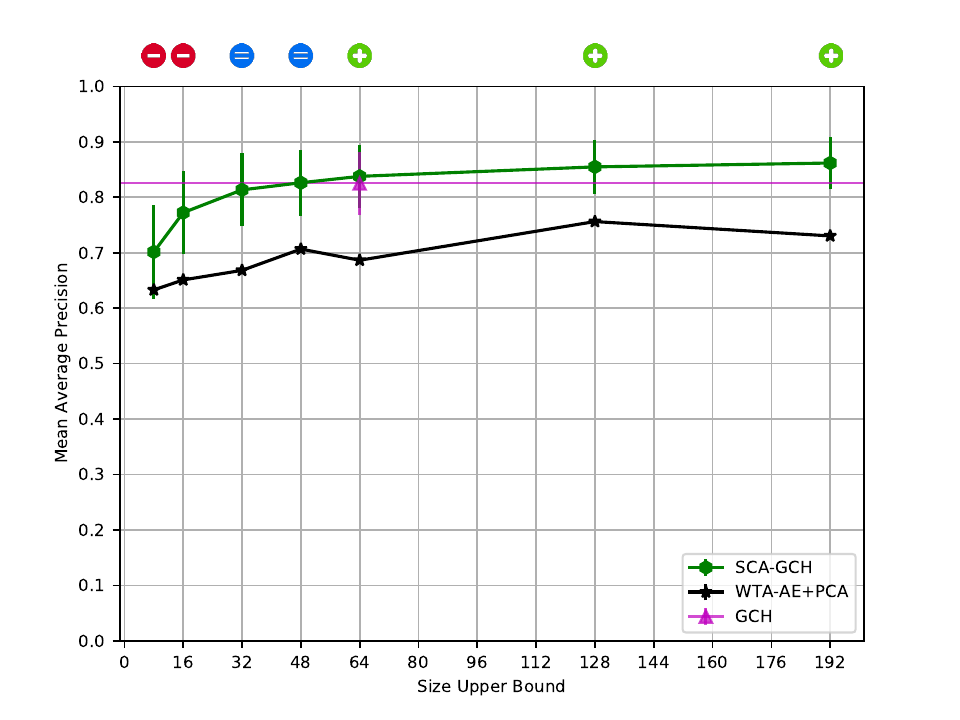}\label{chart:map_gch_la_coil100}}\\
	\subfloat[Corel-1566]{\includegraphics[width=0.5\textwidth]{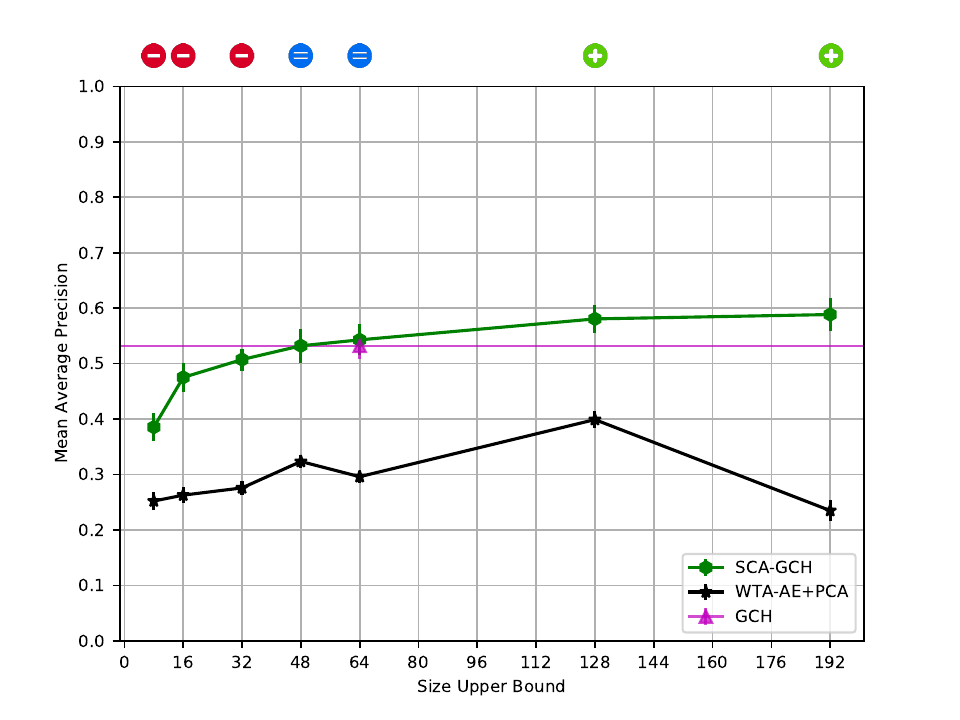}\label{chart:map_gch_la_corel1566}}
	\subfloat[Corel-3906]{\includegraphics[width=0.5\textwidth]{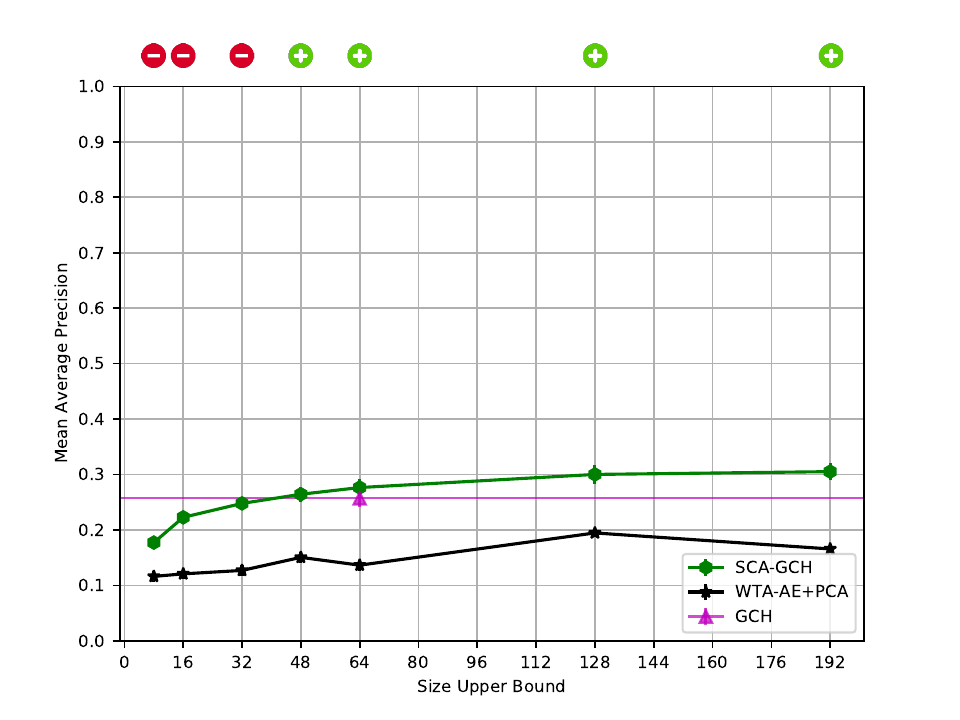}\label{chart:map_gch_la_corel3909}}\\
	\subfloat[ETH-80]{\includegraphics[width=0.5\textwidth]{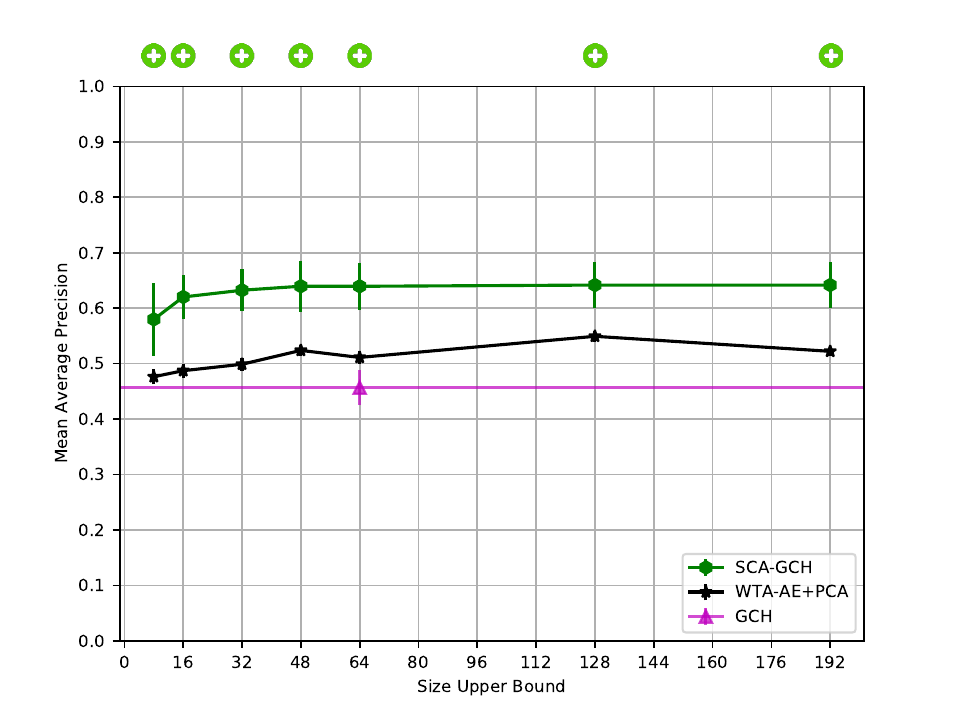}\label{chart:map_gch_la_eth80}}
	\subfloat[Supermarket Produce]{\includegraphics[width=0.5\textwidth]{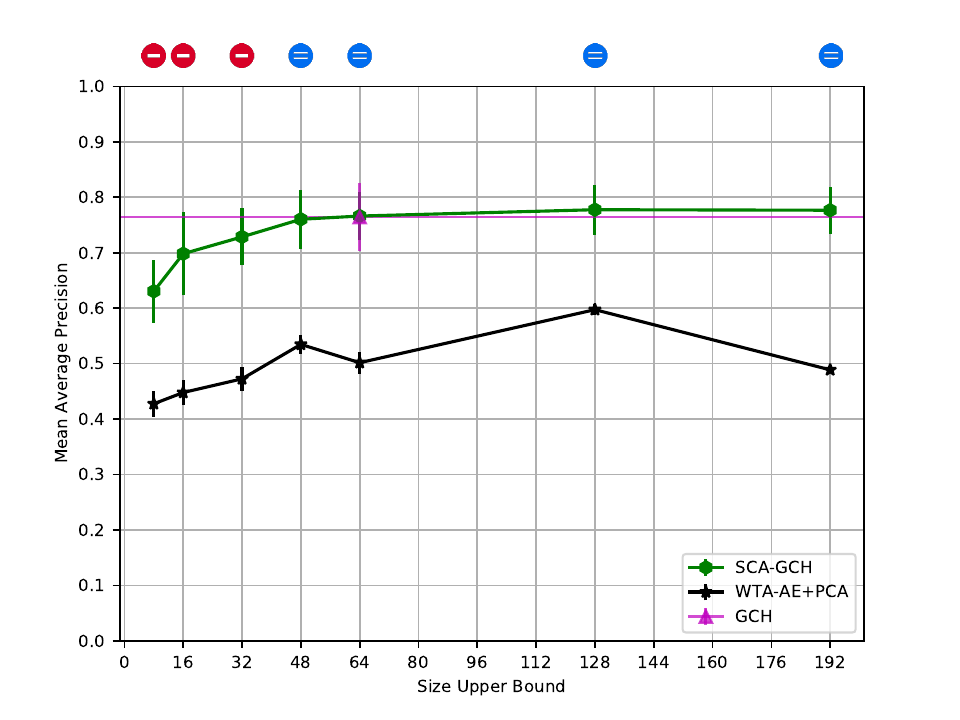}\label{chart:map_gch_la_fruits}}\\
	\subfloat[MSRCORID]{\includegraphics[width=0.5\textwidth]{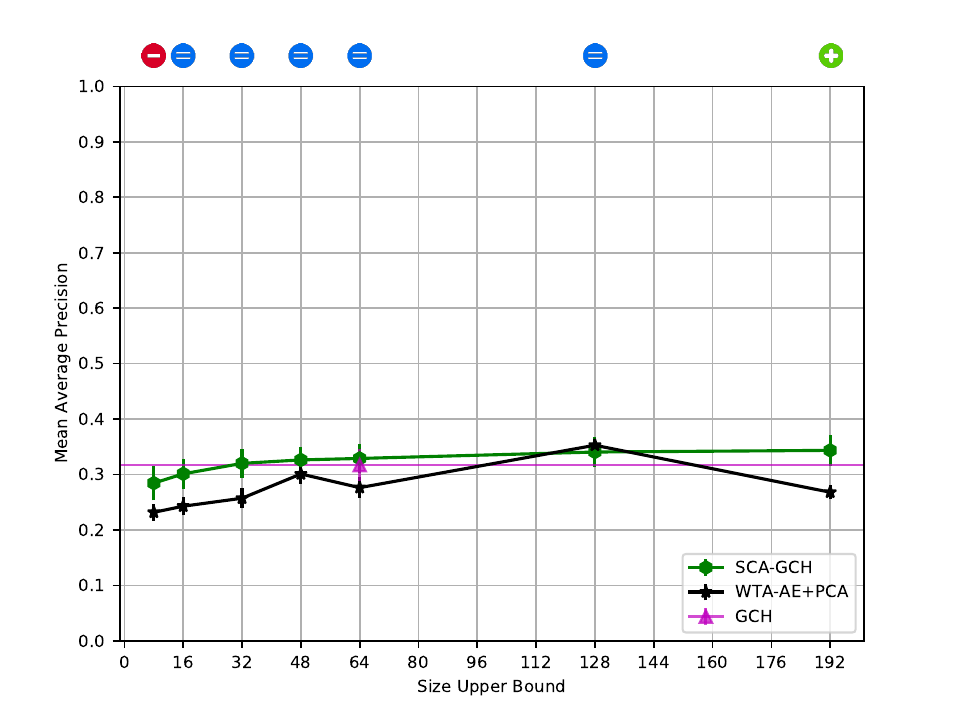}\label{chart:map_gch_la_msrcorid}}
	\subfloat[UCMerced Land-use]{\includegraphics[width=0.5\textwidth]{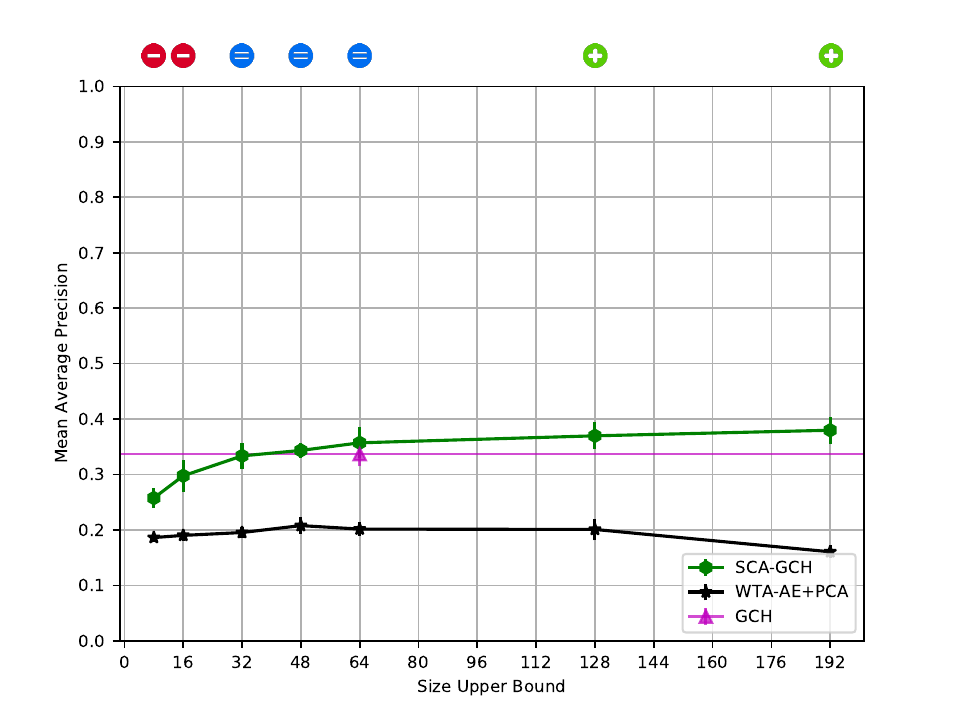}\label{chart:map_gch_la_ucmerced}}
	\caption{Comparison between the MAP results of SCA, WTA Autoencoder and GCH feature extractor}
	\label{chart:map_gch_la}
\end{figure}

Figs.~\ref{chart:precision_bic_la} and~\ref{chart:precision_gch_la} provide the P@10 results for the SCA method when compared with baselines for both BIC and GCH description approaches, respectively. Figs.~\ref{chart:map_bic_la} and~\ref{chart:map_gch_la}, in turn, provide the MAP results for both BIC and GCH description approaches, respectively.
Results related to MAP and P@10 demonstrated that, regardless the feature extraction method considered, the use of the SCA approach is able to create quite effective description approaches, without a high cost in terms of storage requirements, i.e., in terms of the feature vector size.

\begin{figure}[h]
	\centering
	\subfloat[BIC]{\includegraphics[width=0.5\textwidth]{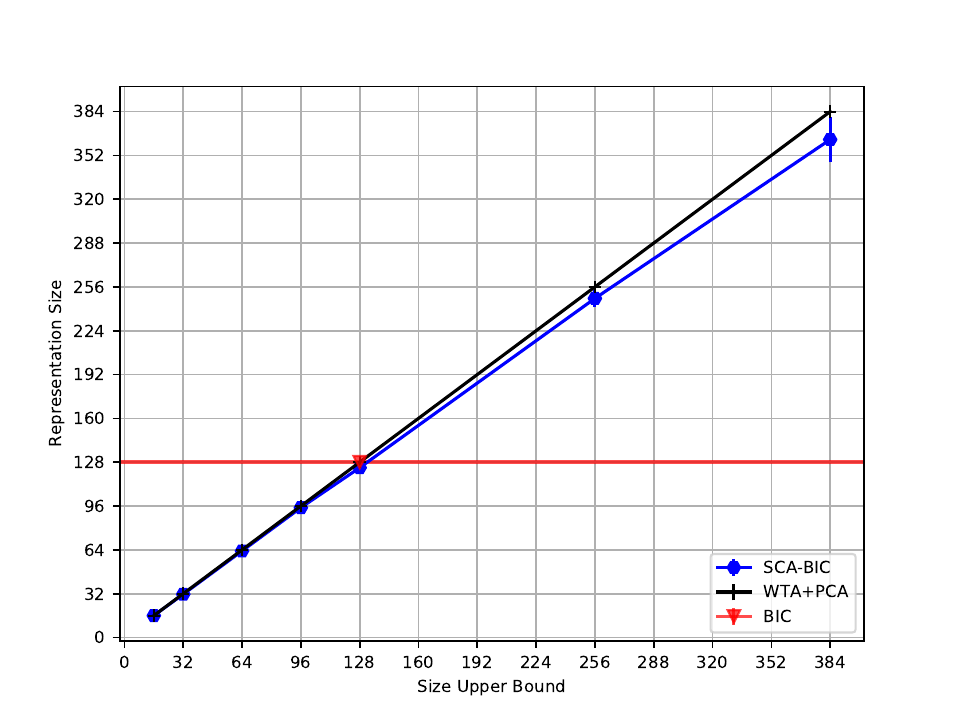}\label{chart:size_bic_la_eth80}}
	\subfloat[GCH]{\includegraphics[width=0.5\textwidth]{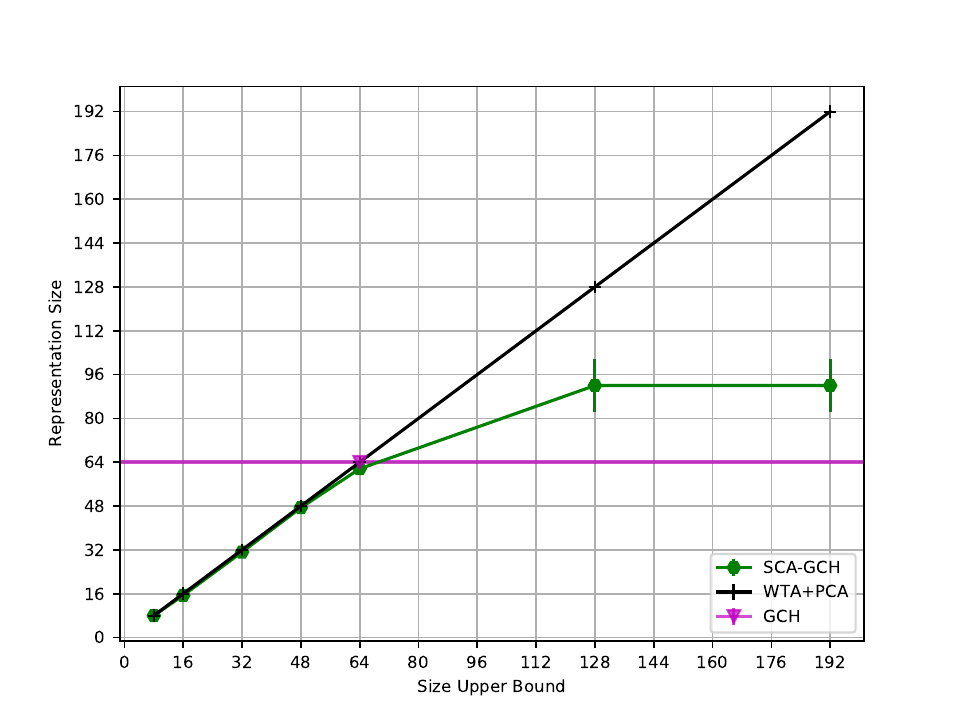}\label{chart:size_gch_la_eth80}}
	\caption{Comparison between the representation size results of SCA, WTA Autoencoder and the feature extractors: (a) BIC and (b) GCH, for the ETH-80 dataset. The supplementary material of this article contains the same comparison for the remaining datasets.}
	\label{chart:size_la}
\end{figure}

Figure~\ref{chart:size_la} shows the sizes of the produced representations given the respective size upper-bounds.
For the BIC approach (Fig. \ref{chart:size_bic_la_eth80})
representations whose size reached or were very close to the imposed upper-bound were produced, showing a tendency for generating quantizations with strong tonality detailing.
In contrast, the results for the GCH approach (Fig.~\ref{chart:size_gch_la_eth80}) are quite different.
For example, for the upper limits 128 and 192, the produced representations were considerably smaller than the maximum size.
In these cases, the more effective representations are not necessarily the ones with the highest possible dimensionality. This finding means that increasing the number of tonalities does not necessarily lead to performance improvements.
In other words, the proposed methods are able to generate representations that are significantly smaller than the predefined upper-bound 
but with high effectiveness.



\newlength{\tempdima}
\newcommand{\rowname}[1]
{\rotatebox{90}{\makebox[\tempdima][c]{#1}}}
\begin{figure}[h]
	\centering
	\settoheight{\tempdima}{\includegraphics[width=.25\linewidth]{example-image-a}}%
	\begin{tabular}{@{}c@{}c@{}c@{}c@{}c@{}}
		\centering
		{}&Groundtruth & Coil-100& Corel-1566& Corel-3906 \\
		\rowname{16}&
		\includegraphics[width=0.25\textwidth]{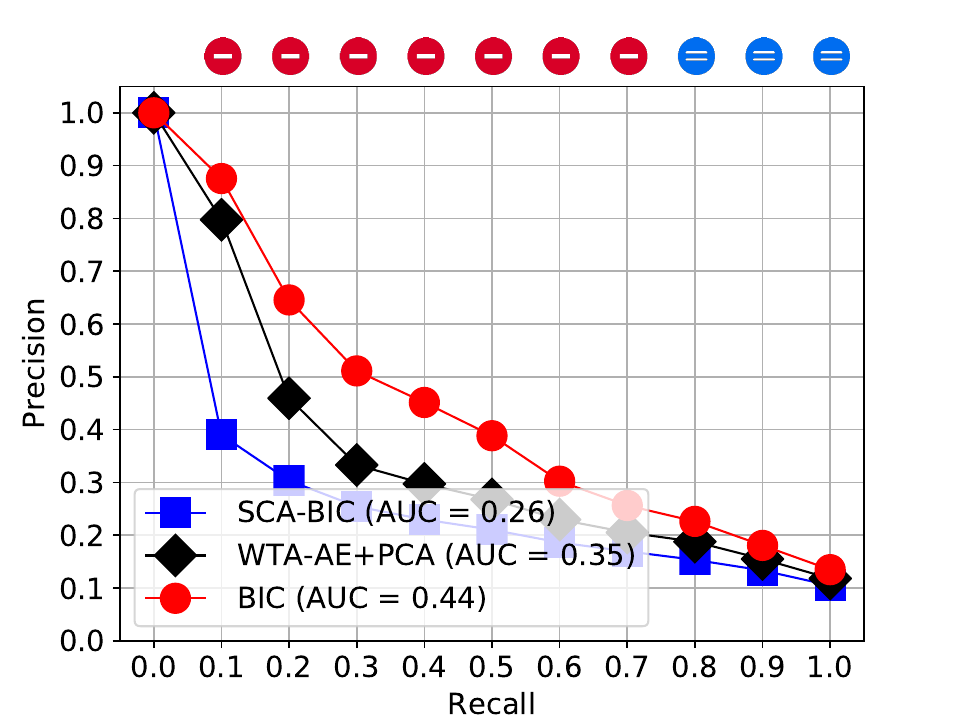}\label{chart:pr_bic_la_coffe_16}&
		\includegraphics[width=0.25\textwidth]{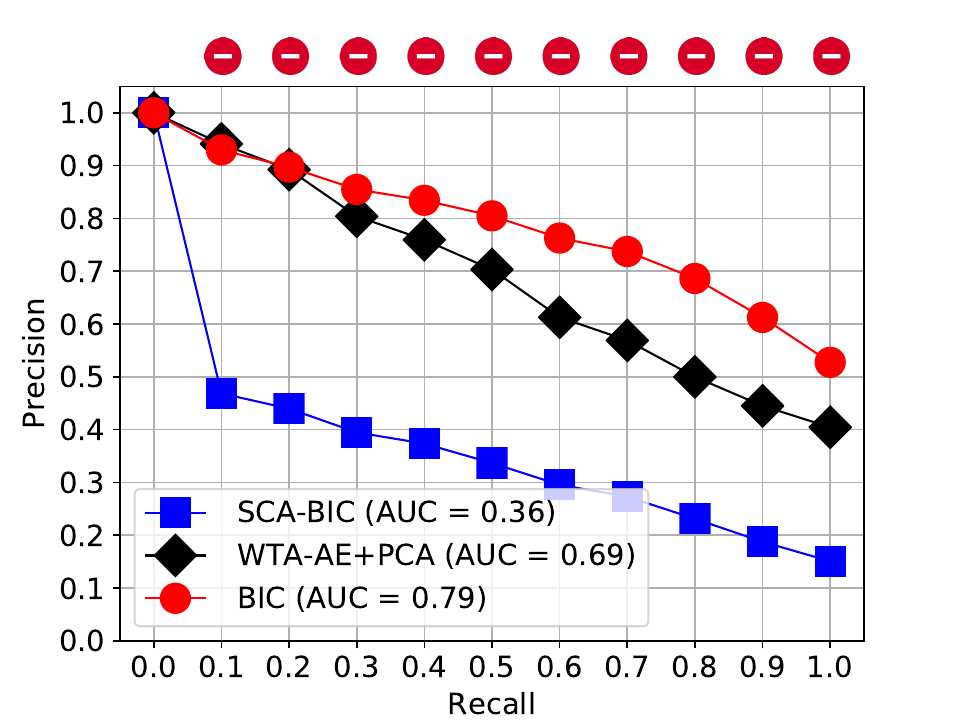}\label{chart:pr_bic_la_coil100_16}&
		\includegraphics[width=0.25\textwidth]{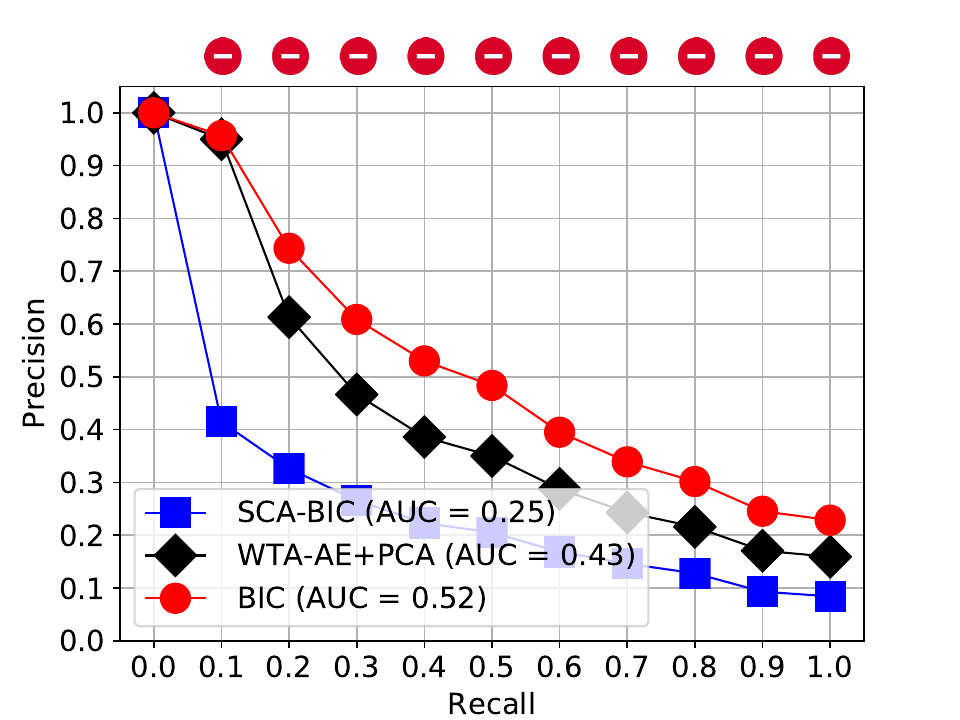}\label{chart:pr_bic_la_corel1566_16}&
		\includegraphics[width=0.25\textwidth]{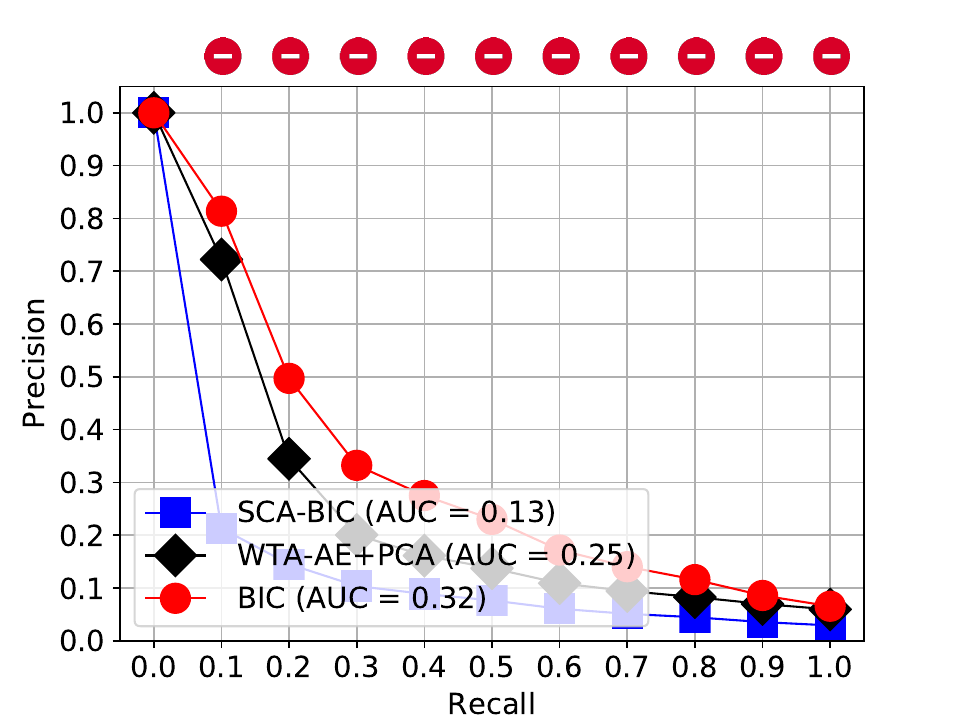}\label{chart:pr_bic_la_corel3909_16}\\
		\rowname{32}&
		\includegraphics[width=0.25\textwidth]{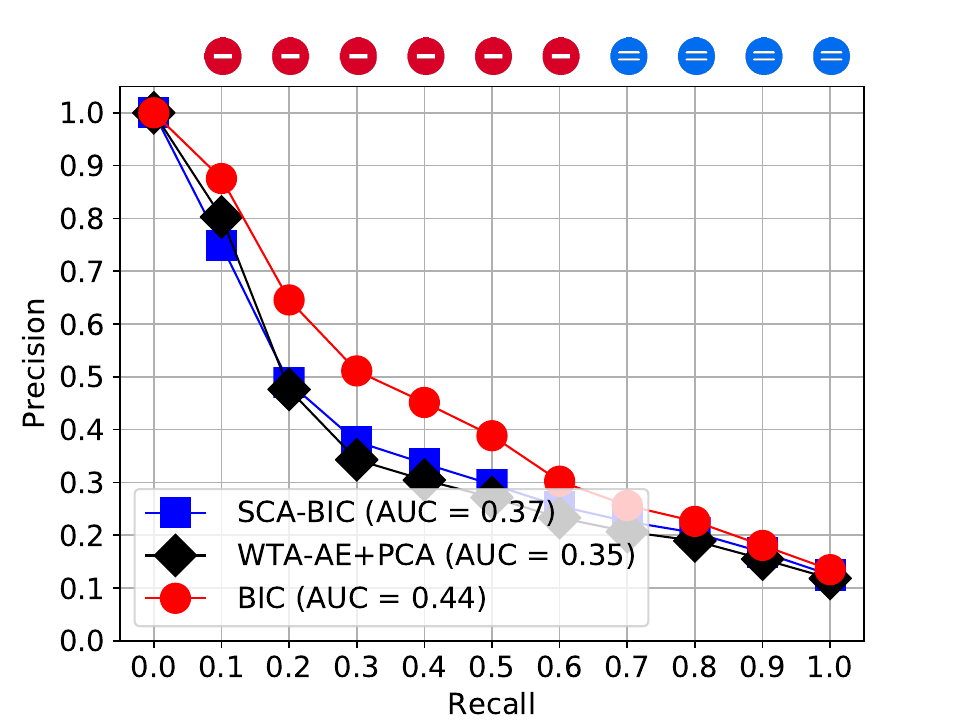}\label{chart:pr_bic_la_coffe_32}&
		\includegraphics[width=0.25\textwidth]{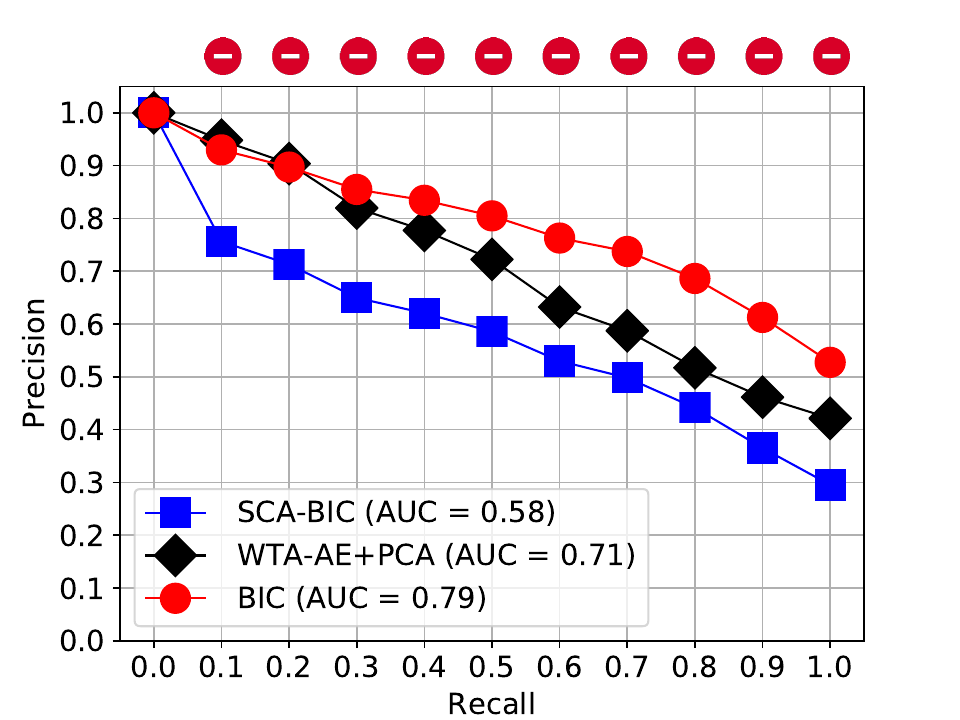}\label{chart:pr_bic_la_coil100_32}&
		\includegraphics[width=0.25\textwidth]{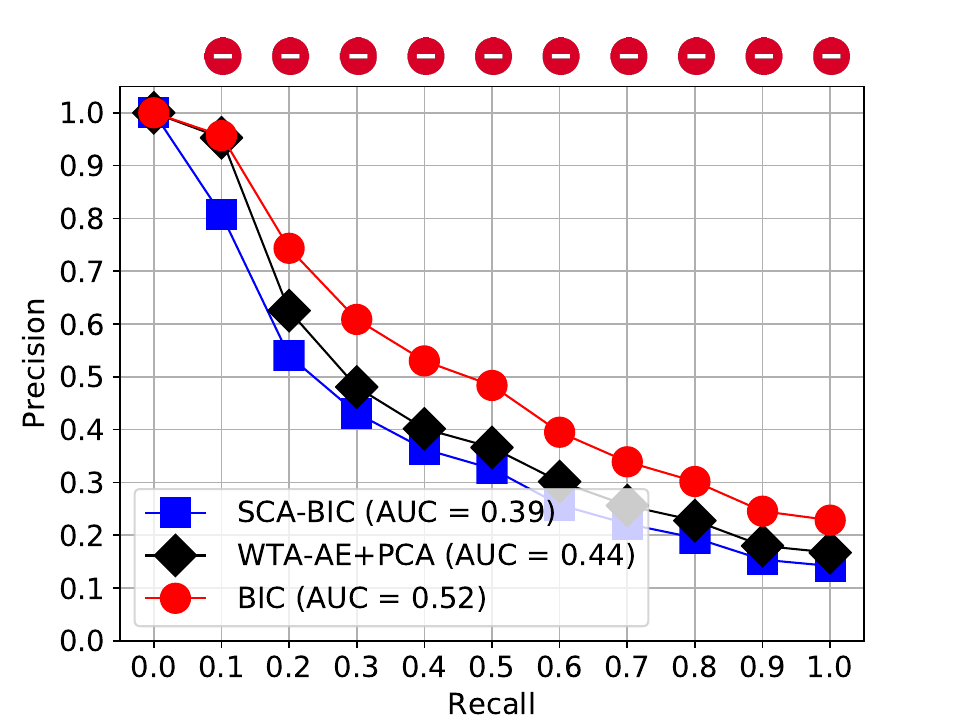}\label{chart:pr_bic_la_corel1566_32}&
		\includegraphics[width=0.25\textwidth]{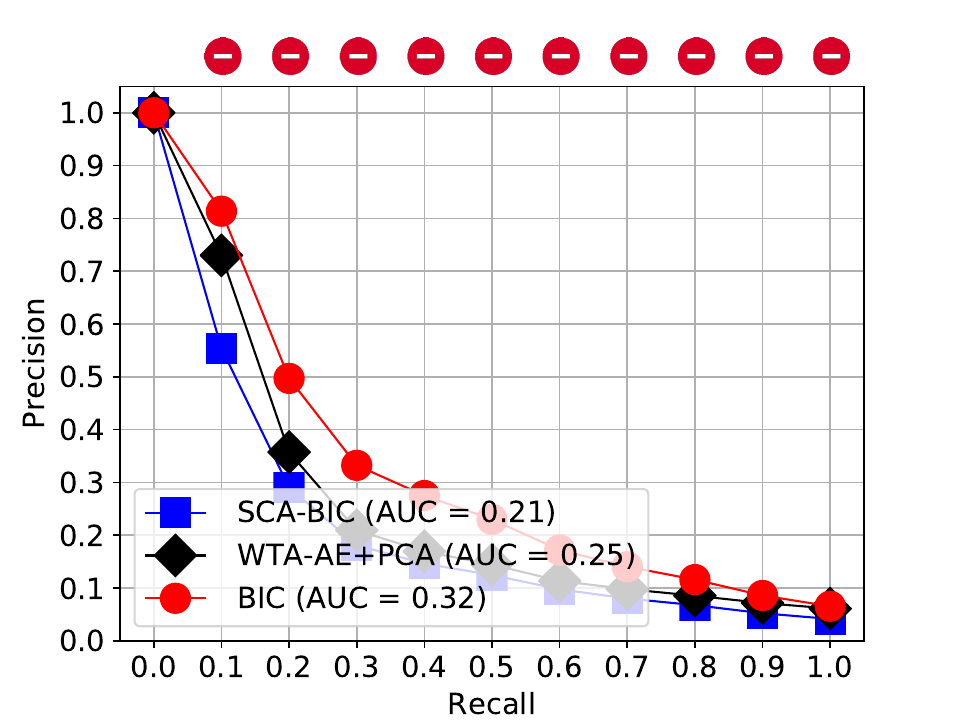}\label{chart:pr_bic_la_corel3909_32}\\
		\rowname{64}&
		\includegraphics[width=0.25\textwidth]{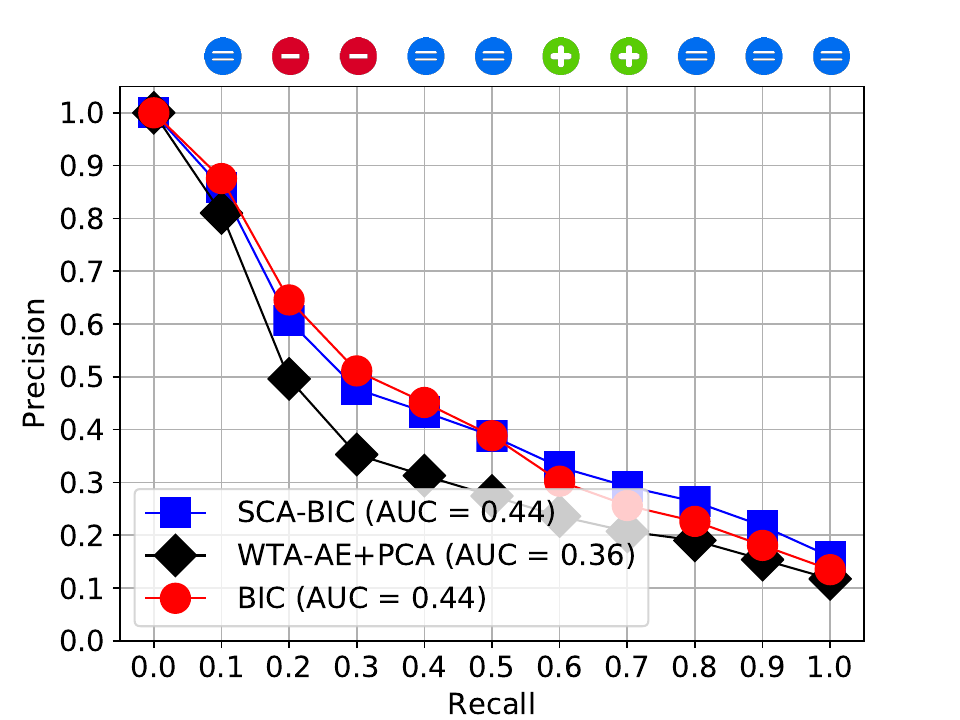}\label{chart:pr_bic_la_coffe_64}&
		\includegraphics[width=0.25\textwidth]{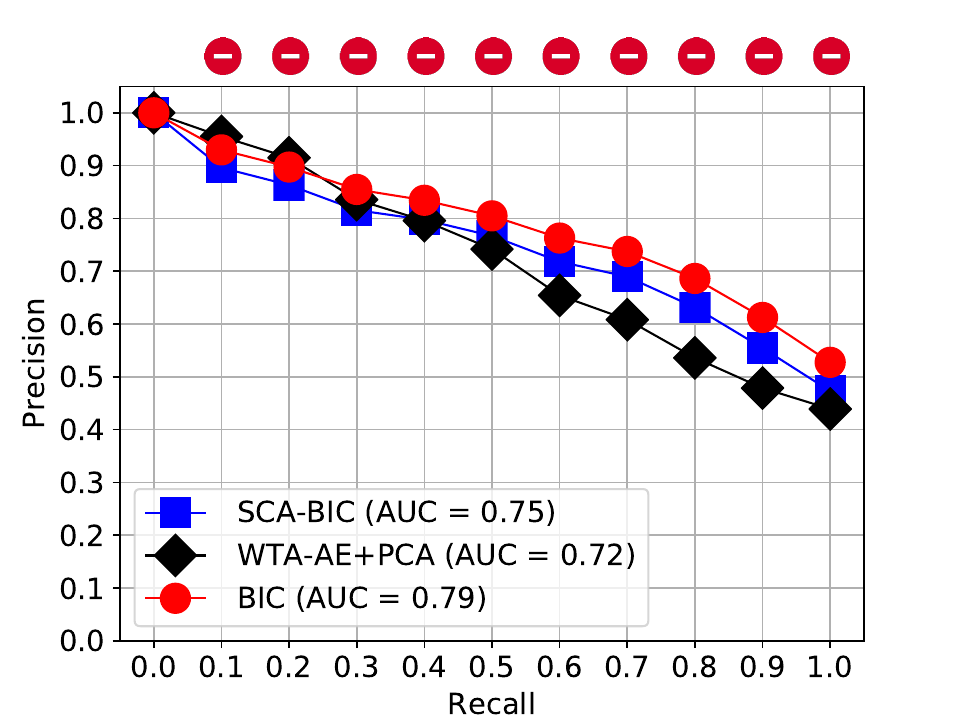}\label{chart:pr_bic_la_coil100_64}&
		\includegraphics[width=0.25\textwidth]{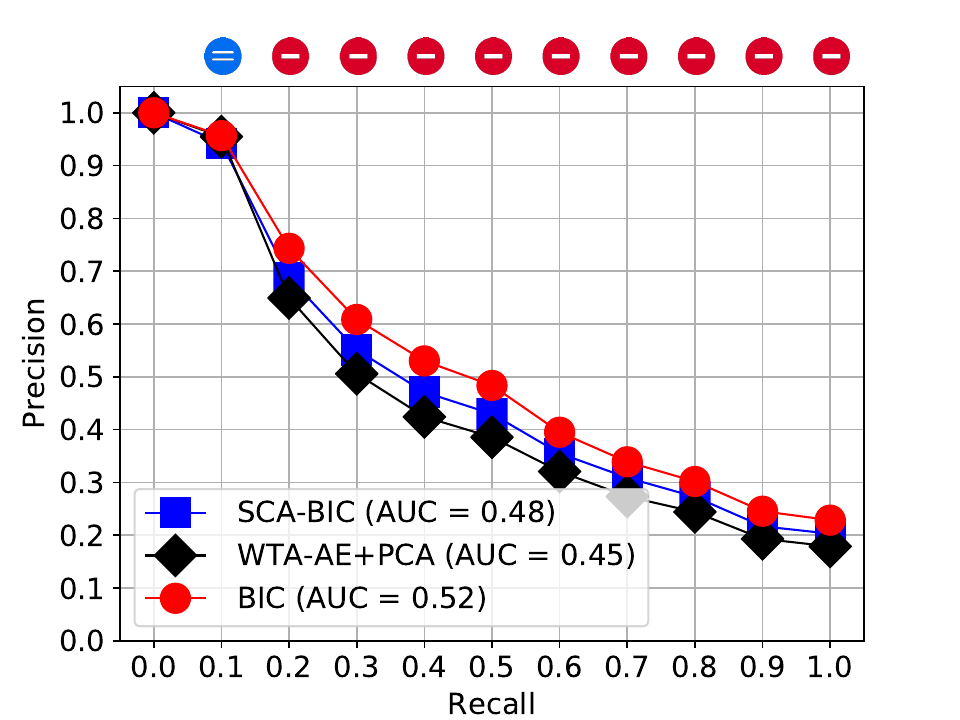}\label{chart:pr_bic_la_corel1566_64}&
		\includegraphics[width=0.25\textwidth]{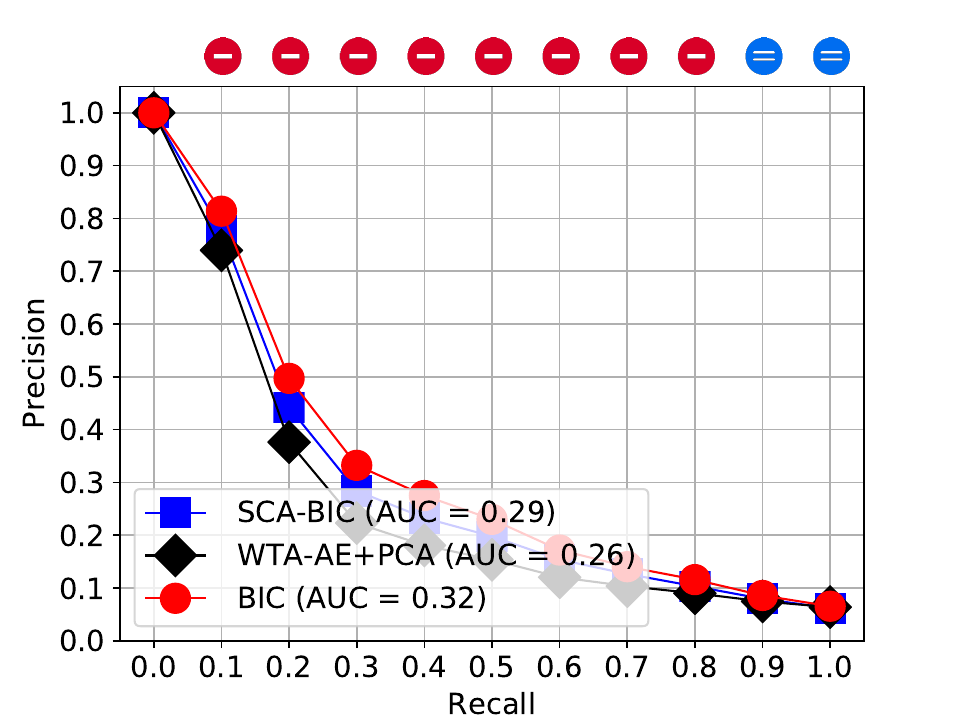}\label{chart:pr_bic_la_corel3909_64}\\
		\rowname{96}&
		\includegraphics[width=0.25\textwidth]{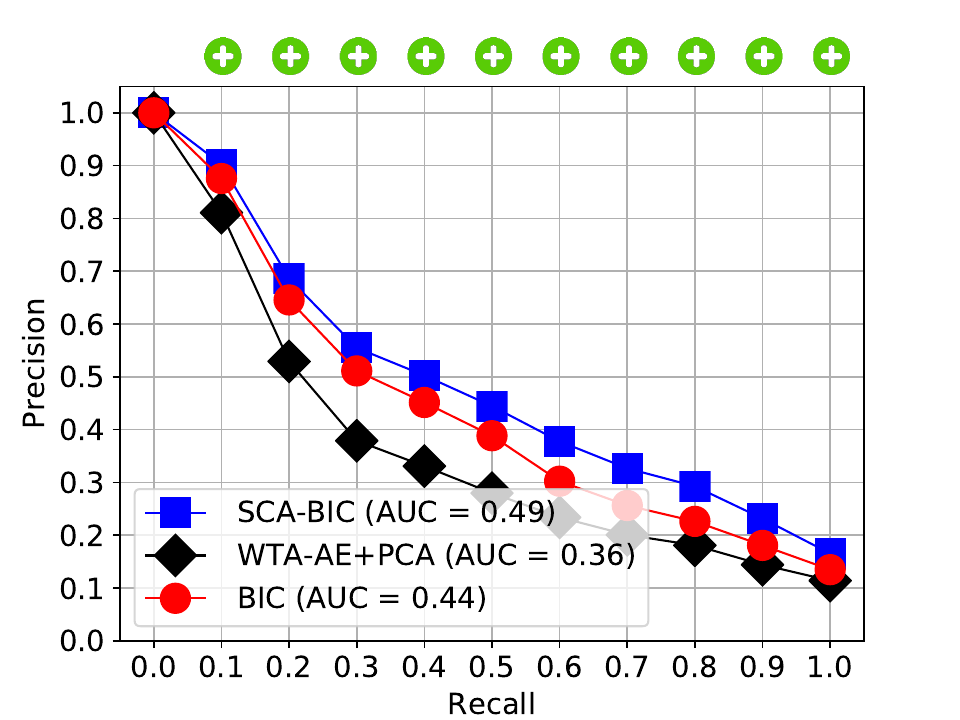}\label{chart:pr_bic_la_coffe_96}&
		\includegraphics[width=0.25\textwidth]{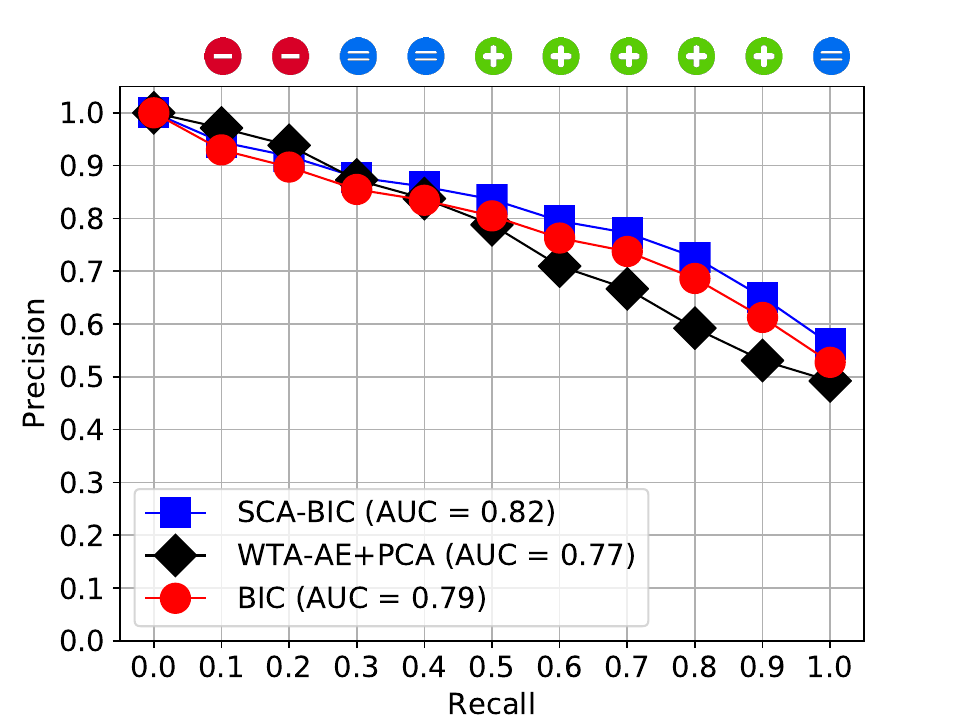}\label{chart:pr_bic_la_coil100_96}&
		\includegraphics[width=0.25\textwidth]{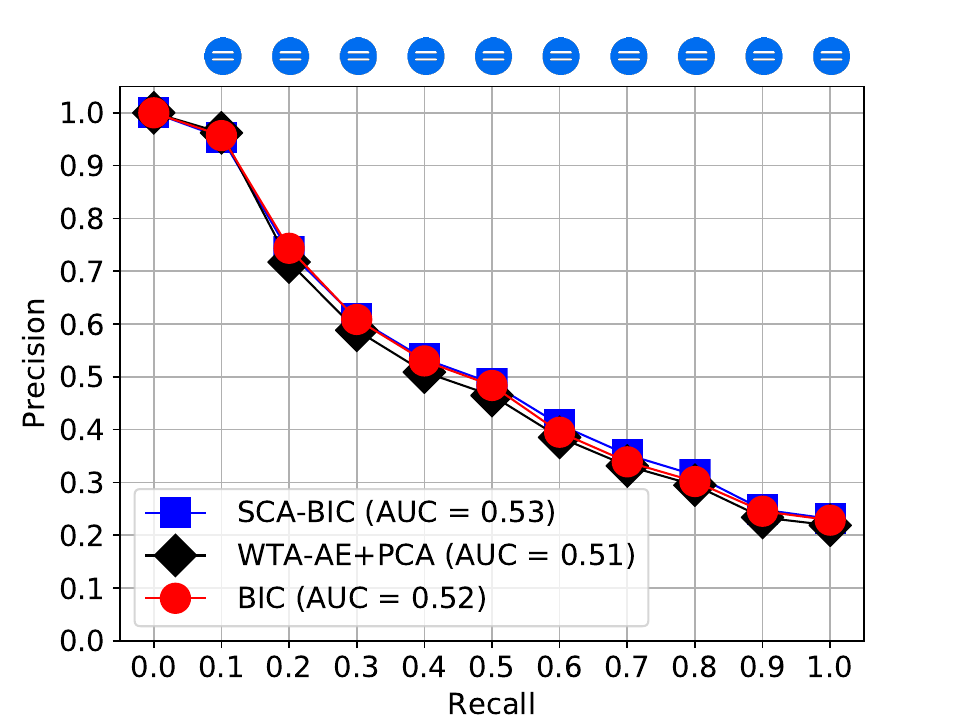}\label{chart:pr_bic_la_corel1566_96}&
		\includegraphics[width=0.25\textwidth]{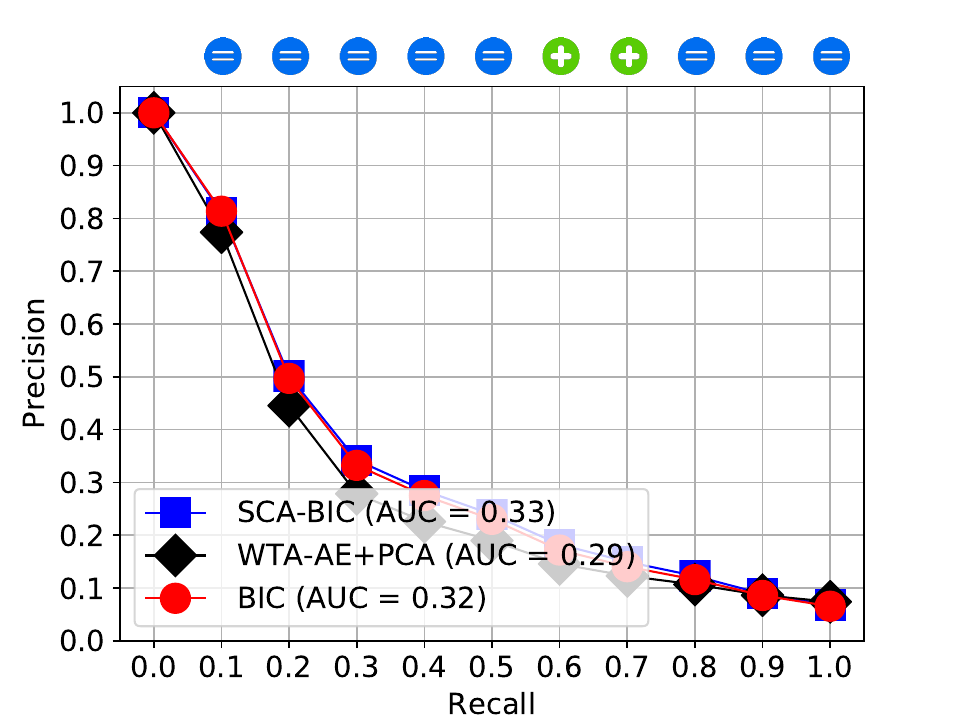}\label{chart:pr_bic_la_corel3909_96}\\
		\rowname{128}&
		\includegraphics[width=0.25\textwidth]{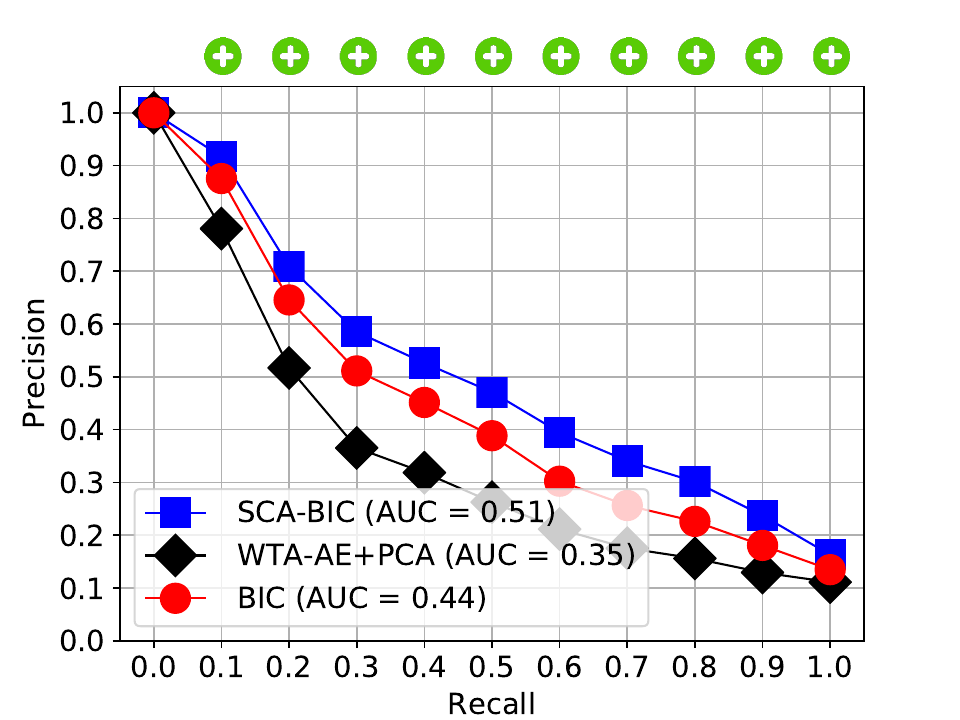}\label{chart:pr_bic_la_coffe_128}&
		\includegraphics[width=0.25\textwidth]{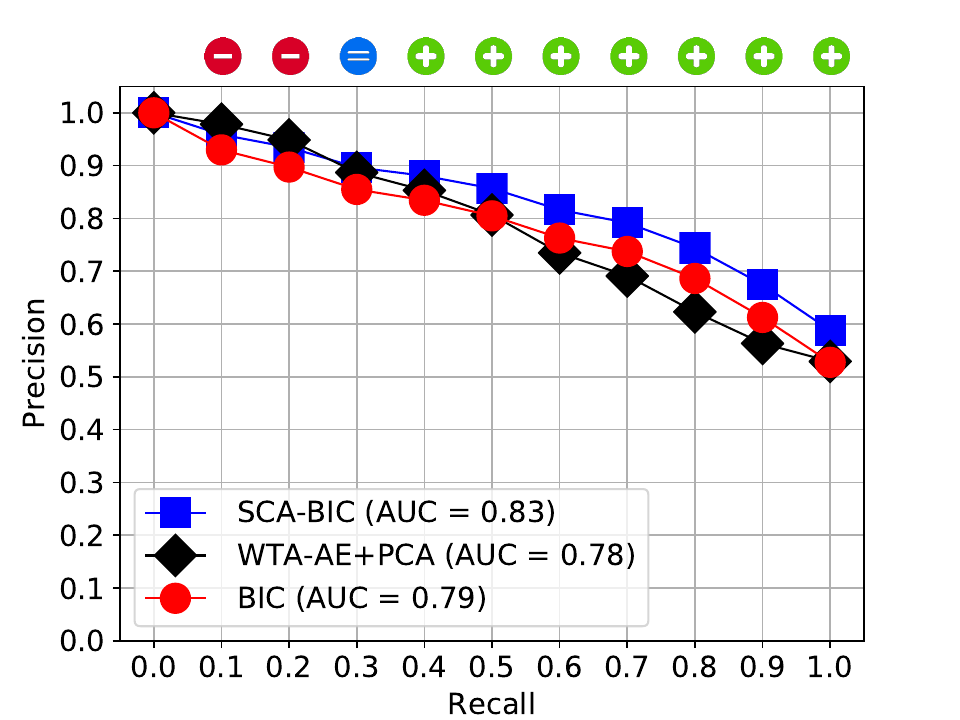}\label{chart:pr_bic_la_coil100_128}&
		\includegraphics[width=0.25\textwidth]{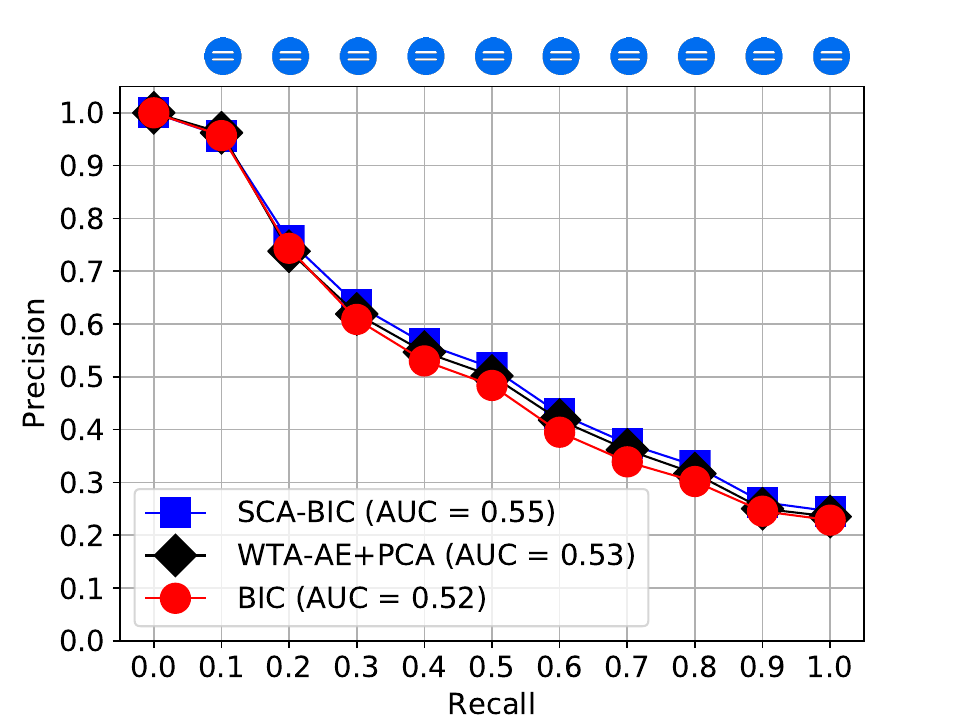}\label{chart:pr_bic_la_corel1566_128}&
		\includegraphics[width=0.25\textwidth]{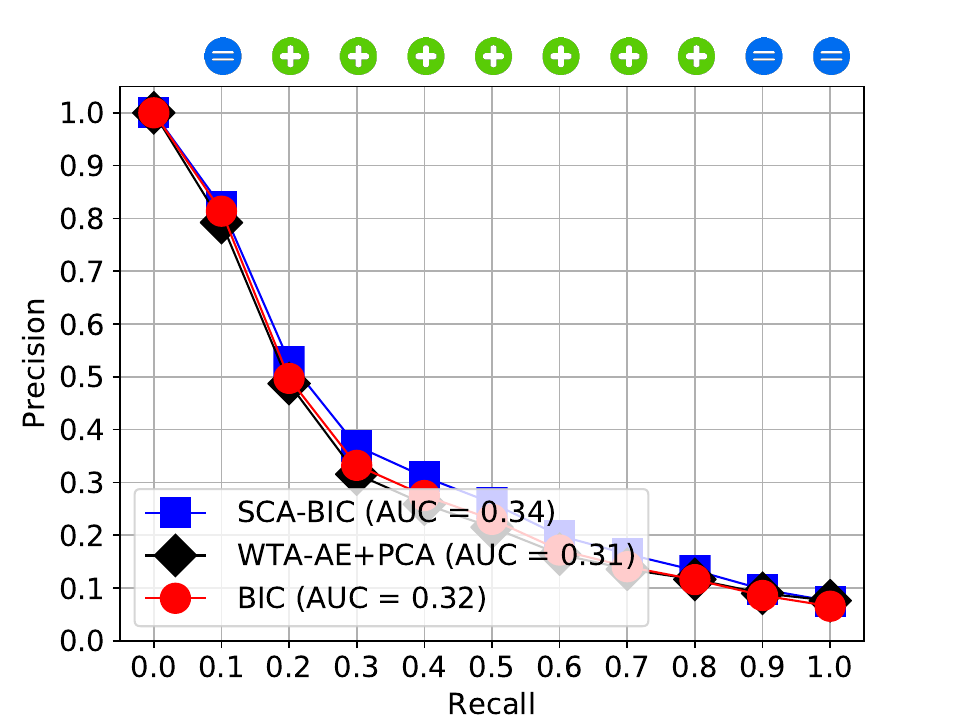}\label{chart:pr_bic_la_corel3909_128}\\
		\rowname{256}&
		\includegraphics[width=0.25\textwidth]{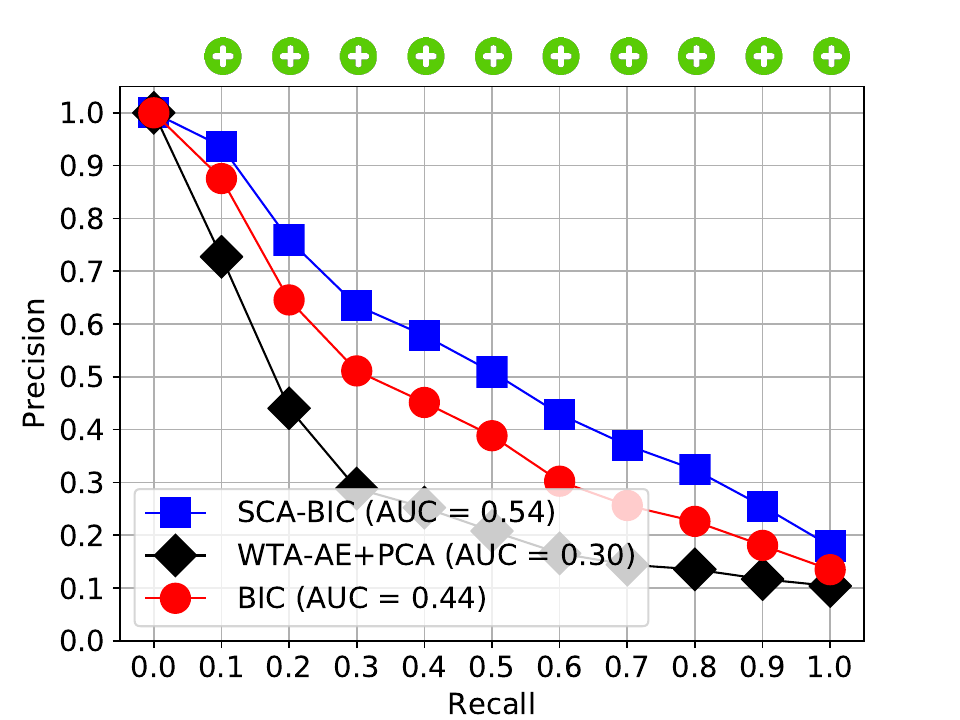}\label{chart:pr_bic_la_coffe_256}&
		\includegraphics[width=0.25\textwidth]{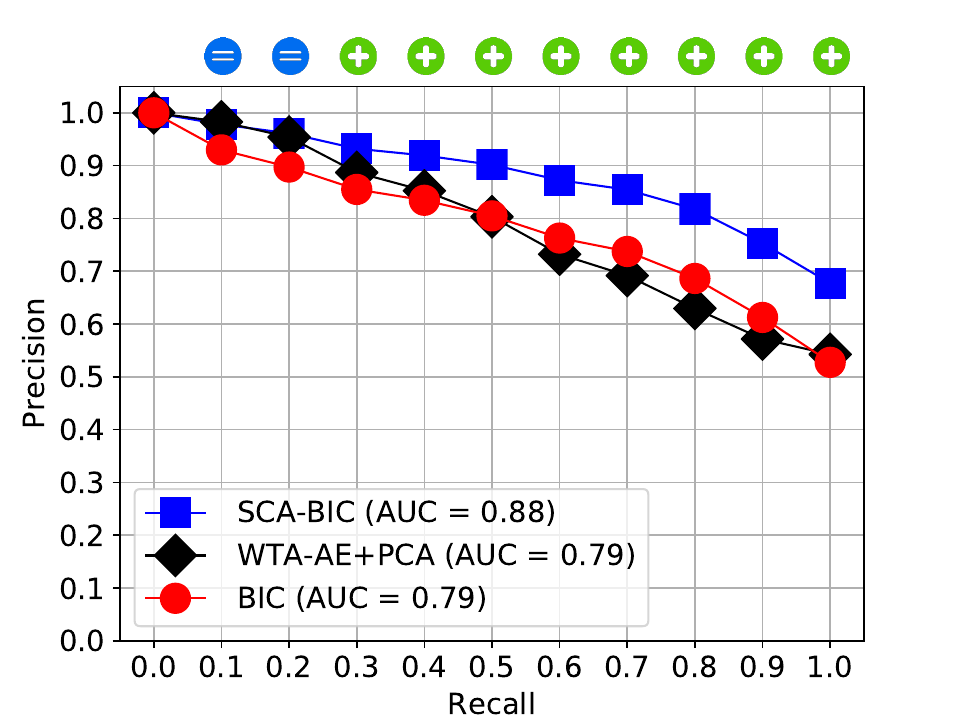}\label{chart:pr_bic_la_coil100_256}&
		\includegraphics[width=0.25\textwidth]{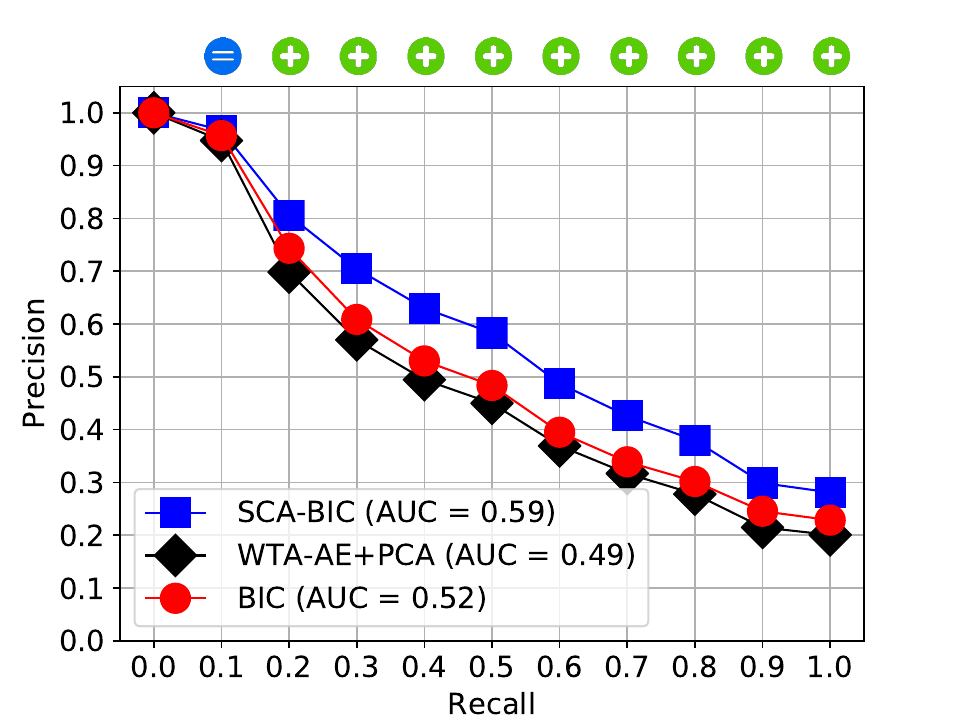}\label{chart:pr_bic_la_corel1566_256}&
		\includegraphics[width=0.25\textwidth]{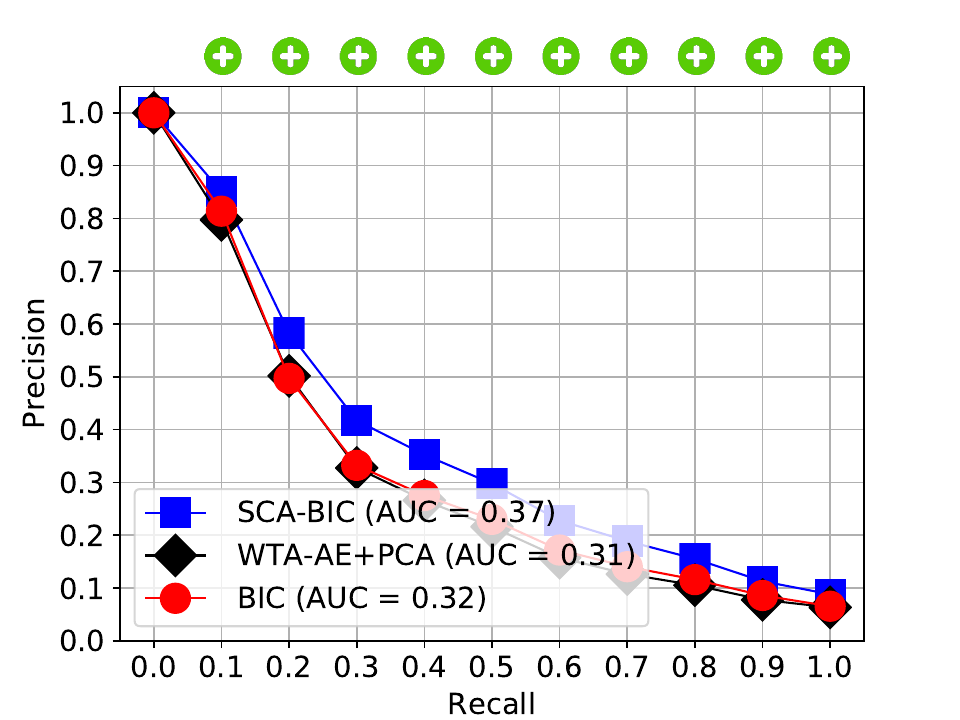}\label{chart:pr_bic_la_corel3909_256}\\
		\rowname{384}&
		\includegraphics[width=0.25\textwidth]{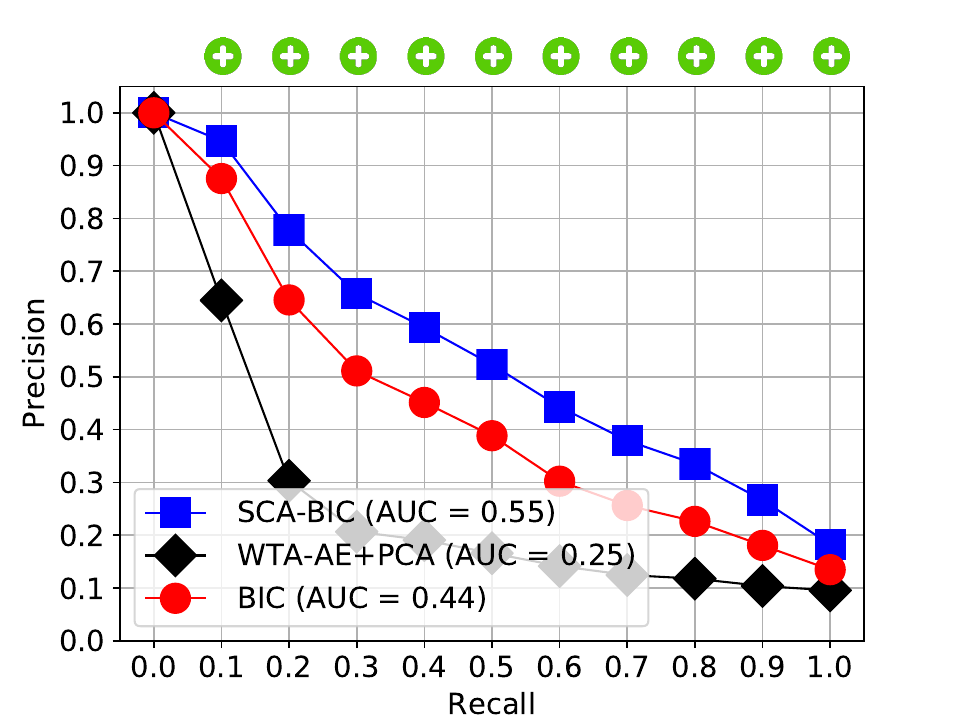}\label{chart:pr_bic_la_coffe_384}&
		\includegraphics[width=0.25\textwidth]{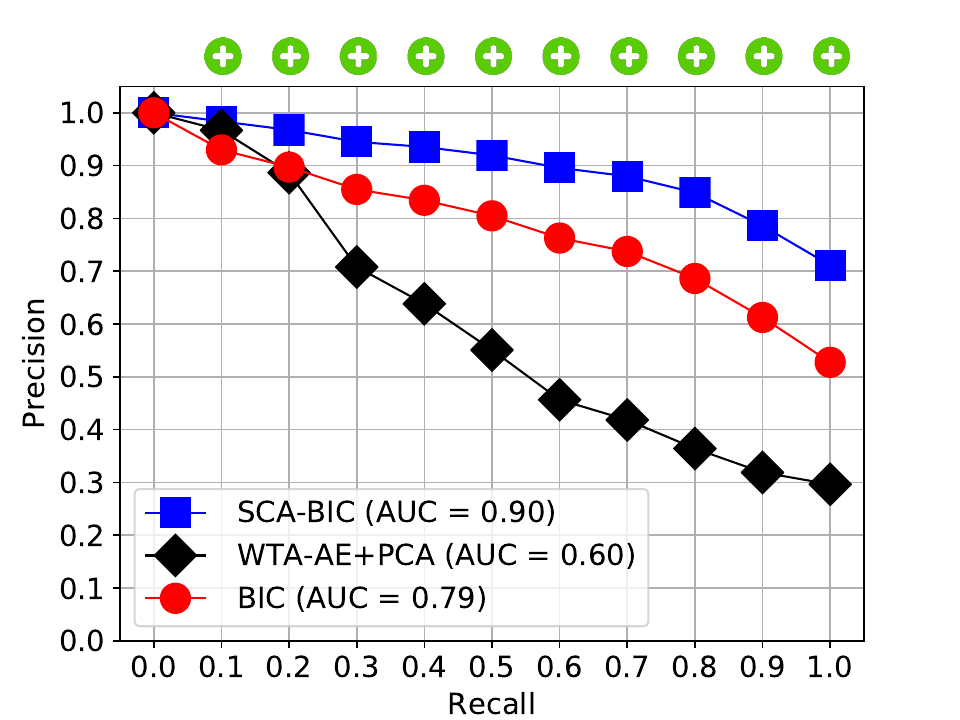}\label{chart:pr_bic_la_coil100_384}&
		\includegraphics[width=0.25\textwidth]{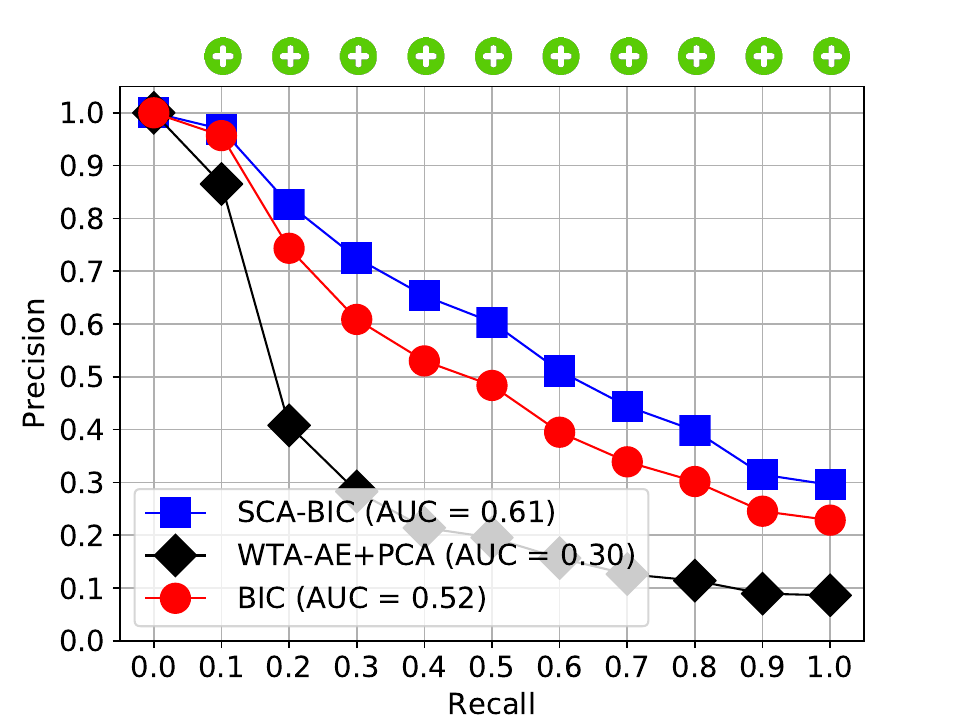}\label{chart:pr_bic_la_corel1566_384}&
		\includegraphics[width=0.25\textwidth]{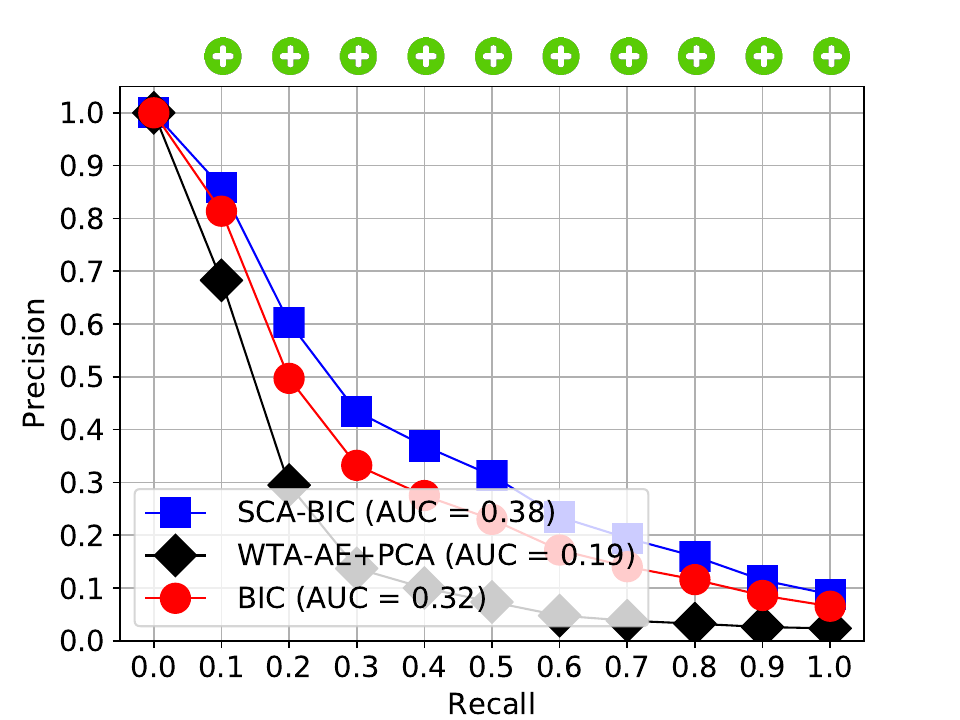}\label{chart:pr_bic_la_corel3909_384}
	\end{tabular}
	\caption{Comparison between the Precision-Recall curves of SCA, WTA Autoencoder and BIC feature extractor considering all representation size limits for the datasets 
		\textit{Groundtruth}, \textit{Coil-100}, \textit{Corel-1566}, and \textit{Corel-3906}. 
	We recommend colourful printing for adequate visualization.}%
		\label{chart:pr_bic_la1}
\end{figure}

\begin{figure}[h]
	\centering
	\settoheight{\tempdima}{\includegraphics[width=.25\linewidth]{example-image-a}}%
	\begin{tabular}{@{}c@{}c@{}c@{}c@{}c@{}}
		\centering
		{}&ETH-80 & Supermarket P. & MSRCORID & UCMerced L. \\
		\rowname{16}&
		\includegraphics[width=0.25\textwidth]{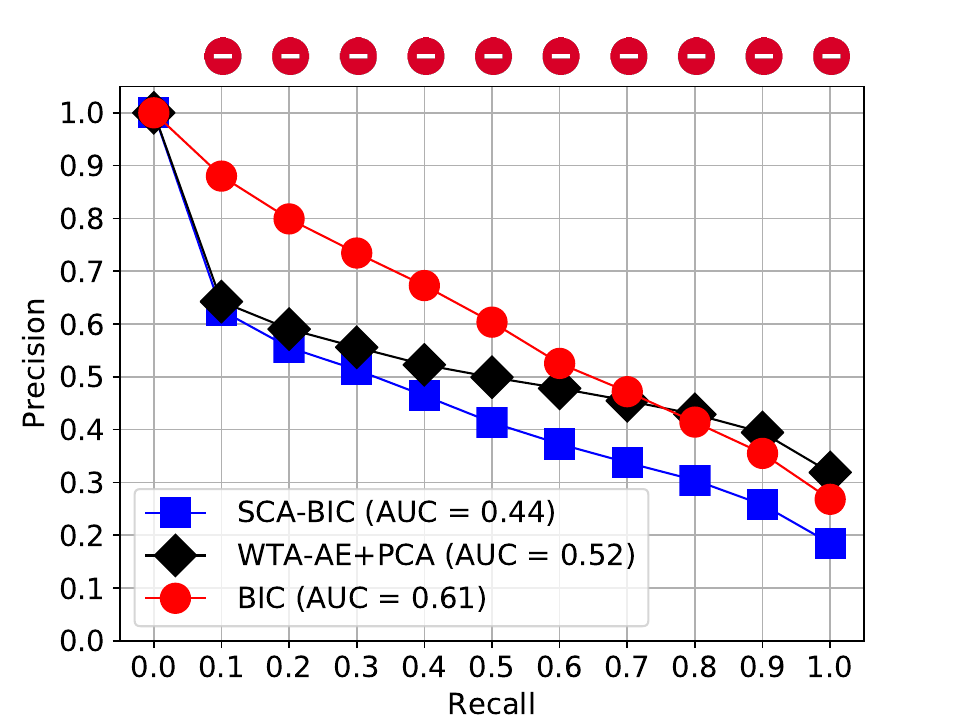}\label{chart:pr_bic_la_eth80_16}&
		\includegraphics[width=0.25\textwidth]{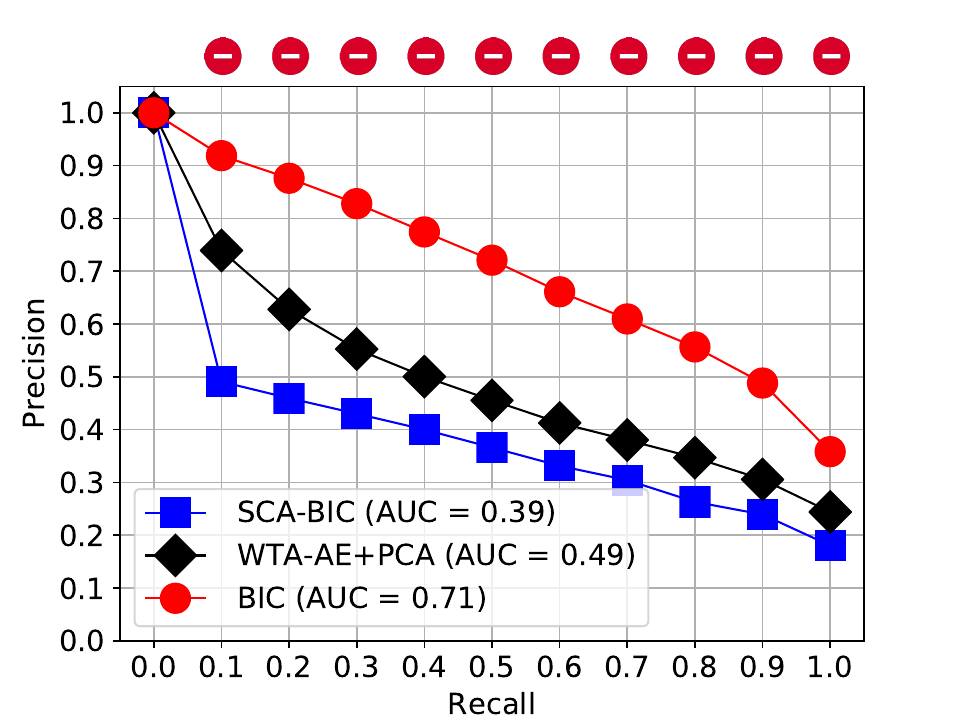}\label{chart:pr_bic_la_fruits_16}&
		\includegraphics[width=0.25\textwidth]{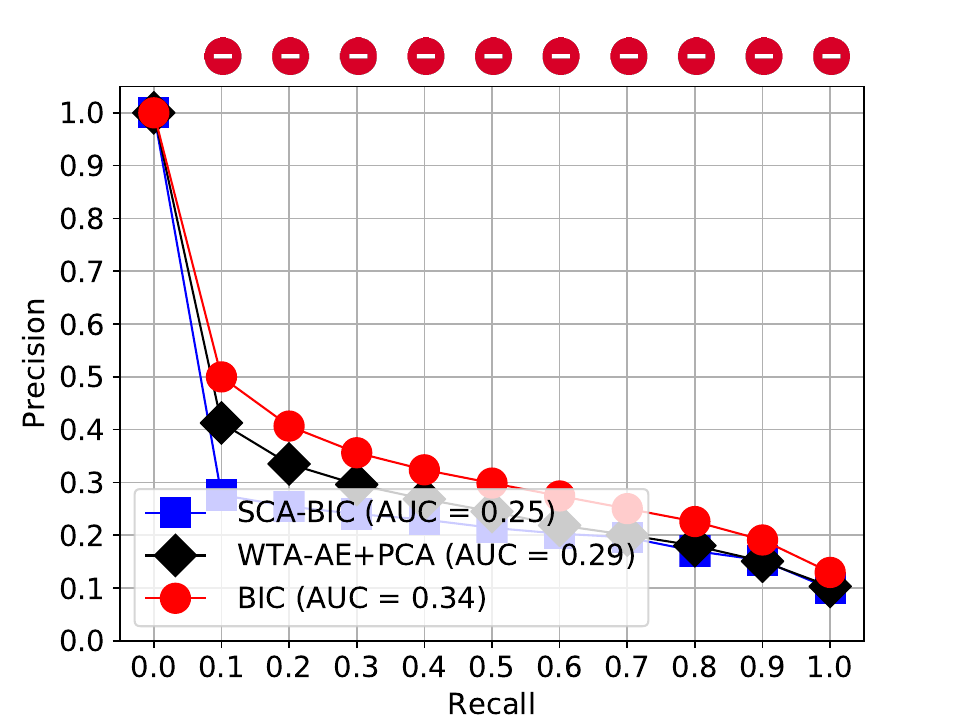}\label{chart:pr_bic_la_msrcorid_16}&
		\includegraphics[width=0.25\textwidth]{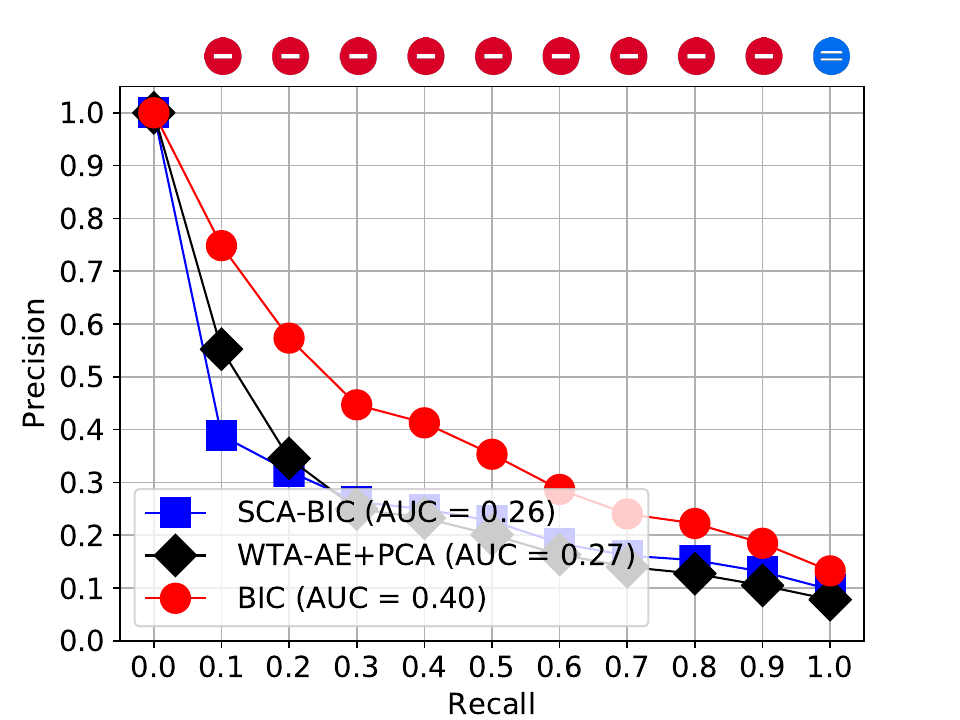}\label{chart:pr_bic_la_ucmerced_16}\\
		\rowname{32}&
		\includegraphics[width=0.25\textwidth]{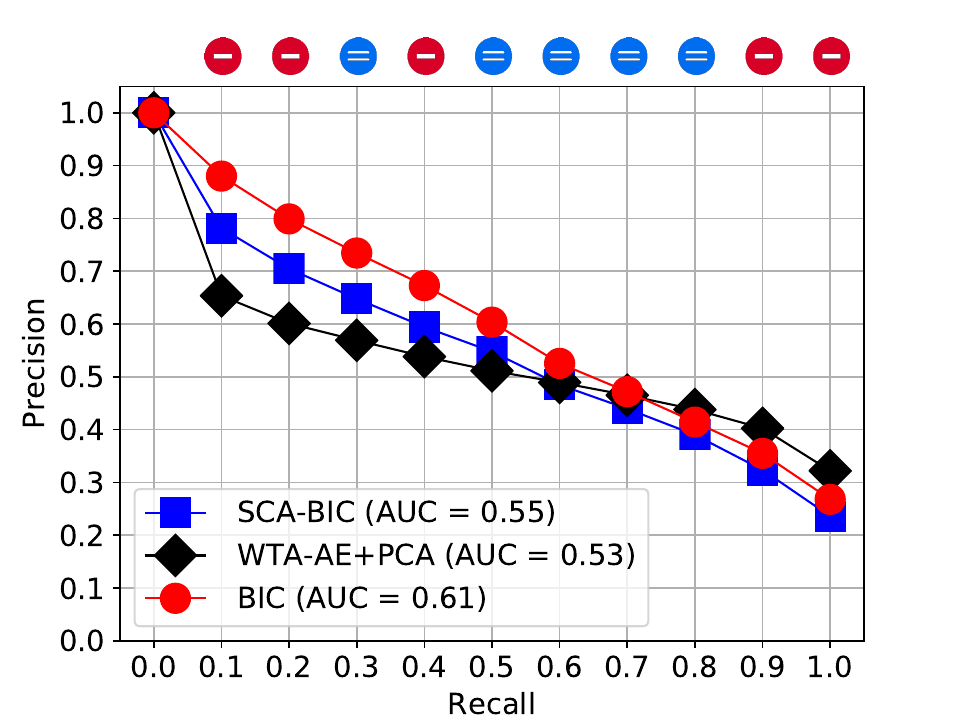}\label{chart:pr_bic_la_eth80_32}&
		\includegraphics[width=0.25\textwidth]{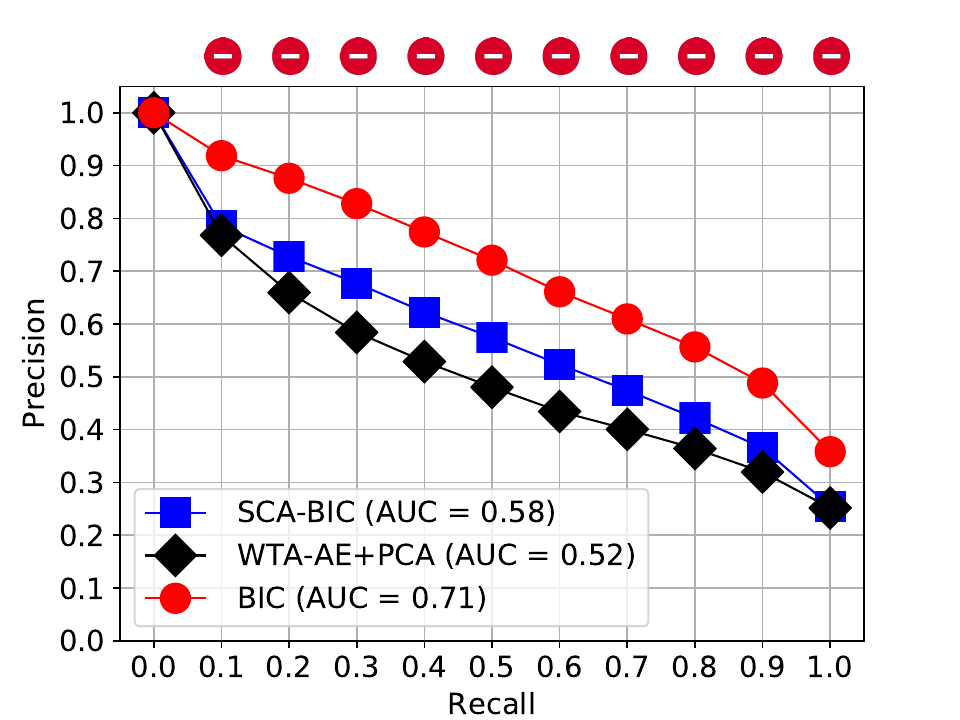}\label{chart:pr_bic_la_fruits__32}&
		\includegraphics[width=0.25\textwidth]{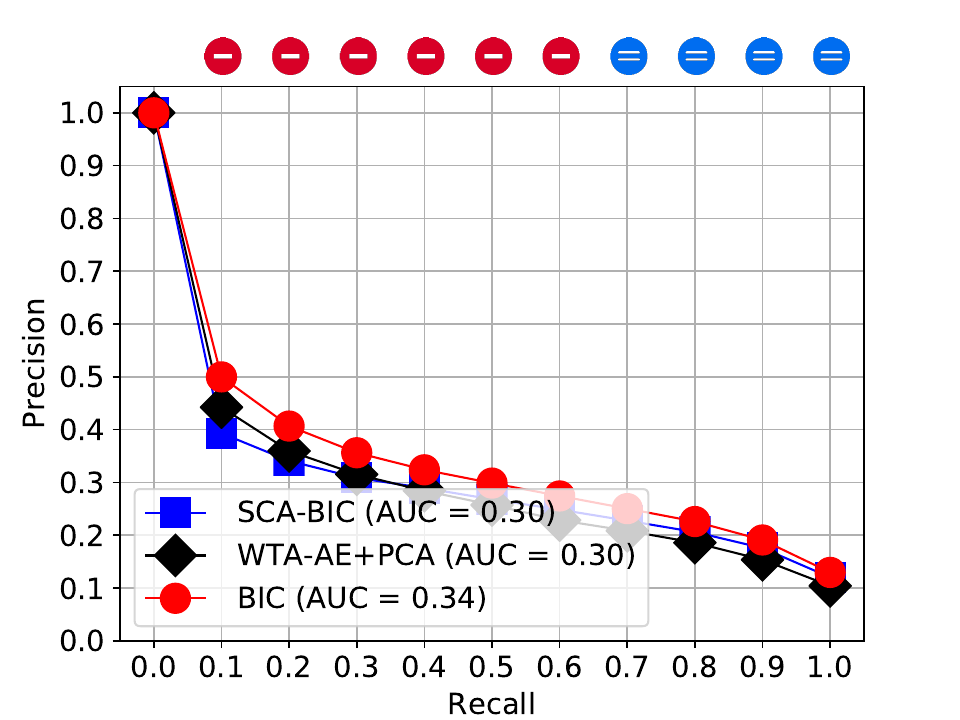}\label{chart:pr_bic_la_msrcorid_32}&
		\includegraphics[width=0.25\textwidth]{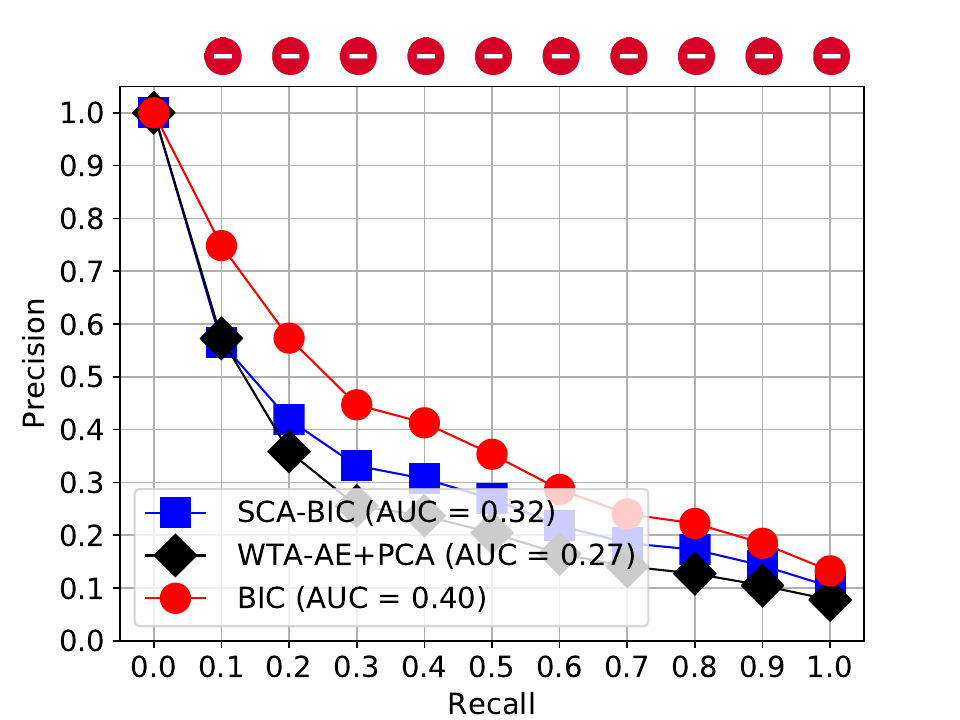}\label{chart:pr_bic_la_ucmerced_32}\\
		\rowname{64}&
		\includegraphics[width=0.25\textwidth]{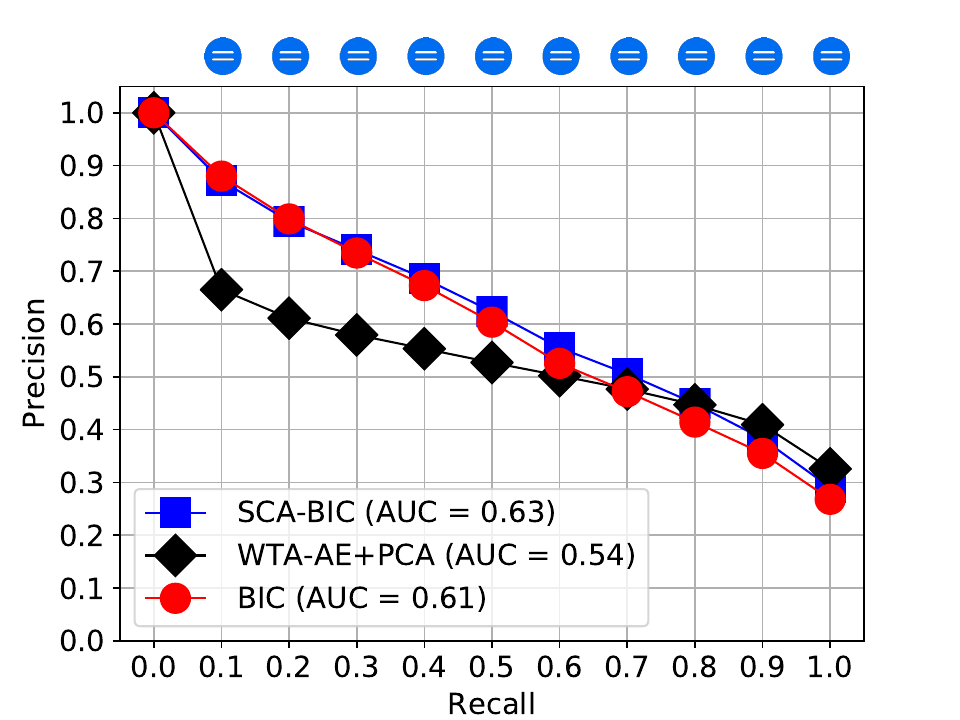}\label{chart:pr_bic_la_eth80_64}&
		\includegraphics[width=0.25\textwidth]{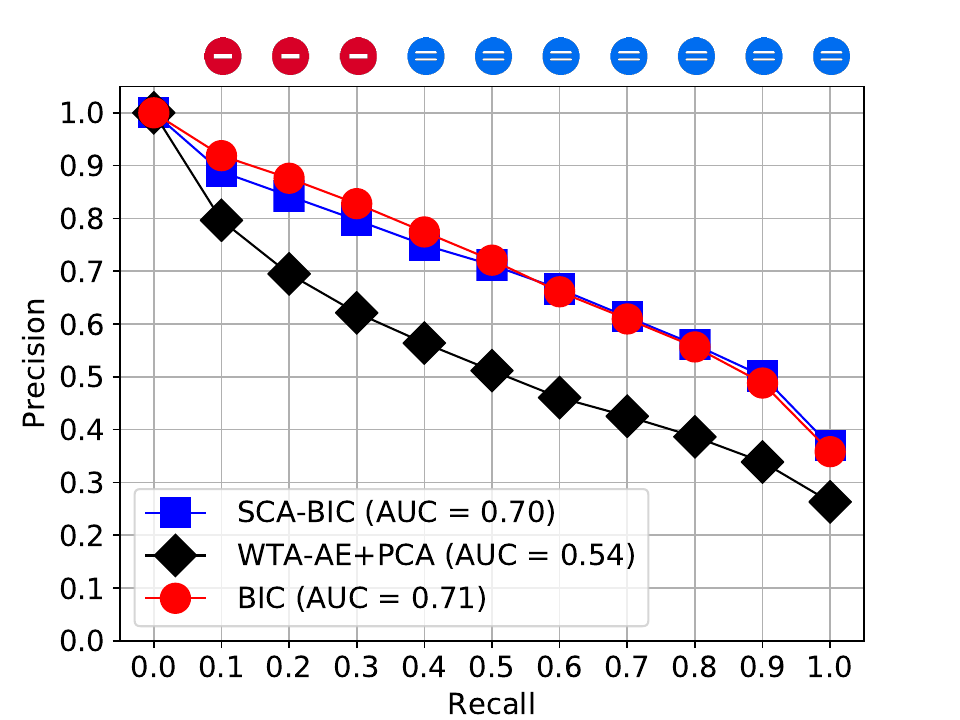}\label{chart:pr_bic_la_fruits__64}&
		\includegraphics[width=0.25\textwidth]{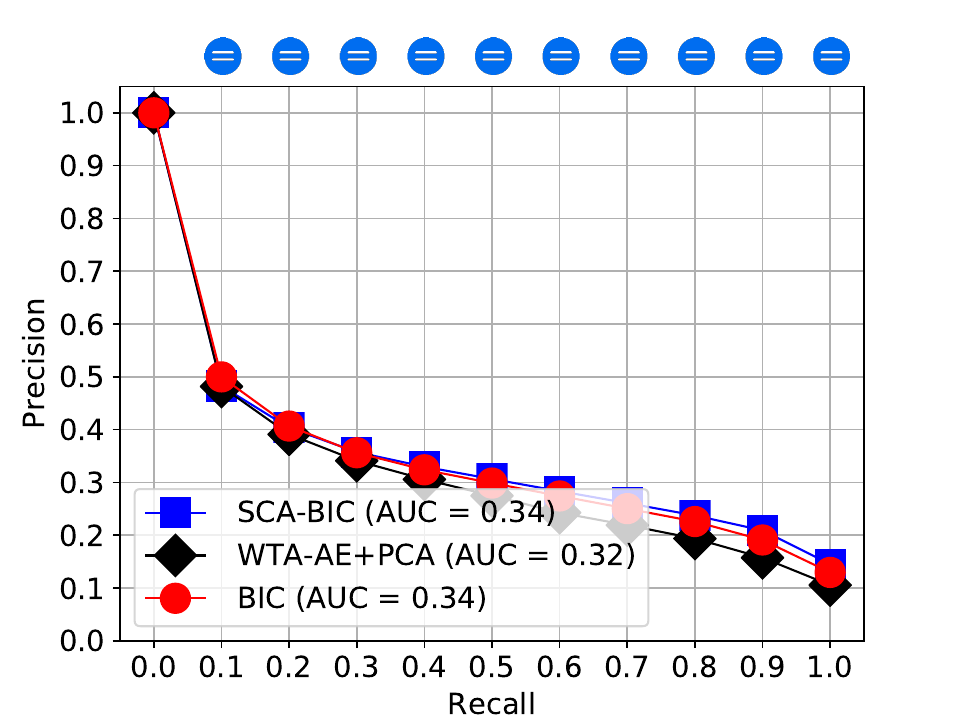}\label{chart:pr_bic_la_msrcorid_64}&
		\includegraphics[width=0.25\textwidth]{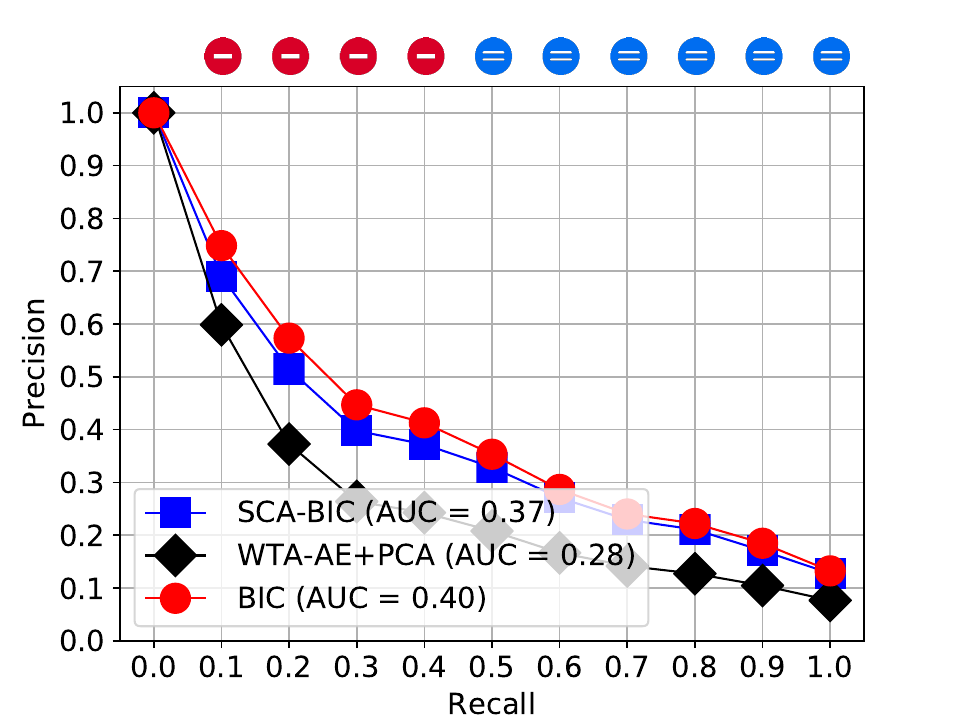}\label{chart:pr_bic_la_ucmerced_64}\\
		\rowname{96}&
		\includegraphics[width=0.25\textwidth]{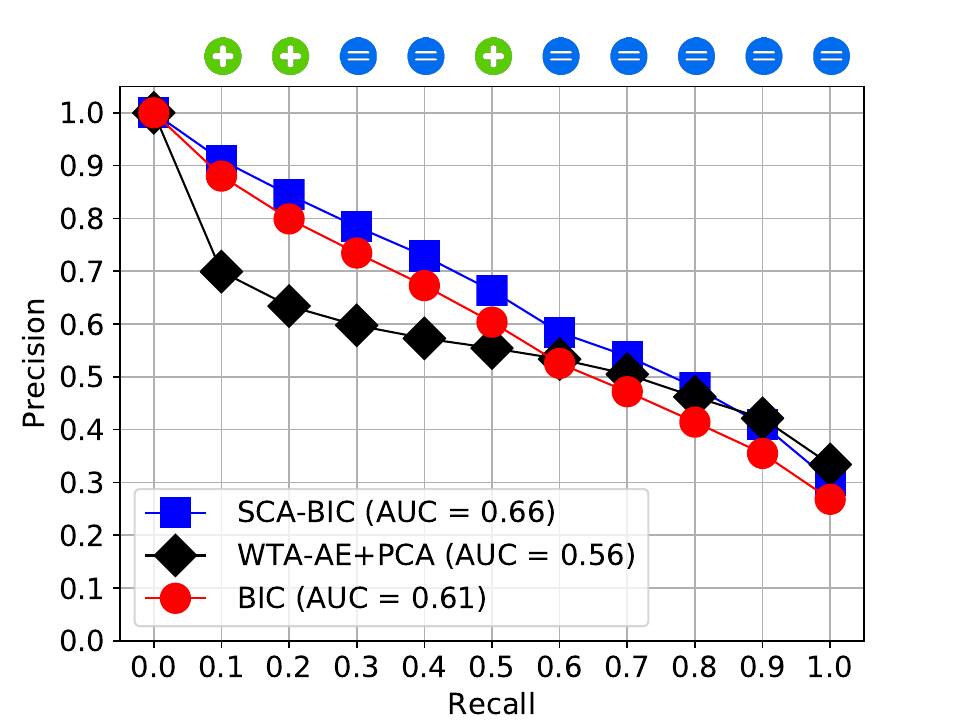}\label{chart:pr_bic_la_eth80_96}&
		\includegraphics[width=0.25\textwidth]{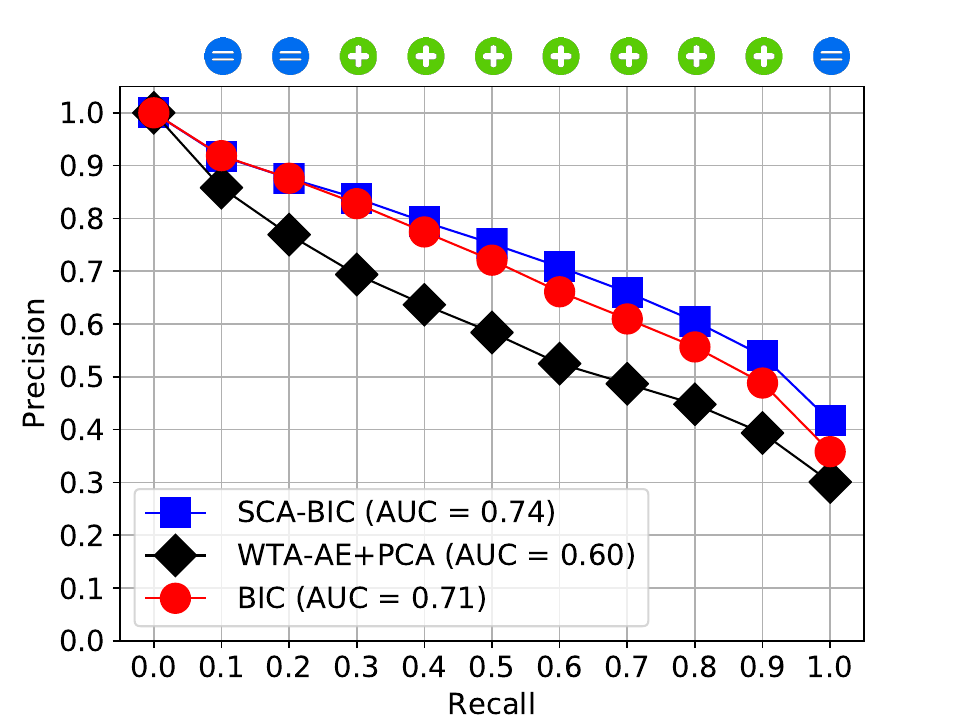}\label{chart:pr_bic_la_fruits__96}&
		\includegraphics[width=0.25\textwidth]{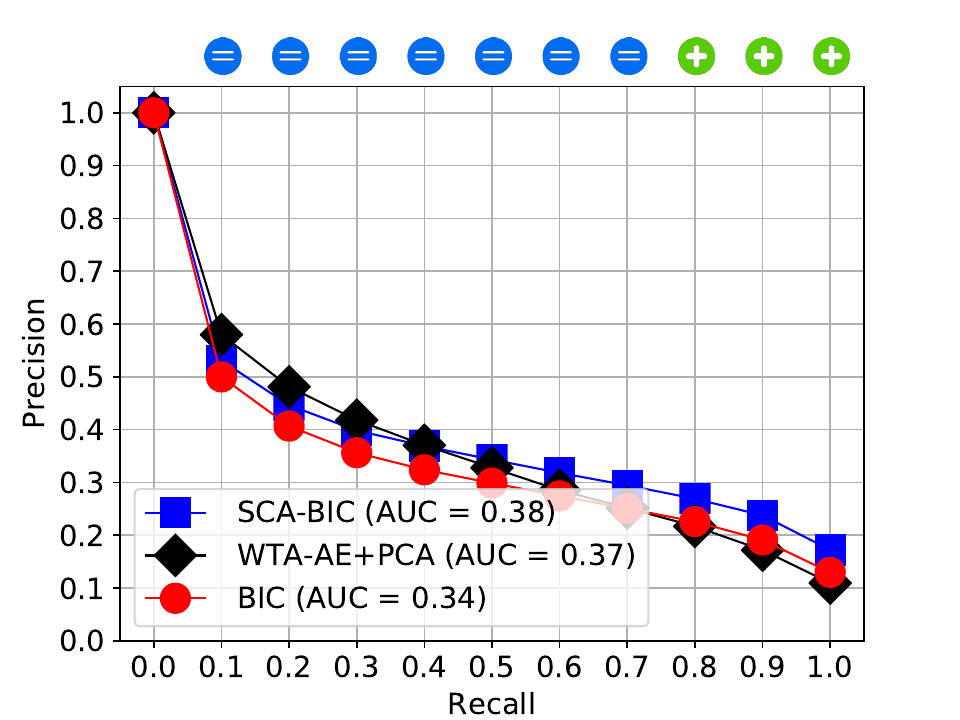}\label{chart:pr_bic_la_msrcorid_96}&
		\includegraphics[width=0.25\textwidth]{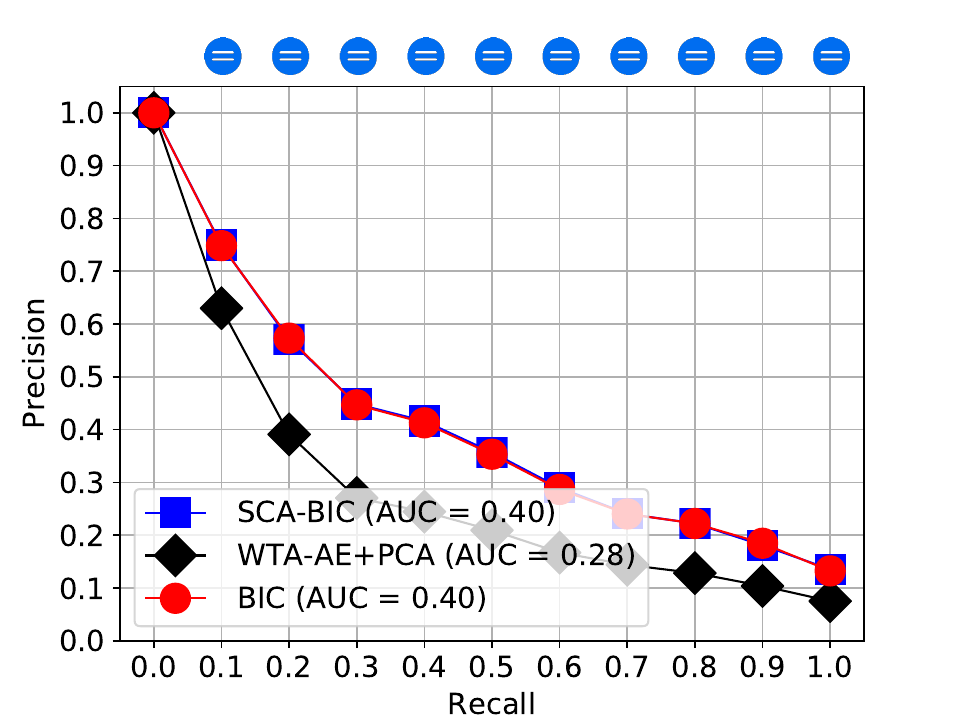}\label{chart:pr_bic_la_ucmerced_96}\\
		\rowname{128}&
		\includegraphics[width=0.25\textwidth]{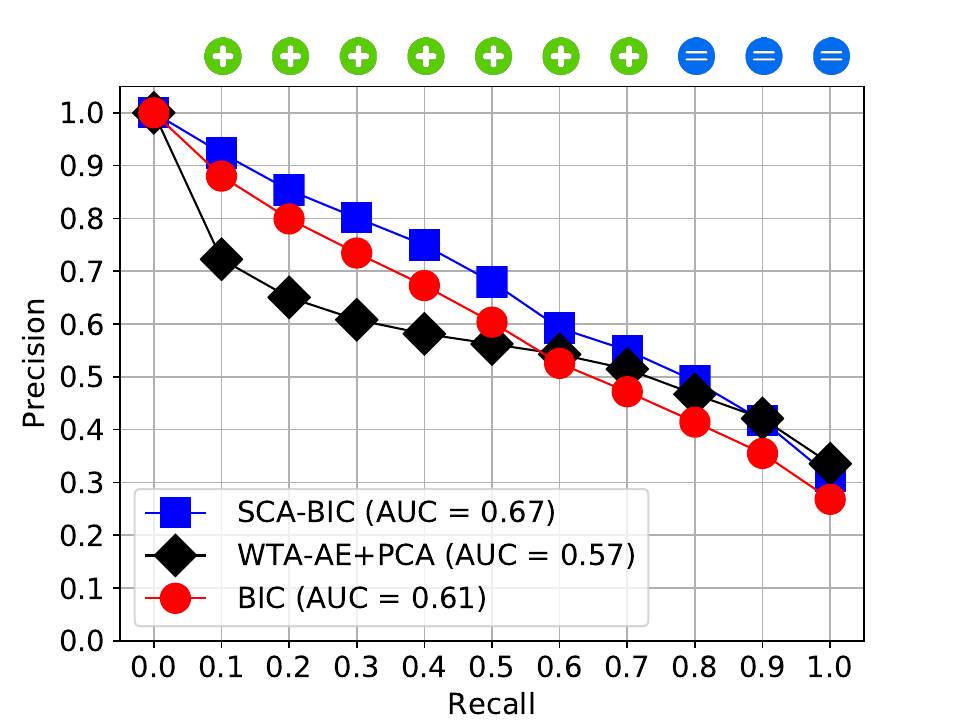}\label{chart:pr_bic_la_eth80_128}&
		\includegraphics[width=0.25\textwidth]{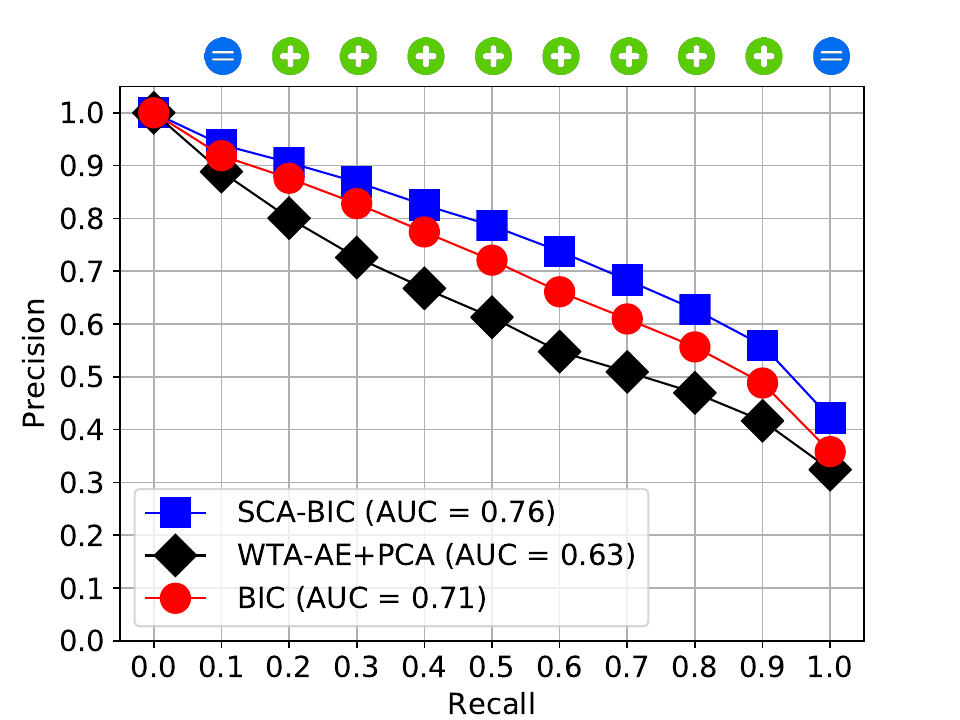}\label{chart:pr_bic_la_fruits__128}&
		\includegraphics[width=0.25\textwidth]{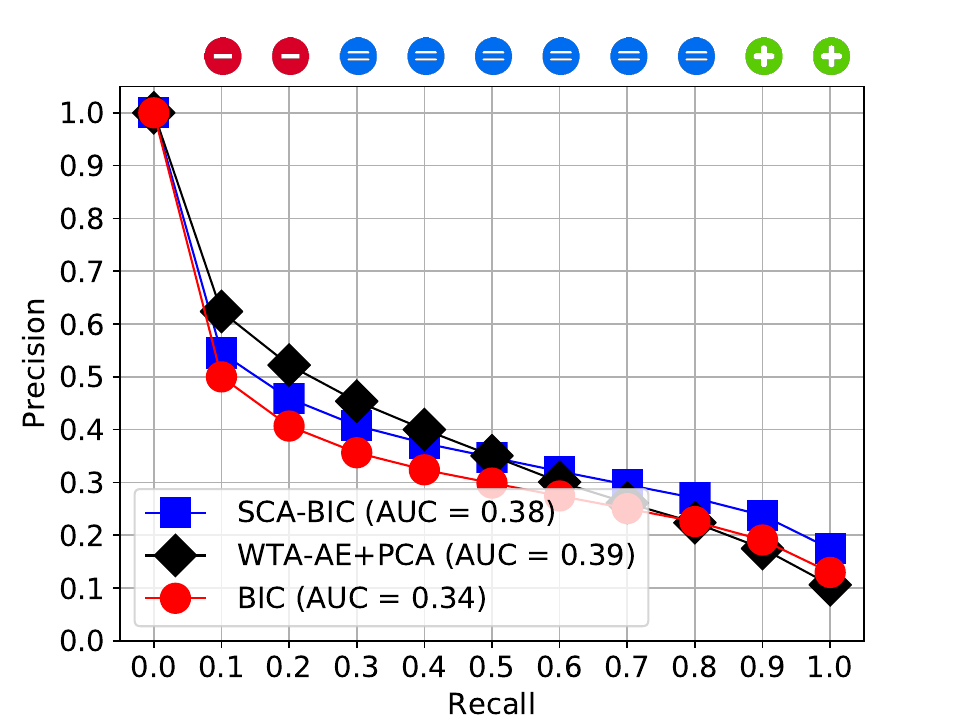}\label{chart:pr_bic_la_msrcorid_128}&
		\includegraphics[width=0.25\textwidth]{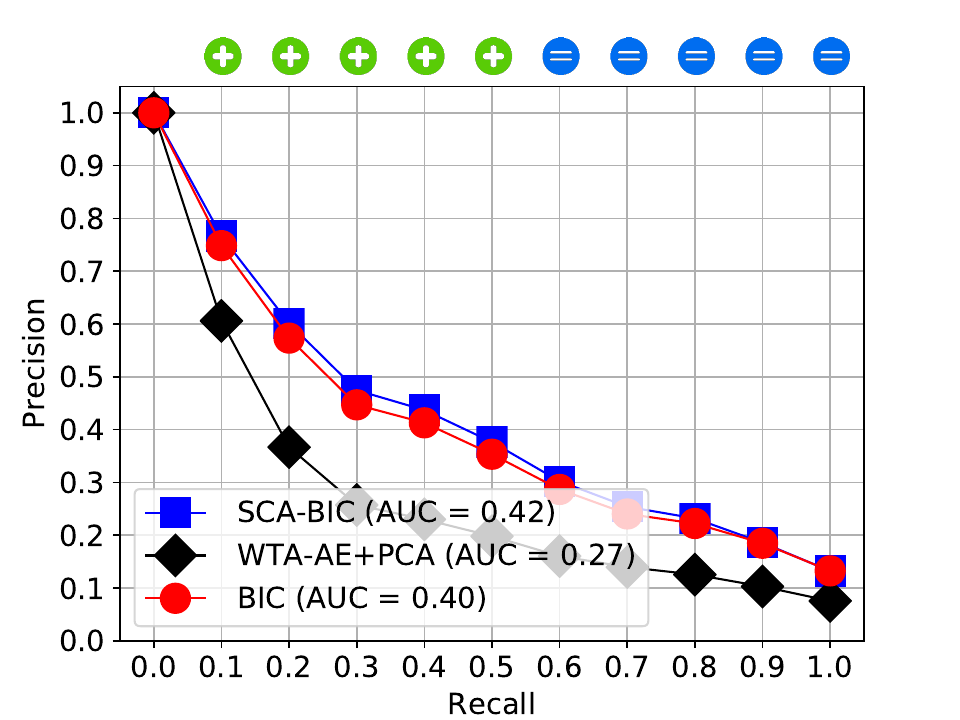}\label{chart:pr_bic_la_ucmerced_128}\\
		\rowname{256}&
		\includegraphics[width=0.25\textwidth]{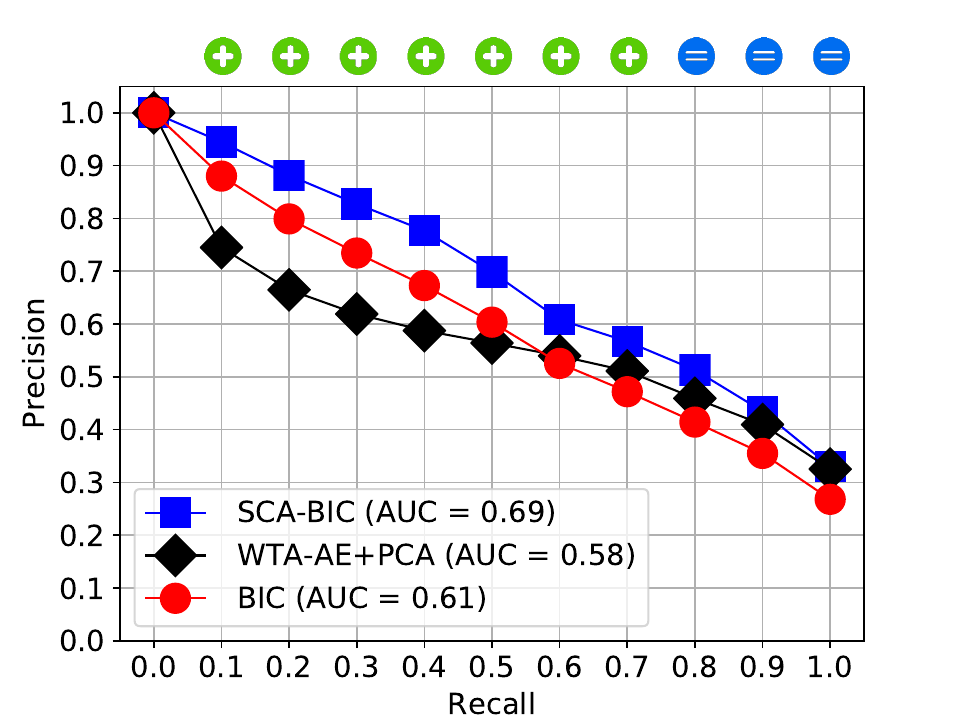}\label{chart:pr_bic_la_eth80_256}&
		\includegraphics[width=0.25\textwidth]{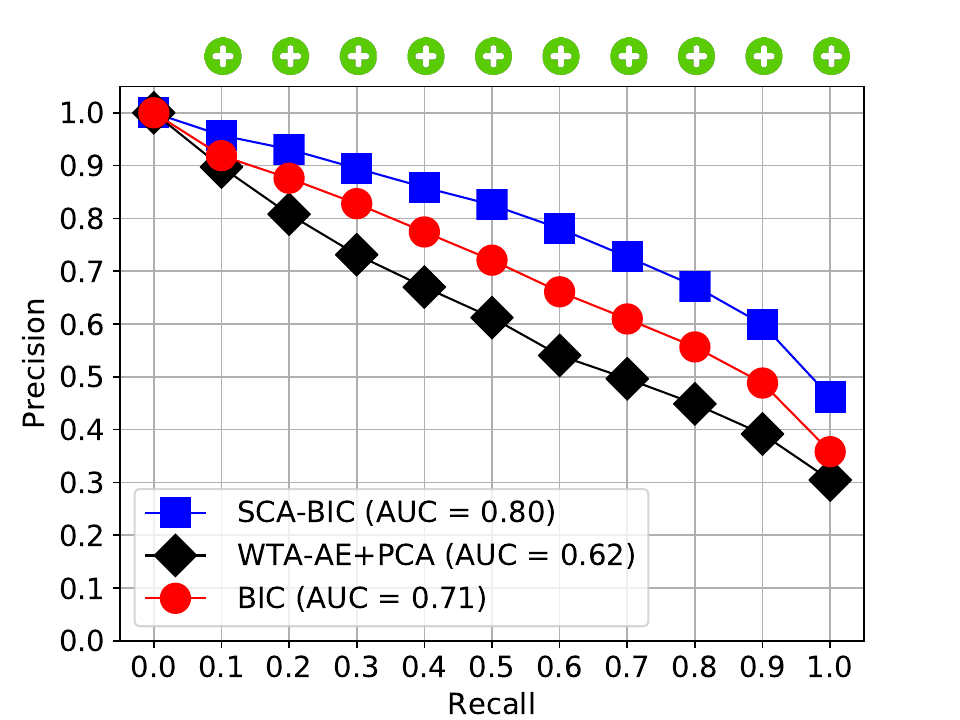}\label{chart:pr_bic_la_fruits__256}&
		\includegraphics[width=0.25\textwidth]{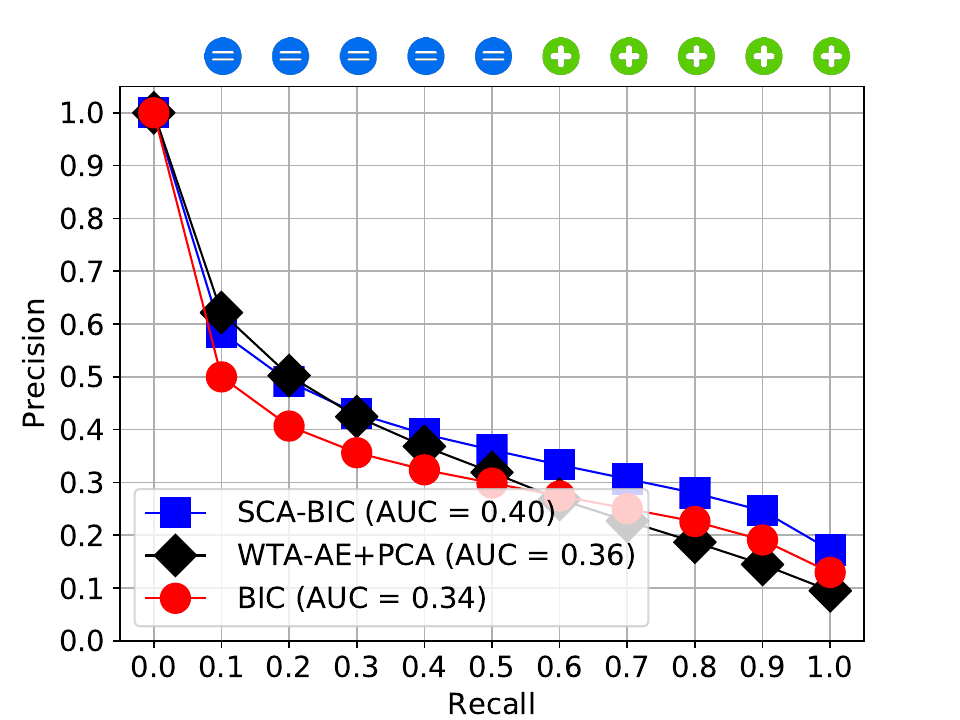}\label{chart:pr_bic_la_msrcorid_256}&
		\includegraphics[width=0.25\textwidth]{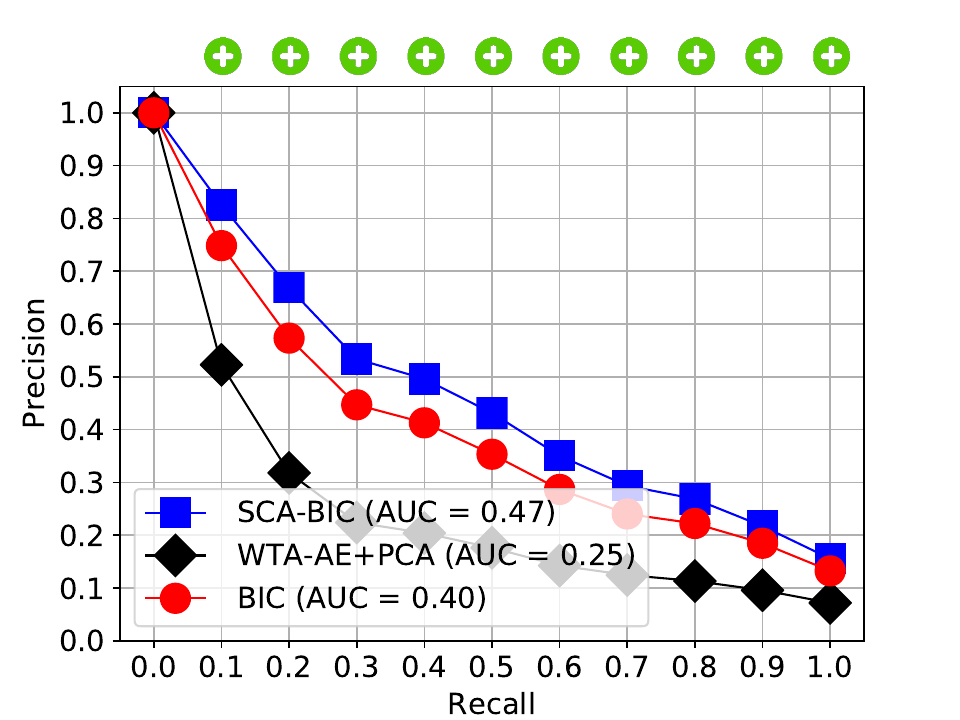}\label{chart:pr_bic_la_ucmerced_256}\\
		\rowname{384}&
		\includegraphics[width=0.25\textwidth]{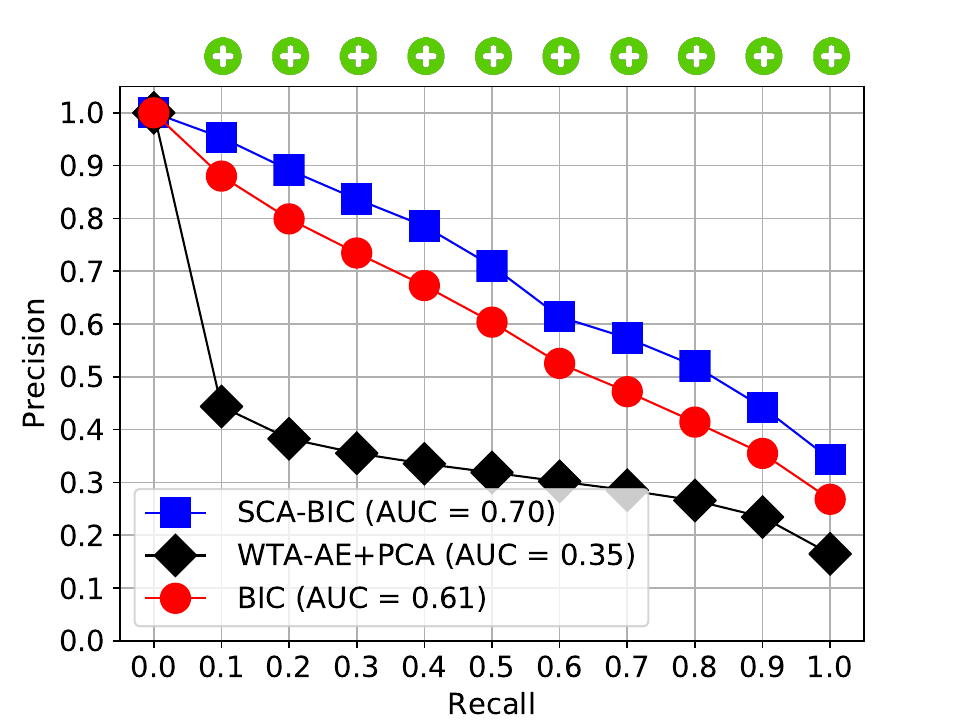}\label{chart:pr_bic_la_eth80_384}&
		\includegraphics[width=0.25\textwidth]{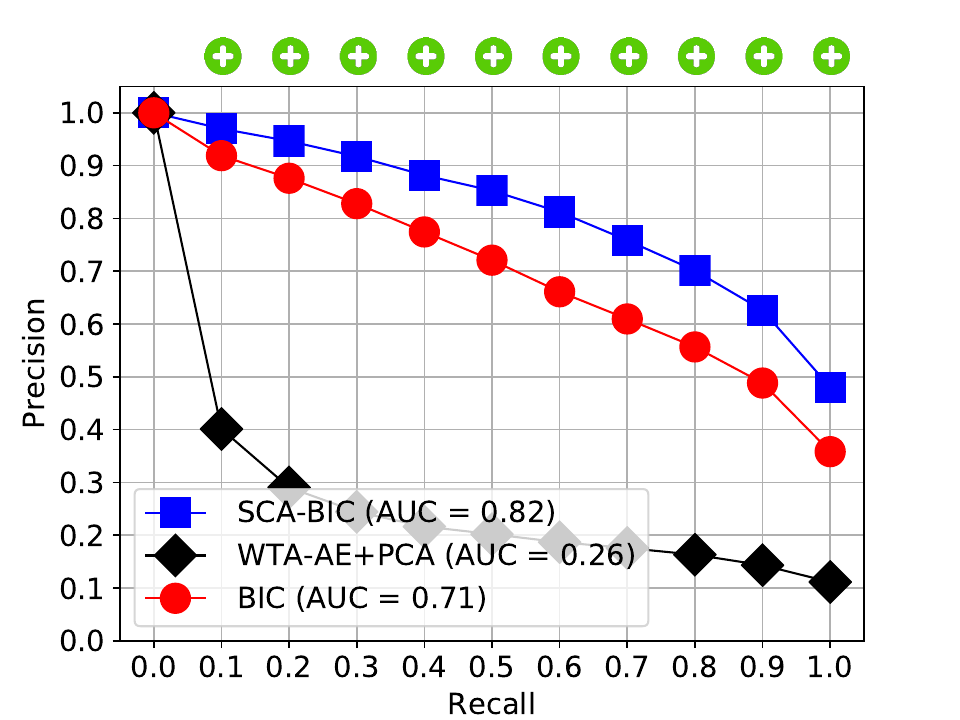}\label{chart:pr_bic_la_fruits__384}&
		\includegraphics[width=0.25\textwidth]{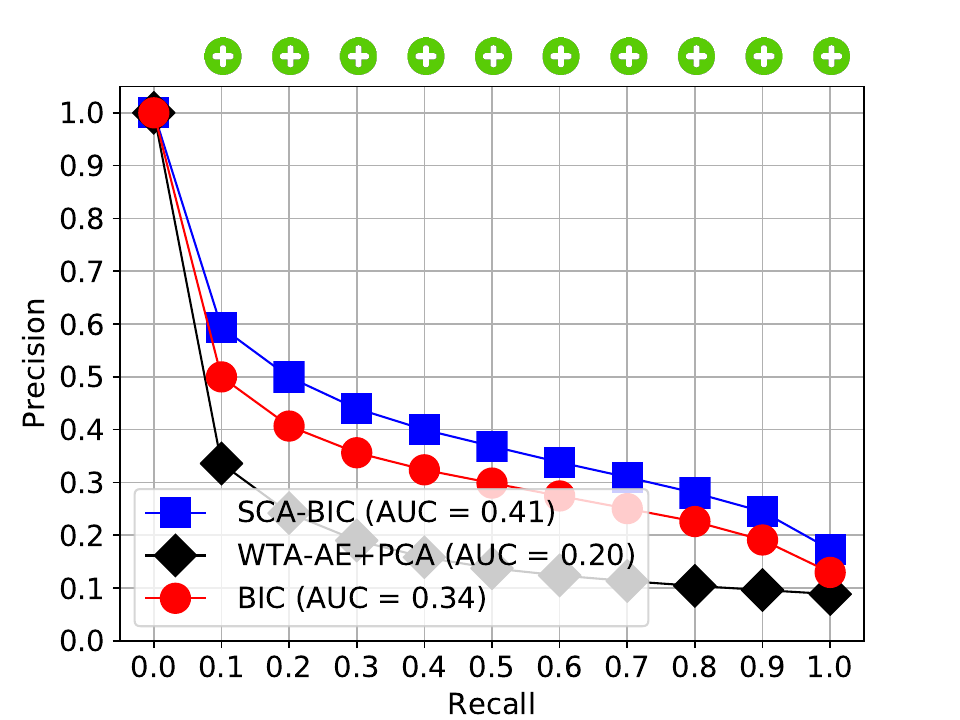}\label{chart:pr_bic_la_msrcorid_384}&
		\includegraphics[width=0.25\textwidth]{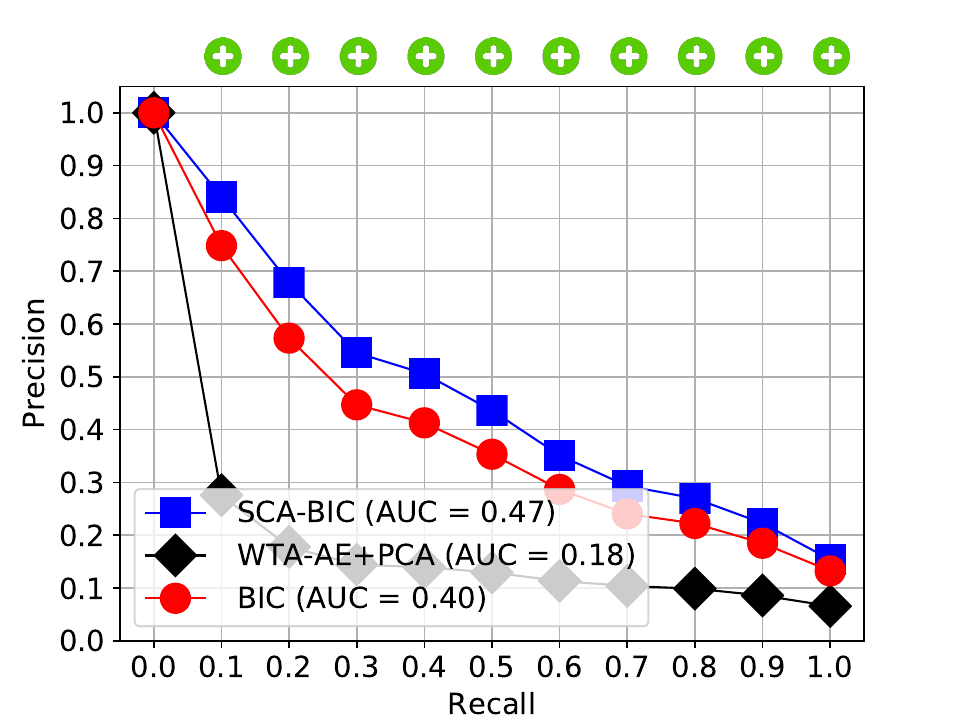}\label{chart:pr_bic_la_ucmerced_384}
	\end{tabular}
	\caption{Comparison between the Precision-Recall curves of SCA, WTA Autoencoder and BIC feature extractor considering all representation size limits for the datasets \textit{ETH-80}, \textit{Supermarket Produce}, \textit{MSRCORID}, and \textit{UCMerced Landuse}. 
	We recommend colourful printing for adequate visualization.}%
	\label{chart:pr_bic_la2}
\end{figure}

\begin{figure}[h]
	\centering
	\settoheight{\tempdima}{\includegraphics[width=.25\linewidth]{example-image-a}}%
	\begin{tabular}{@{}c@{}c@{}c@{}c@{}c@{}}
		\centering
		{}&Groundtruth  & Coil-100& Corel-1566& Corel-3906 \\
		\rowname{8}&
		\includegraphics[width=0.25\textwidth]{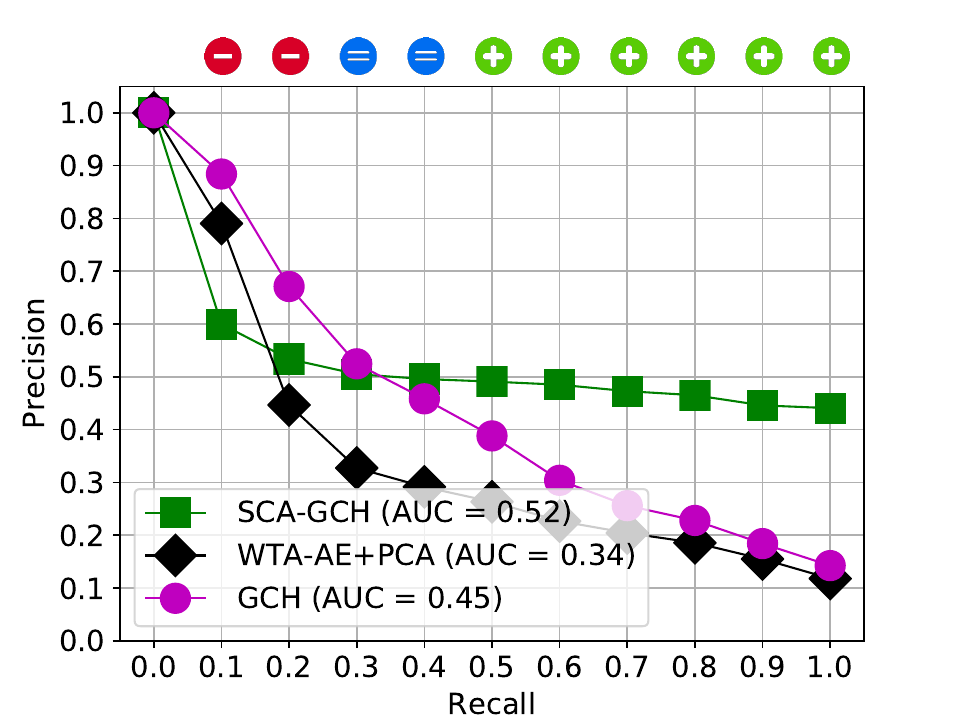}\label{chart:pr_gch_la_coffe_16}&
		\includegraphics[width=0.25\textwidth]{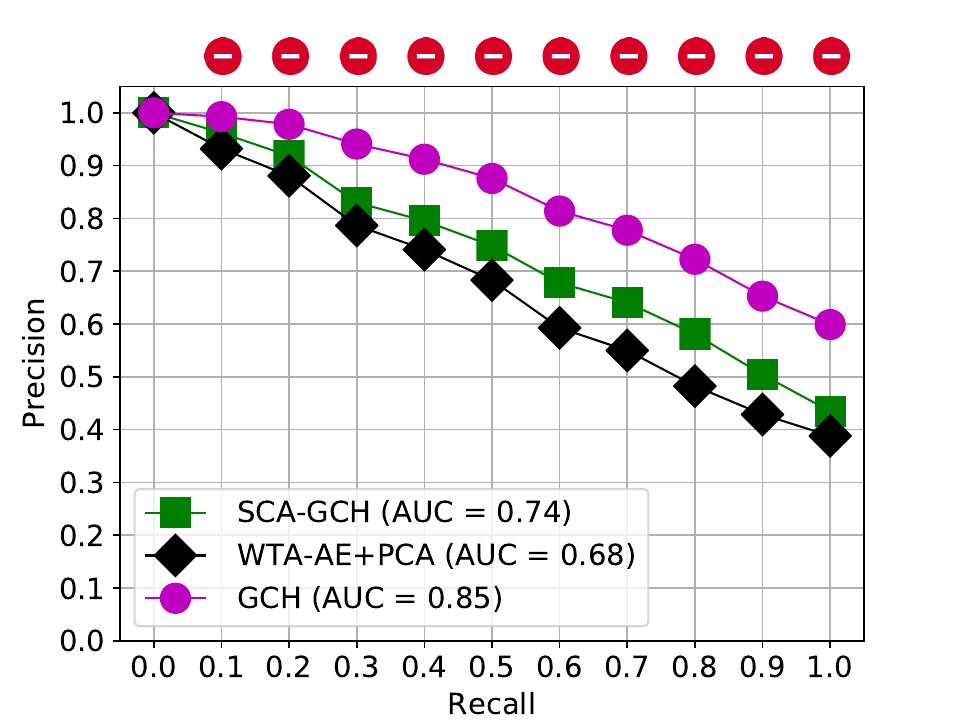}\label{chart:pr_gch_la_coil100_16}&
		\includegraphics[width=0.25\textwidth]{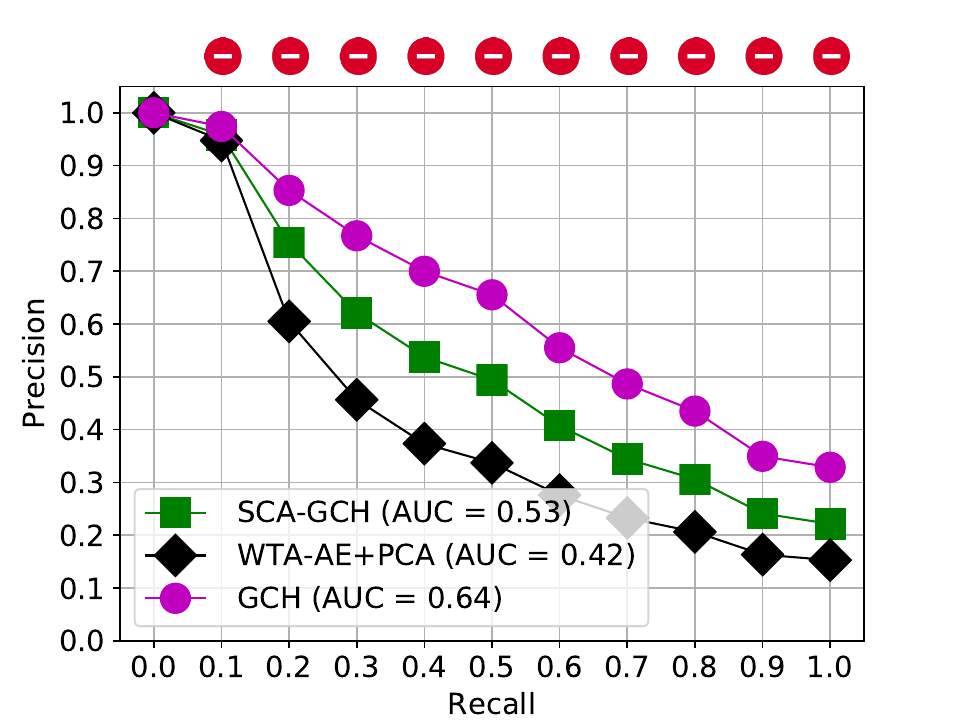}\label{chart:pr_gch_la_corel1566_16}&
		\includegraphics[width=0.25\textwidth]{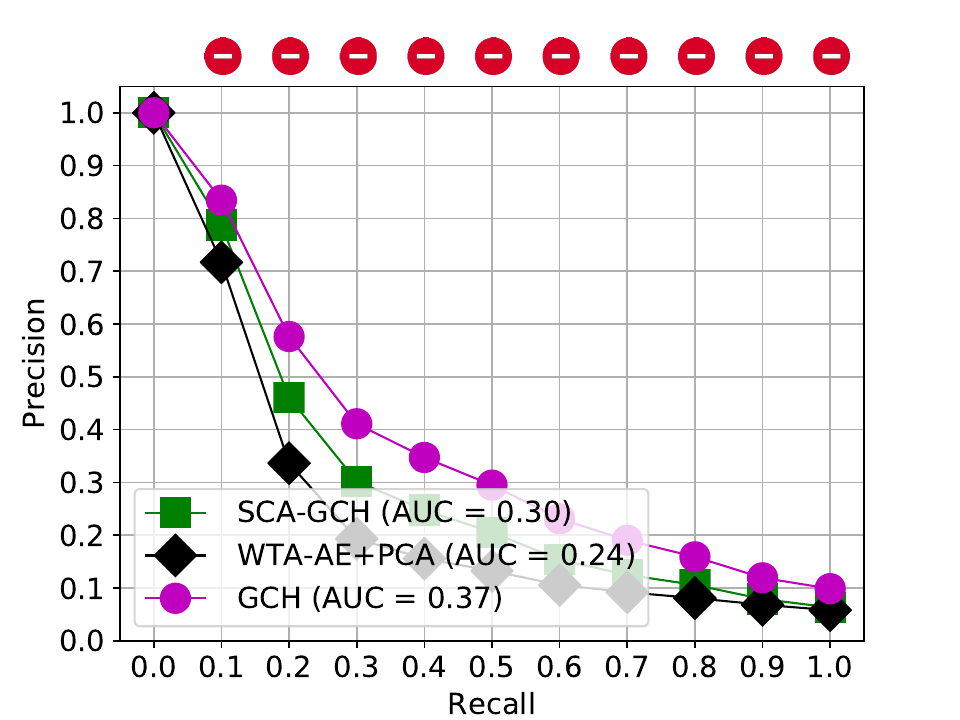}\label{chart:pr_gch_la_corel3909_16}\\
		\rowname{16}&
		\includegraphics[width=0.25\textwidth]{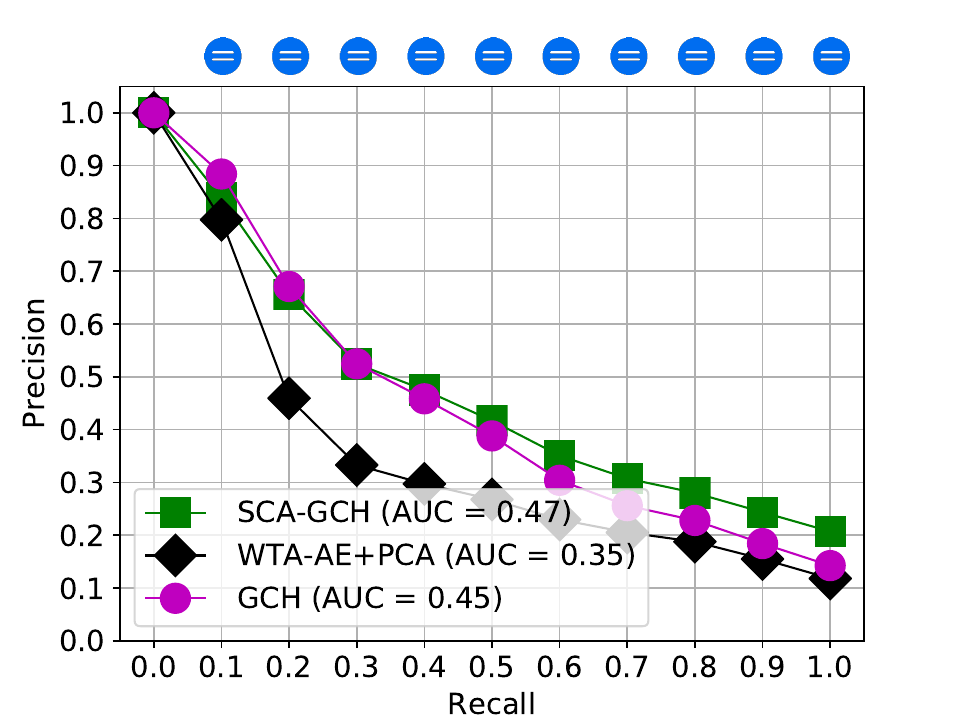}\label{chart:pr_gch_la_coffe_32}&
		\includegraphics[width=0.25\textwidth]{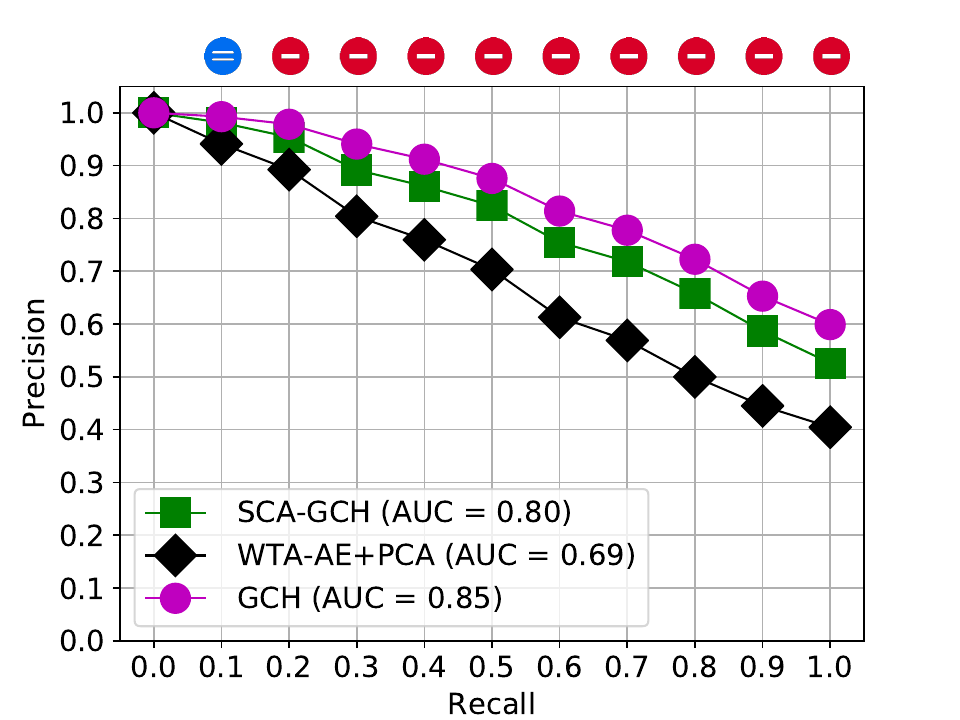}\label{chart:pr_gch_la_coil100_32}&
		\includegraphics[width=0.25\textwidth]{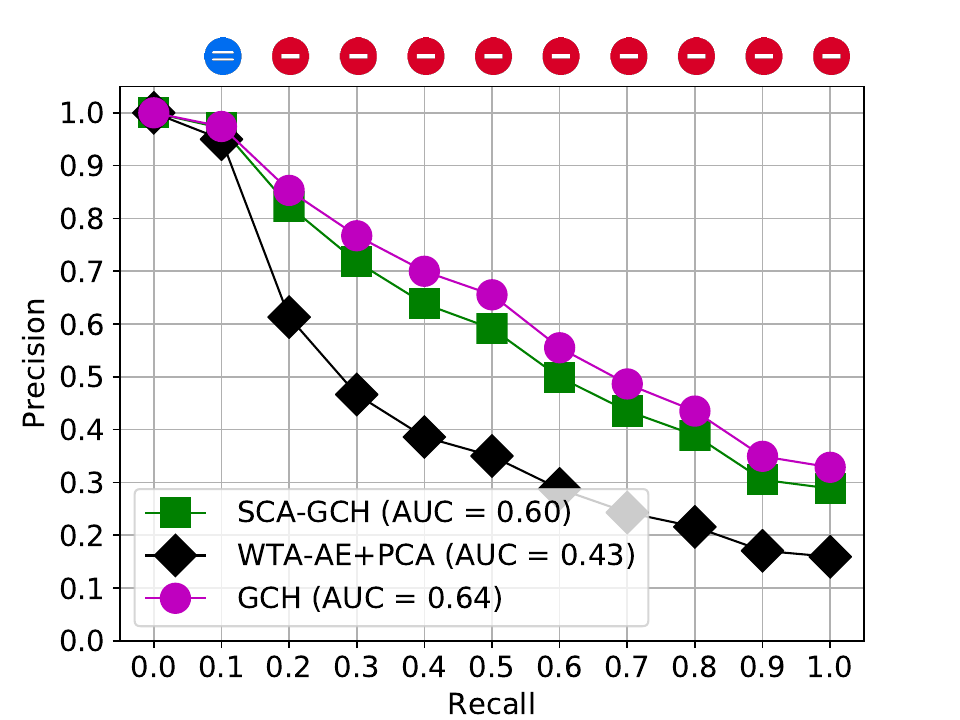}\label{chart:pr_gch_la_corel1566_32}&
		\includegraphics[width=0.25\textwidth]{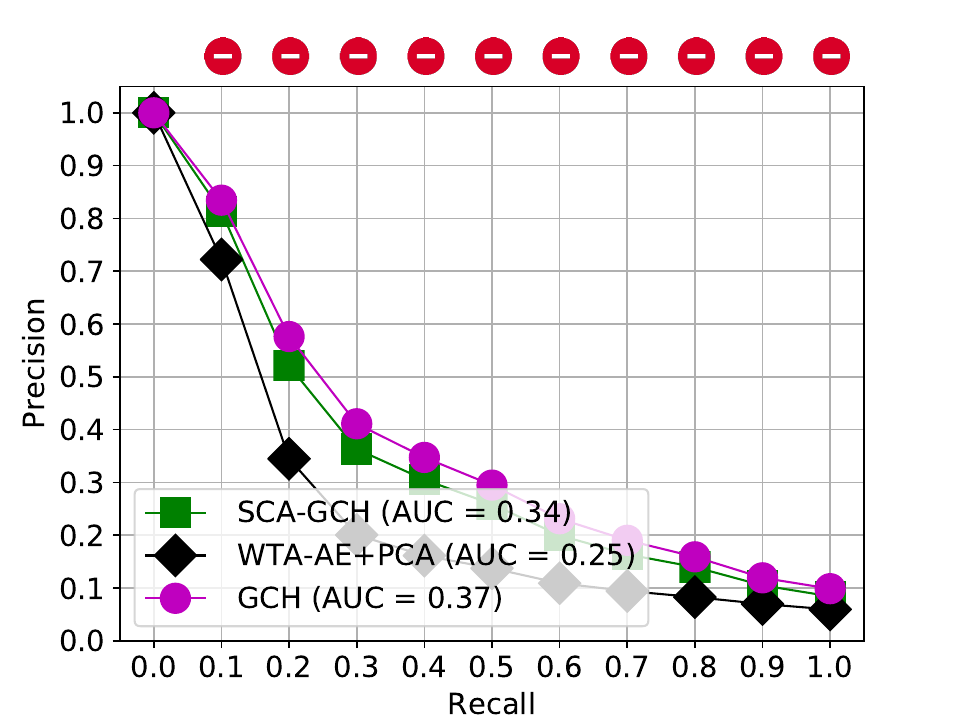}\label{chart:pr_gch_la_corel3909_32}\\
		\rowname{32}&
		\includegraphics[width=0.25\textwidth]{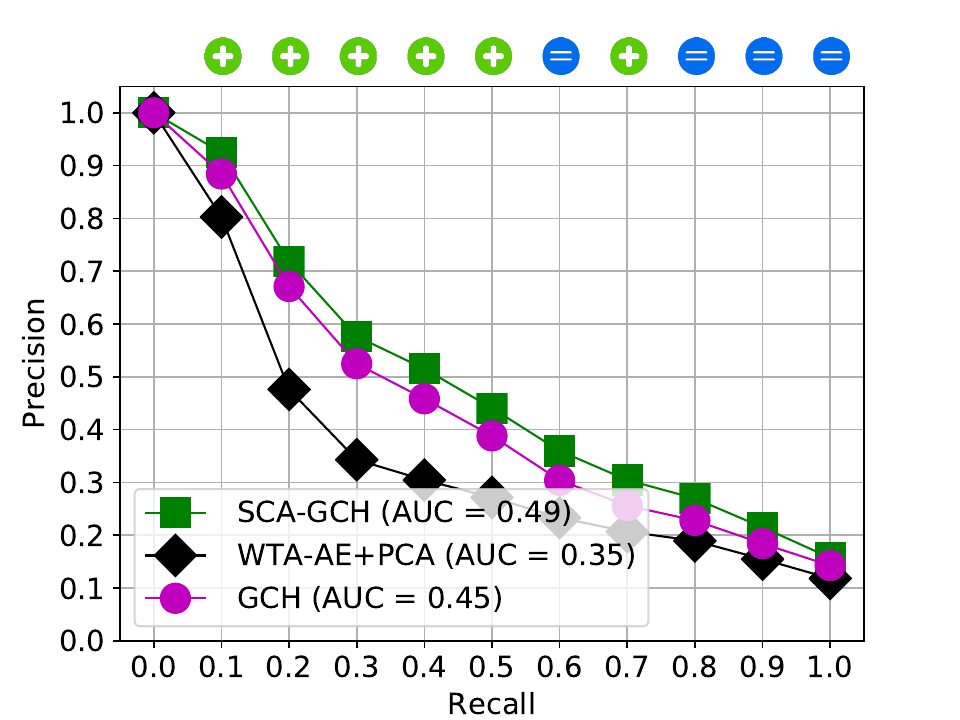}\label{chart:pr_gch_la_coffe_64}&
		\includegraphics[width=0.25\textwidth]{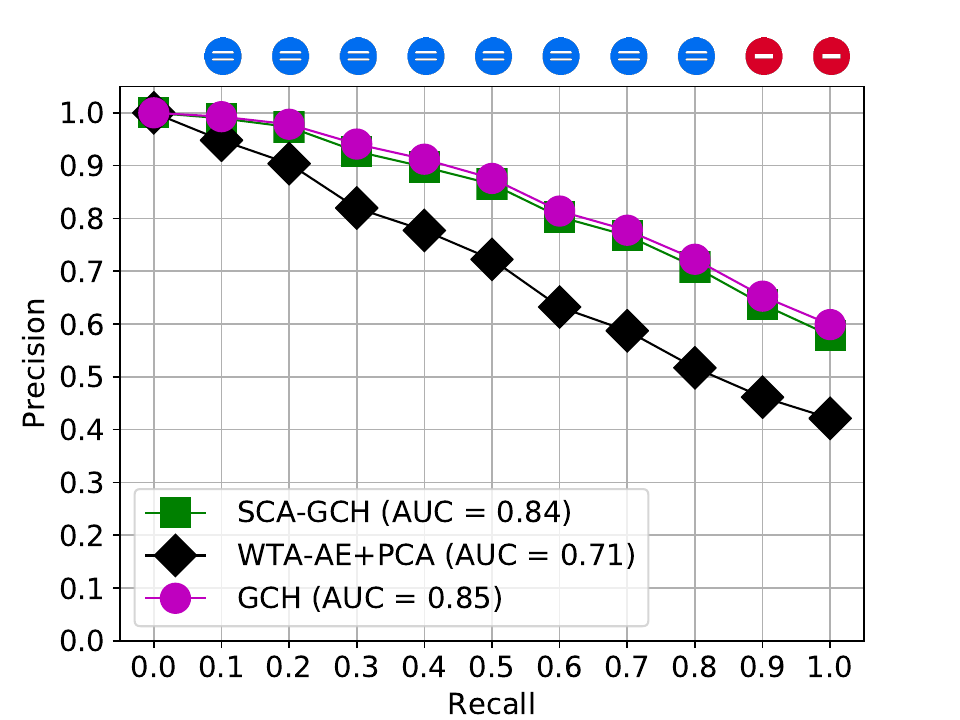}\label{chart:pr_gch_la_coil100_64}&
		\includegraphics[width=0.25\textwidth]{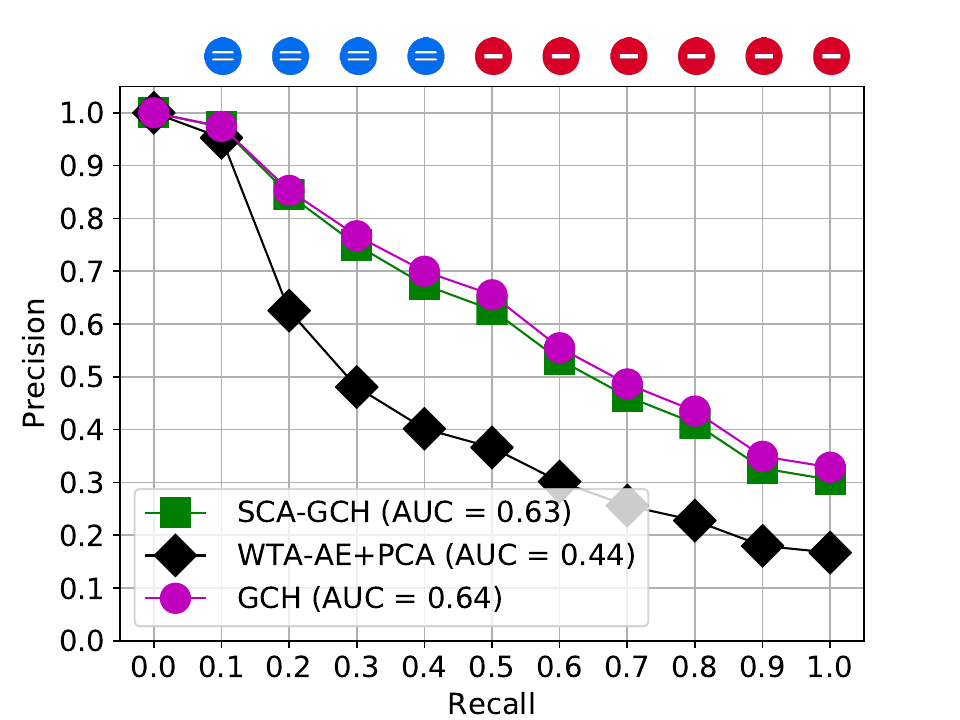}\label{chart:pr_gch_la_corel1566_64}&
		\includegraphics[width=0.25\textwidth]{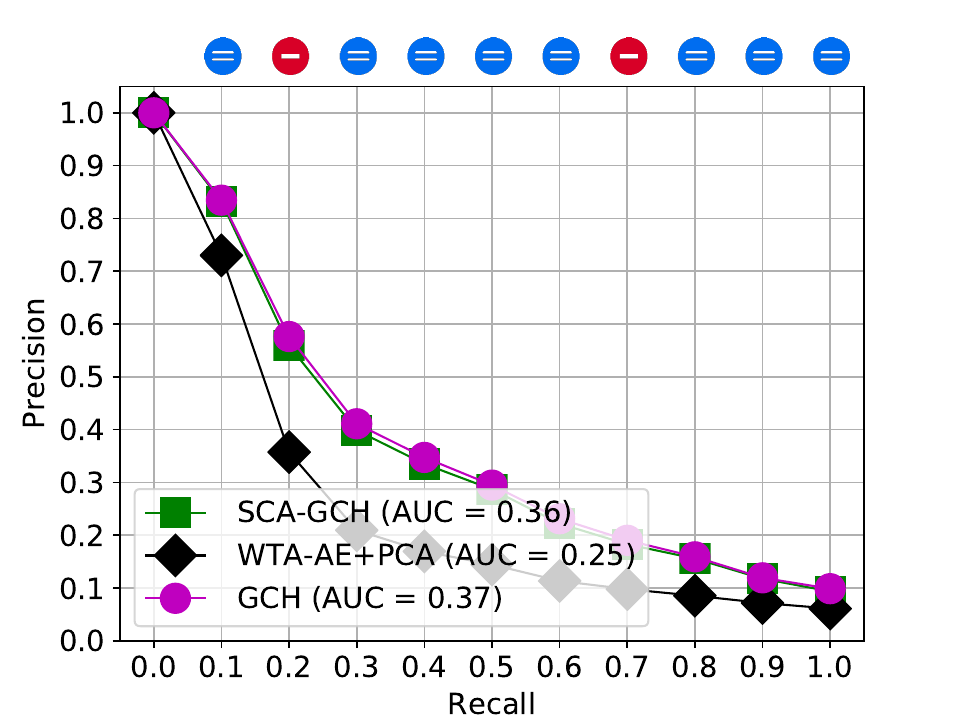}\label{chart:pr_gch_la_corel3909_64}\\
		\rowname{48}&
		\includegraphics[width=0.25\textwidth]{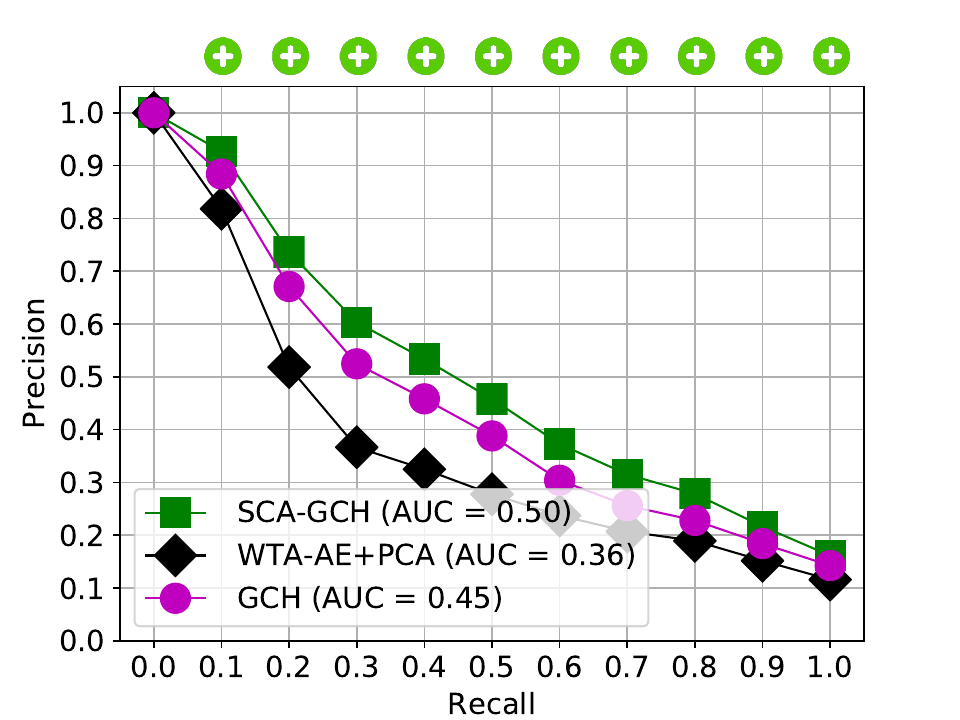}\label{chart:pr_gch_la_coffe_96}&
		\includegraphics[width=0.25\textwidth]{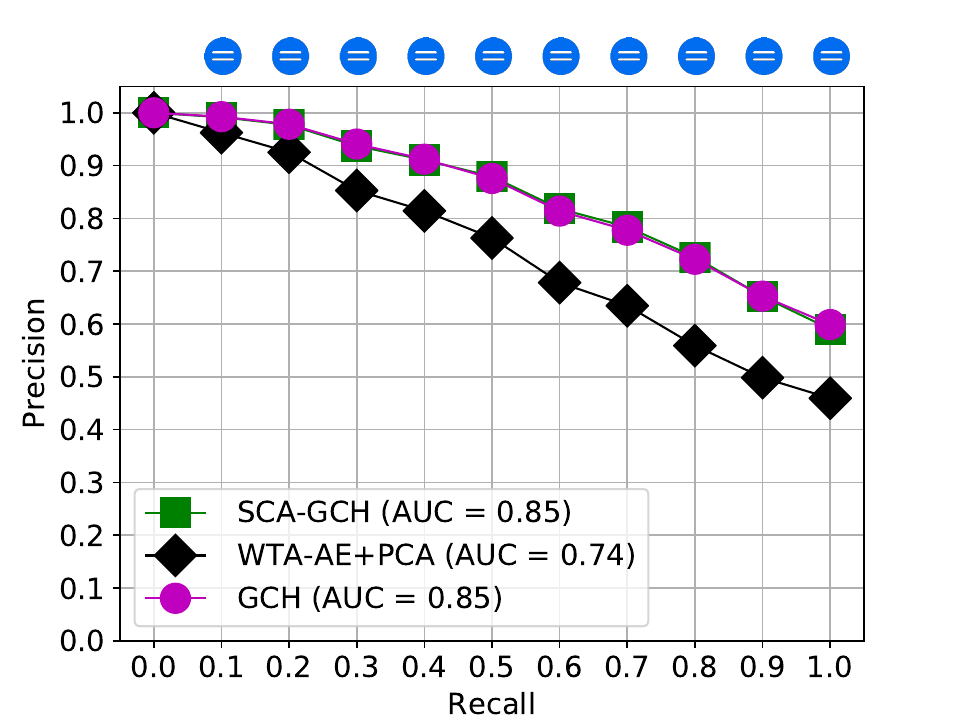}\label{chart:pr_gch_la_coil100_96}&
		\includegraphics[width=0.25\textwidth]{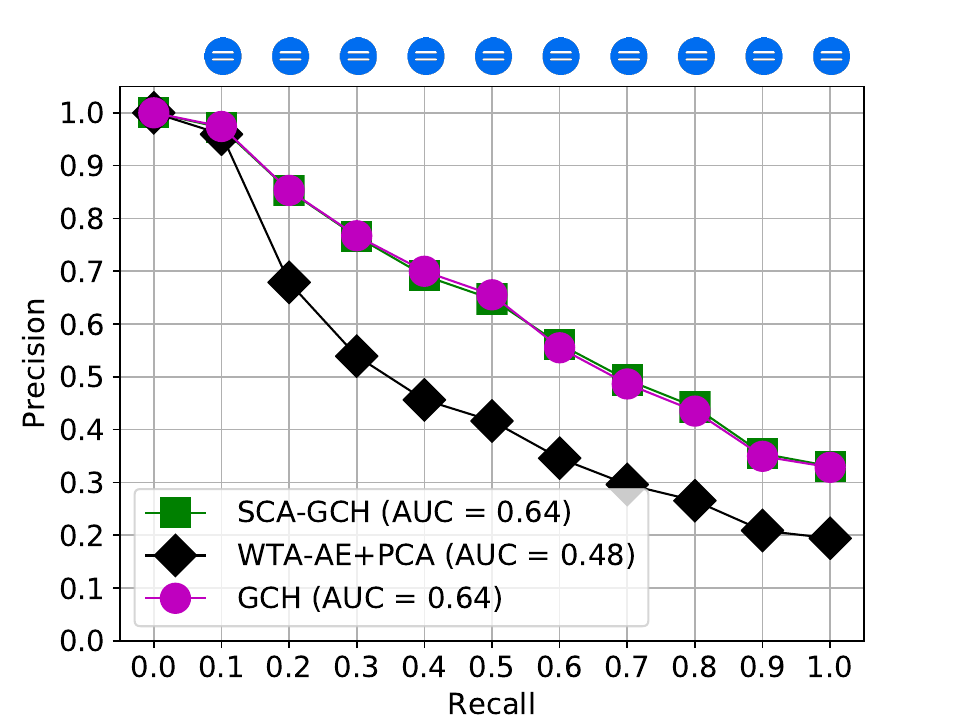}\label{chart:pr_gch_la_corel1566_96}&
		\includegraphics[width=0.25\textwidth]{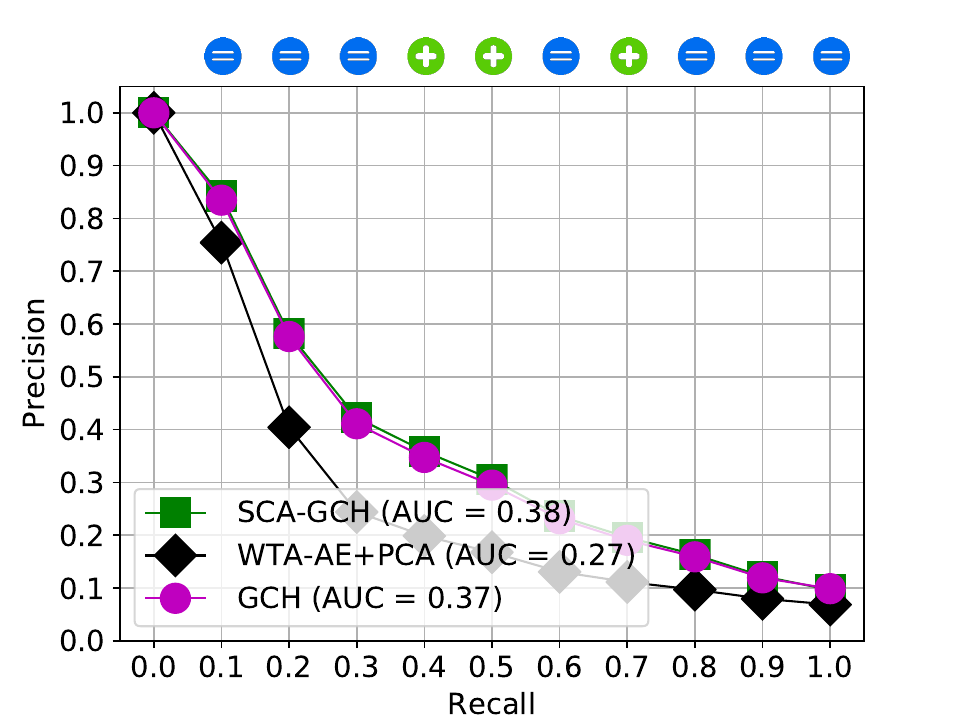}\label{chart:pr_gch_la_corel3909_96}\\
		\rowname{64}&
		\includegraphics[width=0.25\textwidth]{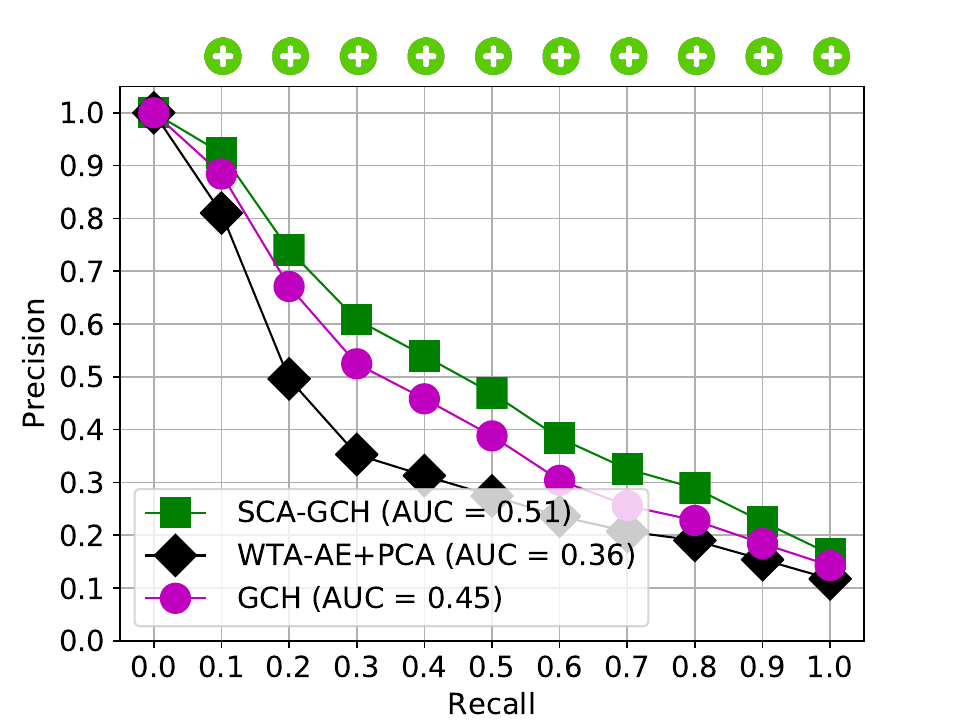}\label{chart:pr_gch_la_coffe_128}&
		\includegraphics[width=0.25\textwidth]{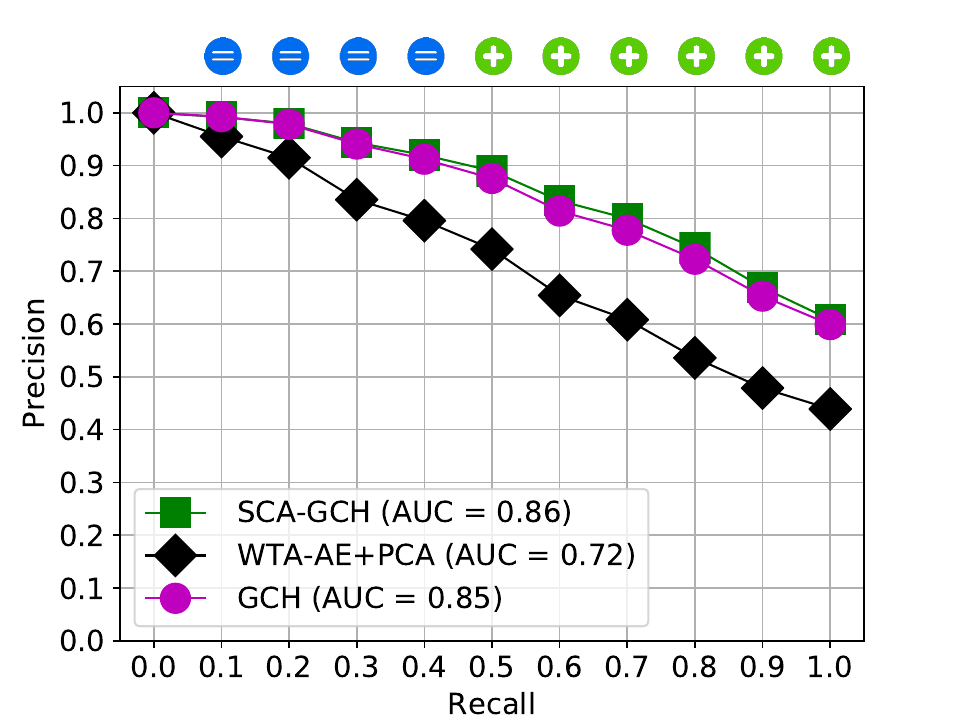}\label{chart:pr_gch_la_coil100_128}&
		\includegraphics[width=0.25\textwidth]{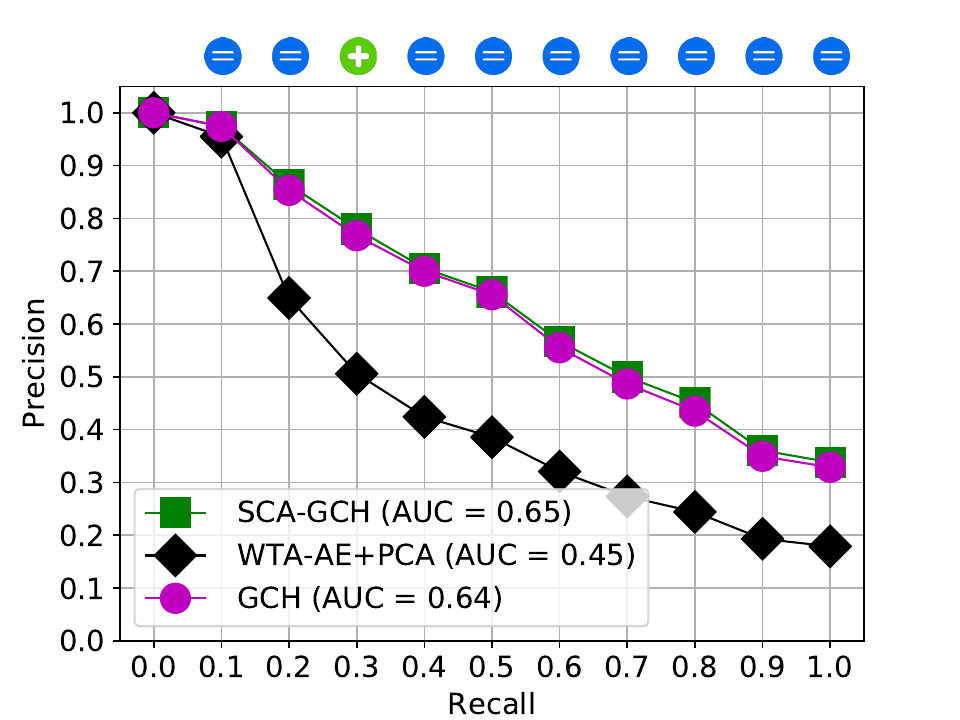}\label{chart:pr_gch_la_corel1566_128}&
		\includegraphics[width=0.25\textwidth]{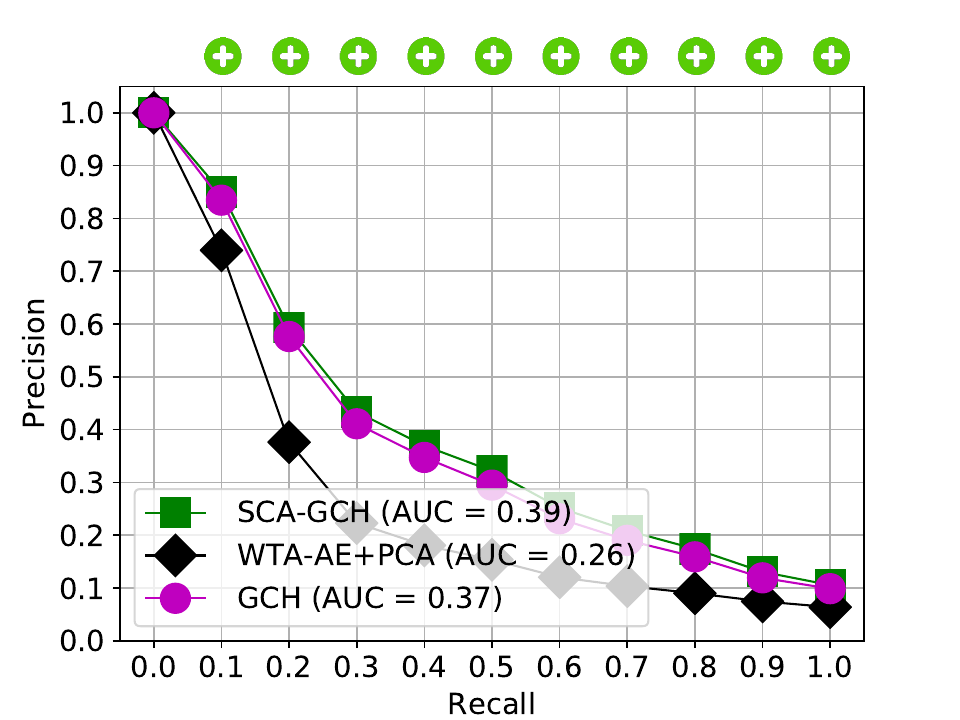}\label{chart:pr_gch_la_corel3909_128}\\
		\rowname{128}&
		\includegraphics[width=0.25\textwidth]{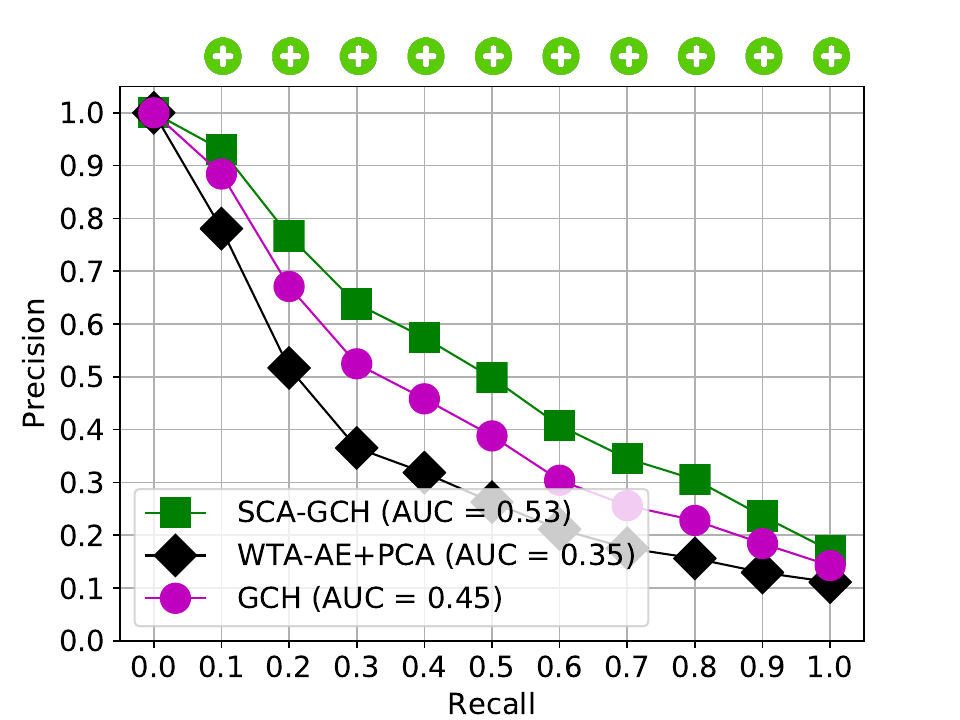}\label{chart:pr_gch_la_coffe_256}&
		\includegraphics[width=0.25\textwidth]{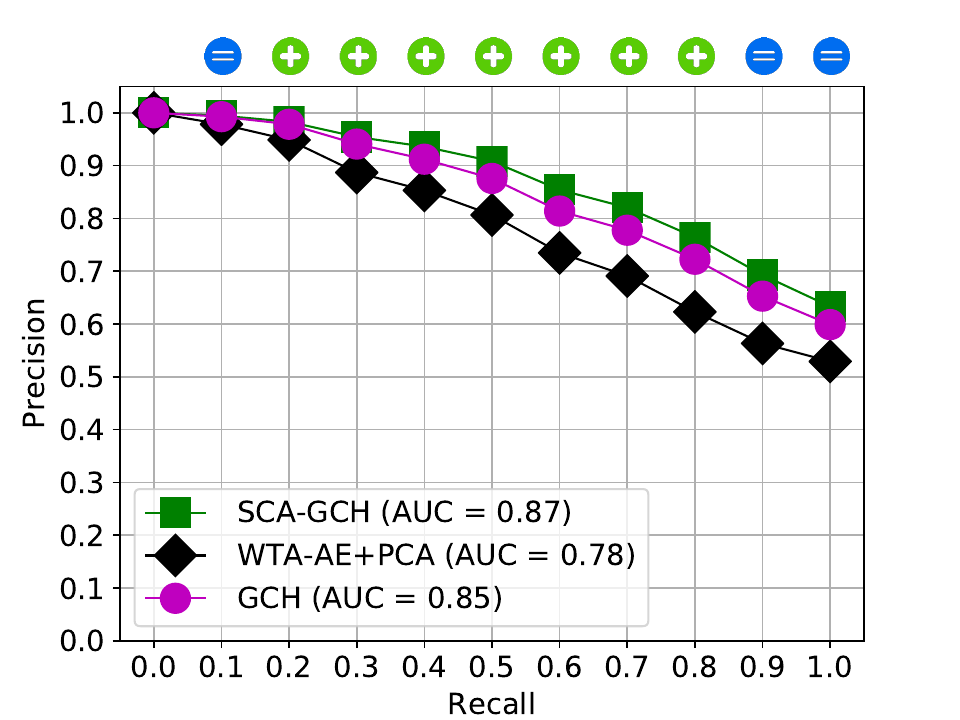}\label{chart:pr_gch_la_coil100_256}&
		\includegraphics[width=0.25\textwidth]{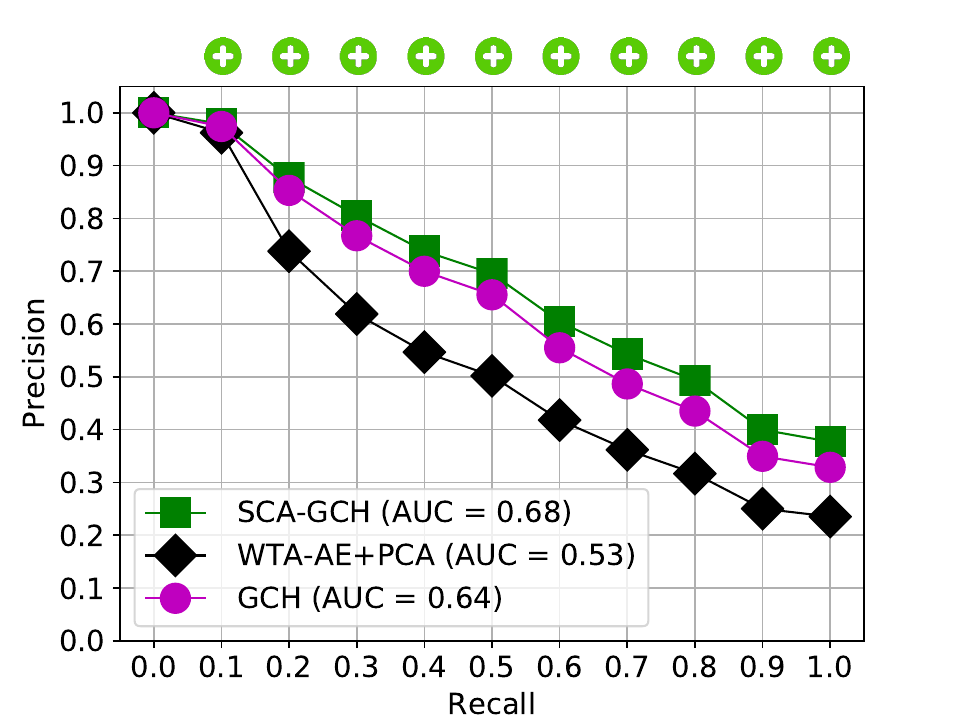}\label{chart:pr_gch_la_corel1566_256}&
		\includegraphics[width=0.25\textwidth]{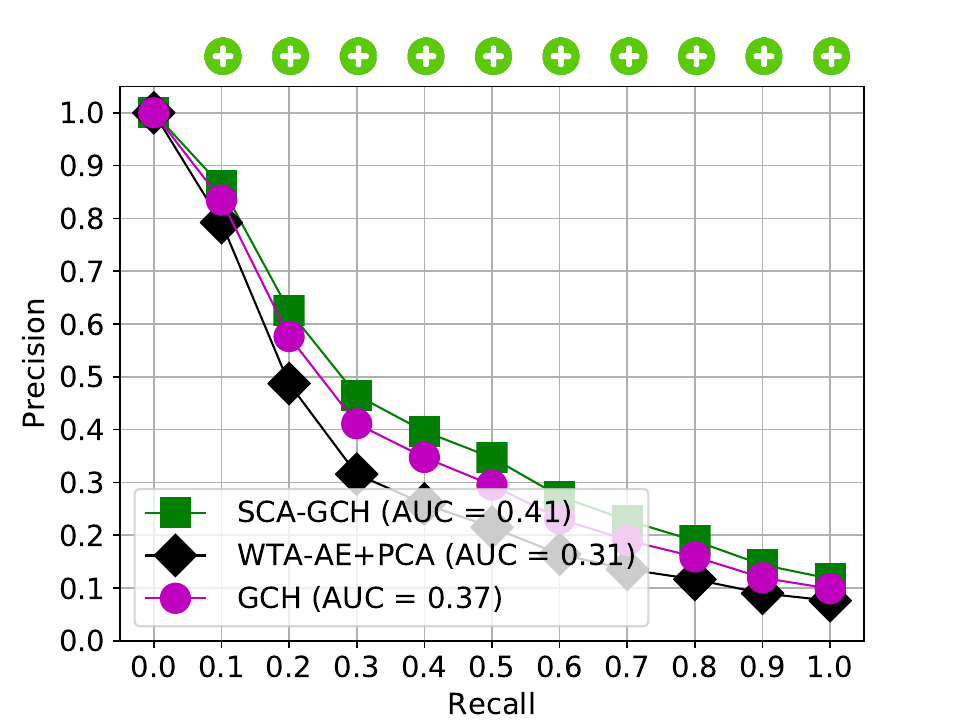}\label{chart:pr_gch_la_corel3909_256}\\
		\rowname{192}&
		\includegraphics[width=0.25\textwidth]{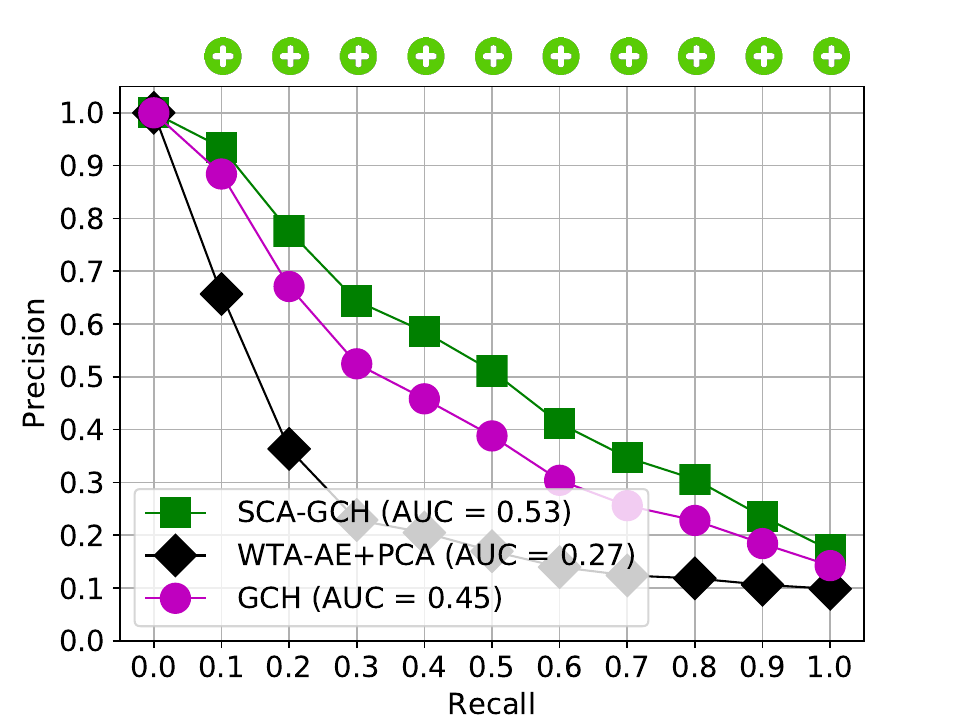}\label{chart:pr_gch_la_coffe_384}&
		\includegraphics[width=0.25\textwidth]{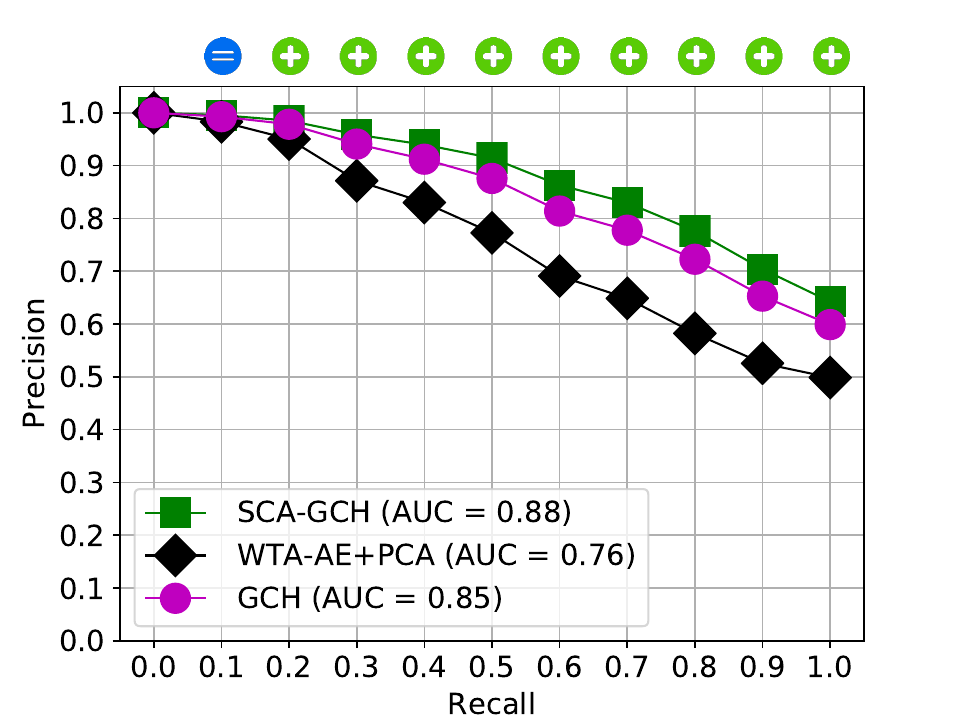}\label{chart:pr_gch_la_coil100_384}&
		\includegraphics[width=0.25\textwidth]{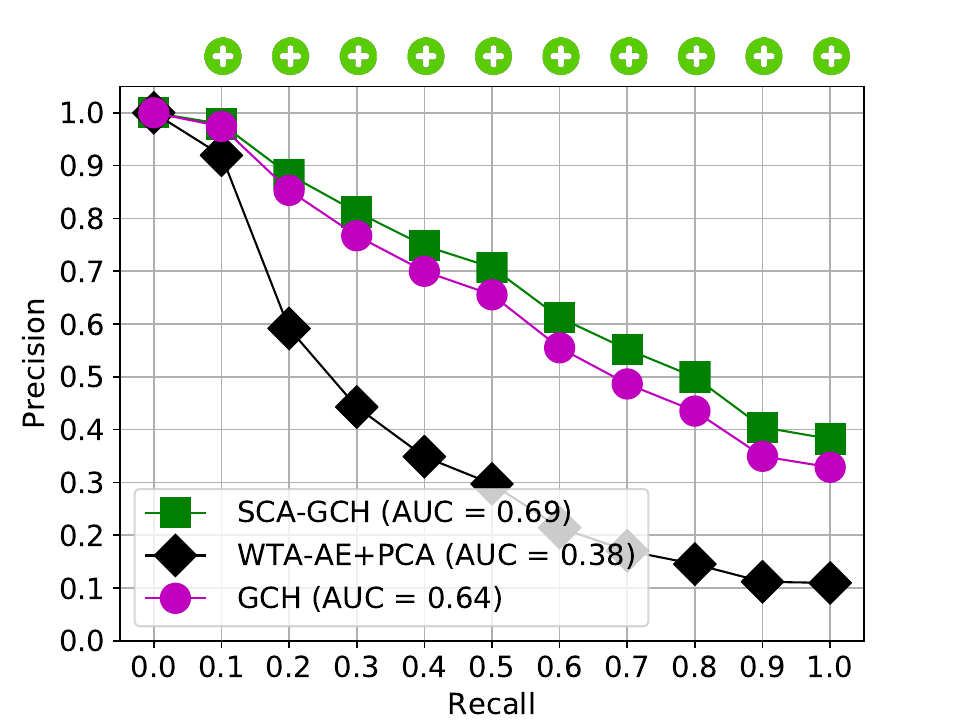}\label{chart:pr_gch_la_corel1566_384}&
		\includegraphics[width=0.25\textwidth]{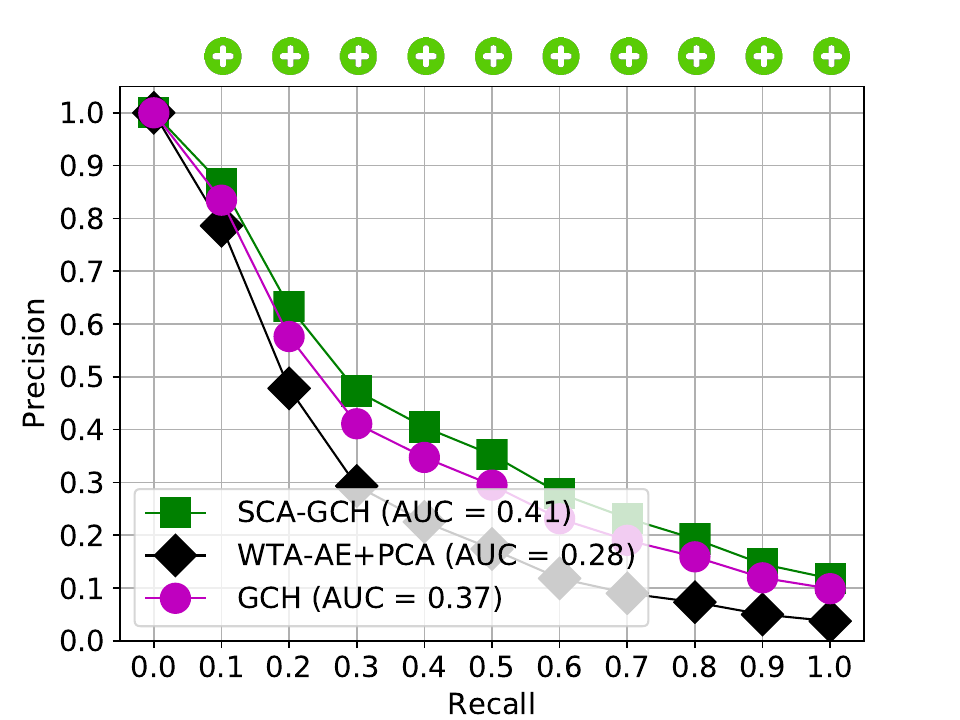}\label{chart:pr_gch_la_corel3909_384}
	\end{tabular}
	\caption{Comparison between the Precision-Recall curves of SCA, WTA Autoencoder and  GCH feature extractor considering all representation size limits for the datasets 
		\textit{Groundtruth}, \textit{Coil-100}, \textit{Corel-1566}, and \textit{Corel-3906}. 
	We recommend colourful printing for adequate visualization.}%
	\label{chart:pr_gch_la1}
\end{figure}

\begin{figure}[h]
	\centering
	\settoheight{\tempdima}{\includegraphics[width=.25\linewidth]{example-image-a}}%
	\begin{tabular}{@{}c@{}c@{}c@{}c@{}c@{}}
		\centering
		{}&ETH-80 & Supermarket P. & MSRCORID & UCMerced L. \\
		\rowname{8}&
		\includegraphics[width=0.25\textwidth]{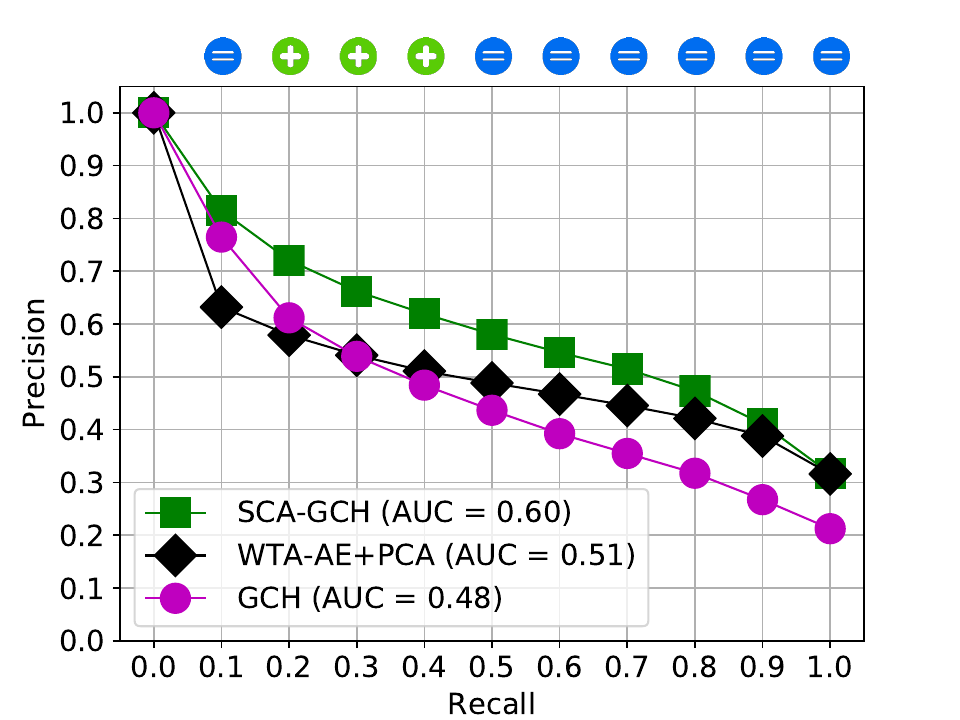}\label{chart:pr_gch_la_eth80_16}&
		\includegraphics[width=0.25\textwidth]{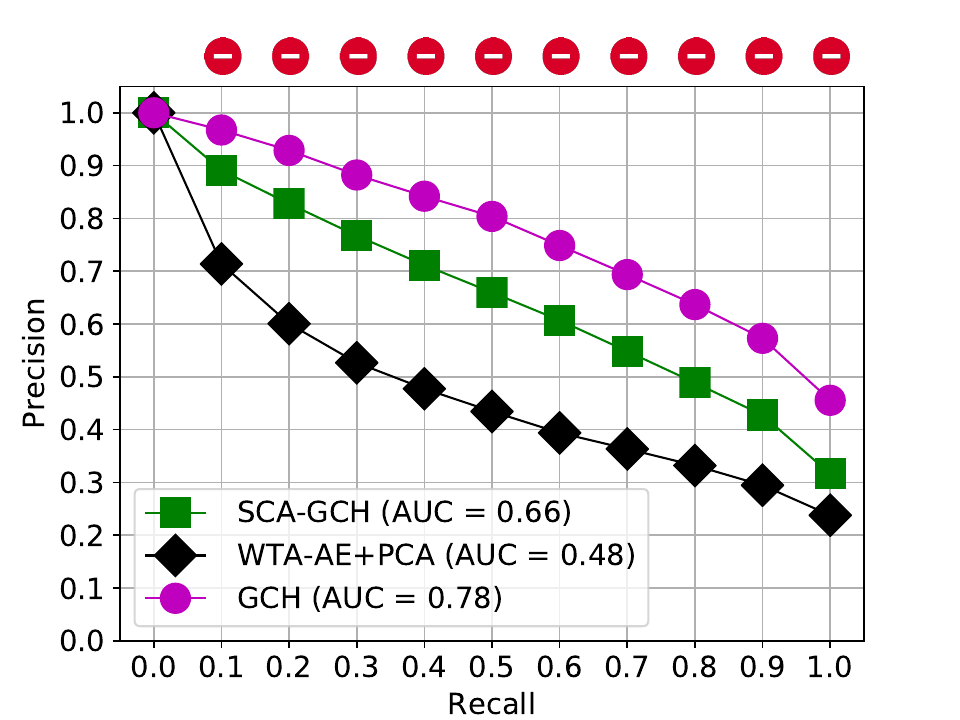}\label{chart:pr_gch_la_fruits_16}&
		\includegraphics[width=0.25\textwidth]{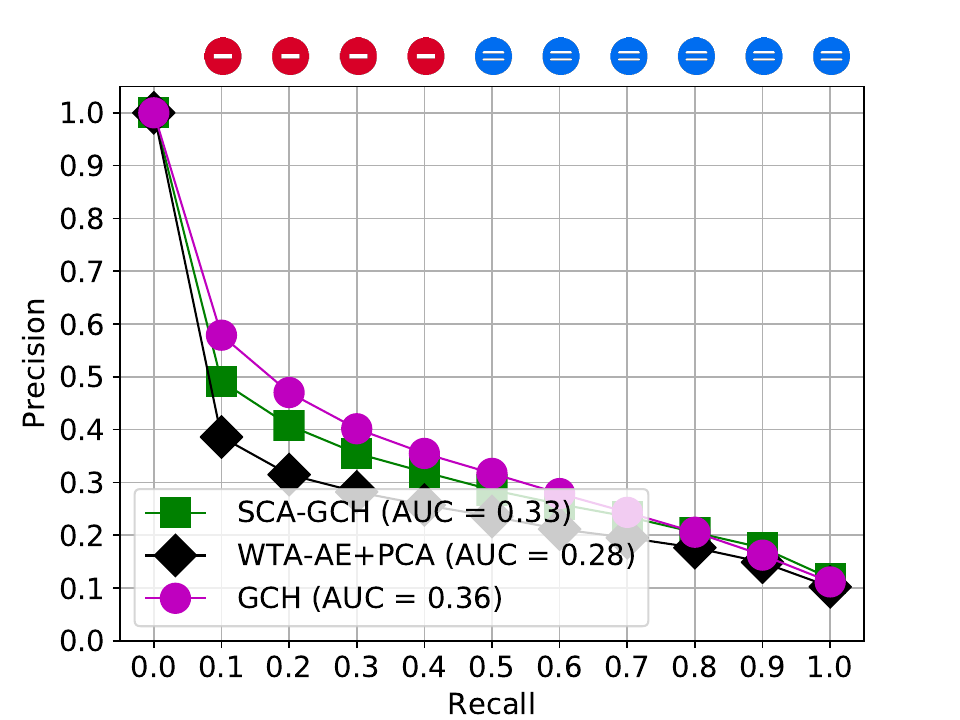}\label{chart:pr_gch_la_msrcorid_16}&
		\includegraphics[width=0.25\textwidth]{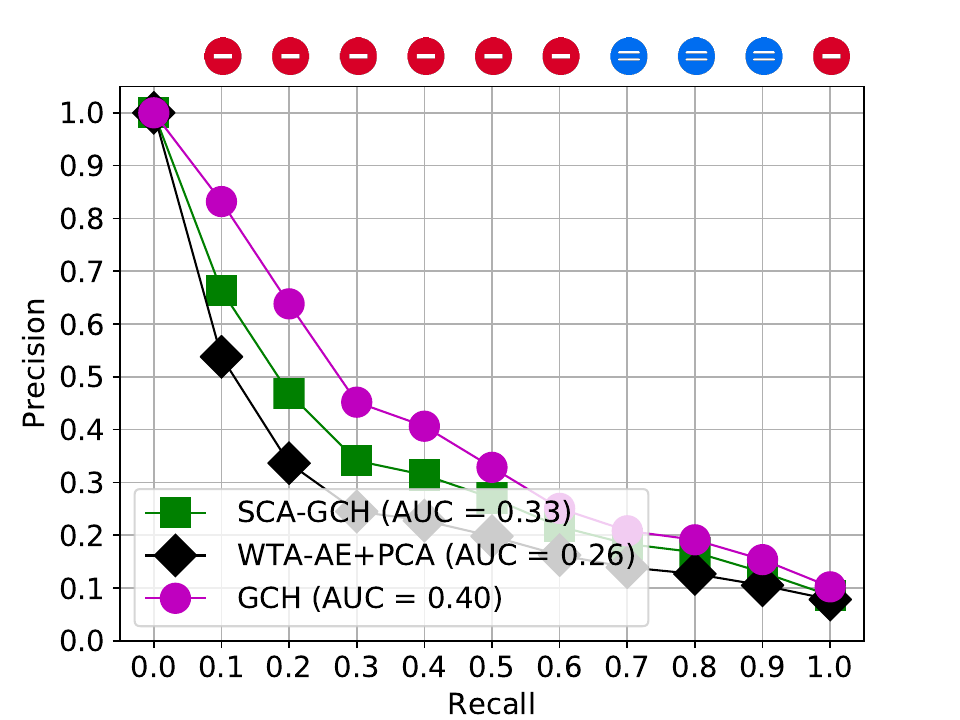}\label{chart:pr_gch_la_ucmerced_16}\\
		\rowname{16}&
		\includegraphics[width=0.25\textwidth]{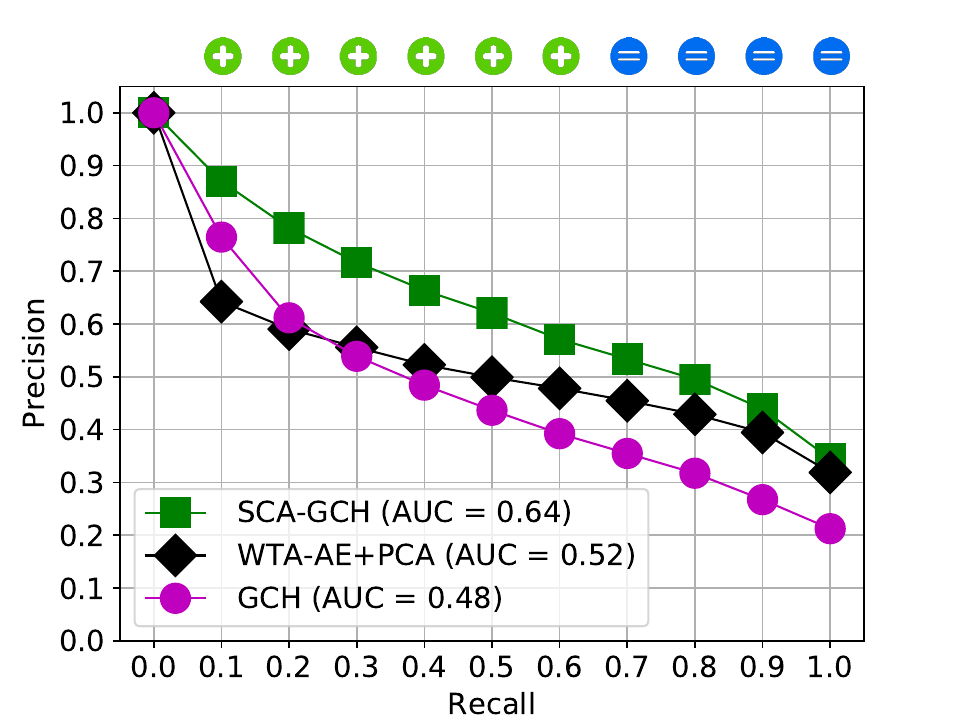}\label{chart:pr_gch_la_eth80_32}&
		\includegraphics[width=0.25\textwidth]{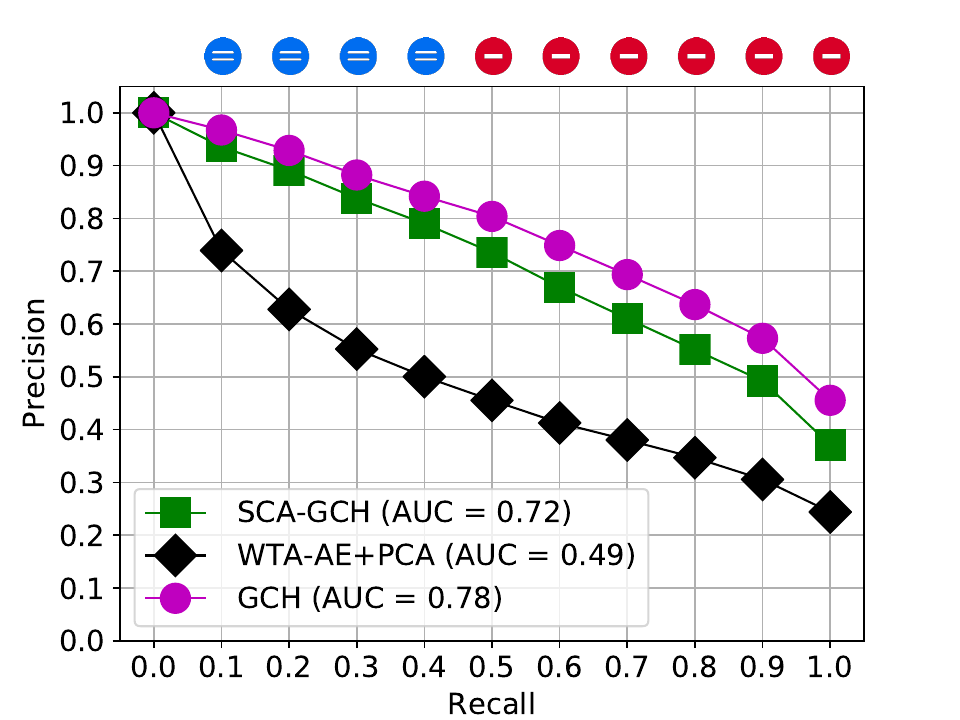}\label{chart:pr_gch_la_fruits__32}&
		\includegraphics[width=0.25\textwidth]{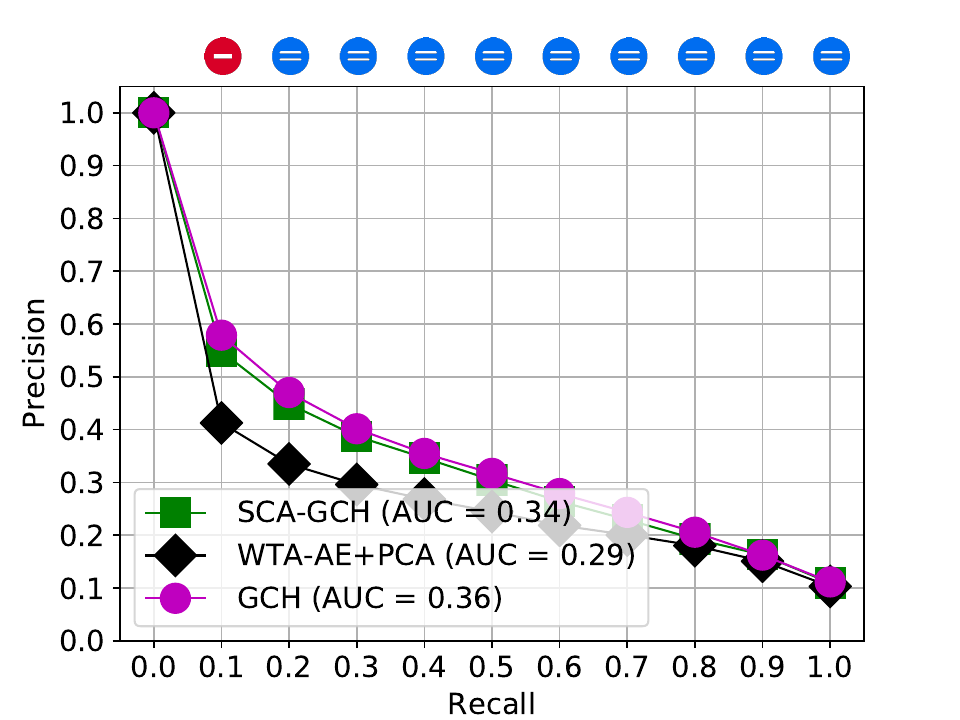}\label{chart:pr_gch_la_msrcorid_32}&
		\includegraphics[width=0.25\textwidth]{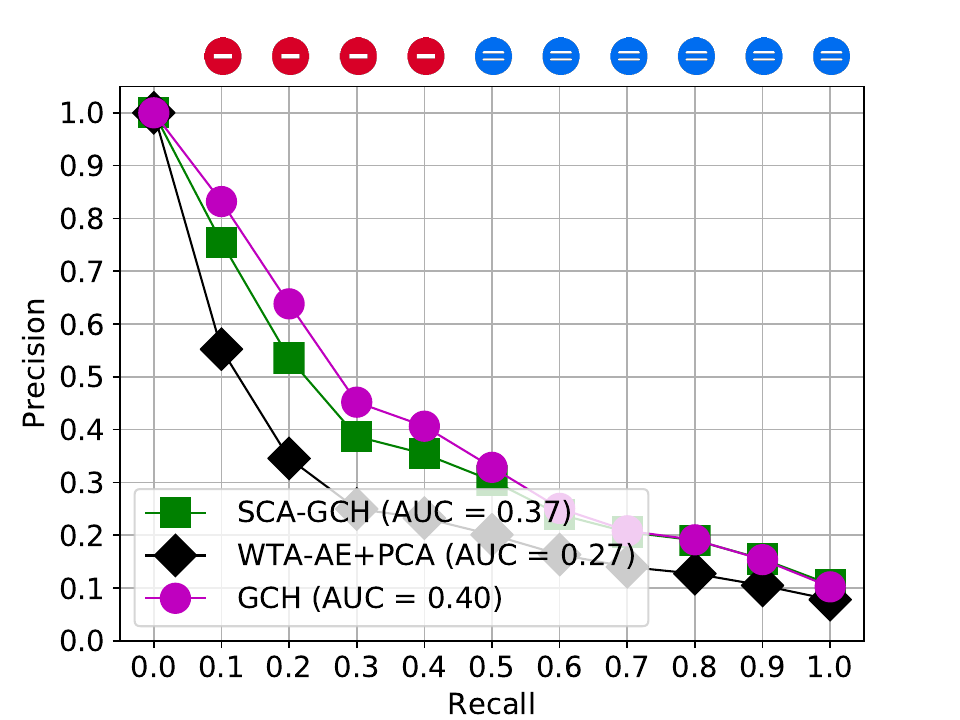}\label{chart:pr_gch_la_ucmerced_32}\\
		\rowname{32}&
		\includegraphics[width=0.25\textwidth]{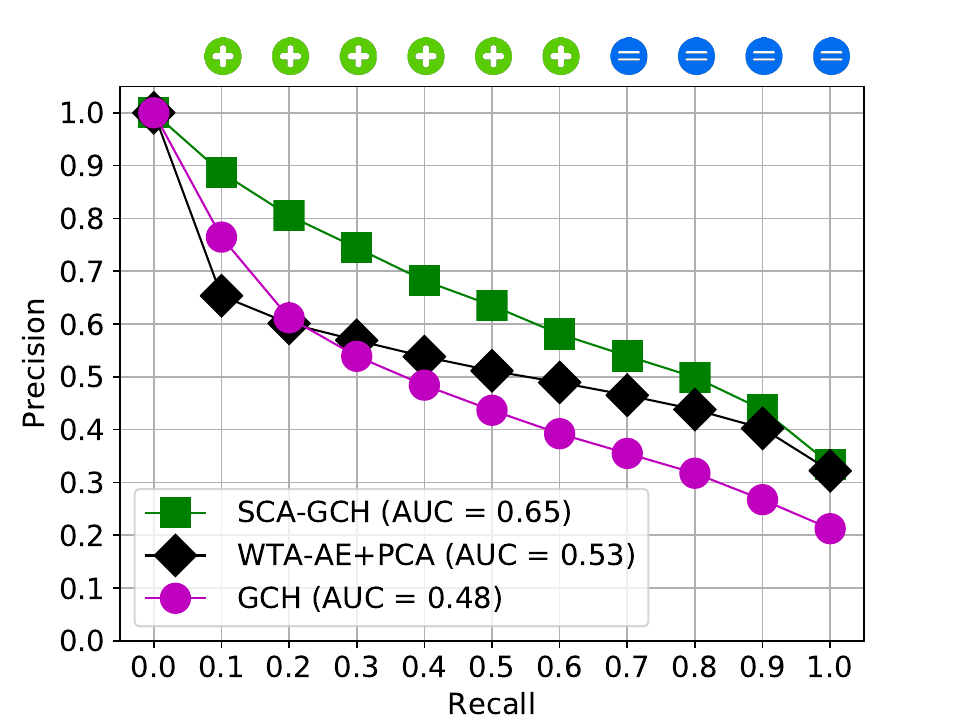}\label{chart:pr_gch_la_eth80_64}&
		\includegraphics[width=0.25\textwidth]{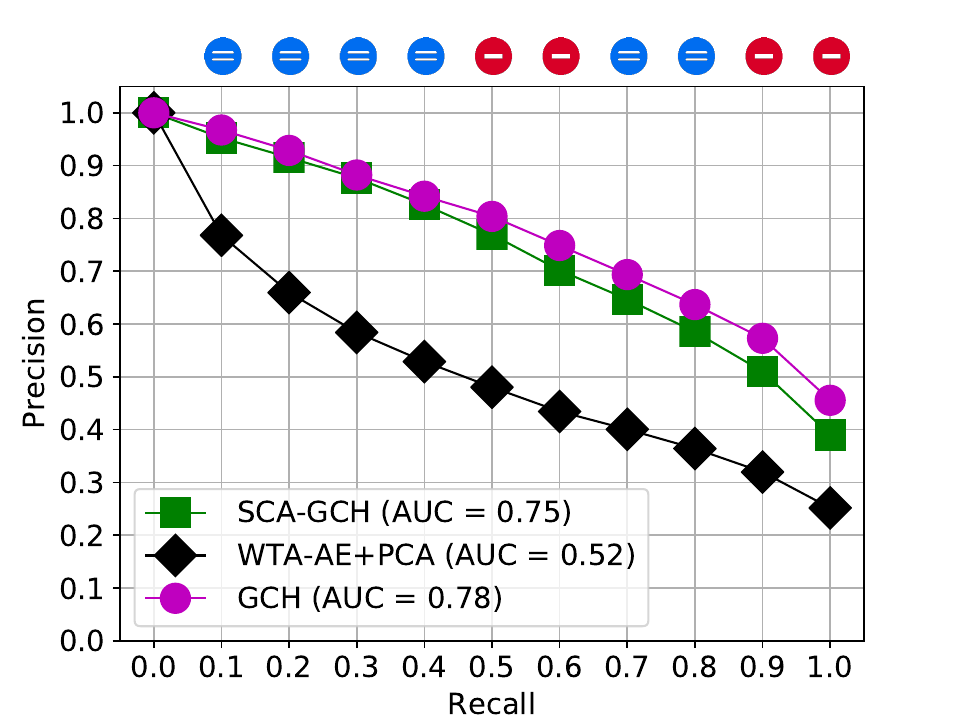}\label{chart:pr_gch_la_fruits__64}&
		\includegraphics[width=0.25\textwidth]{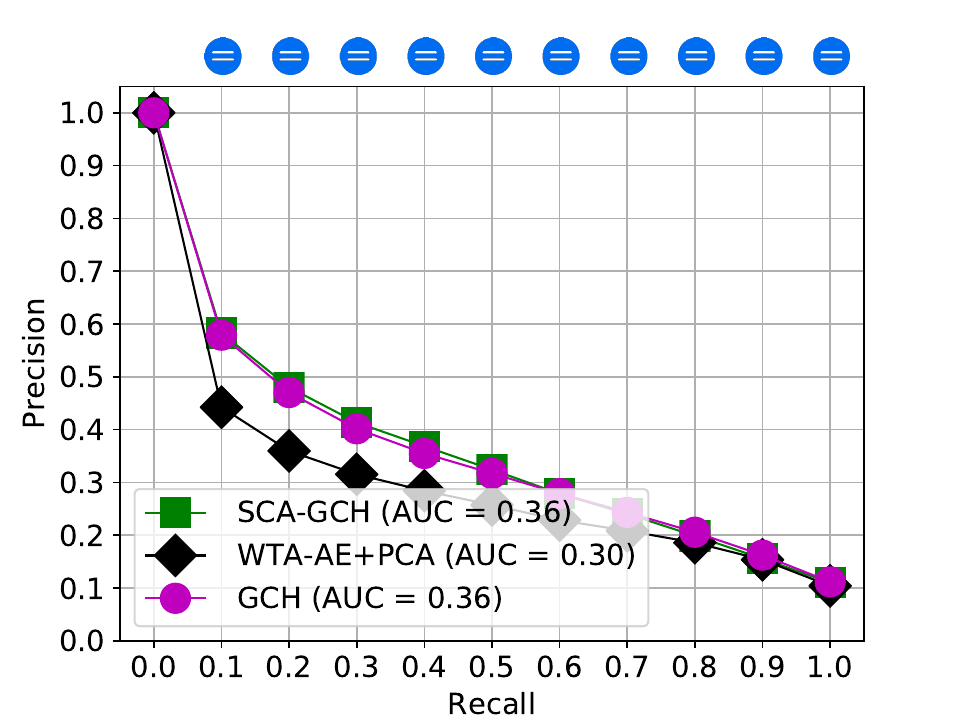}\label{chart:pr_gch_la_msrcorid_64}&
		\includegraphics[width=0.25\textwidth]{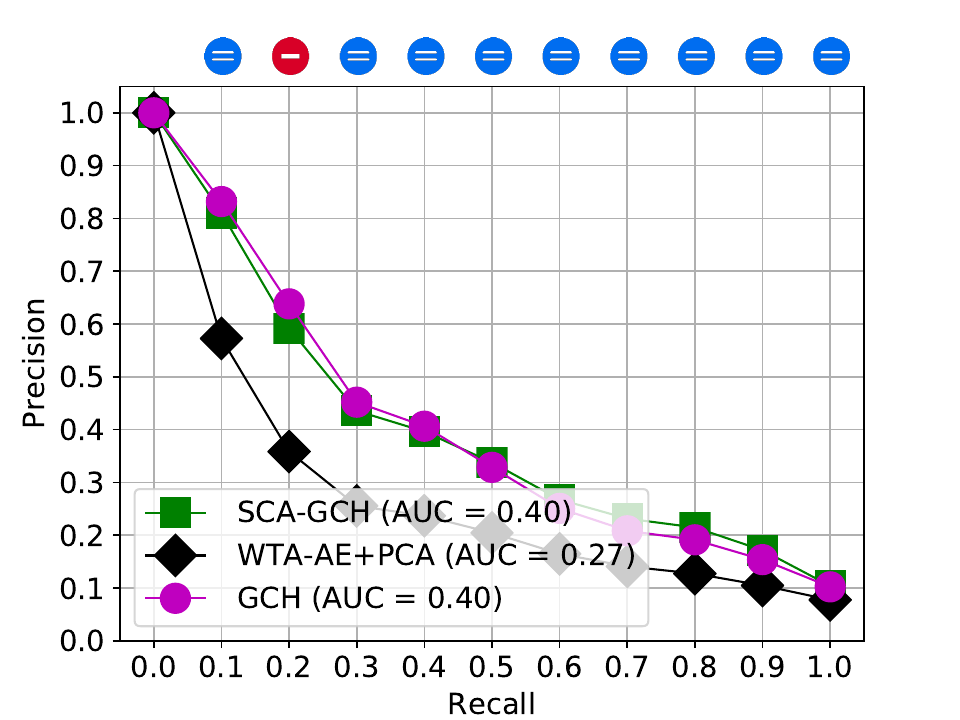}\label{chart:pr_gch_la_ucmerced_64}\\
		\rowname{48}&
		\includegraphics[width=0.25\textwidth]{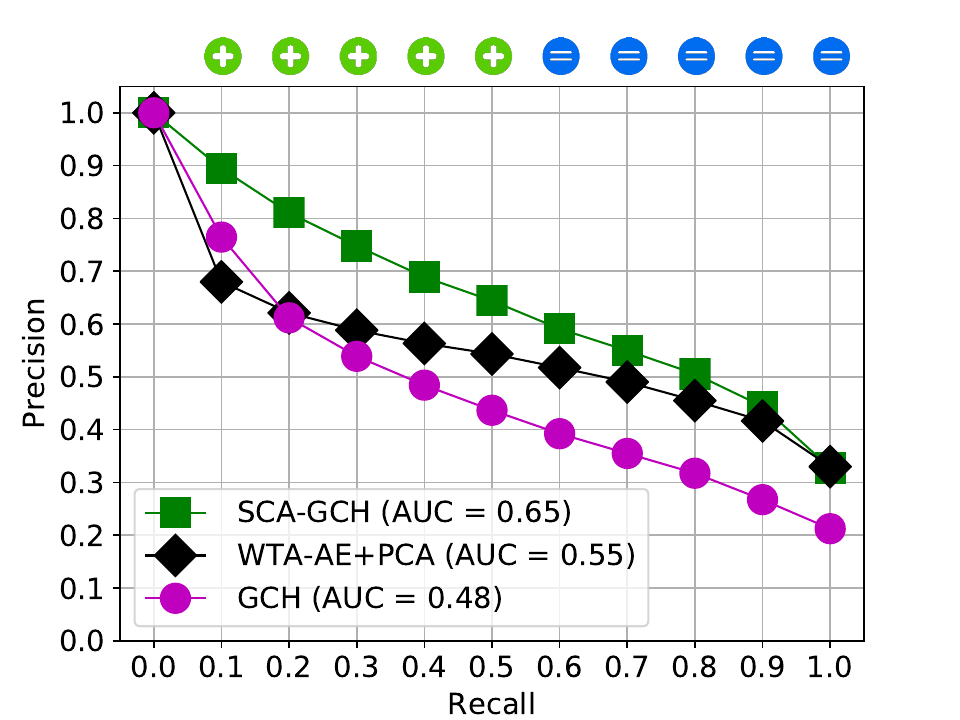}\label{chart:pr_gch_la_eth80_96}&
		\includegraphics[width=0.25\textwidth]{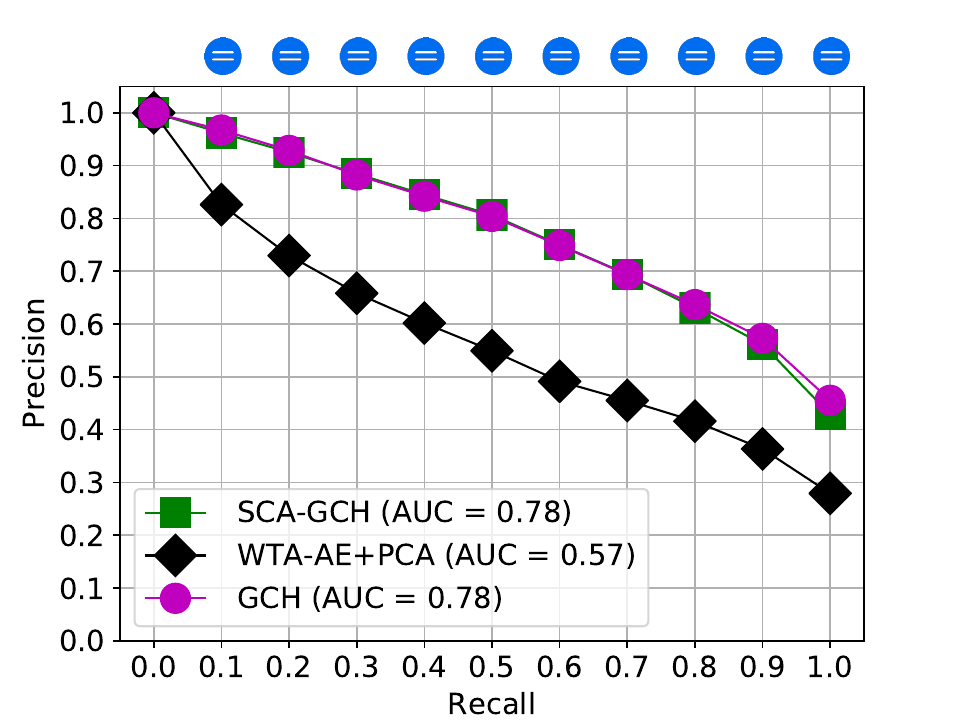}\label{chart:pr_gch_la_fruits__96}&
		\includegraphics[width=0.25\textwidth]{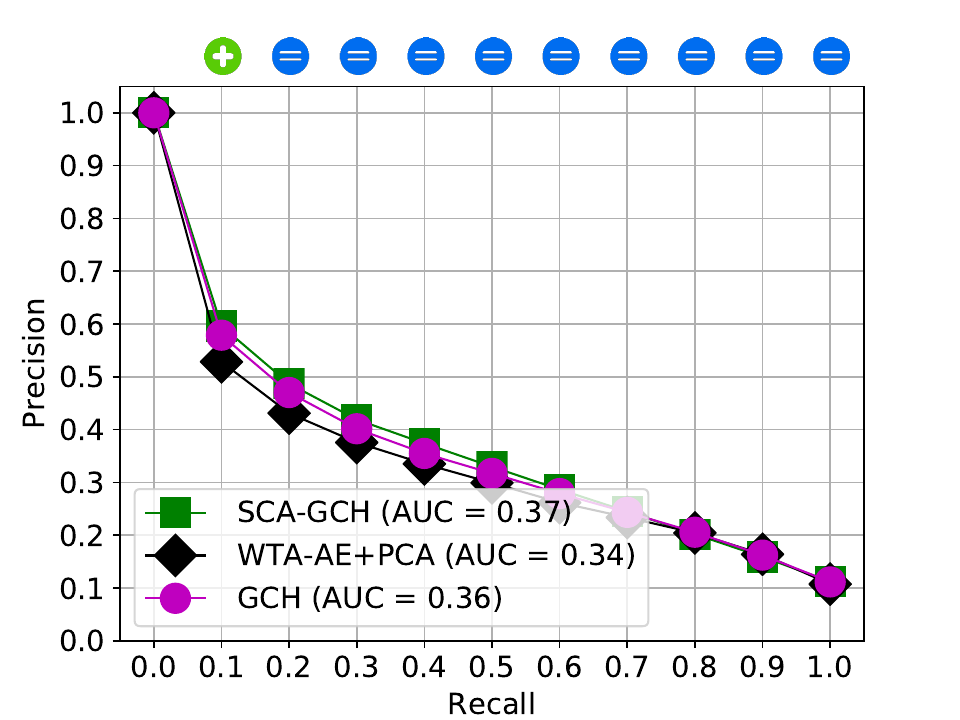}\label{chart:pr_gch_la_msrcorid_96}&
		\includegraphics[width=0.25\textwidth]{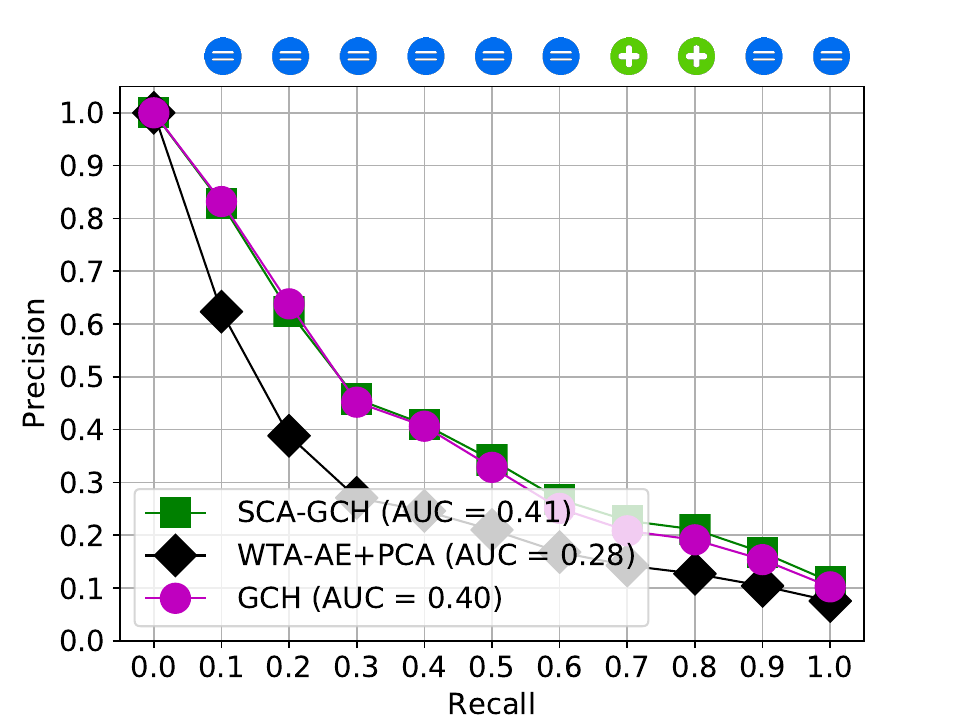}\label{chart:pr_gch_la_ucmerced_96}\\
		\rowname{64}&
		\includegraphics[width=0.25\textwidth]{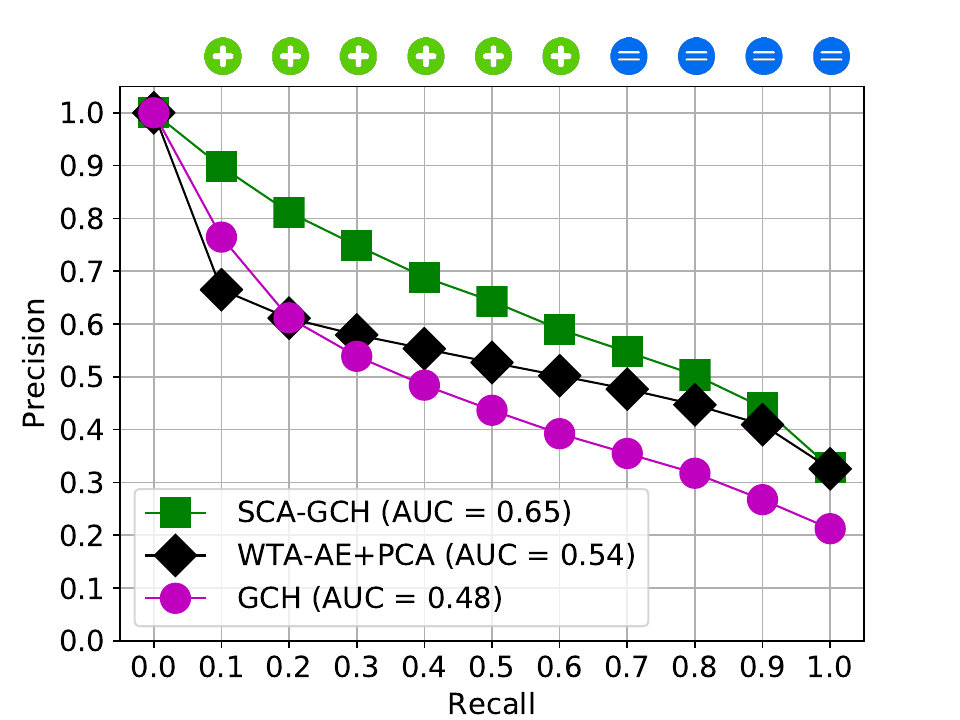}\label{chart:pr_gch_la_eth80_128}&
		\includegraphics[width=0.25\textwidth]{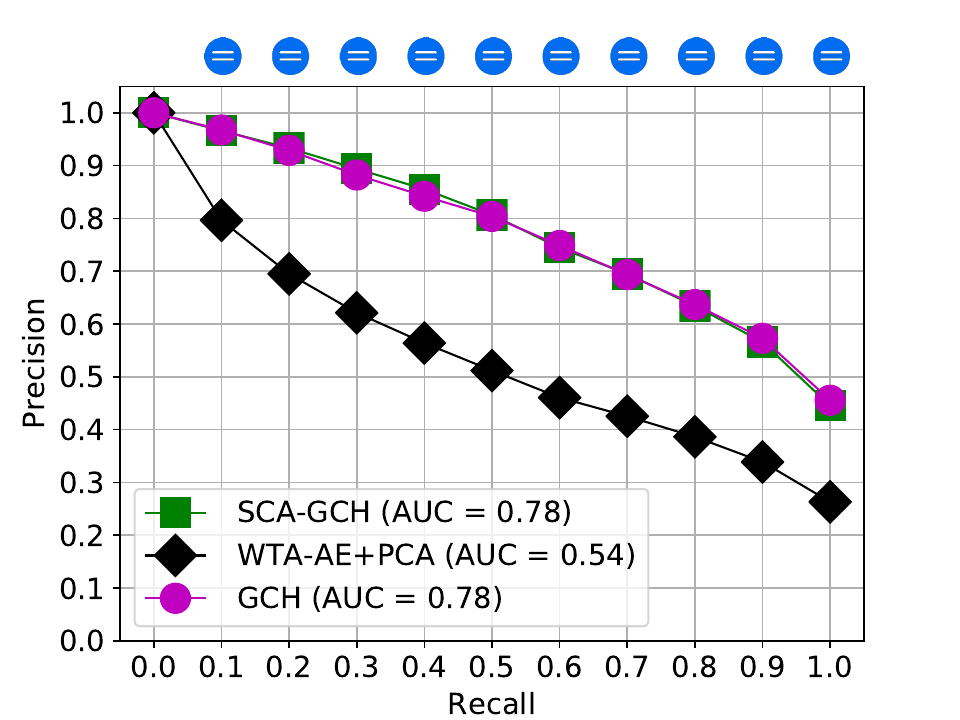}\label{chart:pr_gch_la_fruits__128}&
		\includegraphics[width=0.25\textwidth]{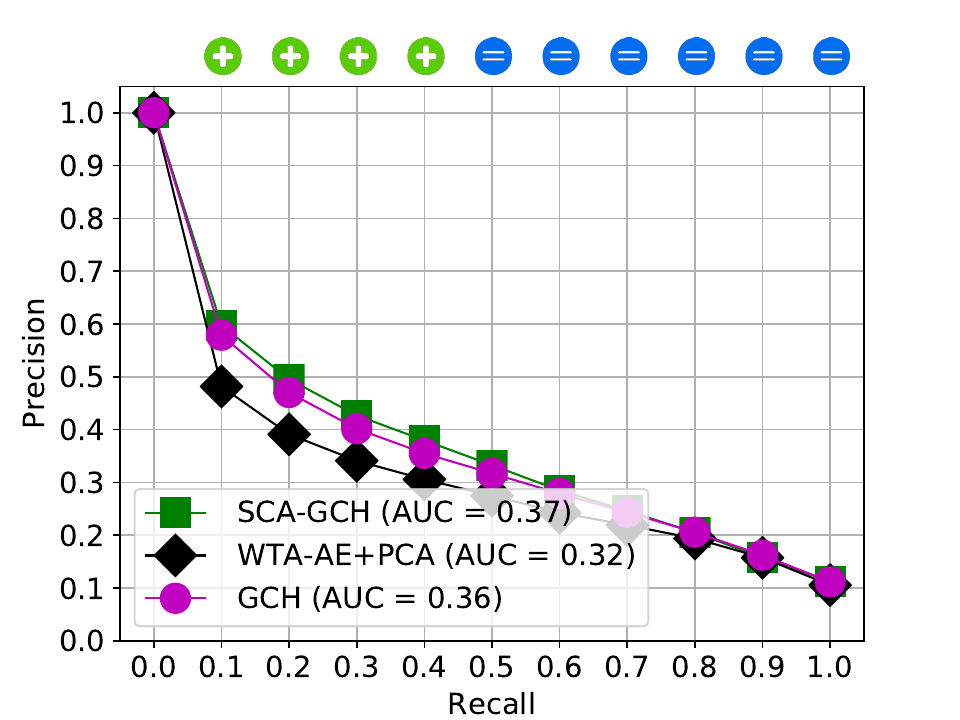}\label{chart:pr_gch_la_msrcorid_128}&
		\includegraphics[width=0.25\textwidth]{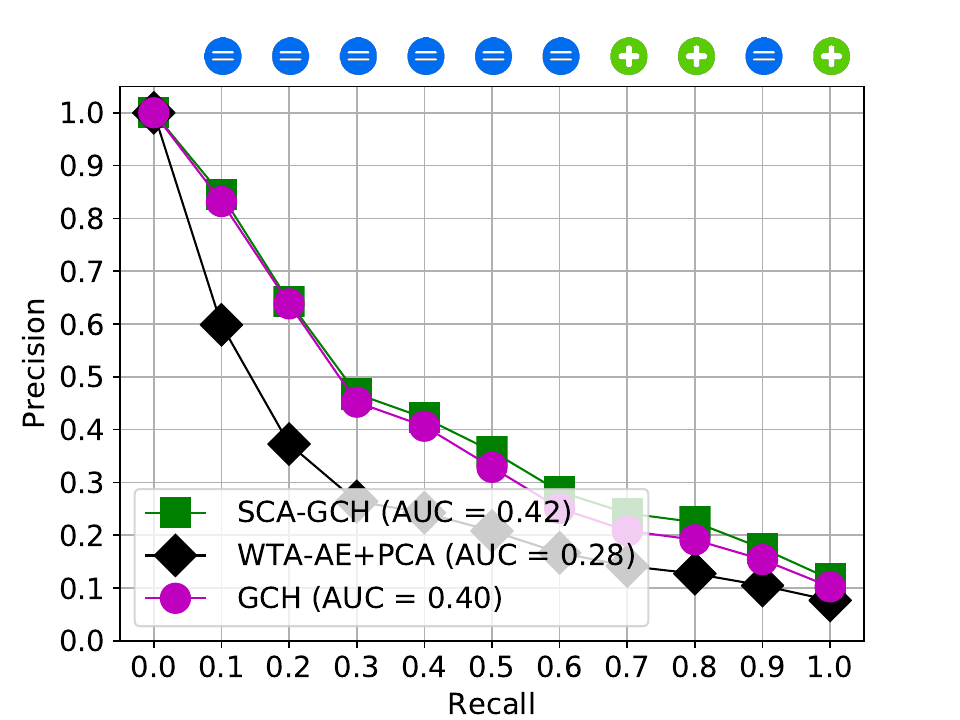}\label{chart:pr_gch_la_ucmerced_128}\\
		\rowname{128}&
		\includegraphics[width=0.25\textwidth]{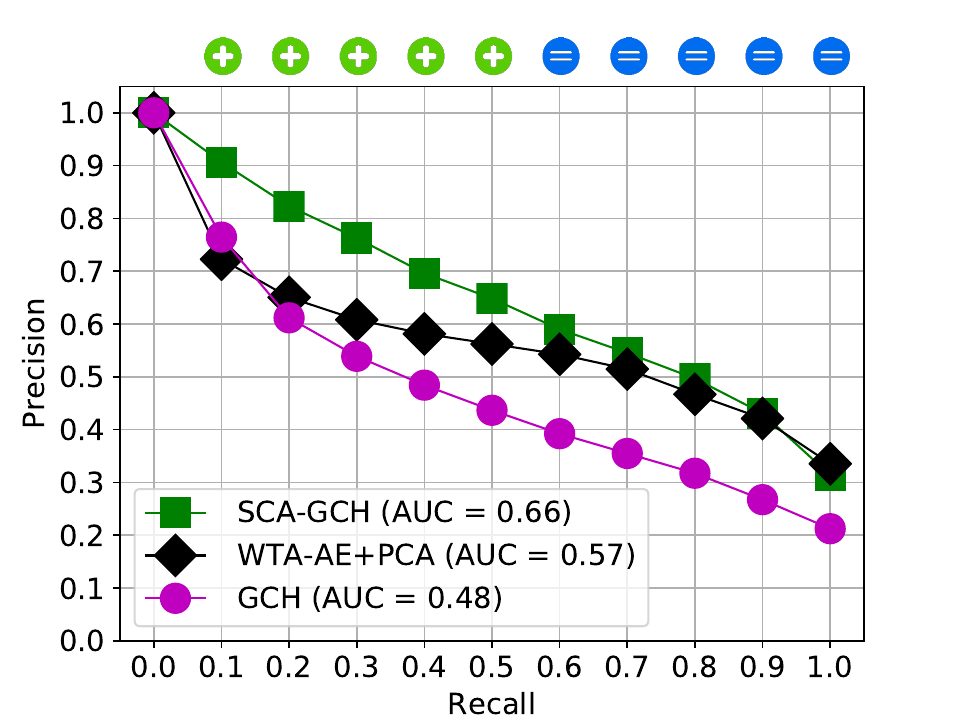}\label{chart:pr_gch_la_eth80_256}&
		\includegraphics[width=0.25\textwidth]{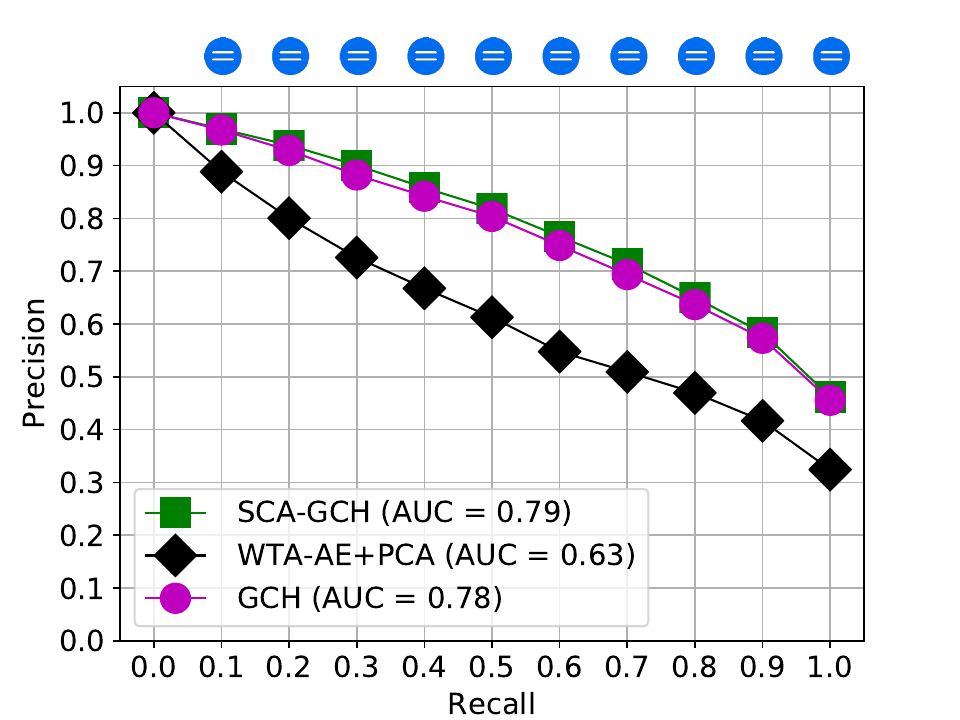}\label{chart:pr_gch_la_fruits__256}&
		\includegraphics[width=0.25\textwidth]{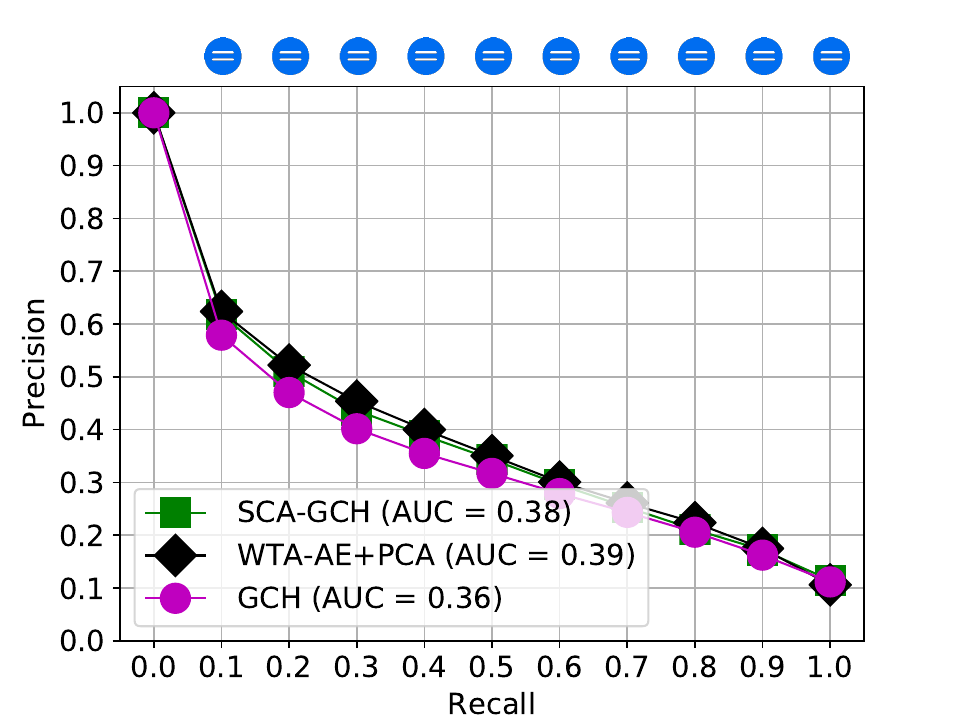}\label{chart:pr_gch_la_msrcorid_256}&
		\includegraphics[width=0.25\textwidth]{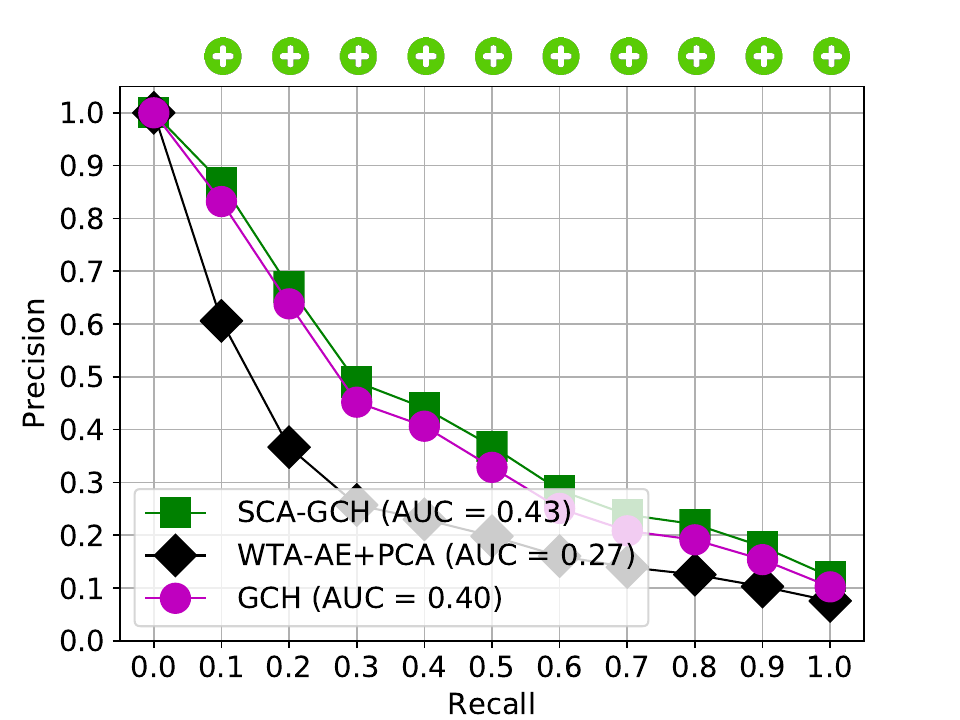}\label{chart:pr_gch_la_ucmerced_256}\\
		\rowname{192}&
		\includegraphics[width=0.25\textwidth]{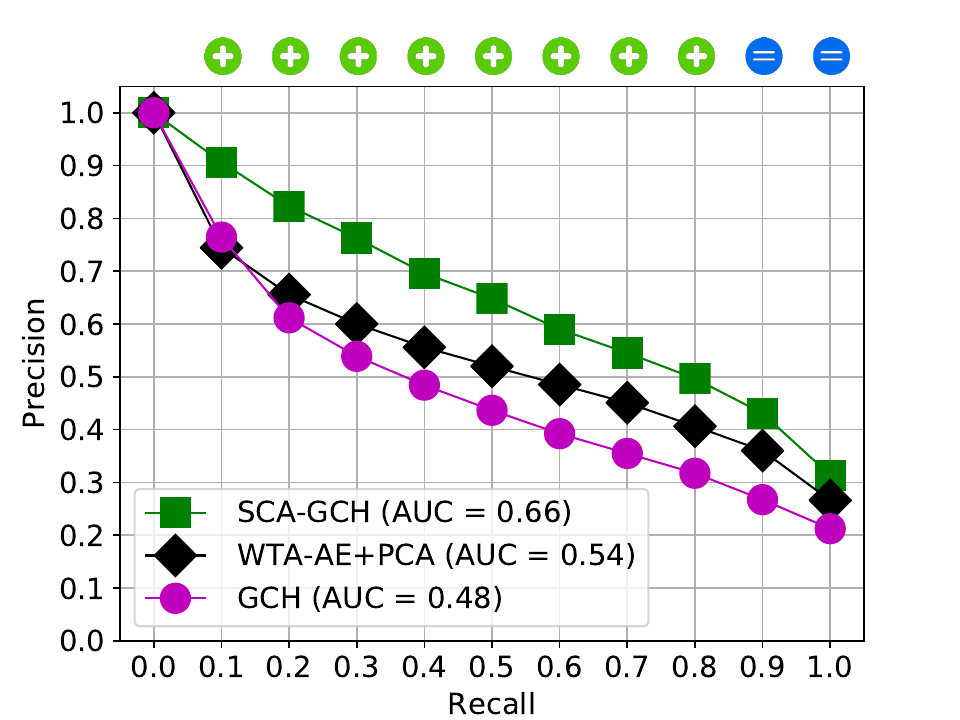}\label{chart:pr_gch_la_eth80_384}&
		\includegraphics[width=0.25\textwidth]{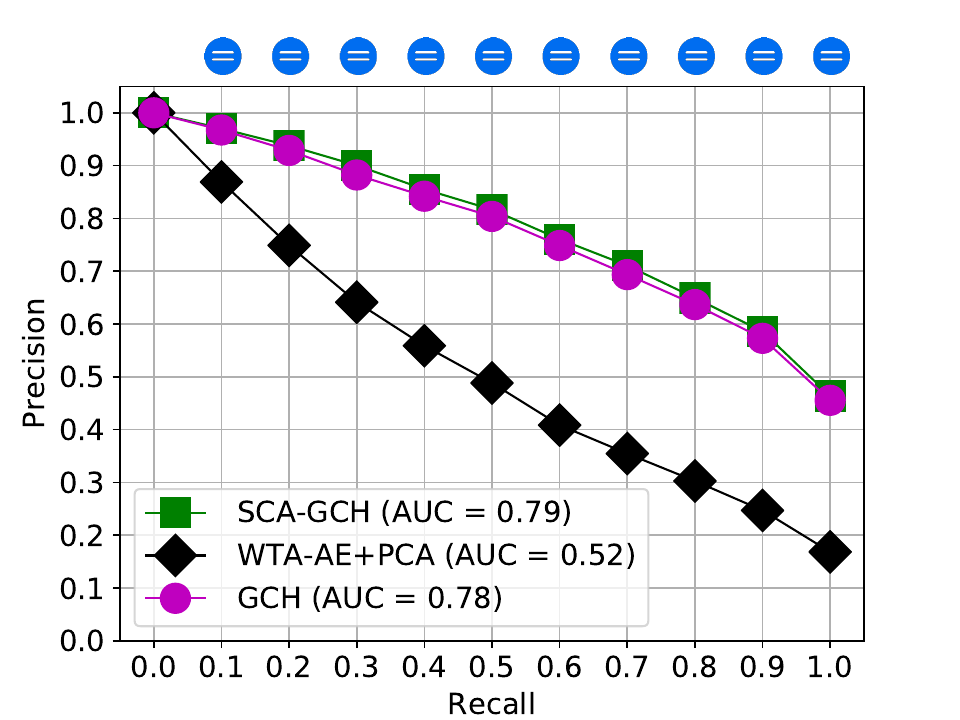}\label{chart:pr_gch_la_fruits__384}&
		\includegraphics[width=0.25\textwidth]{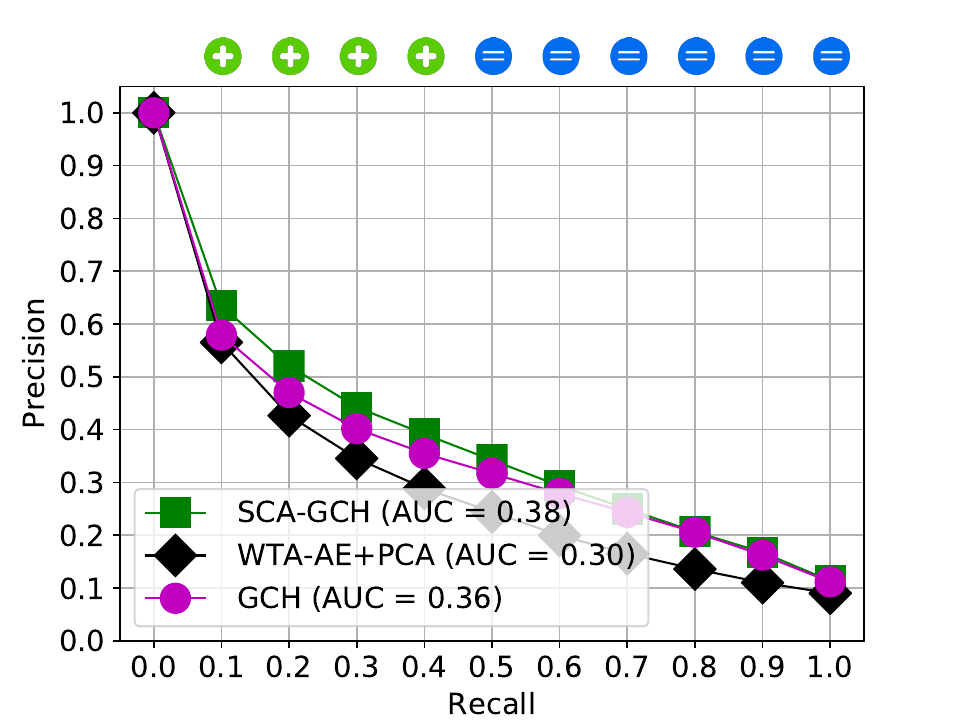}\label{chart:pr_gch_la_msrcorid_384}&
		\includegraphics[width=0.25\textwidth]{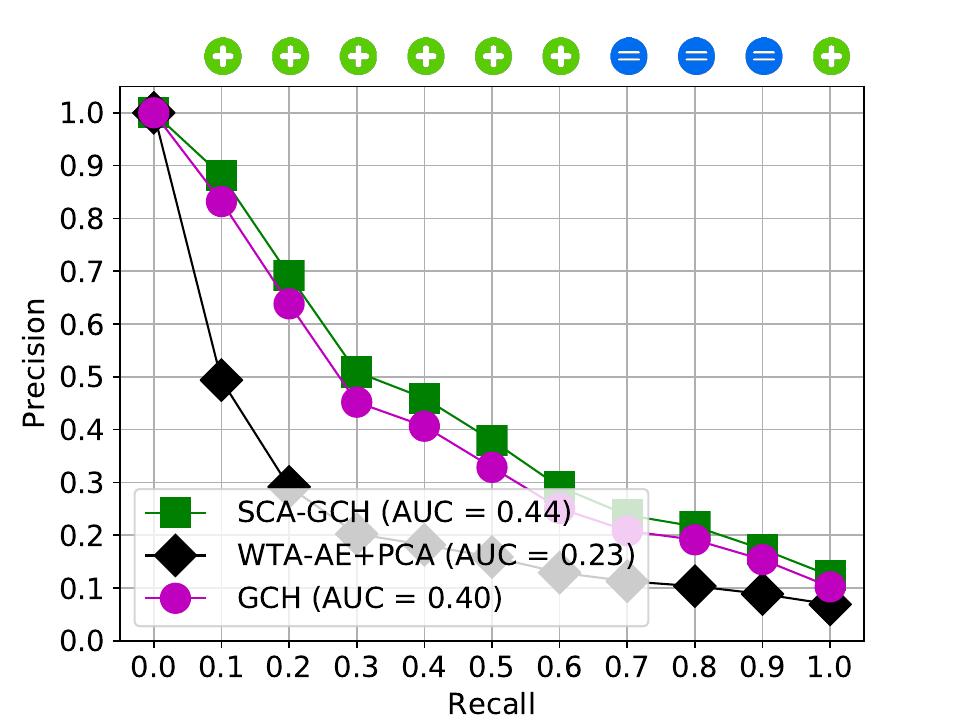}\label{chart:pr_gch_la_ucmerced_384}
	\end{tabular}
	\caption{Comparison between the Precision-Recall curves of SCA, WTA Autoencoder and GCH feature extractor considering all representation size limits for the datasets \textit{ETH-80}, \textit{Supermarket Produce}, \textit{MSRCORID}, and \textit{UCMerced Landuse}. 
	We recommend colourful printing for adequate visualization.}%
	\label{chart:pr_gch_la2}
\end{figure}

\section{Conclusions}
\label{sec:conclusions} 

We proposed two approaches of a representation learning method, which intends to provide more effective and compact image representations by optimizing the colour quantization for the image domain. We performed experiments on eight different image datasets comparing the results with a pre-defined quantization approach and a Sparse Autoencoder in terms of effectiveness performance on content-based image retrieval tasks. Methods are also evaluated in terms of the representation dimensionality.

The first approach, named Unconstrained approach, produced representations that outperformed the hand-crafted baselines in terms of effectiveness but presented feature vectors with several times higher dimensionality.
\edit{It also outperformed the effectiveness of the autoencoder representation, but presenting intensively lower sizes.}
The second approach, which imposes a limitation on the representation dimension (Size-Constrained approach), presented, in general, better effectiveness results for the same dimensionality (e.g., 128 bins). In other situations, this approach reduced the representation size up to 50\%, maintaining statistically comparable performance to the hand-crafted baselines. Finally, the SCA approach also produced results that imposed a reduction of more than 75\% of the storage requirements, but presented poor effectiveness performance, showing the existence of a trade-off between compactness and effectiveness.

Since the representations are based on color histograms, the over- and the sub-sampling of determined color space regions allows for the identification of more effective representations and, consequently, improvements on the search performance. 
Furthermore, a domain-oriented quantization allows for discarding the less contributing tonalities resulting in a possible reduction of the representation size.

In the end, the results confirm our hypothesis, for the tested scenarios, that it was possible to produce more effective and compact fitness by exploring a colour quantization optimized for the image domain. Moreover, our method is capable of improving already existent feature extraction methods by providing descriptions more effective in terms of representation quality and more compact according to a parametric upper bound. This research, therefore, opens novel opportunities for future investigation. \editt{We plan to assess the effects of the proposed quantization approches to other image processing applications such as image classification~{\cite{Li:2019}}, 
image segmentation~{\cite{Garcia:2020}} 
and image dehazing~{\cite{Zhang:2019}}. }
We also plan to investigate the impact of the resulting quantization when combined with deep-learning-based feature extractors.

\section*{Acknowledgments}
	This study was financed in part by: the \emph{Coordena\c{c}\~{a}o de Aperfei\c{c}oamento de Pessoal de N\'{i}vel Superior} - Brasil (CAPES) - Finance Code 001; 
the Brazilian National Council for Scientific and Technological Development (CNPq) - grants \#424700/2018-2 and \#311395/2018-0; and the Minas Gerais Research Foundation (FAPEMIG) - grant APQ-00449-17.
	Authors are also grateful to CAPES (grant
	\#88881.145912/2017-01), S\~ao Paulo Research Foundation -- FAPESP (grants \#2014/12236-1, \#2015/24494-8, \#2016/50250-1, and \#2017/20945-0) and the FAPESP - Microsoft Virtual Institute (grants \#2013/50155-0, \#2013/50169-1, and
	\#2014/50715-9). 

\bibliographystyle{spbasic}      
\bibliography{manuscript}   

\end{document}